\PassOptionsToPackage{dvipsnames}{xcolor}

\def\papersize    {b5paper} % Final papersize (b5paper/a4paper), recommended papersize for DTU is b5paper
\RequirePackage[l2tabu,orthodox]{nag} % Old habits die hard
\documentclass[\papersize,twoside,showtrims,extrafontsizes]{memoir}

\newif\iffull
\fulltrue
\setlrmarginsandblock{2.5cm}{2cm}{*}   % outer/inner
\setulmarginsandblock{3.5cm}{3.3cm}{*}     % top/bottom
\checkandfixthelayout

\usepackage[T1]{fontenc}
\usepackage[utf8]{inputenc}
\usepackage{textcomp}
\showtrimsoff

\checkandfixthelayout                 % Check if errors in paper format!
\sideparmargin{outer}                 % Put sidemargins in outer position (why the fuck is this option not default by the class?)
\usepackage{microtype}
\usepackage{mathtools}
\usepackage{listings}                 % Source code printer for LaTeX
\usepackage{tikz}

\usepackage{amsmath,dsfont,amsfonts,braket}
\usepackage[shortlabels]{enumitem}

\usepackage{acronym}
\usepackage[hang,flushmargin,bottom]{footmisc} % Remove footnote indentation
\usepackage{titlesec}
\usepackage{fancyhdr}

\usepackage[hyphens]{url}             % Allow hyphens in URL's
\usepackage[unicode=false,psdextra]{hyperref}                 % References package

\usepackage{pdfpages}
\usepackage{graphicx}                 % Including graphics and using colours
\usepackage[labelformat=simple]{subcaption}     % To use the environment "subfigure"
\usepackage{xcolor,colortbl}
\usepackage{eso-pic}                  % Watermark and other bag
\usepackage{preamble/dtucolors}
\graphicspath{{graphics/}}

\usepackage{bm}
\usepackage{amssymb}
\usepackage{booktabs}
\usepackage{wrapfig}
\usepackage{multirow}
\usepackage[danish,main=english]{babel}
\usepackage{csquotes}       % language sensitive quotation facilities

\usepackage[nodayofweek]{datetime}
\newdateformat{mydate}{{\THEDAY}{ }\monthname[\THEMONTH] \THEYEAR} % Personal date style

\usepackage{flafter}  % floats is positioned after or where it is defined! 
\usepackage[nameinlink,capitalise]{cleveref} % Clever references. Options: "fig. !1!" --> "!Figure 1!"

\DeclareCaptionFont{dtu}{\normalsize\selectfont}

\usepackage[labelfont={dtu,bf},labelsep=period]{caption}
\captionnamefont{\bfseries}
\subcaptionlabelfont{\bfseries}
\newsubfloat{figure}
\newsubfloat{table}
\crefformat{equation}{(#2#1#3)}
\crefrangeformat{equation}{(#3#1#4) to~(#5#2#6)}
\crefmultiformat{equation}{(#2#1#3)}%
{ and~(#2#1#3)}{, (#2#1#3)}{ and~(#2#1#3)}

\titleformat{\part}[display]{\filcenter\chapnamefont\fontsize{34pt}{36pt}\selectfont\setlength{\parskip}{-1.5cm}}{\vspace{4cm}\color{dtugray}{\chapnamefont\fontsize{32pt}{0pt}\selectfont}
\partname\chapnamefont\color{dtured}\fontsize{38pt}{0pt}\selectfont\hspace{.3em}\thepart}{4ex}{}
\titlespacing{\part}{0pt}{0pt}{0pt}

\makeatletter
\makechapterstyle{mychapterstyle}{
    \chapterstyle{default}
    \def\format{\normalfont\bfseries}  
    \setlength\beforechapskip{10mm}
    \renewcommand*{\chapnamefont}{\format\fontsize{32}{32}}

    \renewcommand*{\printchaptername}{\chapnamefont\MakeUppercase{\@chapapp}}
    \patchcommand{\printchaptername}{\raggedleft\begingroup\color{dtugray}}{\endgroup}
    
    \patchcommand{\printchapternum}{\begingroup\color{dtured}}{\endgroup}
    
    \setlength\midchapskip{3.2ex}   % vskip between "chapter X" and chapter name

}
\makeatother
\chapterstyle{mychapterstyle}

\let\appendixpagenameorig\appendixpagename
\renewcommand{\appendixpagename}{\vspace{-2cm}\normalfont\format\fontsize{42pt}{0pt}\selectfont\appendixpagenameorig} % Change font for the "Appendices" page  % #AD

\newcounter{protocol}

\def\hffont{\small}   % \def\hffont{\sffamily\small}  % #AD
\makepagestyle{myruled}
\makeheadrule{myruled}{\textwidth}{\normalrulethickness}
\makeevenhead{myruled}{\hffont\thepage}{}{\hffont\leftmark}
\makeoddhead{myruled}{\hffont\rightmark}{}{\hffont\thepage}
\makeevenfoot{myruled}{}{}{}
\makeoddfoot{myruled}{}{}{}
\makepagestyle{appruled}
\makeheadrule{appruled}{\textwidth}{\normalrulethickness}
\makeevenhead{appruled}{\hffont\thepage}{}{\hffont\appendixname~\leftmark}
\makeoddhead{appruled}{\hffont\appendixname~\rightmark}{}{\hffont\thepage}
\makeevenfoot{appruled}{}{}{}
\makeoddfoot{appruled}{}{}{}
\makepsmarks{myruled}{
    \nouppercaseheads
    \createmark{chapter}{both}{shownumber}{}{\space}
    \createmark{section}{right}{shownumber}{}{\space}
    \createplainmark{toc}{both}{\contentsname}
    \createplainmark{lof}{both}{\listfigurename}
    \createplainmark{lot}{both}{\listtablename}
    \createplainmark{bib}{both}{\bibname}
    \createplainmark{index}{both}{\indexname}
    \createplainmark{glossary}{both}{\glossaryname}
}
\copypagestyle{cleared}{myruled}      % When \cleardoublepage, use myruled instead of empty
\makeevenhead{cleared}{\hffont\thepage}{}{} % Remove leftmark on cleared pages
\makeevenfoot{plain}{}{}{}            % No page number on plain even pages (chapter begin)
\makeoddfoot{plain}{}{}{}             % No page number on plain odd pages (chapter begin)

\hypersetup{
    pdfdisplaydoctitle,
    bookmarksnumbered=true,
    bookmarksopen,
    breaklinks,
    linktoc=all,
    plainpages=false,
    unicode=true,
    colorlinks=true,
    allcolors=Blue!65!black,
}

\makeatletter
\renewcommand{\@memb@bchap}{%
  \ifnobibintoc\else
    \phantomsection
    \addcontentsline{toc}{part}{\bibname}%
  \fi
  \chapter*{\bibname}%
  \bibmark
  \prebibhook
}
\let\oldtableofcontents\tableofcontents
\newcommand{\newtableofcontents}{
    \@ifstar{\oldtableofcontents*}{
        \phantomsection\addcontentsline{toc}{chapter}{\contentsname}\oldtableofcontents*}}
\let\tableofcontents\newtableofcontents
\makeatother

\newcommand{\prefrontmatter}{
    \pagenumbering{alph}
}

\newcommand{\frieze}{%
    \AddToShipoutPicture*{
        \put(0,0){
            \parbox[b][\paperheight]{\paperwidth}{%
                \includegraphics[trim=130mm 0 0 0,width=0.9\textwidth]{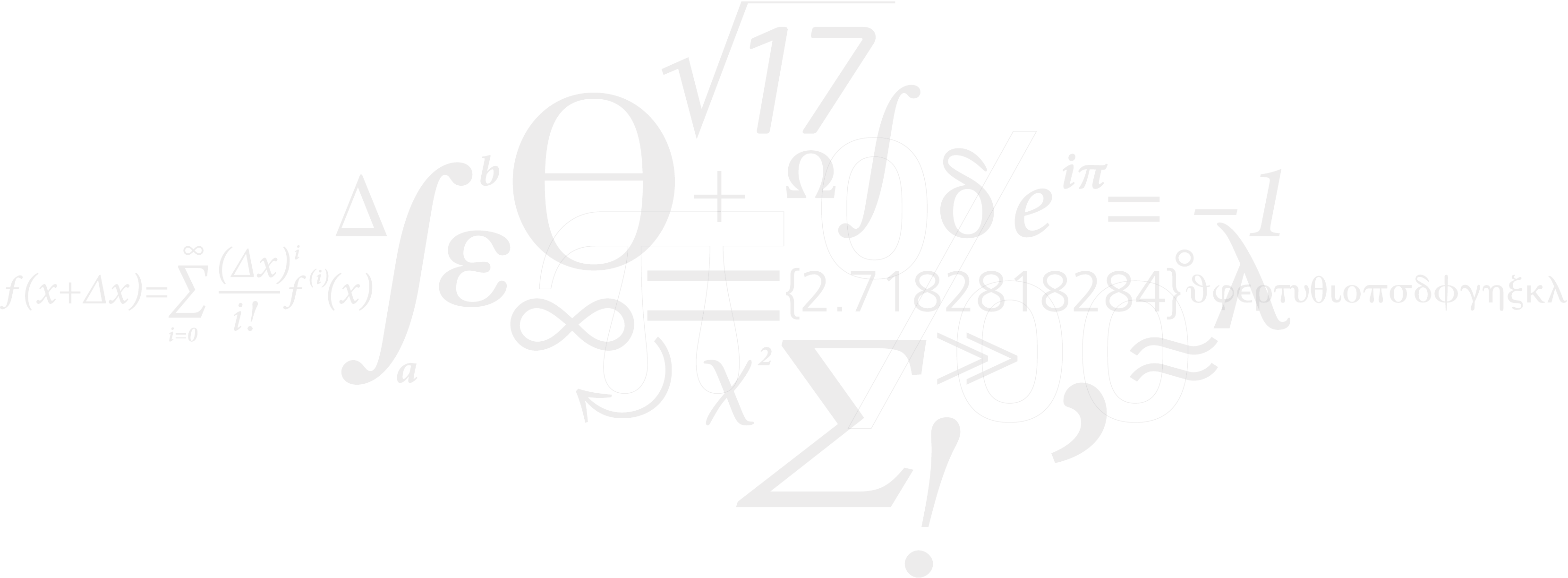}
                \vspace*{2.5cm}
            }
        }
    }
}

\makeatletter
\patchcmd{\leavespergathering}{\ifnum\@memcnta<\tw@}{\ifnum\@memcnta<\@ne}{
    \leavespergathering{1}
    \patchcmd{\@memensuresigpages}{\repeat}{\repeat\frieze}{}{}
}{}
\makeatother

\makeatletter
\def\MT@is@composite#1#2\relax{%
  \ifx\\#2\\\else
    \expandafter\def\expandafter\MT@char\expandafter{\csname\expandafter
                    \string\csname\MT@encoding\endcsname
                    \MT@detokenize@n{#1}-\MT@detokenize@n{#2}\endcsname}%
    \ifx\UnicodeEncodingName\@undefined\else
      \expandafter\expandafter\expandafter\MT@is@uni@comp\MT@char\iffontchar\else\fi\relax
    \fi
    \expandafter\expandafter\expandafter\MT@is@letter\MT@char\relax\relax
    \ifnum\MT@char@ < \z@
      \ifMT@xunicode
        \edef\MT@char{\MT@exp@two@c\MT@strip@prefix\meaning\MT@char>\relax}%
          \expandafter\MT@exp@two@c\expandafter\MT@is@charx\expandafter
            \MT@char\MT@charxstring\relax\relax\relax\relax\relax
      \fi
    \fi
  \fi
}
\def\MT@is@uni@comp#1\iffontchar#2\else#3\fi\relax{%
  \ifx\\#2\\\else\edef\MT@char{\iffontchar#2\fi}\fi
}
\makeatother

\usepackage{bigdelim}
\usepackage{xspace}

\newcommand{\pubinfo}[3]{%
    \textbf{Authors:} #1
    
    \textbf{Status:} #2
    
    \textbf{Abstract.} #3
}

\def\and{ }

\newcommand{\myfnsymbol}[2]{\leavevmode\raise#2\hbox{\normalfont\footnotesize#1}}
\newcommand{\fnstar}{\myfnsymbol{*}{.8ex}}
\newcommand{\fndagger}{\myfnsymbol{\dag}{1.2ex}}
\newcommand{\eqcontrib}{\rlap{\fnstar}\xspace}
\newcommand{\eqadvising}{\rlap{\fndagger}\xspace}
\newcommand{\eqcontribfn}{\fnstar}
\newcommand{\eqadvisingfn}{\fndagger}

\usepackage[
    backend=biber,
    style=authoryear-comp,
    maxbibnames=50,
    minbibnames=20,
    maxcitenames=3,
    mincitenames=1,
    uniquename=false, 
    uniquelist=false,
    sorting=nyt,
    backref=true,
    hyperref=true,
    dashed=true
]{biblatex}
\let\citep\parencite
\let\citet\textcite
\let\citeplain\cite
\let\cite\parencite
\DeclareFieldFormat{citehyperref}{%
  \DeclareFieldAlias{bibhyperref}{noformat}% Avoid nested links
  \bibhyperref{#1}}

\DeclareFieldFormat{textcitehyperref}{%
  \DeclareFieldAlias{bibhyperref}{noformat}% Avoid nested links
  \bibhyperref{%
    #1%
    \ifbool{cbx:parens}
      {\bibcloseparen\global\boolfalse{cbx:parens}}
      {}}}

\savebibmacro{cite}
\savebibmacro{textcite}

\renewbibmacro*{cite}{%
  \printtext[citehyperref]{%
    \restorebibmacro{cite}%
    \usebibmacro{cite}}}

\renewbibmacro*{textcite}{%
  \ifboolexpr{
    ( not test {\iffieldundef{prenote}} and
      test {\ifnumequal{\value{citecount}}{1}} )
    or
    ( not test {\iffieldundef{postnote}} and
      test {\ifnumequal{\value{citecount}}{\value{citetotal}}} )
  }
    {\DeclareFieldAlias{textcitehyperref}{noformat}}
    {}%
  \printtext[textcitehyperref]{%
    \restorebibmacro{textcite}%
    \usebibmacro{textcite}}}
    
\newcommand{\xb}{\mathbf{x}}
\newcommand{\zb}{\mathbf{z}}
\newcommand{\yb}{\mathbf{y}}
\newcommand{\ybpred}{\hat{\mathbf{y}}}
\newcommand{\gb}{\mathbf{g}}
\newcommand{\GGG}{\mathcal{G}}
\newcommand{\thetab}{\boldsymbol{\theta}}
\newcommand{\phib}{\boldsymbol{\phi}}
\newcommand{\epsilonb}{\boldsymbol{\epsilon}}
\newcommand{\dd}{\mathrm{d}}
\newcommand{\given}{\hspace{0.11em}|\hspace{0.06em}}
\newcommand{\ptheta}{p_{\thetab}}
\newcommand{\qphi}{q_{\phib}}
\newcommand{\qphitilde}{\tilde{q}_{\phib}}
\newcommand{\E}{\mathbb{E}}
\newcommand{\kl}{D_\mathrm{KL}}
\newcommand{\dataset}{\mathcal{D}}
\newcommand{\rsq}{$R^2$\xspace}
\newcommand{\up}{$\uparrow$\xspace}
\newcommand{\down}{$\downarrow$\xspace}

\DeclareMathOperator{\diag}{diag}

\newcommand{\code}{\texttt}

\makeatletter
\newcommand{\summary}[1]{%
  \@bsphack
  \@esphack
}
\makeatother

\usepackage{siunitx}
\makeatletter
\renewcommand{\paragraph}{%
  \@startsection{paragraph}{4}%
  {\z@}{3.9ex \@plus .1ex \@minus .1ex}{-1em}%
  {\normalfont\normalsize\bfseries}%
}
\makeatother

\setlist[itemize]{topsep=0pt,itemsep=0pt,partopsep=0pt,parsep=\parskip}
\setlist[enumerate]{topsep=0pt,itemsep=0pt,partopsep=0pt,parsep=\parskip}

\titlespacing*{\section}{0pt}{2.0em plus 0.2em minus 0.1em}{0.4em plus 0.1em minus 0.0em}
\titlespacing*{\subsection}{0pt}{1.6em plus 0.2em minus 0.1em}{0.4em plus 0.1em minus 0.0em}

\DeclareDelimFormat{nameyeardelim}{\addcomma\space}

\usepackage{enumitem}

\makeatletter
\DeclareRobustCommand{\crefnocompress}[1]{%
  \begingroup\@cref@compressfalse\cref{#1}\endgroup
}
\makeatother

\usepackage{blindtext}

\newcommand{\paperOne}{I\xspace}
\newcommand{\paperTwo}{II\xspace}
\newcommand{\paperThree}{III\xspace}

\begin{document}
\prefrontmatter
\thispagestyle{empty}             % No page numbers
\calccentering{\unitlength}

{\setlength{\parskip}{2pt}
\linespread{1.05}\selectfont

% \begin{adjustwidth*}{\unitlength}{-\unitlength}
%     \begin{adjustwidth}{-0.5cm}{-0.5cm}
%         % \sffamily   % #AD
%         \begin{flushright}
%             Doctor of Philosophy\\*[0cm]
%             Doctoral thesis in Photonics Engineering\\
%         \end{flushright}
%         \vspace*{\fill}
%         \noindent
%         \includegraphics[width=0.45\textwidth]{tex_dtu_compute_a_uk}\\*[0.2cm]
%         \HUGE \textbf{Title}\\*[0.6cm]
%         \parbox[b]{0.5\linewidth}{
%         \huge Andrea Dittadi\\*[1.2cm]
%         %\large\thesislocation{}, \the\year
%         } 
        
%         \hfill\includegraphics[scale=0.065]{DTU_Logo.png}
%     \end{adjustwidth}
% \end{adjustwidth*}
\begin{adjustwidth*}{\unitlength}{-\unitlength}
    \begin{adjustwidth}{-0.5cm}{-0.5cm}
        % \sffamily   % #AD
        \begin{flushright}
            % Doctor of Philosophy
            
            % Doctoral Thesis in Machine Learning
        \end{flushright}
        \vspace*{\fill}
        \noindent
        \includegraphics[width=0.6\textwidth]{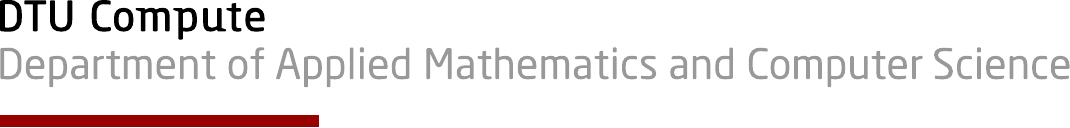}
        
        \vspace{15pt}
        \HUGE \textbf{On the Generalization of Learned Structured Representations}
        
        \vspace{25pt}
        \parbox[b]{0.5\linewidth}{
        \huge Andrea Dittadi
        
        \vspace{2.0cm}
        %\large\thesislocation{}, \the\year
        } 
        
        \hfill\includegraphics[scale=0.065]{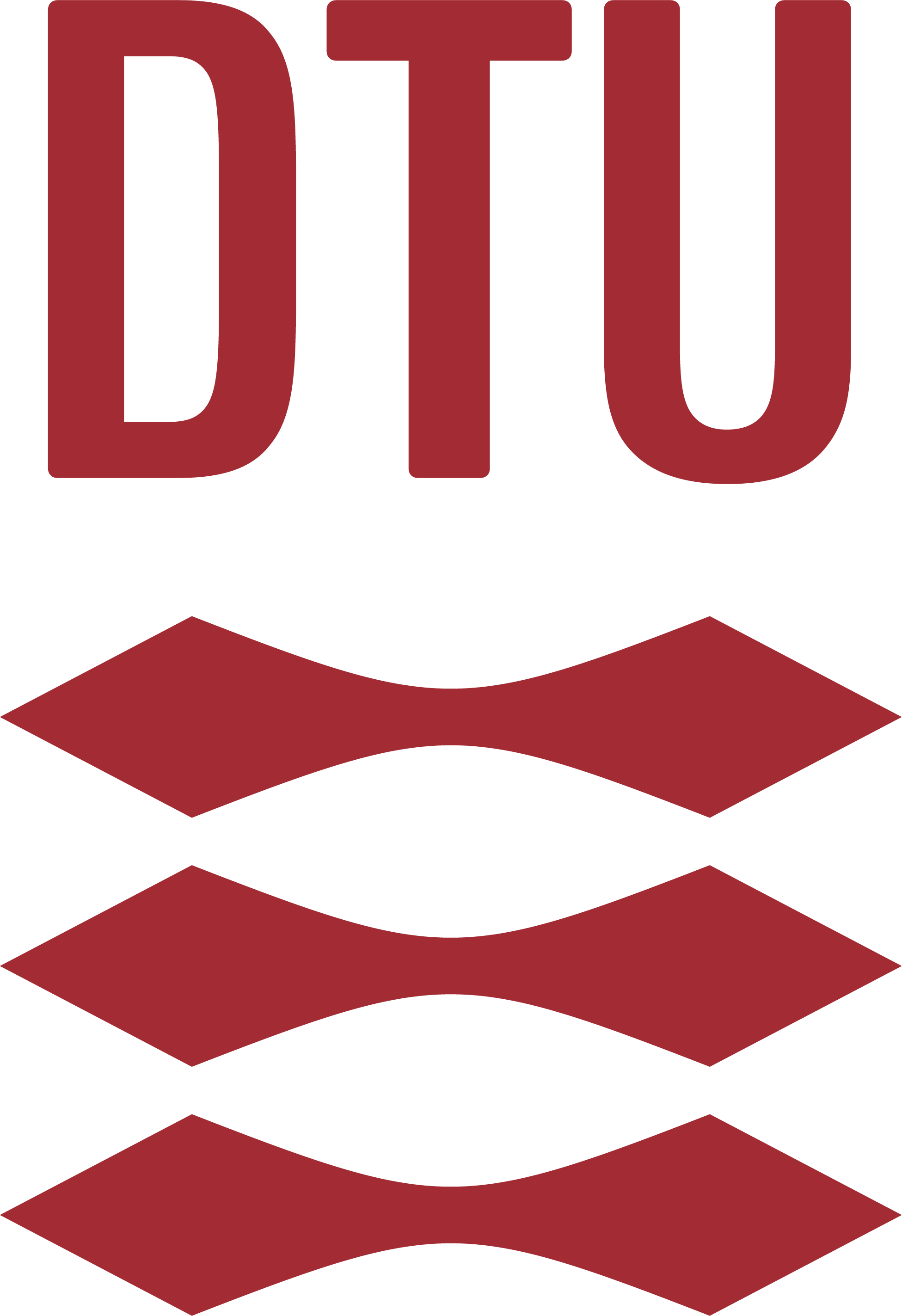}
    \end{adjustwidth}
\end{adjustwidth*}
}
\normalfont
\normalsize

\cleartoevenpage
\thispagestyle{empty} % No page numbers
\frieze
\noindent
% \sffamily

{\setlength{\parskip}{1pt}
\linespread{1.08}\selectfont

\large
\vspace*{\fill}
% \hspace{-0.52cm}
\textbf{Supervisor:} Prof. Ole Winther\\[3pt]

%\hspace{-0.52cm}
\textbf{Co-supervisor:} Prof. Thomas Bolander
\vspace{5cm}

\small
% Leave the following empty spaces to avoid the Warning: "Underfull \hbox (badness 10000)"
\hspace*{\fill}\textbf{DTU Compute}

\hspace*{\fill}\textbf{Department of Applied Mathematics and Computer Science}

\hspace*{\fill}\textbf{Technical University of Denmark}

\hspace*{\fill}Richard Petersens Plads 

\hspace*{\fill}Building 324

\hspace*{\fill}2800 Kongens Lyngby, Denmark

}

\normalsize
\normalfont
\vspace*{5.5cm}

\clearforchapter

\frontmatter
\chapter{Summary}

\vspace{-20pt}

Despite tremendous progress over the past decade, deep learning methods generally fall short of human-level systematic generalization.
It has been argued that explicitly capturing the underlying structure of data should allow connectionist systems to generalize in a more predictable and systematic manner. Indeed, evidence in humans suggests that interpreting the world in terms of symbol-like compositional entities may be crucial for intelligent behavior and high-level reasoning.
Another common limitation of deep learning systems is that they require large amounts of training data, which can be expensive to obtain.
In representation learning, large datasets are leveraged to learn generic data representations that may be useful for efficient learning of arbitrary downstream tasks.

This thesis is about structured representation learning.
We study methods that learn, with little or no supervision, representations of unstructured data that capture its hidden structure.
In the first part of the thesis, we focus on representations that disentangle the explanatory factors of variation of the data. We scale up disentangled representation learning to a novel robotic dataset, and perform a systematic large-scale study on the role of pretrained representations for out-of-distribution generalization in downstream robotic tasks.
The second part of this thesis focuses on object-centric representations, which capture the compositional structure of the input in terms of symbol-like entities, such as objects in visual scenes. Object-centric learning methods learn to form meaningful entities from unstructured input, enabling symbolic information processing on a connectionist substrate. In this study, we train a selection of methods on several common datasets, and investigate their usefulness for downstream tasks and their ability to generalize out of distribution.

\chapter{Resumé}

\vspace{-25pt}

\begin{otherlanguage}{danish}

\setlength{\parskip}{0.95em}
\linespread{1.08}

På trods af enorme fremskridt i løbet af det seneste årti er \textit{deep learning} generelt ikke i stand til at opnå systematisk generalisering på menneskeligt niveau.
Det er en udbredt opffattelse, at eksplicit indfangning af den underliggende struktur i data skulle gøre det muligt for connectionistiske systemer at generalisere på en mere forudsigelig og systematisk måde. 
Faktisk tyder resultater fra eksperimenter med mennesker på,
at det kan være afgørende for intelligent adfærd og ræsonnementer på højt niveau, at fortolke verden i form af symbollignende sammensatte enheder, der kan sammensættes og varieres.
En anden almindelig begrænsning ved deep learning-systemer er, at de kræver store mængder træningsdata, som kan være dyrt at fremskaffe.
I \textit{representation learning} udnyttes store datasæt til at lære generiske datarepræsentationer, som kan være nyttige til effektiv indlæring af vilkårlige efterfølgende opgaver.

Denne afhandling omhandler indlæring af strukturerede repræsentationer.
Vi undersøger metoder, der med lidt eller ingen supervision lærer repræsentationer af ustrukturerede data, som opfanger den skjulte, underliggende struktur.
I den første del af afhandlingen fokuserer vi på repræsentationer, der udreder de forklarende faktorer for variation i dataene. Vi opskalerer indlæringen af udredte repræsentationer (\textit{disentangled representations}) til et nyt robotdatasæt og gennemfører en systematisk, stor-skala undersøgelse af rollen, som forudindlærte repræsentationer spiller for generalisering 
uden for fordelingen 
til efterfølgende robotopgaver.
Den anden del af denne afhandling fokuserer på objektcentrerede repræsentationer, som indfanger inputets kompositoriske struktur i form af symbollignende enheder, såsom objekter i visuelle scener. Objektcentrerede indlæringsmetoder lærer at uddrage meningsfulde enheder fra ustruktureret input, hvilket muliggør symbolsk informationsbehandling på et connectionistisk substrat. I denne undersøgelse træner vi et udvalg af metoder på flere ofte anvendte datasæt og undersøger deres anvendelighed til efterfølgende-opgaver og deres evne til at kunne generalisere %.
uden for fordelingen.

\end{otherlanguage}

\chapter{Acknowledgements}

First and foremost, I'd like to thank my supervisors, Ole Winther and Thomas Bolander. 
We started with an ambitious and rather vague plan that had something to do with combining symbolic AI and deep learning. Although I ended up being (mostly) a ``deep learner'' (sorry, Thomas!), I believe the spirit of this initial idea clearly shows in the topics I have been working on.
Thank you, Ole, for your guidance as my main advisor, for always being unbelievably supportive and leaving me the freedom to explore, but always providing sharp feedback and being a reliable source of knowledge and wisdom. 
Thomas: although regrettably too few, I have learned a lot from our collaborations. I treasure our research discussions as well as our chats about life while having coffee at your fancy machine in building 322 or drinking beer in the nicest bars in Copenhagen.

I'd like to thank Bernhard Schölkopf and Stefan Bauer for warmly welcoming me to the Empirical Inference department at the MPI for Intelligent Systems: this was a turning point in my PhD and resulted in many successful collaborations, some of which are still ongoing.
I'm also very grateful to the Causality group for many insightful discussions.
I would like to especially thank Frederik Träuble for being a fantastic collaborator and friend since my first day at the MPI, and Stefan for being an invaluable mentor in the past couple of years and for always looking out for me.

I would also like to thank other people who have mentored me and helped me grow professionally, in particular Tom Cashman and Ben Lundell at Microsoft, and Peter Gehler and Francesco Locatello at Amazon. Special thanks to Francesco, who has been a great collaborator even outside of my internship at Amazon.

I'm also grateful for the many deep discussions and fun times at work and outside of work with my amazing PhD colleagues at DTU, including Valentin Liévin, Didrik Nielsen, Dimitris Kalatzis, Giorgio Giannone, Anders Christensen, and many more.

I'd like to thank all the collaborators in my projects:
Ben Lundell, Bernhard Sch{\"o}lkopf, Darren Cosker, Felix Widmaier, Francesco Locatello, Frederik Drachmann, Frederik Tr{\"a}uble, Jamie Shotton, Koen Van Leemput, Manuel W{\"u}thrich, Michele De Vita, Ole Winther, Olivier Bachem, Peter Gehler, Samuele Papa, Sebastian Dziadzio, Stefan Bauer, Stefano Cerri, Sveinn P{\'a}lsson, Thomas Bolander, Tom Cashman, Vaibhav Agrawal.
I could not have done this without your invaluable help.

I would also like to thank everyone else I have co-authored a paper with:
Anders Christensen, Anirudh Goyal, Darius Chira, Elliot Creager, Ilian Haralampiev, Lars Maal{\o}e, Niki Kilbertus, Patrick Schwab, Simon Bing, Valentin Li{\'e}vin, Yukun Chen.
Thanks for involving me in your projects---I've learned a lot from all of you. 

Finally, words cannot express how grateful I am to my friends and family---I wouldn't be here if it weren't for all of you. All my friends in Copenhagen, in Italy, and elsewhere, for always being there for me and for the many fun reunions around Europe. My family in Italy for their unconditional support. Anna, for being my pillar and enriching my life more than she knows.

\chapter{Preface}

\vspace{-30pt}

This thesis was prepared at the Cognitive Systems section of the Department of Applied Mathematics and Computer Science (DTU Compute), Technical University of Denmark. It constitutes a partial fulfillment of the requirements for acquiring a PhD at the Technical University of Denmark.

The PhD project was supervised by Ole Winther and Thomas Bolander and it was financed by the Technical University of Denmark. The PhD project was carried out at the Technical University of Denmark in March 2018--April 2022, except for: a six-month stay at the Max Planck Institute for Intelligent Systems in Tübingen, Germany under the supervision of Bernhard Schölkopf and Stefan Bauer; a three-month internship at Microsoft Research in Cambridge, UK under the supervision of Tom Cashman and Ben Lundell; and a four-month internship at Amazon Web Services in Tübingen, Germany under the supervision of Peter Gehler and Francesco Locatello.\looseness=-1

During the course of my PhD, I have worked on a variety of topics including representation learning, generative modeling with variational inference, disentanglement, object-centric learning, out-of-distribution generalization, and classical planning, with applications to robotics, video game playing, human pose modeling, medical imaging, image super-resolution, and medical time series. This work resulted in fifteen research papers (thirteen peer-reviewed publications and two preprints); this dissertation is based on three of them.

\vfill

{
\centering
    \year=2022 \month=5 \day=14
    Kongens Lyngby, \mydate\today\\[.3cm]
    \includegraphics[width=0.2\textwidth]{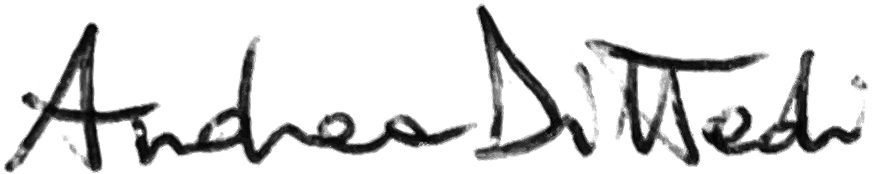}\\[.3cm]
\begin{center}
    Andrea Dittadi
\end{center}
}

\clearforchapter
\phantomsection % This is added to make hyperref point to the right "Contents" page
{\rmfamily  
\tableofcontents
}
\clearforchapter
\newlength{\contribspace}
\setlength{\contribspace}{0.18em}
\newlength{\contribfootnotespace}
\setlength{\contribfootnotespace}{0.4em}

% citation, authors, title, venue
\newcommand{\contribution}[4]{%
    \textbf{#3}\\[\contribspace]
    #2\\[\contribspace]
    #4\\[\contribspace]
    {\citeplain{#1}}%
}
\newcommand{\highlightedMe}{\underline{Andrea Dittadi}}

\chapter{List of publications}

{\linespread{1.08}\selectfont
This doctoral thesis is based on the following contributions:
\begin{enumerate}
    \item \contribution{%
        dittadi2021transfer%
    }{%
        \highlightedMe{}\eqcontrib, Frederik Tr{\"a}uble\eqcontrib, Francesco Locatello, Manuel W{\"u}thrich,\\Vaibhav Agrawal, Ole Winther, Stefan Bauer, Bernhard Sch{\"o}lkopf%
    }{%
        On the Transfer of Disentangled Representations in Realistic Settings%
    }{%
        \textbf{ICLR 2021} (International Conference on Learning Representations)%
    }
    \\[\contribfootnotespace]
    {\small \eqcontribfn Equal contribution.}
    
    \item \contribution{%
        trauble2022role%
    }{%
        \highlightedMe{}\eqcontrib, Frederik Tr{\"a}uble\eqcontrib, Manuel W{\"u}thrich, Felix Widmaier,\\Peter Gehler, Ole Winther, Francesco Locatello, Olivier Bachem,\\Bernhard Sch{\"o}lkopf, Stefan Bauer%
    }{%
        The Role of Pretrained Representations for the OOD Generalization of Reinforcement Learning Agents%
    }{%
        \textbf{ICLR 2022} (International Conference on Learning Representations)%
    }
    \\[\contribfootnotespace]
    {\small \eqcontribfn Equal contribution. The order is random and can be swapped.}
    
    \item \contribution{%
        dittadi2021generalization%
    }{%
        \highlightedMe{}, Samuele Papa, Michele De Vita, Bernhard Sch{\"o}lkopf,\\Ole Winther, Francesco Locatello%
    }{%
        Generalization and Robustness Implications in Object-Centric\\Learning%
    }{%
        \textbf{ICML 2022} (International Conference on Machine Learning)%
    }
\end{enumerate}

\clearpage

% \vspace*{0pt}

The PhD student has also contributed to the following publications that are not included in this dissertation:
\begin{enumerate}[resume]
    \item \contribution{%
        papa2022inductive%
    }{%
        Samuele Papa, Ole Winther, \highlightedMe{}%
    }{%
        Inductive Biases for Object-Centric Representations in the Presence of Complex Textures%
    }{%
        Preprint: arXiv:2204.08479, 2022%
    }
    
    \item \contribution{%
        chira2022image%
    }{%
        Darius Chira\eqcontrib, Ilian Haralampiev\eqcontrib, Ole Winther, \highlightedMe{}\eqadvising, Valentin Liévin\eqadvising%
    }{%
        Image Super-Resolution With Deep Variational Autoencoders%
    }{%
        Preprint: arXiv:2203.09445, 2022%
    }
    \\[\contribfootnotespace]
    {\small \eqcontribfn Equal contribution, alphabetical order. \eqadvisingfn Equal advising, alphabetical order.}
    
    \item \contribution{%
        bing2022conditional%
    }{%
        Simon Bing, \highlightedMe{}, Stefan Bauer\eqadvising, Patrick Schwab\eqadvising%
    }{%
        Conditional Generation of Medical Time Series for Extrapolation to Underrepresented Populations%
    }{%
        % Preprint: arXiv:2201.08186, 2022%
        To appear in \textbf{PLOS Digital Health}, 2022%
    }
    \\[\contribfootnotespace]
    {\small \eqadvisingfn Equal advising.}
    
    \item \contribution{%
        chen2021boxhead%
    }{%
        Yukun Chen, Frederik Tr{\"a}uble, \highlightedMe{}, Stefan Bauer, Bernhard Sch{\"o}lkopf%
    }{%
        Boxhead: A Dataset for Learning Hierarchical Representations%
    }{%
        \textbf{NeurIPS 2021 Workshop} on Shared Visual Representations in Human and Machine Intelligence%
    }
    
    \item \contribution{%
        dittadi2021full%
    }{%
        \highlightedMe{}, Sebastian Dziadzio, Darren Cosker, Ben Lundell,\\Tom Cashman, Jamie Shotton%
    }{%
        Full-Body Motion From a Single Head-Mounted Device: Generating SMPL Poses From Partial Observations%
    }{%
        \textbf{ICCV 2021} (IEEE/CVF International Conference on Computer Vision)%
    }
    
    \clearpage
    
    \item \contribution{%
        dittadi2021planning%
    }{%
        \highlightedMe{}\eqcontrib, Frederik Drachmann\eqcontrib, Thomas Bolander%
    }{%
        Planning from Pixels in Atari with Learned Symbolic Representations%
    }{%
        \textbf{AAAI 2021} (AAAI Conference on Artificial Intelligence)%
    }
    \\[\contribfootnotespace]
    {\small \eqcontribfn Equal contribution.}
    
    \item \contribution{%
        trauble2021disentangled%
    }{%
        Frederik Träuble, Elliot Creager, Niki Kilbertus, Francesco Locatello,\\\highlightedMe{}, Anirudh Goyal, Bernhard Schölkopf, Stefan Bauer%
    }{%
        On disentangled representations learned from correlated data%
    }{%
        \textbf{ICML 2021} (International Conference on Machine Learning)%
    }
    
    \item \contribution{%
        lievin2020optimal%
    }{%
        Valentin Li{\'e}vin, \highlightedMe{}, Anders Christensen, Ole Winther%
    }{%
        Optimal Variance Control of the Score Function Gradient Estimator for Importance Weighted Bounds%
    }{%
        \textbf{NeurIPS 2020} (Advances in Neural Information Processing Systems)%
    }
    
    \item \contribution{%
        palsson2019semi%
    }{%
        Sveinn P{\'a}lsson\eqcontrib, Stefano Cerri\eqcontrib, \highlightedMe{}\eqcontrib, Koen Van Leemput%
    }{%
        Semi-supervised variational autoencoder for survival prediction%
    }{%
        \textbf{MICCAI 2019 Workshop} on Brain Lesion%
    }
    \\[\contribfootnotespace]
    {\small \eqcontribfn Equal contribution.}
    
    \item \contribution{%
        dittadi2019lavae%
    }{%
        \highlightedMe{}, Ole Winther%
    }{%
        LAVAE: Disentangling Location and Appearance%
    }{%
        \textbf{NeurIPS 2019 Workshop} on Perception as Generative Reasoning%
    }
    
    \item \contribution{%
        lievin2019towards%
    }{%
        Valentin Li{\'e}vin, \highlightedMe{}, Lars Maal{\o}e, Ole Winther%
    }{%
        Towards Hierarchical Discrete Variational Autoencoders%
    }{%
        \textbf{AABI 2019} (Symposium on Advances in Approximate Bayesian Inference)%
    }
    
    \item \contribution{%
        dittadi2018learning%
    }{%
        \highlightedMe{}, Thomas Bolander, Ole Winther%
    }{%
        Learning to Plan from Raw Data in Grid-based Games%
    }{%
        \textbf{GCAI 2018} (Global Conference on Artificial Intelligence)%
    }
    
\end{enumerate}

}

\clearforchapter
\mainmatter

%%%%%%%%%%%%%%%%%%%%%%%%%%%%%%%%%%%%%%%%%%%%%%%%%%%%%%%%%

\chapter{Introduction}

Despite the extraordinary progress made in the last decade in deep learning, human-level intelligence still seems out of reach.
Major limitations of most contemporary methods include poor data efficiency and a lack of systematic generalization.
\emph{Representation learning} provides a possible way to alleviate the efficiency issue by learning to extract meaningful patterns in the data and represent them in a compact and reusable way, which should facilitate solving arbitrary downstream tasks, such as classification, abstract reasoning, planning, or robotic manipulation.
In particular, representations that explicitly capture some of the \emph{structure} in the data are believed to be beneficial for interpretability, efficiency of downstream learning, fairness, and human-level systematic generalization.
In this dissertation, we will focus on areas of representation learning that are concerned with learning this structure: \textit{disentangled} and \textit{object-centric} representation learning.

\emph{Disentangled representations} separately encode the ground-truth generative factors of variation of the data in a compact and reusable manner.
Although recent years have seen several experimental studies on disentangled representation learning for image data, some questions remain unanswered. For example, (i) the generalization in downstream tasks is rarely investigated thoroughly, and (ii) these studies largely focus on toy datasets.
One contribution of this thesis is a rigorous analysis of the generalization of disentangled representations in more realistic settings, including a large-scale study on the role of pretrained representations for the generalization in downstream robotic tasks.
\cref{chapter:contribs_disentanglement} provides an overview of these studies and a discussion of the key results. The original publications (\citeplain{dittadi2021transfer,trauble2022role}; referred to as Papers~\paperOne and~\paperTwo in this dissertation) are included in \cref{chapter:iclr2021,chapter:iclr2022}.

Following an analogous line of reasoning, in a multi-object setting, representations should ideally capture the compositional structure of visual scenes in terms of objects. Separately representing objects and interpreting them as compositional building blocks should, in principle, enable complex symbolic reasoning, causal inference, and human-like systematic generalization.
However, since \emph{object-centric learning} developed relatively recently as a subfield of representation learning, there is still limited understanding of (i) the inductive biases for learning good object representations without supervision, (ii) their usefulness for downstream tasks, and (iii) their out-of-distribution generalization. 
In this dissertation, we present a systematic empirical study that thoroughly investigates how useful these representations are in practice, and how well common object-centric representation learning methods generalize out of distribution (points (ii) and (iii) above).
\cref{chapter:contribs_objects} provides an overview of the study and a discussion of the major results. The original publication (\citeplain{dittadi2021generalization}; Paper~\paperThree in this dissertation) is included in \cref{chapter:objects}.
In an additional paper not included in this thesis but briefly discussed in \cref{sec:contribs_objects_discussion} \cite{papa2022inductive}, we investigate architectural inductive biases that may help successfully separate objects with complex textures (point (i) above).

\section{Thesis outline}

The main body of this dissertation consists of three parts: 
In the first part (\cref{chapter:background}), we provide relevant background on the topics underlying the remainder of the thesis. 
In the second part (\cref{chapter:contribs_disentanglement,chapter:contribs_objects}), we summarize our contributions, present the most salient results, and discuss them in depth.
The third part (\cref{chapter:iclr2021,chapter:iclr2022,chapter:objects}) contains Papers~\paperOne, \paperTwo, and~\paperThree.
We outline the structure of this thesis more in detail below.\looseness=-1

\textbf{\cref{chapter:background}} introduces the relevant background for this dissertation.
After a brief overview of the history and motivation of representation learning, we discuss some desirable properties of learned representations. We focus in particular on their format, and introduce the concept of disentanglement.
We then introduce \emph{variational autoencoders}, a popular approach for generative modeling.
We continue with a review of common approaches for disentangled representation learning based on variational autoencoders, and present the disentanglement metrics used in this thesis.
Finally, we introduce unsupervised object-centric representation learning, and present a selection of popular methods that are relevant for this thesis.

\textbf{\cref{chapter:contribs_disentanglement}} summarizes the main contributions of Papers~\paperOne and~\paperTwo. These papers focus on learning (disentangled) representations in a robotic setting. In two large-scale studies, we investigate the relationship between properties of the representations, the performance on downstream tasks from simple factor prediction to challenging robotic tasks, and the generalization on these downstream tasks.
We start by discussing our proposed dataset and the robotics context on which these publications are based. We then motivate and present the experimental studies, and discuss the key results and takeaways, leaving a more detailed analysis for \cref{chapter:iclr2021,chapter:iclr2022}.

\textbf{\cref{chapter:contribs_objects}} summarizes the main contributions of Paper~\paperThree, where we perform an empirical study on unsupervised object-centric representation learning. In this study, we formulate three hypotheses based on common assumptions in the literature that are however typically left implicit, and systematically test them.
More specifically, we investigate the usefulness of object representations for downstream models solving prediction tasks, and analyze their generalization under different types of distribution shifts at test time.
Similarly to \cref{chapter:contribs_disentanglement}, in this chapter we discuss the key results and takeaways, leaving a more detailed analysis for \cref{chapter:objects}.

\textbf{\crefnocompress{chapter:iclr2021,chapter:iclr2022,chapter:objects}}
contain Papers~\paperOne, \paperTwo, and~\paperThree, respectively. Finally, we conclude in \textbf{\cref{chapter:conclusion}} by summarizing and discussing the main contributions of this dissertation.

%%%

% \vspace{1cm}

% \hrule height 0.1pt depth 0pt width \textwidth \relax

% \warn{(Num references: \total{citenum})}

% Number of references on 11/05/2022: 401

%%%%

\chapter{Background}\label{chapter:background}

This chapter provides the relevant background for the contributions of this dissertation.
We start in \cref{sec:background - representation learning} by discussing the main motivations for representation learning, some key challenges in terms of generalization, and properties that are widely believed to be desirable in learned representations, such as disentanglement.
In \cref{sec:background - vaes}, we then provide an overview of variational autoencoders, which form the basis for some of the methods employed in this thesis, especially in the context of disentangled representation learning.
We then continue in \cref{sec:background - disentanglement} with a review of disentangled representation learning, including common disentanglement metrics and a weakly supervised learning method that we will employ in Papers~\paperOne and~\paperTwo (see \cref{chapter:contribs_disentanglement,chapter:iclr2021,chapter:iclr2022}).
Finally, in \cref{sec:background - objects} we motivate and introduce object-centric representations and discuss a selection of recent methods for learning them without supervision (these will be used in Paper~\paperThree; see \cref{chapter:contribs_objects,chapter:objects}).

\section{Representation learning}\label{sec:background - representation learning}

\subsection{The importance of data representation}\label{sec:background - representation learning - importance of representation}

\summary{Input representation is crucial.}
The performance of machine learning algorithms is strongly affected by the way their input data is represented \cite{murphy2012machine,bengio2013representation,john1994irrelevant,ragavan1993complex}.
Representations that focus on relevant aspects of the data are easier to learn from, because the learning algorithm is spared the burden of having to infer which information is relevant for the task, and isolate this information from irrelevant content.
Moreover, such representations should be more robust to changes in the input that are irrelevant to the task at hand. These irrelevant changes are not necessarily noise in the classical signal processing sense \cite{rabiner1975theory,oppenheim2009discrete}; they can also be alterations in nuisance variables (or nuisance factors), i.e., proper factors of variation in the data that are coincidentally irrelevant for the task under consideration but may be relevant for others \parencites{meehl1970nuisance}{fawzi2016measuring}[Section~2.2]{achille2018emergence}.

\summary{So feature engineering was important and is still relevant in some domains.}
Traditionally, most machine learning applications involved linear models on top of hand-engineered features \cite{lecun2015deep,bengio2013representation,lecun1998gradient}, whose purpose was to enrich or replace parts of the data in order to facilitate solving the task at hand. \emph{Feature engineering} can be time-consuming as it typically relies on significant domain expertise and is often carried out manually by humans \parencites{lecun2015deep}[Section~1.2]{murphy2012machine}. For this reason, automated feature engineering has been the focus of a considerable amount of research in the past \cite{markovitch2002feature,
% vision:
dalal2005histograms,belongie2001matching,freeman1995orientation,freeman1996computer,lowe1999object,viola2005detecting,%
% text:
scott1999feature,%
% domain-specific
sutton1991learning,hirsh1994bootstrapping,%
% domain-independent
aha1991incremental,matheus1989constructive%
},
% Feature eng for vision: \cite{dalal2005histograms,belongie2001matching,freeman1995orientation,freeman1996computer,lowe2004distinctive,mohan2001example,viola2005detecting,lowe1999object}.
% Automated feature engineering: \cite{markovitch2002feature,nargesian2017learning,khurana2016cognito}.
% Examples of domain-independent: \cite{bagallo1990boolean,aha1991incremental,matheus1989constructive}.
% Examples of domain-specific: \cite{sutton1991learning,hirsh1994bootstrapping}.
and is still relevant in some domains \cite{nargesian2017learning,khurana2016cognito,zoller2021benchmark,lam2017one,khurana2018feature}.

\subsection{Deep neural networks}
\summary{Deep networks replace feature engineering, and flexibly learn abstract representations.}
\emph{Deep learning} approaches \cite{lecun2015deep,schmidhuber2015deep} do away with this expensive, often problem-specific feature engineering, and instead learn to extract data representations that are suitable for the task, together with the task itself.
Such methods are based on \emph{Deep Neural Networks} (\emph{DNNs}), which consist of compositions of non-linear transformations, typically called \emph{layers}. Each transformation computes a more abstract representation of the data using the less abstract, lower-level representations from the previous layers \cite{ballard1987modular}. The deeper the layer---i.e., the further removed from the data---the more abstract and useful its representation of the data can be, as ``more abstract concepts can be often be constructed in terms of less abstract ones'' \cite{bengio2013representation}.\footnote{This abstraction can be explicitly built into the architecture (e.g., pooling in convolutional neural networks), but it should also naturally occur when the task to be solved ultimately requires abstraction (e.g., classification of high-level concepts from raw unstructured data).} 
Furthermore, the complexity of the set of functions learnable by deep networks can grow exponentially with the network's depth (\citeplain{montufar2014number,pascanu2013number,safran2017depth,raghu2017expressive,telgarsky2016benefits,eldan2016power}; but see also \citet{haastad1986almost,haastad1991power,wegener1987complexity} for related results before the deep learning era) since the features at different layers can be \emph{re-used and composed} in exponentially many ways \cite[Section~3.4]{bengio2013representation}.

\summary{Thanks to these advantages, DNNs are SOTA in many domains.}
Therefore, in addition to eliminating the need for the labor-intensive feature engineering on which traditional machine learning methods rely, DNNs can learn very flexible, abstract representations that are potentially even more suitable for a given task than problem-specific hand-engineered features---and they can do this directly from unstructured data.
Their increased flexibility and ease of applicability have led DNNs to 
redefine the state of the art in a broad range of domains, with notable early successes in, e.g., 
image and video recognition \cite{ciresan2011flexible, krizhevsky2012imagenet, sermanet2013overfeat, zeiler2014visualizing, simonyan2014two}, natural language processing \cite{mikolov2011strategies,collobert2011natural,bordes2014question,jean2014using,sutskever2014sequence,seide2011conversational,schwenk2012large,socher2011dynamic}, speech recognition \cite{mohamed2011acoustic,hinton2012deep,sainath2013improvements,le2012structured}, generation of images \cite{goodfellow2014generative,denton2015deep,mirza2014conditional} and raw audio \cite{oord2016wavenet}, and reinforcement learning from unstructured visual input \cite{mnih2015human,lillicrap2015continuous,schulman2015trust}.

\subsection{Learning generic representations for downstream tasks}\label{sec:bgr - task-agnostic representations}

\summary{DNNs can be used for generic representation learning. The representations are then used for downstream tasks, and they can be learned with different degrees of supervision.}
Deep networks have also been employed to learn \emph{generic} representations that are not tailored to a specific task and can be later used for arbitrary \emph{downstream tasks} \cite{bengio2013representation}. 
The learned representation function $r$, which maps a data point $\xb$ to its representation $r(\xb)$, is then typically used as a black box for downstream learning \cite{stooke2020decoupling,finn2016deep,ha2018recurrent,peters2018deep,brown2020language,kingma2014semi} or {adapted} (\emph{fine-tuned}) to the task at hand \cite{hinton2006reducing,kolesnikov2020big,zhai2019large,zbontar2021barlow}.
This approach is simply called \emph{representation learning} \cite{bengio2013representation,wang2020understanding,lesort2018state,hamilton2020graph,tschannen2018recent,rumelhart1986learning}, with the implication that no further objective is of interest at this stage, other than learning $r$ itself.\footnote{In these cases, even if the representations are learned by training a DNN to solve a specific supervised (or self-supervised) task---such as classification on a massive dataset of real-world images---the ultimate goal is not to solve these (auxiliary) tasks as much as it is to learn meaningful representations for later use.}
These generic {representation functions} can be learned without supervision \cite{hinton2006reducing,bengio2006greedy,ranzato2006efficient,ranzato2007unsupervised,donahue2019large,chen2020generative,radford2018improving},
with self-supervision---i.e., technically unsupervised but trained with supervised learning using, e.g., auxiliary tasks \cite{ahmed2008training,kolesnikov2019revisiting,doersch2015unsupervised,gidaris2018unsupervised,dosovitskiy2014discriminative} or contrastive methods \cite{oord2018representation,tian2020makes,chen2020simple,zbontar2021barlow,he2020momentum}---or with supervision on another dataset where numerous labeled examples are available \cite{zeiler2014visualizing,girshick2014rich,razavian2014cnn,van2014transfer,joulin2016learning,sun2017revisiting,kolesnikov2020big}.

\summary{How is representation learning actually useful for downstream tasks? 1: train with fewer labels.}
A primary reason to do representation learning is that learning from raw data typically requires a large amount of labeled examples, which may be expensive or impossible to obtain. One way to alleviate this issue is to leverage other sources of data that at least partially share the structure of the target task.
This paradigm is broadly referred to as \emph{transfer learning} \cite{pratt1996survey,pan2009survey,tan2018survey} as it involves transferring knowledge learned on one or more source tasks to a target task.
The source task may or may not be supervised, and there may be a covariate distribution shift, a change in the task definition (e.g., unsupervised to supervised, or a different conditional distribution of the labels, or even a different label space), or both.

For example, \citet{zhai2019large} conduct a study of transfer learning in vision, where deep networks are trained on supervised, unsupervised, or self-supervised tasks on ImageNet \cite{deng2009imagenet}, and the target tasks differ from the source tasks in terms of both the covariate distribution and the label space.
Another example is domain adaptation \cite{pan2010domain,wang2018deep}, where the source and target tasks are exactly the same (e.g., the conditional distribution of the classification labels $p(\yb \given \xb)$ is unchanged) but the covariates $\xb$ undergo a distribution shift.
It may also be the case that the source and target tasks differ while the data distribution $p(\xb)$ does not: for example, the source task could be unsupervised or self-supervised, and the target task may be classification or reinforcement learning \cite{laskin2020curl,grill2020bootstrap}. This can be seen as an instance of \emph{semi-supervised learning} \cite{chapelle2006semi,zhu2005semi,van2020survey}.\footnote{Roughly falling into the \emph{unsupervised preprocessing} category of semi-supervised learning, more specifically \emph{feature extraction} and \emph{pretraining} \cite[Section~5]{van2020survey}.} 
Examples of this scenario include Papers~\paperOne, \paperTwo, and~\paperThree \cite{dittadi2021transfer,dittadi2021generalization,trauble2022role} as well as some of our works not included in this dissertation \cite{trauble2021disentangled,dittadi2021planning,papa2022inductive}; however, in some of these works we also include experiments where the data distribution changes, e.g. with unseen values of the factors of variation \cite{dittadi2021generalization,trauble2022role,dittadi2021transfer}, synthetic image manipulations \cite{dittadi2021generalization}, or from simulated data to the real world \cite{trauble2022role,dittadi2021transfer}.
Finally, an increasingly common approach in transfer learning is to train very large DNNs on massive generic datasets and adapt them to a wide range of downstream tasks (\citeplain{bert,gpt3,clip,chen2020simple,henaff2021efficient}; and see a discussion on the so-called \emph{foundation models} in \citet{bommasani2021opportunities}).

\summary{How is representation learning actually useful for downstream tasks? 2: it can be way more efficient to learn tasks from pretrained representations.}
Even if data from the target task is available, learning from raw data may nevertheless be very inefficient when the task to be solved is particularly difficult, or when the function we are trying to learn is complex and highly-varying \cite[Section~2]{bengio2009learning}. This issue lies at the heart of the challenge of training deep architectures \cite[Section~10.1]{bengio2013representation}, which until the early 2010s was mostly based on layerwise unsupervised pretraining \cite{erhan2010does}.
For example, even after recent significant significant progress in training very deep models, reinforcement learning algorithms are often trained from learned low-dimensional representations \cite{stooke2020decoupling,finn2016deep,trauble2022role} or from the ground-truth state of the environment, if available \cite{ahmed2020causalworld,yu2020meta,lee2021ikea,vinyals2019grandmaster}.
Finally, even if learning end-to-end from raw data were practically feasible, it would still be advantageous (e.g., in terms of data and compute) to reuse a representation function that captures all the salient information in the data, rather than learn every new task from scratch.

\subsection{Desirable properties of learned representations}\label{sec:properties of learned representations}

\summary{Intro: consider properties of representations in terms of information and structure.}
We introduced and motivated representation learning, and argued that
\emph{pretrained} representations can be beneficial for efficient learning of downstream tasks, in particular when obtaining target task data is costly.
We will now consider desirable properties of the learned representation functions, more specifically in terms of their \emph{information content} and their \emph{structure}.

\subsubsection{Information content}\label{sec:properties learned representations - content}

\summary{Sufficiency and minimality.}
When learning a representation to be used for a downstream task, the representation should be \emph{sufficient} for the task, i.e., it should retain all necessary information to solve it. On the other hand, we would like the representation to be \emph{minimal}---i.e., to omit any information about the data that is not relevant for the task---and therefore invariant to irrelevant changes in the input \cite{bengio2013representation}.
\citet[Section~2.1]{achille2018emergence} define a representation $\zb$ of the data $\xb$ to be sufficient for a task $\yb$ (the target variable we wish to predict) if $I(\zb ; \yb) = I(\xb ; \yb)$, i.e., the representation $\zb$ contains as much information about the task as the data $\xb$ itself (or, equivalently, $\xb$ and $\yb$ are conditionally independent given $\zb$). Among the sufficient representations, the one that minimizes $I(\xb;\zb)$ is said to be minimal, because it contains just enough information about the data to allow solving the task, and is therefore invariant to the effect of nuisance factors \cite[Section~3]{achille2018emergence}.\looseness=-1

\summary{These concepts are not enough if we want generic representations: we want a task-agnostic representation.}
The problem is that these notions are task-dependent: A sufficient representation for a task may not be sufficient for another task if the representation omits information that is irrelevant for the first task but necessary for the second. Even in a multi-task setting \cite{caruana1997multitask,ruder2017overview}, a representation that is sufficient for \emph{all} training tasks while minimizing $I(\xb;\zb)$ might be insufficient for a test task.
This points to a fundamental trade-off between generality (usefulness for as many tasks as possible) and usefulness for specific tasks (potentially giving up usefulness for other tasks).
Since future tasks are not necessarily known ahead of time, a sensible approach is to learn generic \emph{task-agnostic} representations that make it possible to solve any downstream task that might be reasonably expected.\footnote{Different tasks ``provide different views on the same underlying reality'' \cite{bengio2009learning}.}

\subsubsection{Structure}\label{sec:properties learned representations - structure}

\summary{Assume generative process.}
We now shift our focus from \emph{what} information is in the representation to \emph{how} this information is represented. To do so, we consider the generative aspect of the data.
A common assumption is that the data is the result of a \emph{generative process} where the \emph{factors of variation} of the data tend to vary independently of each other, and typically few of them at a time vary in consecutive real-world inputs \cite{bengio2013representation,scholkopf2021toward}. These explanatory variables of the data, which we denote by the vector $\gb$, are then mixed in a potentially highly complex and non-linear way according to a conditional distribution $p(\xb \given \gb)$, to give rise to the observed data $\xb$. Ideally, given a data point $\xb$, we would like to find the ground-truth factors $\gb$ that generated it---or, following a Bayesian approach, the posterior distribution
\begin{equation}
    p(\gb\given\xb) = \frac{p(\gb) p(\xb \given \gb)}{\int_{\gb} p(\gb) p(\xb \given \gb) \dd \gb} 
    \label{eq:posterior of GT generative model}
\end{equation}
of such factors.
As we typically do not have access to the true generative model, nor to the ground-truth generative factors $\gb$ underlying each observed data point $\xb$, we learn representations $\zb = r(\xb)$, with $\xb \sim p(\xb\given\gb)$, that are in general different from $\gb$.

\summary{Define disentanglement.}
Under the assumptions above, a good task-agnostic representation should preserve as much information as possible about the data, while separating, or \emph{disentangling}, its explanatory factors of variation \cite[Section~3.5]{bengio2013representation} in a compact and reusable manner (see \citet{devereux2014centre} for evidence in humans). Although there is no generally agreed-upon definition of disentanglement, the core idea is that each specific factor $z_j$ in a \emph{disentangled representation} $\zb = r(\xb)$ should only reflect the state of one ground-truth generative factor of variation $g_i$.\footnote{Sometimes, the definition of disentanglement is taken to be the converse, i.e., a change in one factor $g_i$ should only affect one dimension of the representation \cite{chen2018isolating}. Arguably, this does not reflect the intuitive notion of disentanglement, since it allows multiple explanatory factors to be reflected in a single dimension of the representation. It has been termed \emph{completeness} by \citet{eastwood2018framework}, since a representation satisfying this definition would consist of elements that \emph{completely} describe one (or more) factors of variation of the data.}
Intuitively, this should enable downstream processing systems (e.g., a classifier) to flexibly access the subset of factors in $\gb$ that are relevant for the given task.
In practice, disentangled representations have proven useful, e.g., for interpretability \cite{adel2018discovering,higgins2018scan}, fairness of downstream models \cite{locatello2019fairness,trauble2021disentangled}, efficient downstream learning for reasoning tasks \cite{van2019disentangled}, and generalization \cite{locatello2020weakly,dittadi2021transfer}.\looseness=-1

Computer graphics provides a related perspective, where the data is generated by a known simulator given the generative factors $\gb$, i.e., $p(\xb \given \gb)$ is available. These graphics codes present a compact description of a scene that aligns well with the desired representational properties discussed above. From this point of view, the task of inferring the generative factors of the data is sometimes also known as \emph{inverse graphics} \cite{kulkarni2015deep}, and has also been applied to inverting a \emph{known} generative process such as a graphics engine \cite{wu2017learning,wu2017neural,eslami2016air}.

\subsection{Challenges for generalization}

\summary{Challenges for generalization.}
Despite the benefits in terms of efficient learning of downstream tasks, especially when obtaining target task data is costly, there are practical issues when learning representations from raw data.
First, there may be correlations or biases in the data that will be captured by the representations \cite{mehrabi2021survey,trauble2021disentangled,torralba2011unbiased}, thereby limiting their generality and applicability.\footnote{This is also related to ``shortcut learning'' of surface statistics and spurious correlations. Relevant work in computer vision includes \citet{jo2017measuring,ilyas2019adversarial,xiao2021noise,hendrycks2021natural}.}
Second, it might be desirable for learned representations to be invariant to some changes in the input, e.g., in terms of nuisance (task-irrelevant) variables in classification under domain shift \cite{pei2018multi,anselmi2016unsupervised} or in downstream reinforcement learning tasks \cite{zhang2020learning}; if such invariances cannot be inferred from the training data distribution---e.g., if a nuisance variable such as lighting conditions in a scene is constant across the entire dataset---the learned representation function will not naturally exhibit them.

These issues can be problematic for generalization.
For example, if a representation function is not invariant to lighting conditions, the representation $r(\xb)$ of an image $\xb$ with unseen lighting conditions at test time may inaccurately represent other relevant information in the image. This issue is tackled, for example, in Papers~\paperOne and~\paperTwo \cite{dittadi2021transfer,trauble2022role}.
Regarding correlations, a training dataset may for example contain images of a mountain landscape only in daylight, while city scenes are uniformly collected during day and night. Representation functions trained on such a dataset might not generalize well to photographs taken in the mountains at night, even though both mountain images and nighttime images are included in the training data.
In \citet{trauble2021disentangled}, we discuss training data correlations in the context of disentangled representation learning.

\summary{Possible ways to solve them.}
A conceptually simple solution is to collect more training data \cite{sutton2019bitter,kaplan2020scaling,henighan2020scaling}, assuming that in the limit of infinite data the spurious correlations and biases will disappear. However, despite the impressive generalization results of recent large-scale models trained on massive datasets \cite{brown2020language,ramesh2021zero,radford2021learning,ramesh2022,chowdhery2022palm}, their significance is sometimes called into question \parencites[Section~6.1]{bender2021dangers}{avocado}{deep_learning_wall}, while bias \parencites{abid2021persistent}[Section~6.2]{bender2021dangers} and training data memorization \cite{carlini2021extracting} appear to persist.\footnote{It should be noted that the concept of generalization itself may be ill-posed as the training distributions become ever wider. What appear to be strong generalization capabilities might in fact be explained by (arguably very complex) interpolation. On the other hand, it becomes increasingly more difficult to define meaningful generalization tasks, since the training distributions are so wide.}
An alternative solution is to attempt to directly resolve these biases during data collection and preprocessing, or in the learning algorithm itself \cite[Section~5.1]{mehrabi2021survey}.\footnote{Note that there are other sources of bias and discrimination than the data itself \cite[Section~3.1]{mehrabi2021survey}.}
Finally, we can mitigate some of these issues by introducing \emph{inductive biases} \cite{mitchell1980need}, e.g.,  in the model architecture, in the training objective, or in the optimization procedure \cite{geirhos2020shortcut}.
Intuitively, among the many possible ways an algorithm could generalize, inductive biases should help choose one (an inductive bias is ``any basis for choosing one generalization over another, other than strict consistency with the observed training instances'' \cite{mitchell1980need}). They typically derive from assumptions we make about the data or from potential constraints or requirements in the learning problem.

As discussed in \cref{sec:properties of learned representations}, it has been argued that capturing and disentangling all factors of variation underlying the data---rather than attempting to build invariances in the representations---should lead to robust representations that generalize better \cite{bengio2013representation,scholkopf2021toward}.
In \cref{sec:background - disentanglement}, we will discuss the learning and evaluation of disentangled representations. Many popular disentanglement learning methods, including those in this dissertation, are based on variational autoencoders, which are covered next.

\section{Variational autoencoders}\label{sec:background - vaes}

This section introduces variational autoencoders, a framework for variational inference in generative models that can be used for representation learning. In \cref{sec:background - disentanglement}, we will then discuss common disentangled representation learning methods that are based on variational autoencoders.

\paragraph{Latent variable models.}
\emph{Latent variable models} (LVMs) are probabilistic models with unobserved variables. In the following, we will denote the \emph{observed variables} by $\xb$, and the unobserved \emph{latent variables} by $\zb$. For simplicity, both $\xb$ and $\zb$ can be assumed to be vectors.
The marginal distribution over the observed data is:
\begin{equation}
    \ptheta (\xb) = \int_{\zb} \ptheta (\xb, \zb) \dd \zb 
    \label{eq:lvm_marginal}
\end{equation}
where $\ptheta (\xb, \zb)$ is the \emph{joint distribution} over observed and latent variables, and $\thetab$ is a vector of model parameters. This probability distribution is typically called \emph{marginal likelihood} or \emph{model evidence}. 
Note that this is the marginal likelihood of a \emph{single} data point. We are typically interested in the expectation of this quantity over the true data distribution, or more pragmatically, over the \emph{empirical data distribution} $q(\xb)$:
\begin{equation}
    \E_{q(\xb)} \left[ \ptheta (\xb)\right] =  \frac{1}{N} \sum_{i=1}^N \ptheta (\xb) = \frac{1}{N} \sum_{i=1}^N \int_{\zb^{(i)}} \ptheta (\xb^{(i)}, \zb^{(i)}) \dd \zb^{(i)}
\end{equation}
where $N$ is the size of the data set and the superscript denotes different data points. We will omit this outer expectation in the following, unless noted otherwise.

A simple and rather common structure for LVMs is:
\begin{equation}
    \ptheta (\xb, \zb) = \ptheta (\xb \given \zb) \ptheta(\zb)
    \label{eq:lvm_joint}
\end{equation}
where $\ptheta(\zb)$ is the \emph{prior distribution} over the latent variables and $\ptheta (\xb \given \zb)$ is the conditional distribution of a data point given the latent variables, typically called \emph{likelihood}.
This structure suggests a generative interpretation of latent variable models: a data point $\xb$ arises from a \emph{generative process} (see \cref{sec:properties learned representations - structure}) whereby latent variables $\zb$ are first sampled from the prior $\ptheta(\zb)$ and then used for conditional sampling of an observation $\xb$ according to the conditional distribution $\ptheta (\xb \given \zb)$.

\paragraph{Posterior inference.}
From the representation learning point of view (\cref{sec:background - representation learning}), it is typically of interest to invert this generative process and \emph{infer} the posterior distribution of the latent variables given the observed data, i.e., intuitively, find the value of the latent variables that gave rise to the observation (see \cref{sec:properties learned representations - structure}, where in \cref{eq:posterior of GT generative model} we write the posterior of the latent variables of the \emph{ground-truth} generative model). This is typically known as \emph{posterior inference}.

The \emph{posterior distribution} of the latent variables:
\begin{equation}
    \ptheta(\zb \given \xb) = \dfrac{\ptheta(\xb\given\zb) \ptheta(\zb)}{\ptheta(\xb)} = \dfrac{\ptheta(\xb\given\zb) \ptheta(\zb)}{\int_{\zb} \ptheta(\xb\given\zb) \ptheta(\zb) \dd \zb } 
    \label{eq:lvm_true_posterior}
\end{equation}
is often intractable---e.g., when the likelihood is a non-linear function parameterized by a deep neural network---due to the lack of a practical estimator of, or an analytical solution to the integral in \cref{eq:lvm_marginal}, which appears as denominator in \cref{eq:lvm_true_posterior}.

\paragraph{Approximating the posterior with variational inference.}
The issue of the intractability of the latent posterior is addressed by approximating it with \emph{sampling methods} or \emph{variational inference}. 
Methods in the former class, such as Markov Chain Monte Carlo (MCMC), yield samples from the exact posterior distribution in the limit of infinite samples, so the approximation simply derives from having finite computational resources. The main limitation of these methods is that they do not scale well with dataset size and model complexity. 

On the other hand, \emph{variational inference} trades sampling for optimization, and provides a {deterministic approximation} to the posterior. In variational inference, we define a \emph{variational distribution} $\qphi (\zb)$, with parameters $\phib$, that approximates the posterior $\ptheta(\zb \given \xb)$, and is therefore also known as \emph{approximate posterior}. The variational distribution is then optimized to minimize a divergence from the true posterior, typically the Kullback-Leibler (KL) divergence \cite{kullback1951information}:
\begin{equation}
    \kl(\qphi (\zb)\,\|\,\ptheta(\zb \given \xb)) = \E_{\qphi (\zb)} \left[ \log \dfrac{\qphi (\zb)}{\ptheta(\zb \given \xb)} \right]
    \label{eq:kl_div_posteriors}
\end{equation}
which is a non-negative quantity and is equal to $0$ if and only if $\qphi (\zb) = \ptheta(\zb \given \xb)$ almost everywhere.
Note that, in the standard case, $\qphi (\zb)$ approximates the posterior distribution for a \emph{single} data point $\xb$, and is derived or optimized separately for each example.
In practice, the approximate posterior is restricted to a family of distributions that are flexible enough to yield a good approximation, but simple enough to allow for efficient optimization.

For any choice of $\qphi$, we have:
\begin{equation}
    \log \ptheta(\xb) 
    = \log \E_{\qphi(\zb)} \left[ \frac{\ptheta(\xb , \zb)}{\qphi(\zb)} \right] 
    \geq \E_{\qphi(\zb)}  \left[ \log \frac{\ptheta(\xb, \zb)}{\qphi(\zb)} \right] 
    = \mathcal{L}^{\text{ELBO}}_{\thetab,\phib} (\xb)
    \label{eq:elbo_derivation_jensen_not_amortized}
\end{equation}
using Jensen's inequality and the fact that the logarithm is a concave function.
The quantity $\mathcal{L}^{\text{ELBO}}_{\thetab,\phib} (\xb)$ is called \emph{Evidence Lower BOund} (ELBO) as it is a lower bound to the marginal log-likelihood (or model evidence).

Crucially, maximizing this lower bound with respect to the variational parameters $\phib$ is equivalent to minimizing the KL divergence between the true and approximate posterior distributions in \cref{eq:kl_div_posteriors}. This can be shown by decomposing the log-likelihood as follows:
\begin{align}
    \log \ptheta(\xb) 
    &= \E_{\qphi(\zb)} \left[ \log \ptheta(\xb) \right] \nonumber\\
    &= \E_{\qphi(\zb)}  \left[ \log \frac{\ptheta(\xb, \zb)}{\ptheta(\zb \given \xb)}    \frac{\qphi(\zb)}{\qphi(\zb)} \right] \nonumber\\
    &= \E_{\qphi(\zb)}  \left[ \log \frac{\ptheta(\xb, \zb)}{\qphi(\zb)}\right] + \kl(\qphi (\zb)\,\|\,\ptheta(\zb \given \xb))\ . \label{eq:decomposition_likelihood}
\end{align}
Now, observe that the l.h.s. $\log \ptheta(\xb)$ is fixed as we are only optimizing the approximate posterior. We can therefore conclude that maximizing the ELBO \cref{eq:elbo_derivation_jensen_not_amortized}, which is the first term in \cref{eq:decomposition_likelihood}, is equivalent to minimizing the KL divergence in the second term. 
Since the KL divergence is non-negative, this also provides an alternative derivation of the ELBO without using Jensen's inequality (cf. Eq. \eqref{eq:elbo_derivation_jensen_not_amortized}).

\paragraph{Variational autoencoders.}
In contrast to traditional variational inference methods, where each data point has its own variational parameters that are optimized separately, \emph{amortized variational inference} \cite{gershman2014amortized} uses function approximators like neural networks to share variational parameters across data points. Beside improving learning efficiency and allowing variational inference to scale to massive datasets, this amortization enables efficient inference on new data points at test time, whereas in traditional variational inference this would require an expensive optimization of the ELBO.
In amortized variational inference, the approximate posterior of the latent variables is conditional on the observed data and denoted by $\qphi (\zb \given \xb)$, so the ELBO for one data point can be written as follows:
\begin{equation}
    \mathcal{L}^{\text{ELBO}}_{\thetab,\phib} (\xb) 
    =\E_{\qphi(\zb \given \xb)}  \left[ \log \frac{\ptheta(\xb, \zb)}{\qphi(\zb \given \xb)} \right] 
    \leq
    \log \ptheta(\xb) \ .
    \label{eq:elbo}
\end{equation}

Variational Autoencoders (VAEs)~\cite{kingma2013auto,rezende2014stochastic} are a framework for \emph{amortized stochastic variational inference}, in which the expectation of the ELBO over the dataset
\begin{equation}
    \E_{q(\xb)} \left[ \mathcal{L}^{\text{ELBO}}_{\thetab, \phib}(\xb) \right] \label{eq:expected_elbo}
\end{equation}
is maximized by jointly optimizing the inference model and the LVM (i.e., $\phib$ and $\thetab$, respectively) with stochastic gradient ascent.
The ELBO \cref{eq:elbo} to be maximized can be decomposed as follows (for one data point):
\begin{align}
    \mathcal{L}^{\text{ELBO}}_{\thetab,\phib} (\xb) 
    &= \E_{\qphi(\zb \given \xb)}  \left[ \log \ptheta(\xb \given \zb) \right] - \E_{\qphi(\zb \given \xb)}  \left[ \log \frac{\qphi(\zb \given \xb)}{\ptheta(\zb)} \right] \nonumber \\
    &= \E_{\qphi(\zb \given \xb)}  \left[ \log \ptheta(\xb \given \zb) \right] - \kl(\qphi(\zb \given \xb) \,\|\, \ptheta(\zb))
    \label{eq:elbo_decomposed}
\end{align}
where the first term can be interpreted as negative expected \emph{reconstruction error}, and the second term is the KL divergence from the prior $\ptheta(\zb)$ to the approximate posterior.
In this setting, $\qphi (\zb \given \xb)$ is often called \emph{inference model} or \emph{encoder}, while the likelihood $\ptheta(\xb \given \zb)$ is called \emph{decoder}.
Typically, the prior is a fixed, isotropic Gaussian distribution with unit variance, such that the dependency on the parameters $\thetab$ can be dropped:
\begin{equation}
    p(\zb) = \mathcal{N}(\zb;\, \mathbf{0}, \mathbf{I}) \ .
\end{equation}
The approximate posterior $\qphi(\zb \given \xb)$ is also a Gaussian with diagonal covariance matrix, but here the means and variances of each component are the output of an encoder network that takes the data as input:
\begin{equation}
    \qphi(\zb \given \xb) = \mathcal{N}(\zb;\, \mu_{\phib}(\xb), \diag ( \sigma_{\phib}(\xb))) \ .
\end{equation}

\paragraph{VAE optimization.}
A crucial issue with the approach presented above is that the gradients of the ELBO cannot be naively backpropagated through the sampling step. However, for a rather large class of probability distributions \cite{kingma2013auto}, a random variable can be expressed as a differentiable, deterministic transformation of an auxiliary variable with an independent marginal distribution. 
For example, if $\epsilonb \sim \mathcal{N}(0,1)$ and $\zb = \sigma_{\phib}(\xb)\, \epsilonb + \mu_{\phib}(\xb)$, then $\zb$ is a sample from a Gaussian random variable with mean $\mu_{\phib}(\xb)$ and standard deviation $\sigma_{\phib}(\xb)$. Thanks to this \emph{reparameterization}, $\zb$ can be differentiated with respect to $\phib$ by standard backpropagation.
This widely used approach, called \emph{pathwise gradient estimator}, tends to exhibit lower variance than the alternatives, which are typically based on the score function gradient estimator \cite{williams1992simple,ranganath2014black}.\footnote{However, the variance of this estimator can be reduced via \emph{control variates} \cite{glasserman2004monte}---see, e.g., \citet{mnih2014neural,mnih2016variational,tucker2017rebar}. Additionally, there is a line of work that focuses on using multiple stochastic samples of the ELBO \cite[\emph{importance-weighted ELBO}]{burda2015importance} to improve gradient estimates \cite{rainforth2018tighter,roeder2017sticking}; this includes our own work on optimal control variates for importance-weighted bounds \cite{lievin2020optimal}.}\looseness=-1

In the standard case, $\zb$ follows a simple multivariate probability distribution such as a Gaussian with diagonal covariance. However, we can incorporate knowledge or assumptions about the generative process by defining a more structured probabilistic model. We apply this to object-centric generative modeling (see \cref{sec:background - objects}) in \citet{dittadi2019lavae}.

\paragraph{Rate--distortion tradeoff.}
From an information theory perspective, optimizing the variational lower bound~\eqref{eq:elbo_decomposed} involves a tradeoff between rate and distortion, where the reconstruction term represents distortion and the divergence term represents rate \cite{alemi2018fixing}.
A straightforward way to control the rate--distortion tradeoff is to use the $\beta$-VAE framework \cite{higgins2016beta}, in which the training objective~\eqref{eq:elbo} is modified by scaling the KL term with a scalar $\beta>0$:
\begin{equation}
    \mathcal{L}_{\thetab,\phib, \beta} (\xb) 
    = \E_{\qphi(\zb \given \xb)}  \left[ \log \ptheta(\xb \given \zb) \right] 
    - \beta \kl(\qphi(\zb \given \xb) \,\|\,  p(\zb))\ . \label{eq:beta_objective_function}
\end{equation}

In \cref{sec:background - disentanglement - methods}, we will review more fine-grained modulations of the KL term that have been proposed to encourage learning disentangled representations.

\section{Disentanglement}\label{sec:background - disentanglement}

In \cref{sec:background - representation learning}, we have discussed the importance of data representations and some desirable properties such as disentanglement.
A substantial body of work spanning multiple decades has argued or demonstrated that learning disentangled representations is beneficial for a variety of purposes (e.g., \citeplain{bengio2013representation,peters2017elements,barlow1989unsupervised,schmidhuber1992learning,tschannen2018recent,lake2017building,scholkopf2021toward,kulkarni2015deep,locatello2019fairness,van2019disentangled}).
In this section, we provide an overview of common modern methods for learning disentangled representations, as well as popular metrics to assess their degree of disentanglement.

\subsection{Learning disentangled representations}\label{sec:background - disentanglement - methods}

Most state-of-the-art methods for unsupervised disentanglement learning are based on the variational autoencoder (VAE) framework introduced in \cref{sec:background - vaes}.
The relative success of VAEs in disentanglement may be explained by the stochasticity of the encoder, which promotes local orthogonality due to its diagonal covariance structure \cite{rolinek2018variational}.
While approaches based on generative adversarial networks (GANs) exist, these typically only disentangle style and content and require some form of weak supervision \cite{chen2016infogan,mathieu2016disentangling,lee2018diverse}.
In the following, we present several VAE-based approaches for learning disentangled representations. 

\paragraph{$\beta$-VAE.} The $\beta$-VAE (see Eq.~\eqref{eq:beta_objective_function}) modulates the bottleneck capacity---i.e., the amount of information the model is allowed to represent in the latent variables---by modifying the KL divergence term of the ELBO \cref{eq:elbo_decomposed} which intuitively acts as a regularizer for the approximate posterior. Increasing $\beta$ encourages more structured \cite{burgess2018understanding} but less informative representations, and corresponds to a lower rate and a higher distortion from the point of view of the rate--distortion tradeoff \cite{alemi2018fixing}. Decreasing $\beta$, on the other hand, leads to a higher rate and a lower distortion---i.e., more information in the latent variables and more accurate data reconstructions---but too weak a regularization may result in less structured and less useful representations \cite{bengio2013representation,higgins2016beta}.
Indeed, in our work we have observed that higher values of $\beta$ tend to encourage disentanglement \cite{dittadi2021transfer,trauble2021disentangled,trauble2022role,chen2021boxhead}, confirming previous results \cite{higgins2016beta,burgess2018understanding}. However, this behavior is not robust and it appears to strongly depend on the dataset  \cite[Fig.~15]{locatello2018challenging}.

\paragraph{AnnealVAE.} \label{label:annealVAE}
While this approach may be promising, a crucial issue is that the rate--distortion trade-off is, in some sense (and with some vigorous handwaving), independently present for each factor of variation. More concretely, let us assume there are two discrete factors of variation with the same number of possible values, and both affecting the color of an object. If the sizes of these objects differ significantly (e.g., one object constitutes the entire background while the other is a small cube in the foreground), the two factors will affect the reconstruction term $\ptheta(\xb \given \zb)$ in the objective function to dramatically different extents. Given a bottleneck with limited capacity (i.e., a strong regularization in latent space, attainable for example by choosing $\beta \gg 1$) the factor with a minimal effect on the likelihood would be ignored. In general, the inference model will simply ignore the factors of variation that are not worth being modeled---in other words, those that require too high a rate compared to the modest reduction in distortion that modeling them would bring. 
We have observed this while running experiments for \citet[not shown in the paper]{dittadi2021transfer}: some of the $\beta$-VAEs trained with $\beta=4$ did not model the rotation of the cube, and the learned generative model produced a cylinder (with the correct color and location), which can be interpreted as a cube averaged over all its possible rotations.\looseness=-1

Noticing this issue, \citet{burgess2018understanding} proposed to address it by introducing the \emph{AnnealVAE}, where the bottleneck capacity is progressively increased. The objective function then becomes:
\begin{equation}
    \mathcal{L}_{\thetab,\phib, \gamma} (\xb) 
    = \E_{\qphi(\zb \given \xb)}  \left[ \log \ptheta(\xb \given \zb) \right] 
    - \gamma \left| \kl(\qphi(\zb \given \xb) \,\|\,  p(\zb)) - C\right| \label{eq:annealedvae_objective_function}
\end{equation}
where $\gamma>0$ plays a similar role to $\beta$, and $C$ is annealed from 0 (which corresponds to a $\beta$-VAE with $\beta=\gamma$) to a potentially large positive value.

\paragraph{$\beta$-TCVAE and FactorVAE.} Let us now consider the KL term of the ELBO \cref{eq:elbo_decomposed}, and decompose its expectation with respect to the data distribution $q(\xb)$ as follows:
\begin{align}
    \E_{q(\xb)} &\left[\kl(\qphi (\zb \given \xb)\,\|\, p(\zb))\right] \nonumber \\
    &= \E_{q(\xb)} \E_{\qphi (\zb \given \xb)} \left[\log \frac {\qphi (\zb \given \xb) }{  p(\zb)}\right] \nonumber\\
    &= \E_{q(\xb)} \E_{\qphi (\zb \given \xb)} \left[\log \frac {\textcolor{BrickRed}{\qphi (\zb \given \xb)} }{ \textcolor{OliveGreen}{ p(\zb)}} \frac{\textcolor{BrickRed}{q(\xb)} \textcolor{NavyBlue}{\qphi (\zb)} \textcolor{OliveGreen}{\prod_i \qphi (z_i)}}{\textcolor{BrickRed}{q(\xb) \qphi(\zb)} \textcolor{NavyBlue}{\prod_i \qphi(z_i)}} \right] \nonumber\\
    &= \textcolor{BrickRed}{\underbrace{
            \kl(\qphi(\zb \given \xb) q(\xb)\,\|\,\qphi(\zb) q(\xb) )
            }_{
            \text{or } \E_{q(\xb)} \left[\kl(\qphi(\zb \given \xb)  \given \qphi(\zb) )\right]  }
       }  \nonumber\\
       &\qquad +\textcolor{NavyBlue}{
            \E_{\qphi(\zb)} \left[ \log \frac{\qphi(\zb)}{\prod\nolimits_i \qphi(z_i)}\right]
       }
       +\textcolor{OliveGreen}{
            \E_{\qphi(\zb)}  \left[ \log \frac{\prod_i \qphi (z_i)}{\prod_i p(z_i)} \right]
       } \nonumber\\
    &= \textcolor{BrickRed}{
            I(\xb; \zb)
       }
       +\textcolor{NavyBlue}{
            \kl\left( \qphi(\zb) \,\Big\|\,\prod\nolimits_i \qphi(z_i)\right)
       }
       +\textcolor{OliveGreen}{
            \sum\nolimits_i \kl\left( \qphi(z_i)\,\|\,p(z_i)\right)
       } 
       \label{eq:vae_elbo_tcvae_decomposition}
\end{align}
where $z_i$ denotes the $i$th dimension of the latent variable $\zb$.\footnote{In fact, this decomposition holds for any partition of the dimensions of $\zb$, but we focus on the case where each $z_i$ is a scalar variable.}
This expression appears with a negative sign in the VAE objective function (i.e., the expected ELBO over the entire dataset \eqref{eq:expected_elbo}), and is therefore \emph{minimized}.
The \textcolor{BrickRed}{first term} is the \emph{mutual information} between $\xb$ and $\zb$ with $\xb,\zb \sim \qphi(\zb\given\xb)q(\xb)$. It is zero when $\qphi(\zb \given \xb) = \qphi(\zb)$ almost everywhere, i.e., when the approximate posterior does not depend on the input.
Penalizing this mutual information through the information bottleneck has been found to encourage compact and disentangled representations \citep{achille2018emergence,burgess2018understanding}. On the other hand, it has also been argued that this term should not be penalized at all \cite{makhzani2015adversarial,zhao2017infovae,kim2018disentangling}.
The \textcolor{NavyBlue}{second term} is the \emph{total correlation} (TC) of the variables $\{z_i\}$ under the distributions $\{\qphi(z_i)\}$. This is a generalization of the mutual information to multiple variables \cite{watanabe1960information} and in this case penalizes \emph{aggregate posteriors}
\begin{equation}
    \qphi(\zb) = 
    \E_{q(\xb)} \left[\qphi(\zb \given \xb) \right]
    = \frac{1}{N} \sum_{i=1}^N \qphi(\zb^{(i)} \given \xb^{(i)})
\end{equation}
that do not factorize.
Finally, the \textcolor{OliveGreen}{third term} is the dimension-wise KL divergence from the prior to the aggregate posterior, which encourages each component of $\qphi(\zb \given \xb)$ to be close to its prior (which is typically $\mathcal{N}(z_i;\, 0, 1)$).

\citet{chen2018isolating} argue that the \textcolor{NavyBlue}{total correlation term} is what should be regularized in order to encourage disentanglement, and is the reason why $\beta$-VAEs tend to learn more disentangled representations when $\beta$ is increased. Both the \emph{$\beta$-TCVAE} \cite{chen2018isolating} and the \emph{FactorVAE} \cite{kim2018disentangling} modify the standard ELBO objective by scaling the {total correlation} by a factor $\gamma > 1$, albeit using different estimators for this quantity. The objective function to be maximized can then be rewritten in terms of the original ELBO objective \cref{eq:elbo_decomposed} as follows:
\begin{align}
    \E_{q(\xb)} \left[\mathcal{L}_{\thetab, \phib, \gamma}(\xb)\right] 
    &= \E_{q(\xb)} \E_{\qphi(\zb \given \xb)} \left[ \log \ptheta(\xb \given \zb) \right]
        - \textcolor{BrickRed}{
            I(\xb; \zb)
       } \nonumber \\
       & \quad - \gamma \, \textcolor{NavyBlue}{
            \kl\left( \qphi(\zb) \,\Big\|\,\prod\nolimits_i \qphi(z_i)\right)
       }
       - \textcolor{OliveGreen}{
            \sum\nolimits_i \kl\left( \qphi(z_i)\,\|\,p(z_i)\right)
       } \nonumber \\
       &= \E_{q(\xb)} \left[ \mathcal{L}^{\text{ELBO}}_{\thetab, \phib}(\xb) \right] - (\gamma - 1) \, \textcolor{NavyBlue}{
            \kl\left( \qphi(\zb) \,\Big\|\,\prod\nolimits_i \qphi(z_i)\right)
       }
\end{align}
where we explicitly include the expectation over the dataset, which is necessary for the decomposition in Eq.~\eqref{eq:vae_elbo_tcvae_decomposition}. The \textcolor{NavyBlue}{additional regularizer} on the total correlation vanishes when $\gamma=1$, resulting in the standard VAE objective.

\paragraph{DIP-VAE.} Finally, \citet{kumar2017variational} claim that the standard VAE objective is not sufficient to encourage disentanglement, and propose to explicitly regularize a divergence between the prior and the aggregate posterior, to encourage the latter to be disentangled. The desired objective function (including the expectation over the data distribution) is then:
\begin{equation}
    \E_{q(\xb)} \left[ \mathcal{L}_{\thetab, \phib, \gamma}(\xb)\right] 
    = \E_{q(\xb)} \left[ \mathcal{L}^{\text{ELBO}}_{\thetab, \phib}(\xb) \right]  
    - \gamma \, D\left( \qphi(\zb)\,\|\,p(\zb)\right)
\end{equation}
where $D$ is an arbitrary divergence. Note that, when $D$ is the KL divergence, $D\left( \qphi(\zb)\,\|\,p(\zb)\right)$ is equal to the sum of the \textcolor{NavyBlue}{second} and \textcolor{OliveGreen}{third} terms in \cref{eq:vae_elbo_tcvae_decomposition}. \citet{kumar2017variational} introduce two ways of approximating the additional divergence term, corresponding to two distinct optimization objectives: DIP-VAE-I and DIP-VAE-II.

\subsection{Measuring disentanglement}\label{sec:background - disentanglement - metrics}

Although a consensus has yet to be reached on a precise definition of disentanglement, various quantitative metrics have been proposed in an attempt to measure disentanglement based on the intuitive notions discussed earlier.
Crucially, common disentanglement metrics do not always agree with each other in practice \cite{locatello2019disentangling}. While it is clear that some of them measure different notions than others, not all discrepancies are easily explained, especially in terms of the different results observed across datasets. \citet{locatello2020sober} present a thorough analysis and discussion, and provide recommendations for practitioners.

A first distinction to be made is whether a metric relies on \emph{interventional data}---where we are allowed to perform interventions on the ground-truth factors of variation and assess how these affect the representations---or \emph{observational data}---where we must estimate the statistical relationships between the learned representations and the ground-truth factors given a set of (annotated) examples.

In the interventional setting, two properties that are typically assessed are \emph{consistency} and \emph{restrictiveness} as defined by \citet{shu2019weakly}, which measure the effect that intervening on a single factor of variation (or group of factors) has on the representation. In a \emph{consistent} representation function, \emph{fixing} a factor (or group of factors) and varying the others corresponds to fixing a subset of dimensions in the representation. In a \emph{restrictive} representation function, \emph{changing} a factor (or group of factors) with the others fixed corresponds to varying only a subset of dimensions in the representation.

In the observational setting, the metrics presented in this section typically focus on \emph{disentanglement} and \emph{completeness} in the sense of \citet{eastwood2018framework}. Each dimension of a \emph{disentangled} representation captures at most one factor of variation of the data: one factor may be encoded into multiple dimensions, but each of these dimensions must not encode any other factor. 
Conversely, in a \emph{complete} representation, each factor of variation corresponds to at most one dimension in the representation: different factors may be mixed in one representation dimension, but no factor may be split over multiple dimensions. \label{label:completeness}

The four properties outlined above are illustrated in \citet[Fig.~11]{locatello2020sober}.
In the following, we introduce some of the most popular disentanglement metrics.

\paragraph{BetaVAE.} The \emph{BetaVAE} metric \cite{higgins2016beta} is computed as the accuracy of a classifier that predicts which factor of variation has been fixed in a batch of image pairs. More specifically, for each batch, we choose a factor of variation $i$ at random and sample a batch of pairs $(\xb_1, \xb_2)$ such that the $i$th factor has the same value in $\xb_1$ and $\xb_2$. All pairs in the batch have the same factor fixed, but the value can differ across pairs. For each image pair, we then compute the absolute value of the difference between the encoded representations of the two images, and finally average over the batch. The resulting vector:
\begin{equation}
    \frac{1}{B} \sum_{n=1}^B |r(\xb_1^{(n)}) - r(\xb_2^{(n)})| \ ,
\end{equation}
where $B$ is the batch size, is the input to the logistic regression model; the regression target is the index $i$ of the factor that has been fixed in all the pairs in the batch. Each batch yields one data point for the downstream training of the regressor.

\paragraph{FactorVAE.} \citet{kim2018disentangling} discuss some weaknesses of the BetaVAE metric and propose to address them with the \emph{FactorVAE} score, computed as follows:
We first estimate the variance of each latent dimension and exclude unused dimensions (those with a small variance) from all subsequent computations.
Then, we generate batches of samples where one randomly chosen factor is constant in each batch.
For each generated batch, we estimate the representation dimension that encodes the fixed factor as the one that has the smallest variance (normalized by the global variance across the entire training set). The estimated dimension is one training sample for a majority vote classifier, and the (known) fixed factor is the corresponding target.
The classifier rule is then defined as taking for each latent dimension the ground-truth factor that has the most votes from the training set.\footnote{This can be achieved by repeating the step above for many batches, and constructing a matrix of size $K \times D$ (with $K$ the number of factors and $D$ the latent space dimensionality) where each entry denotes the number of times a given dimension was estimated to correspond to a specific factor. For each dimension $d$, the factor with most votes is the argmax of the $d$th column.}
The FactorVAE score is then the accuracy of this classifier on a held-out test set.

\paragraph{DCI.} \citet{eastwood2018framework} argue that three distinct notions are relevant in this context: \emph{disentanglement}, \emph{completeness}, and \emph{informativeness}. All three metrics are based on classifiers such as random forests or gradient boosted trees, one per factor of variation, each trained to predict the ground-truth factor value from the data representation $r(\xb)$. Informativeness is simply the average accuracy of the classifiers on a held-out test set. 
Disentanglement and completeness are based on the \emph{importance matrix} obtained by concatenating the feature importances of each classifier. This matrix consists of the predictive importance of each representation dimension for each ground-truth factor.
Disentanglement is maximized when each latent dimension has a positive importance for only one factor. Completeness measures the converse: it is maximized when only one latent dimension has a positive predictive importance for any given factor, i.e., each factor is only captured by one dimension in the representation.
This framework is named \emph{DCI} after the three metrics introduced. For simplicity, we will refer to the disentanglement metric defined by \citet{eastwood2018framework} as DCI.

\paragraph{MIG.} \citet{chen2018isolating} propose to measure disentanglement with the Mutual Information Gap (MIG), based on the mutual information between each ground-truth factor of variation and each latent dimension. The mutual information gap for one ground-truth factor is defined as the difference between the highest and second-highest mutual information, normalized by the estimated entropy of that factor. When a factor is only represented by one latent dimension, this quantity is 1. The MIG score is then obtain by averaging over all ground-truth factors.
Note that this metric in fact measures \emph{completeness} (see p.~\pageref{label:completeness}).

\paragraph{Modularity and explicitness.} \citet{ridgeway2018learning} introduce \emph{modularity} and \emph{explicitness}. Modularity is computed in a similar fashion to the MIG, except that the gap is measured over factors of variation and for a given latent dimension, rather than the opposite. Modularity intuitively corresponds to disentanglement as defined in \cref{sec:properties learned representations - structure} and in \citet{eastwood2018framework}. However, empirically, it appears to measure a different notion of disentanglement than other metrics \cite[Section~6.1]{locatello2020sober}.
Explicitness, on the other hand, measures how easily the factors of variation can be predicted from the representation (loosely corresponding to informativeness in \citet{eastwood2018framework}). It is computed for each factor of variation as the ROC-AUC (area under the receiver operating characteristic curve) of a logistic regression classifier trained to predict the ground-truth value of that factor. The global explicitness is the average of this quantity over all factors of variation.

\paragraph{SAP.} The \emph{Separated Attribute Predictability (SAP)} score proposed by \citet{kumar2017variational} is computed as follows:
First, for each combination of factor of variation and latent dimension, we train a regression model or a classifier (depending on whether the factor is continuous or discrete) to predict the factor from the latent dimension. These result in a $K \times D$ \emph{score matrix} containing the R$^2$ score (for continuous factors) or the accuracy (for discrete ones) for all combinations, computed on a test set.
For each factor of variation, we then compute the difference between the highest and second-highest score. The SAP score is the average difference over all factors of variation. This difference will be maximal when each factor is only predictable from one dimension of the representation. Therefore, like the MIG, it measures \emph{completeness} in the sense of \citet{eastwood2018framework} (see also p.~\pageref{label:completeness}).

\paragraph{IRS.} The \emph{Interventional Robustness Score (IRS)} proposed by \citet{suter2019robustly}
performs interventions on the factors of variation and measures resulting changes in the representation. 
The \emph{post-interventional disagreement} in a representation component $z_k$ due to a generative factor $g_j$ given a fixed value of $g_i$ is defined as the distance (e.g., the $\ell_2$-norm) between the expectation of $z_k$ when we only fix $g_i$ and when we also fix $g_j$ (with $i \neq j)$. Intuitively, we fix $g_i$ and observe how robust $z_k$ is when $g_j$ changes. 
The IRS score then measures the (normalized) expected maximum disagreement over all factors of variation and their distributions, to assess the worst-case effect a change in nuisance factors (such as $g_j$) might have on the representation of $g_i$.
Note that the interventional setting (see Pearl's do-calculus \cite{pearl2009causality}) is not necessary when there are no confounding correlations in the generative process, since in that case interventions are equivalent to regular conditioning. \citet[Section~5]{suter2019robustly} also propose a method for estimating the IRS from purely observational data.
\citet[Section~6.1]{locatello2020sober} note that the IRS is not consistently correlated with the other disentanglement metrics.

\subsection{Disentangling with limited supervision}\label{bgr:disentanglement partial supervision} % partial/limited

\citet[Theorem~1]{locatello2018challenging} show that disentangled representation learning is impossible without supervision or appropriate inductive biases. They support this empirically in a large-scale experimental study, where they train the models presented in \cref{sec:background - disentanglement - methods} on several synthetic datasets for disentanglement learning. In this study, they observe that hyperparameters and random seed appear to matter significantly more than the model type. Furthermore, since unsupervised model selection is particularly challenging, it is necessary to directly evaluate disentanglement with the ad hoc metrics introduced in \cref{sec:background - disentanglement - metrics}, which require ground-truth annotations of the factors of variation. 

However, when some label information is available, it may be reasonable to use it instead as a direct supervision signal for learning disentangled representations: When a few observations are fully-labeled, this corresponds to the semi-supervised setting (\citeplain{locatello2019disentangling,khemakhem2020variational,sorrenson2020disentanglement}; but see also \citet{paige2017learning,klys2018learning,reed2014learning} for previous related work). In \citet[Section~5]{trauble2021disentangled}, we explore a different instantiation of semi-supervised learning (see \cref{sec:bgr - task-agnostic representations}) where representations learned in an unsupervised fashion are adapted (or aligned) to reflect new ground-truth information about the factors.
On the other hand, when only weak labels are available (typically for the entire dataset), we can exploit them to learn disentangled representations by constructing a weakly supervised learning setting \cite{locatello2020weakly,shu2019weakly,bouchacourt2017multi,hosoya2019group}, discussed below.

\paragraph{Weakly supervised disentanglement.} We will briefly present here the weakly supervised approach proposed by \citet{locatello2020weakly} that we will employ in \cref{chapter:iclr2021,chapter:iclr2022} (Papers~\paperOne and~\paperTwo).
The key idea behind this method is that, while the ground-truth factors of variation are provably not identifiable in the i.i.d. case, they become identifiable given pairs of observations that differ in a subset of factors of size $k$. This subset of changing factors may differ from pair to pair, and even the number $k$ of changing factors need not be fixed across all pairs.
A possible justification for this setting is that changes in natural environments are caused by changes in a few factors of variation at a time \cite{foldiak1991learning,wiskott2002slow}, which is related to the \emph{sparse mechanism shift} hypothesis \cite{scholkopf2021toward,scholkopf2019causality}.
Unlike previous weakly supervised approaches that rely on group information (e.g., knowing which factors are changing between two observations), the method by \citet{locatello2020weakly} only requires the number $k$ of changing factors for each pair of observations to be known. In fact, they propose a practical algorithm that relaxes even this assumption by estimating $k$ via a simple heuristic. 

More concretely, the method computes approximate posterior distributions of the latent variables for a pair of images $(\xb^{(1)}, \xb^{(2)})$, and then selects the $k$ latent dimensions that \emph{differ the most} in the posteriors of the two images, in terms of the KL divergence $\kl(\qphi(z_i \given \xb^{(1)}) \,\|\, \qphi(z_i \given \xb^{(2)}))$. We assume fully factorized approximate posteriors, and $z_i$ denotes the $i$th dimension of $\zb$.
The $k$ dimensions with a large KL divergence are considered to be changing between the two observations, while the others are considered to be unchanged. Since the KL divergence introduces an unnatural ordering in the input pair, in Papers~\paperOne and~\paperTwo we modify the original definition by \citet{locatello2020weakly} and use the \emph{symmetrized} KL divergence (see \cref{app:implementation_details}).

Then, given a symmetric averaging function $a$, the modified posterior distribution for the latent dimension $j$, and for $i\in\{1,2\}$, is defined as follows:
\begin{equation*}
    \qphitilde^{(i)}( z_j \given \xb^{(1)},\xb^{(2)}) = 
    \begin{dcases}
        \qphi( z_j \given \xb^{(i)})      &\text{if $z_j$ is inferred to be changing}\\
        a(\qphi( z_j \given \xb^{(1)}), \qphi( z_j \given \xb^{(2)})) & \text{if $z_j$ is inferred to be shared}\\
    \end{dcases}
\end{equation*}
such that the posteriors of the dimensions that are inferred to be unchanged between the two inputs are collapsed into the same distribution.
The averaging function forces the approximate posterior of the shared latent variables to be the same for the two observations. \citet{locatello2020weakly} propose to use the averaging functions from the \emph{Multi-Level VAE} (ML-VAE) \cite{bouchacourt2017multi} or from \emph{Group-VAE} (GVAE) \cite{hosoya2019group}, and call the resulting methods \emph{Ada-ML-VAE} and \emph{Ada-GVAE}, respectively. In this dissertation, we will focus on Ada-GVAE (with the minor difference that we use a symmetrized KL divergence, as mentioned above), where the averaging function consists of averaging the means and variances of the (Gaussian) posteriors.
The objective function for the pair of observations is a straightforward modification of the standard $\beta$-VAE objective:
\begin{align}
  \sum_{i\in\{1,2\}}  \left(
    \E_{\qphitilde^{(i)}(\zb \given \xb^{(1)},\xb^{(2)})} \log(\ptheta(\xb^{(i)} \given \zb)) - \beta \kl \Big(\qphitilde^{(i)}\big(\zb \given \xb^{(1)}, \xb^{(2)}\big) \,\Big\|\, p(\zb)\Big)
  \right)
  \label{eq:weak-supervision-elbo}
\end{align}
and it is optimized by drawing samples $(\xb^{(1)}, \xb^{(2)})$ from a non-i.i.d. data distribution that accounts for the weak supervision assumptions discussed above.\footnote{More formally, we can define the empirical joint distribution of each pair as follows:
\begin{equation}
    q(\xb^{(1)}, \xb^{(2)}) = q(\xb^{(1)}, \xb^{(2)} \given k) q(k)\ ,
\end{equation}
where $q(\xb^{(1)}, \xb^{(2)} \given k)$ selects a random pair of images that differ in $k$ factors of variation and $q(k)$ is a distribution over $k$. In \cref{chapter:iclr2021,chapter:iclr2022}, we take $k=1$ deterministically, or equivalently $q(k) = \delta(k-1)$.
}

The authors empirically show that this approach (in both its variants) significantly improves the disentanglement of the learned representations, and that model selection can be successfully performed without explicit label supervision (i.e., the metrics from \cref{sec:background - disentanglement - metrics} are not needed). 
We remark, however, that in practice it seems to be necessary to show several pairs of observations with \emph{only one} changing factor. This is evident from the original publication itself \cite{locatello2020weakly}, where half of the training pairs are always differing only in one factor: when $k>1$, the observations actually still have $k=1$ with probability 0.5. We reached a similar conclusion while running experiments in Papers~\paperOne and~\paperTwo \cite{dittadi2021transfer,trauble2022role}, 
where $k=1$ often yielded perfectly disentangled factors, while even $k=2$ (without forcing $k=1$ for half of the pairs) resulted in a dramatic performance drop.

\section{Object-centric representations}\label{sec:background - objects}

\subsection{Motivation}

In \cref{sec:background - disentanglement}, we introduced disentanglement and discussed how disentangled representations should be beneficial for downstream learning and generalization.
The underlying assumption is that the data comes from a structured generative process with a few underlying factors of variation, and we wish to invert such a process by \emph{disentangling} these factors.
Although there is currently no precise definition of disentanglement, the consensus is that distinct factors of variation in the data should be represented separately from each other.

In general, however, visual scenes may contain a variable number of objects, which makes it less straightforward to define a plausible generative model of the data within the framework introduced so far. For example, if we learn a disentangled representation of visual scenes with one object such as the robotic setup in Papers~\paperOne and~\paperTwo (\cref{chapter:iclr2021,chapter:iclr2022}; see \cref{fig:robot_dataset_sim_real}), what should the representation of a scene with two objects be? When a second object is placed into the arena, either the representation stops being disentangled, or new factors of variation have to be introduced in order to represent the new object.
One might argue that, if the task were to learn a representation of scenes where no more than one object is ever observed, it would be unreasonable to expect a model to generalize to multiple objects. However, the issue persists even when training on data with a variable number of objects: since there can always be more objects than ever observed in the training distribution, a robust model must have learned a mechanism to cope with the compositional structure of data that has discrete object-like building blocks. It is unclear how this may be possible unless the representation has a modular structure and its size adapts to the corresponding observation.\footnote{Since the human working memory has a limited capacity and cannot hold more than a few objects simultaneously \cite{kibbe2019conceptually,oberauer2019working,cowan2001magical,miller1956magical,fukuda2010discrete}, we could also assume that the number of objects to be represented is bounded. However, here we focus on the general problem of representation learning in the multi-object setting, without necessarily limiting ourselves to scenarios that are realistic for human learning.}

A second motivation for learning object representations can be found in the successes of \textit{symbolic artificial intelligence} (AI) methods.
Symbolic AI revolves around the idea that the abstractions necessary for reasoning and intelligent behavior are best represented by symbols.
Historically, this approach has been at the basis of many of artificial intelligence's early successes, e.g., in automated planning \cite{fikes1971strips,ghallab2004automated}, theorem proving \cite{gelernter1959realization,newell1956logic}, and knowledge-based systems \cite{buchanan1969heuristic}.
Despite their advantages, such as theoretical guarantees, interpretability, and systematic generalization, symbolic AI methods still require symbolic inputs and often rely on expert domain knowledge: Symbols have to be defined and grounded in the real world---this is known as the \emph{symbol grounding} problem \cite{harnad1990symbol,searle1980minds,steels2008symbol}---and knowledge, in terms of domain-specific facts and rules, must be entered by humans into the system using a formal language (e.g., STRIPS \cite{fikes1971strips} and PDDL \cite{mcdermott1998pddl} in automated planning, or description logics \cite{baader2003basic}).\footnote{Although in some cases this knowledge can be learned, observations are typically still required to be symbolic. This is the case, e.g., in action model learning for automated planning \cite{mourao2012learning,cresswell2013acquiring,yang2007learning,pasula2007learning,walsh2008efficient,zhuo2010learning}. Note that, in \citet{dittadi2018learning}, we do in fact learn a simple neural transition model in Sokoban from non-symbolic observations and successfully use it in a tree search algorithm for planning. However, (1) this approach relies on assumptions on the transition function for the domain, and (2) having a non-symbolic transition model prevents us from using optimized off-the-shelf planners that can automatically derive powerful heuristics to significantly improve planning efficiency.} While this approach may be reasonable in some cases, it is not acceptable in more general contexts that heavily involve learning and low-level perception.
Because of these limitations, symbolic AI research has fallen out of favor in the deep learning era. However, many of these approaches are used to this day as part of the standard computer science toolbox, and many argue that symbol manipulation capabilities are necessary for overcoming the challenges of modern machine learning in terms of systematic generalization \cite{marcus2003algebraic,marcus2018deep,lake2017building,greff2020binding,pearl2018theoretical,scholkopf2021toward,battaglia2018relational}.

On a similar note, machine learning methods have also been shown to benefit from symbol-like observations and from explicitly incorporating structure in a connectionist model. Recent successes can be found, e.g., in reinforcement learning \cite{openai2019dota,vinyals2019grandmaster,ahmed2020causalworld} or in physical reasoning \cite{sanchez-gonzalez2020learning,battaglia2016interaction} where structured observations are available via the internal state of a simulator. This access to ground-truth structured data is often necessary to solve complex tasks that require high-level skills such as reasoning and planning. Looking forward, it will be crucial to relax the assumption that ground-truth information about the state of the world is available, and learn to extract it instead.\looseness=-1

Furthermore, there is evidence in cognitive psychology and neuroscience that humans perceive the world in a structured way, in terms of objects and their interactions \cite{spelke1990principles,teglas2011pure,wagemans2015oxford}. 
In fact, learning and reasoning about objects has been shown to develop in humans at an early age \cite{spelke2007core,dehaene2020how,baillargeon1985object}. 
Objects constitute compositional building blocks for higher-level cognitive tasks, and naturally enable systematic generalization outside of prior experiences \cite{dehaene2020how}. 

Taking once again inspiration from human cognition, it has also been proposed that artificial intelligence should take a hybrid, \emph{neuro-symbolic} approach, where direct sensory information is integrated with complex abstractions that allow for reasoning and planning \cite[Section~5.2]{marcus2018deep}.
In this context, symbol manipulation can be performed by purely symbolic methods \cite{asai2018classical,yi2018neural,mao2019neuro,dittadi2021planning,ayton2021policy} or within a connectionist framework \cite{evans2018learning,smolensky1990tensor,pollack1990recursive,battaglia2018relational,greff2020binding}.

In general, object-centric representations are likely to play a crucial role in artificial learning systems that are able to reason, plan, and generalize systematically beyond their experience. In the remainder of this thesis, we will focus on how these representations are obtained from perceptual data alone, without any supervision, and regardless of how higher-level cognitive functions that manipulate these representations may be implemented.
Note, however, that these issues are more generally related to the binding problem in neural networks, i.e., the ``inability of contemporary neural networks to effectively form, represent, and relate symbol-like entities'' \cite{greff2020binding}.

\subsection{Object-centric learning}

In \emph{object-centric representation learning}, we assume the data comes from a structured generative process based on discrete entities, their relationships, and a set of properties defining such entities and relationships. While disentangled representation learning (\cref{sec:background - disentanglement}) is concerned with learning a uniform collection of factors of variation underlying the data, here we shift our focus to the compositional structure of data in terms of building blocks that we call \emph{objects}. In this dissertation, we will focus on the most natural and intuitive case in the image domain, where objects are indeed \emph{concrete} objects that are \emph{visually} perceived in the world. However, the term ``object'' could be interpreted more generally as, e.g., spoken words or utterances, remembered entities, or abstract concepts and categories \cite[Section~2.3]{greff2020binding}. 

The goal of object-centric learning is to obtain data representations that combine the richness of neural representations with the compositionality of symbols: for example, new objects can be created from unseen feature combinations, and objects can be composed in novel ways without their features interfering with each other (this is related to the ``superposition catastrophe'' \cite{von1986thinking,bowers2014neural}).
Although, intuitively, object-centric learning is strictly related to disentanglement learning---or may even be considered a special case thereof \cite[Section 6]{scholkopf2021toward}---the traditional disentanglement framework cannot be straightforwardly applied in this case, as it assumes flat representations with a fixed vector format that imposes an arbitrary ordering of the dimensions. 

The typical setting in {object-centric representation learning} is to assume the generative factors are a \emph{set} of latent vectors $\{\zb_i\}_{i=1}^N$ where $N$ is the number of objects and each $\zb_i$ contains the representation of one object in the observation $\xb$. This \emph{separation} \cite[Section~3.1.1]{greff2020binding} of object representations is crucial for compositionality at the scene level. 
The number of objects could even be treated as a (discrete) generative factor of variation that defines the number of factors of variation for a given data point---e.g., in \citet{dittadi2019lavae}, a discrete random variable controls the number of latent vectors (\emph{slots}) representing objects in each observation.
Note that this \emph{slot-based} approach to object separation is only one of the possible ways to tackle the problem---see \citet[Section~3.3]{greff2020binding} for a discussion of more sophisticated strategies.

Although this thesis focuses on the common \emph{autoencoding} setting for static images, object representations can also be learned via contrastive \cite{kipf2019contrastive} or adversarial \cite{van2020investigating,chen2019unsupervised} methods, 
or as intermediate representations in a larger model that is trained on a supervised task \cite{locatello2020object}. 
Additional inductive biases can also be introduced through data, e.g., exploiting temporal information by observing sequences or by interacting with an environment \cite{kipf2019contrastive,lowe2020learning,kipf2021conditional,kabra2021simone}.

In the slot-based case considered here, all object representations $\zb_i$ have a \emph{common representational format} achieved by weight sharing across slots \cite[Section~3.1.2]{greff2020binding}. The shared format allows generalization between objects when using the representations downstream. This should lead, e.g., to the generalization of relations between objects independently of the context, which is relevant for reasoning, planning, and other high-level tasks.
Finally, following the arguments in \cref{sec:background - disentanglement}, it may be desirable for each object's features to be disentangled. This, together with the common representational format, is hypothesized to be beneficial for generalization to unseen feature combinations \cite[Section~3.1.3]{greff2020binding}.

\subsection{Slot-based methods relevant for this thesis}\label{sec:background - slot-based-models}

In this section, we provide a brief overview of slot-based models and a high-level description of the methods relevant for this thesis. This section is based on parts of Paper~\paperThree (\cref{chapter:objects}) and its supplementary material (\cref{chapter:appendix:objects}).

Slot-based methods for object-centric learning can be categorized according to their approach to object separation \cite{greff2020binding}.
In models that use \emph{instance slots}, each slot is used to represent a different part of the input. This introduces a routing problem, because all slots are identical but they cannot all represent the same object, so a mechanism needs to be introduced to allow slots to communicate with each other.
In models based on \emph{sequential slots}, the representational slots are computed in a sequential fashion, which solves the routing problem and allows to dynamically change the number of slots, but introduces dependencies between slots.
In models based on \emph{spatial slots}, a spatial coordinate is associated with each slot, introducing a dependency between slot and spatial location. 
In this work, we focus on four scene-mixture models as representative examples of approaches based on instance slots (Slot Attention), sequential slots (MONet and GENESIS), and spatial slots (SPACE). 
What follows is a high-level overview of these four models.

\paragraph{MONet.} In MONet~\cite{burgess2019monet}, attention masks are computed by a recurrent segmentation network that takes as input the image and the current \emph{scope}, which is the still unexplained portion of the image. For each slot, a variational autoencoder (the \emph{component VAE}) encodes the full image and the current attention mask, and then decodes the latent representation to an image reconstruction and mask. The reconstructed images are combined using the \emph{attention} masks (\emph{not} the masks decoded by the component VAE) into the final reconstructed image. 
The reconstruction loss is the negative log-likelihood of a spatial Gaussian mixture model (GMM) with one component per slot, where each pixel is modeled independently. The overall training loss is a (weighted) sum of the reconstruction loss, the KL divergence of the component VAEs, and an additional mask reconstruction loss for the component VAEs.\looseness=-1

\paragraph{GENESIS.} Similarly to MONet, GENESIS~\cite{engelcke2020genesis} models each image as a spatial GMM. The spatial dependencies between components are modeled by an autoregressive prior distribution over the latent variables that encode the mixing probabilities.
From the image, an encoder and a recurrent network are used to compute the latent variables that are then decoded into the mixing probabilities. The mixing probabilities are pixel-wise and can be seen as attention masks for the image. Each of these is concatenated with the original image and used as input to the component VAE, which finds latent representations and reconstructs each scene component. These are combined using the mixing probabilities to obtain the reconstruction of the image.
While in MONet the attention masks are computed by a deterministic segmentation network, GENESIS defines an autoregressive prior on latent codes that are decoded into attention masks. GENESIS is a proper probabilistic generative model, and it is trained by maximizing a modification of the ELBO introduced by \citet{rezende2018taming}, which adaptively trades off the log-likelihood and KL terms in the ELBO.\looseness=-1

\paragraph{Slot Attention.} As our focus is on the object discovery task, we use the Slot Attention autoencoder model proposed by \citet{locatello2020object}.
The encoder consists of a CNN followed by the Slot Attention module, which maps the feature map to a set of {slots} through an iterative refinement process. 
At each iteration, dot-product attention is computed with the input vectors as keys and the current slot vectors as queries. The attention weights are then normalized over the slots, introducing competition between the slots to explain the input. Each slot is then updated using a GRU that takes as inputs the current slot vectors and the normalized attention vectors. After the refinement steps, the slot vectors are decoded into the appearance and mask of each object, which are then combined to reconstruct the entire image. The model is optimized by minimizing the MSE reconstruction loss.
While MONet and GENESIS use sequential slots to represent objects, Slot Attention employs instance slots.

\paragraph{SPACE.}
Spatially Parallel Attention and Component Extraction (SPACE; \citeplain{lin2020space}) combines the approaches of scene-mixture models and spatial attention models. The foreground objects are segregated using bounding boxes computed through a parallel spatial attention process. The parallelism allows for a larger number of bounding boxes to be processed compared to previous related approaches. The background elements are instead modeled by a mixture of components. The use of bounding boxes for the foreground objects could lead to under- or over-segmentation if the size of the bounding box is not tuned appropriately. An additional boundary loss tries to address the over-segmentation issue by penalizing splitting objects across bounding boxes.

\subsection{Measuring object separation}

In this section, we define the segmentation metrics which we use in this dissertation to measure object separation. This section is based on the supplementary material for Paper~\paperThree (see \cref{chapter:appendix:objects}).

\paragraph{Adjusted Rand Index (ARI).}

The Adjusted Rand Index (ARI) \cite{hubertComparingPartitions1985} measures the similarity between two partitions of a set (or clusterings). Interpreting segmentation as clustering of pixels, the ARI can be used to measure the degree of similarity between two sets of segmentation masks. Segmentation accuracy is then assessed by comparing ground-truth and predicted masks.
The expected value of the ARI on random clustering is 0, and the maximum value is 1 (identical clusterings up to label permutation).
As in prior work \cite{burgess2019monet,engelcke2020genesis,locatello2020object}, we only consider the ground-truth masks of foreground objects when computing the ARI.
Below, we define the Rand Index and the Adjusted Rand Index in more detail.

The Rand Index is a symmetric measure of the similarity between two partitions of a set~\citep{rand1971objective,hubertComparingPartitions1985,wagner2007comparing}. It is inspired by traditional classification metrics that compare the number of correctly and incorrectly classified elements. 
The Rand Index is defined as follows:
Let $S$ be a set of $n$ elements, and let $A = \{A_1, \dots, A_{n_A}\}$ and $B = \{B_1, \dots, B_{n_B}\}$ be partitions of $S$. Furthermore, let us introduce the following quantities:%
\begin{itemize}
	\item $m_{11}$: number of pairs of elements that are in the same subset in both $A$ and $B$,
	\item $m_{00}$: number of pairs of elements that are in different subsets in both $A$ and $B$,
	\item $m_{10}$: number of pairs of elements that are in the same subset in $A$ and in different subsets in $B$,
	\item $m_{01}$: number of pairs of elements that are in different subsets in $A$ and in the same subset in $B$.
\end{itemize}
The Rand Index is then given by:
\begin{equation}
	\mathrm{RI} (A,B) = \cfrac{m_{11}+m_{00}}{m_{11}+m_{00}+m_{10}+m_{01}} = \cfrac{2(m_{11}+m_{00})}{n(n-1)}
\end{equation}
and quantifies the number of elements that have been correctly classified over the total number of elements. 

The Rand Index ranges from 0 (no pair classified in the same way under $A$ and $B$) to 1 ($A$ and $B$ are identical up to a permutation). However, the result is strongly dependent on the number of clusters and on the number of elements in each cluster. If we fix $n_A$, $n_B$, and the proportion of elements in each subset of the two partitions, then the Rand Index will increase as $n$ increases, and even converge to 1 in some cases \cite{fowlkes1983method}. The expected value of a random clustering also depends on the number of clusters and on the number of elements $n$.

The Adjusted Rand Index (ARI) \cite{hubertComparingPartitions1985} addresses this issue by normalizing the Rand Index such that, with a random clustering, the metric will be 0 in expectation.
Given the same conditions as above, let $n_{i,j} = |A_i \cap B_j|$, $a_i = |A_i|$, and $b_i = |B_i|$, with $i=1,\ldots,n_A$ and $i=1,\ldots,n_B$. The ARI is then defined as:
\begin{equation}
	\mathrm{ARI}(A,B)  = \cfrac{\sum_{i,j} \binom{n_{i,j}}{2} - \cfrac{\sum_{i} \binom{a_i}{2}\sum_{j} \binom{b_j}{2}}{\binom{n}{2}}}{\frac{1}{2}\left[\sum_i\binom{a_i}{2}+\sum_j\binom{b_j}{2}\right]-\cfrac{\sum_i\binom{a_i}{2}\sum_j\binom{b_j}{2}}{\binom{n}{2}}}
\end{equation}
which is 0 in expectation for random clusterings, and 1 for perfectly matching partitions (up to a permutation). Note that the ARI can be negative.

\paragraph{Segmentation covering metrics.}
Segmentation Covering (SC) \cite{arbelaez2010contour} uses the intersection over union (IOU) between pairs of segmentation masks from the sets $A$ and $B$. How the segmentation masks are matched depends on whether we are considering the covering of $B$ by $A$ (denoted by $A \rightarrow B$) or vice versa ($B \rightarrow A$).
We use the slightly modified definition by \citet{engelcke2020genesis}:
\begin{equation}
    \mathrm{SC}(A\rightarrow B) = \frac{1}{\sum_{R_B \in B} \left|R_B\right|} \sum_{R_B \in B} \left|R_B\right| \max_{R_A \in A} \operatorname{\textsc{iou}}(R_A, R_B)\ ,
\end{equation}
where $|R|$ denotes the number of pixels belonging to mask $R$, and the intersection over union is defined as:
\begin{equation}
    \operatorname{\textsc{iou}}(R_A, R_B) = \frac{\left|R_A \cap R_B\right|}{\left|R_A \cup R_B\right|}\ .
\end{equation}

While standard (weighted) segmentation covering weights the IOU by the size of the ground truth mask, mean (or unweighted) segmentation covering (mSC) \cite{engelcke2020genesis} gives the same importance to masks of different size:
\begin{equation}
    \mathrm{mSC}(A\rightarrow B) = \frac{1}{|B|} \sum_{R_B \in B}  \max_{R_A \in A} \operatorname{\textsc{iou}}(R_A, R_B)\ ,
\end{equation}
where $|B|$ denotes the number of non-empty masks in $B$.
Since a high SC score can still be attained when small objects are not segmented correctly, mSC is considered to be a more meaningful and robust metric across different datasets~\cite{engelcke2020genesis}.\looseness=-1

Note that neither SC nor mSC are symmetric: Following \citet{engelcke2020genesis}, we consider $A$ to be the predicted segmentation masks and $B$ the ground-truth masks of the foreground objects.
As observed by \citet{engelcke2020genesis}, both SC and mSC penalize over-segmentation (segmenting one object into separate slots), unlike the ARI.
Both SC and mSC take values in $[0,1]$.

\chapter{Study on representation learning in a robotic setting}
\label{chapter:contribs_disentanglement}

After having introduced the necessary background in \cref{chapter:background}, we briefly outline the main contributions of this thesis. 
Here we take a more focused approach than in the original papers, and discuss the key motivations, conclusions, and limitations of the studies.

In this chapter, we introduce the motivation and experimental setting of Papers~\paperOne and~\paperTwo (\cref{chapter:iclr2021,chapter:iclr2022}), summarize a few main results, and discuss the key takeaways.
The most relevant background for this chapter is in \cref{sec:background - representation learning,sec:background - vaes,sec:background - disentanglement}.
We take a similar approach in \cref{chapter:contribs_objects}, which is concerned with the experimental study in Paper~\paperThree (\cref{chapter:objects}).

\section{Introduction and study design}

In Papers~\paperOne and~\paperTwo (\cref{chapter:iclr2021,chapter:iclr2022}), we focus on representation learning for downstream tasks in the context of robotics. 
We scale disentangled representation learning to a robotic setting and analyze the link between properties of the learned representations and different flavors of generalization in downstream tasks, spanning from ground-truth factor prediction to robotic pushing.

\begin{figure}[tb]
    \centering
    \includegraphics[height=0.38\textwidth]{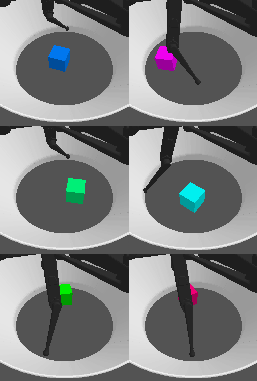}
    \hspace{0.05\textwidth}
    \includegraphics[height=0.38\textwidth]{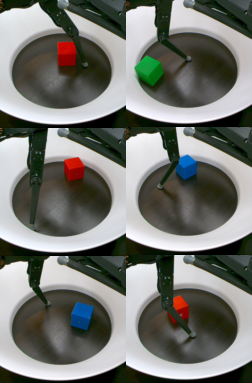}
    \caption{Random samples from the simulated (left) and real (right) dataset proposed in Paper~\paperOne \cite{dittadi2021transfer}.}
    \label{fig:robot_dataset_sim_real}
\end{figure}

The setting is a simulated robotic platform consisting of a bowl-shaped stage with a flat floor, a monochromatic cube with a range of possible colors, and a robotic arm with three degrees of freedom (\cref{fig:robot_dataset_sim_real}, left). 
We consider two tasks: reaching the cube, or pushing it to a given target location.
The platform has a real-world counterpart which we use for sim-to-real generalization experiments.\footnote{Our setting is derived from a more general setup with three robotic fingers that enables studying a wide range of robotic tasks from reaching to dexterous manipulation. This corresponds to CausalWorld \cite{ahmed2020causalworld} in simulation and the TriFinger robot \cite{wuthrich2020trifinger} in the real world.} Images from the real-world platform are shown in the right panel in \cref{fig:robot_dataset_sim_real}.

\subsection{Dataset}
Our first contribution, introduced in Paper~\paperOne (\cref{chapter:iclr2021}), is a new annotated dataset for learning and evaluating disentangled representations in the robotic setting introduced above.
A main advantage of this dataset is that it has a \emph{direct downstream application to real-world robotics}. This has the beneficial side effect that on this dataset it is more difficult to learn useful representations that accurately capture the underlying structure of the data. Consequently, our dataset also provides a valuable testbed to challenge state-of-the-art methods for disentangled representation learning.

There are several aspects to the difficulty of our dataset:
\begin{enumerate}
    \item It has a higher resolution (128x128) than most other datasets commonly used in disentangled representation learning \cite{higgins2016beta,reed2015deep,fidler20123d,kim2018disentangling,gondal2019transfer}, with the exception of SmallNORB \citep{lecun2004learning} that has the same resolution. Note that in this comparison we only consider datasets with precise annotations of the factors of variation: while there are datasets with labeled factors, such as CelebA \cite{liu2015deep}, these have only qualitative annotations and, therefore, do not allow for quantitative evaluations or fine-grained control over the factors.
    
    \item Our dataset has seven fully-annotated, fine-grained factors of variation; other datasets have fewer, except for MPI3D \cite{gondal2019transfer} that also has seven.
    
    \item Some of these factors of variation have correlations due to the interactions of the finger with the cube and the stage floor: the finger cannot be completely extended vertically and it cannot go through the cube. 
    In another work not included in this dissertation \cite{trauble2021disentangled}, although with simpler correlation structures, we show theoretically and empirically that disentangled representation learners might struggle when some factors of variation are correlated. However, we observe that the weakly supervised approach introduced in \cref{bgr:disentanglement partial supervision}, which we also use in Papers~\paperOne and~\paperTwo, may resolve this issue.
    
    \item Some factors of variation have a significantly larger impact on the pixel-wise reconstruction loss. This could make it challenging to find the ``sweet spot'' for regularization strength in autoencoder-based disentanglement learners. In fact, we observe that the cube rotation---the factor with the smallest impact---is sometimes not captured by the representations when the regularization is too strong. See the related discussion in \cref{sec:background - disentanglement - methods}, in the paragraph that introduces AnnealVAE (p.~\pageref{label:annealVAE}).
    
    \item Unlike most previous datasets, this dataset presents heavy occlusions. E.g., the tip of the finger may be hidden by the cube (or even exit the field of view, although this is not technically an occlusion), or the cube might be almost entirely hidden by the finger (see \cref{fig:robot_dataset_sim_real}).
    
    \item The factors of variations in the dataset are significantly more fine-grained than other datasets. This results in over $1.5$ billion possible combinations of the seven factors, while the largest among the common disentanglement datasets only has one million combinations.
    
    \item The fine granularity of the factors implies that our representations are trained on a rather small portion (less than 0.1\%) of the space of possible combinations. By contrast, previous works on other datasets report training on the entire datasets, i.e., on all factor combinations. 
\end{enumerate}
A further advantage of our dataset is that it enables sim-to-real evaluation: in addition to one million annotated images of the simulated platform, it also includes over 1,800 annotated images of the real platform (\cref{fig:robot_dataset_sim_real}, right).

\subsection{Learning representations}
To learn compact representations of simulated camera observations from our robotic platform, we choose the $\beta$-VAE for its simplicity. As discussed in \cref{sec:background - disentanglement - methods}, this is an extension to the variational autoencoder (VAE) that allows for the modulation of the information bottleneck capacity, thereby encouraging disentanglement. 
In addition to training $\beta$-VAEs in the standard unsupervised setting, we also use the Ada-GVAE approach proposed by \citet{locatello2020weakly} to introduce weak supervision in the training procedure (see \cref{bgr:disentanglement partial supervision}).

As mentioned above, this dataset is challenging for common disentanglement methods, which in our preliminary experiments failed to reconstruct the input images.
For this reason, we increased the encoder and decoder depths and ran a hyperparameter sweep to determine the best configuration (including depth and width of the networks, presence and parameterization of residual connections, weight initialization of the networks, batch normalization, and dropout). The final architecture is over 4 times as deep as standard convolutional autoencoders for disentanglement learning \cite{locatello2020sober}, and it has over 20 times as many parameters (see \cref{sec:implementation_details}). Proposing a practical way to scale up disentanglement learning to more challenging settings constitutes our second contribution.

\subsection{Study on generalization in downstream tasks}

As a third contribution of Paper~\paperOne (\cref{chapter:iclr2021}), we perform a reproducible large-scale study in which we train 1,080 variational autoencoders varying several hyperparameters, including supervision method (unsupervised or weakly supervised), bottleneck capacity, and presence of noise in the input. The latter was an attempt---which turned out to be relatively successful---to learn encoders that would be more robust to strong distribution shifts such as sim-to-real. Further details on the hyperparameter search are presented in \cref{sec:scaling,sec:implementation_details}.

In both Papers~\paperOne and~\paperTwo, we \emph{leverage the learned representation functions (the encoders) to learn downstream tasks, and investigate the relationship between representation properties and performance on downstream tasks, with a particular focus on out-of-distribution generalization}. The representation functions are pretrained and frozen, and a \emph{downstream model} is trained to solve a specific task given the (learned) data representation as input. In Paper~\paperOne, the task is to predict the ground-truth factors of variation; in Paper~\paperTwo, the task is either to reach the cube with the robotic finger, or to move it to a given location.

We formulate a framework for measuring generalization based on two scenarios:
\begin{itemize}
    \item \emph{OOD1}: The downstream model is evaluated out of distribution with respect to its training distribution, but \emph{the representation functions are still in distribution}. This means that the representation $r(\xb)$ of an input $\xb$ in the OOD1 set will be as good as representations of data in the downstream models' training distribution. Therefore, here we \emph{purely test the generalization of downstream models} trained on representations with different structures and properties.
    
    \item \emph{OOD2}: The downstream model is evaluated out of distribution with respect to both its own training distribution and that of the representation function. The key observation here is that, \emph{since the representation functions are deep neural networks, they are prone to well-known generalization issues}. Therefore, for a data point $\xb$ in the OOD2 set (when the encoders are out of distribution), the corresponding representation $r(\xb)$ \emph{may not faithfully represent} $\xb$.
\end{itemize}
Although often overlooked, this distinction is crucial when discussing generalization in representation learning, as it allows us to \emph{separately study (1) the structure and properties of representations, and (2) the generalization of representation functions}.

\begin{figure}[tb]
    \centering
    \includegraphics[width=0.6\textwidth]{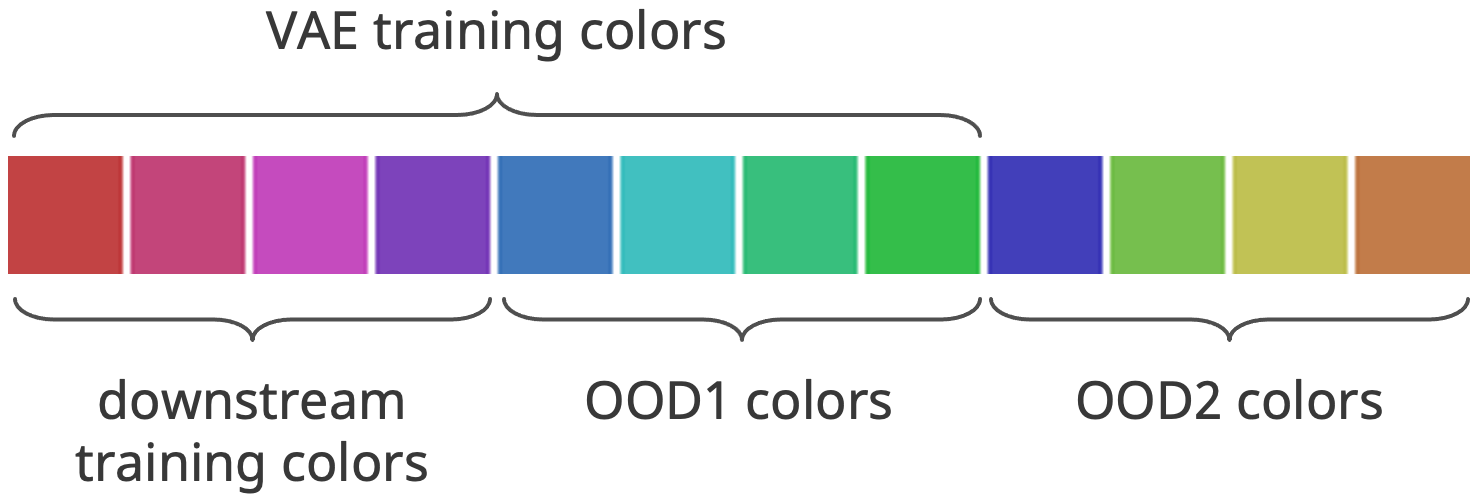}
    \caption{Illustration of the cube colors in the studies from Papers~\paperOne and~\paperTwo.}
    \label{fig:OOD-splits-visualization}
\end{figure}

For the sake of clarity, we now briefly introduce the concrete distribution shifts studied in \cref{chapter:iclr2021,chapter:iclr2022} in the simulated setting.\footnote{Although we also consider sim-to-real shifts, these cannot be precisely characterized in terms of factors of variations of the dataset. This could be solved, e.g., by introducing a binary factor of variation that denotes simulation or real world, or by describing simulated and real settings via a set of complex factors of variation, such as surface characteristics and lighting conditions. This is, however, out of the scope of this dissertation and the research papers it is based on.}
We treat the cube color as a nuisance factor (see \cref{sec:background - representation learning - importance of representation}) and consider distribution shifts affecting only this factor.
In our setup, the cube color can take 12 possible values (with uniformly distributed hue in the HSV space, and maximum saturation and value). As shown in \cref{fig:OOD-splits-visualization}, 
we train the representation functions on 8 of these colors, chosen at random before running the study, and kept fixed. The held-out colors are used for OOD2 evaluation, i.e., with the representation function out of distribution. The downstream tasks---e.g., predicting the cube's position, or pushing the cube to a target location---are then trained on a \emph{subset} of the colors that were used when training the representations (the 4 leftmost colors in the example in \cref{fig:OOD-splits-visualization}). When evaluating on the OOD1 colors, we gauge the generalization abilities of the downstream task, since in this case the representations should be accurate. When evaluating on the OOD2 colors, by contrast, we measure the generalization of the encoders as well.
In \cref{chapter:iclr2021}, we consider three different splits of the VAE training colors. The ``extrapolation'' split shown in \cref{fig:OOD-splits-visualization} (from downstream training colors to OOD1 colors) is the one we focus on in \cref{chapter:iclr2022}.

\section{Key findings on disentanglement and factor prediction}

In this section, we present and discuss the main results from Paper~\paperOne on learning disentangled representations and solving downstream factor prediction tasks.

\begin{figure}
    \centering
    \begin{subfigure}[b]{0.45\textwidth}
        \centering
        \includegraphics[height=0.7\textwidth]{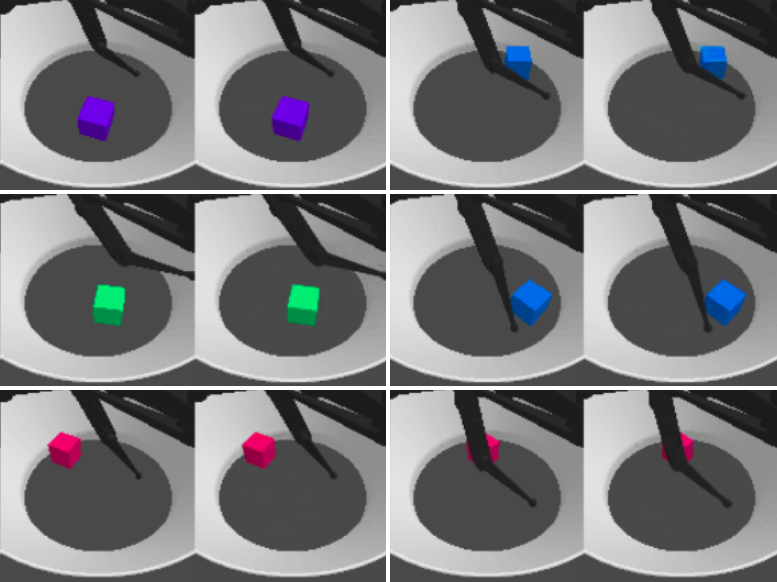}
        \caption{Inputs and reconstructions.}
        \label{fig:maintext-model-recons}
    \end{subfigure}
    \hspace{0.02\textwidth}
    \begin{subfigure}[b]{0.45\textwidth}
        \centering
        \includegraphics[height=0.7\textwidth]{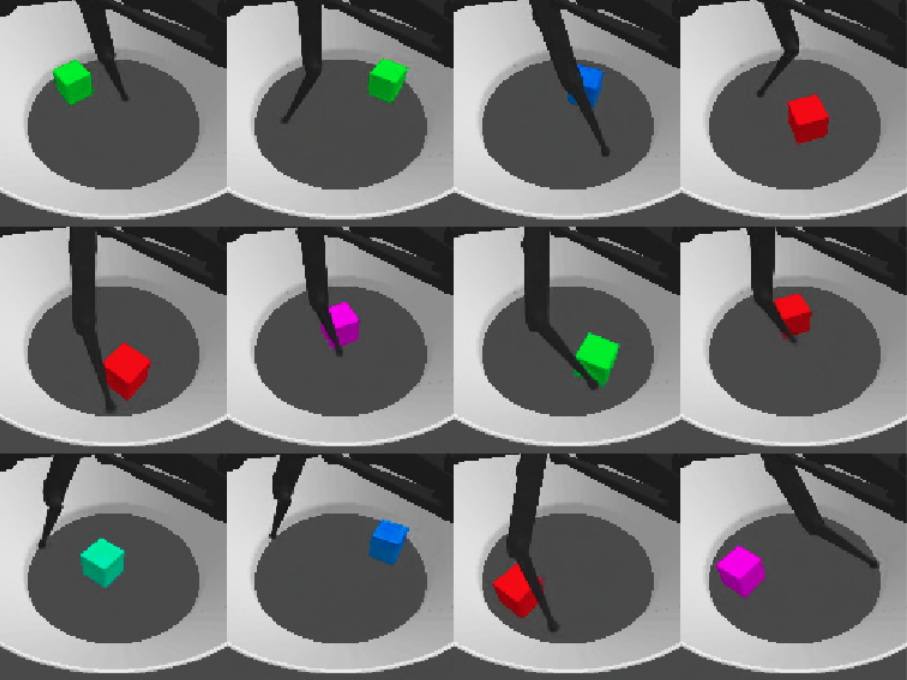}
        \caption{Random samples.}
        \label{fig:maintext-model-samples}
    \end{subfigure}
    \caption{Input--reconstruction pairs and random samples of a VAE from our study.}
\end{figure}

First, we demonstrate that the proposed architecture allows VAEs to reconstruct all relevant details of input images (see input--reconstruction pairs in \cref{fig:maintext-model-recons}). 
This suggests that the latent representations in the trained models accurately capture all relevant information in the data (see \cref{sec:properties learned representations - content}). Arguably, this is necessary if we want to draw meaningful conclusions about the usefulness of representations with diverse structures (see \cref{sec:properties learned representations - structure}): if relevant information is not retained by some representation functions, these will likely be less useful regardless of the structure of the information they do retain.
In addition, although generative modeling is not a goal of this research effort, we remark that most of the learned models can also sample new high-quality images, as shown in \cref{fig:maintext-model-samples}.

Another key finding is that weak supervision as defined in Ada-GVAE (see \cref{bgr:disentanglement partial supervision}) can successfully learn \emph{fully disentangled} representations, as seen by visual inspection and from the remarkably high DCI Disentanglement scores (often above $0.99$). The failure cases (when representations are not fully disentangled) typically occur when the latent space regularization is too strong and therefore one factor---the cube rotation, which contributes the least to the reconstruction error---is ignored (see discussion in \cref{sec:background - disentanglement - methods}).
Conveniently, this allows for model selection in the weakly supervised case, since the entangled models are very likely to have worse unsupervised metrics (reconstruction loss and ELBO). 
Even in the cases where one factor of variation is ignored, those that are not ignored are often disentangled. In contrast, none of the unsupervised models learn disentangled representations, and almost all of them have lower DCI and MIG scores than any weakly supervised model (of the four disentanglement metrics considered in this study,\footnote{We can only compute \emph{observational} metrics (DCI Disentanglement, MIG, Modularity, and SAP) on this dataset, as opposed to \emph{interventional} metrics (see \cref{sec:background - disentanglement - metrics}), because it does not provide interventional capabilities (it consists of a fixed set of images that is a rather small subset of all possible combinations of the factors of variation).}
these are the one that we empirically observed, via visual inspection, to better correspond to our intuitive notion of disentanglement).

Thanks to this combination of unsupervised and weakly supervised approaches, we learn a collection of encoders that extract highly diverse representations. This allows us to draw sound conclusions from the observed relationships between relevant metrics. 
We will now focus in particular on disentanglement metrics and downstream factor prediction: How does the disentangling capability of the upstream representation function $r$ affect the OOD performance of a downstream model that predicts the ground-truth factors of variation?
A key insight from our findings---which are discussed below---is that {the effect of disentanglement on generalization is not as straightforward as typically assumed}. 

\begin{figure}
    \centering
    \includegraphics[width=\linewidth]{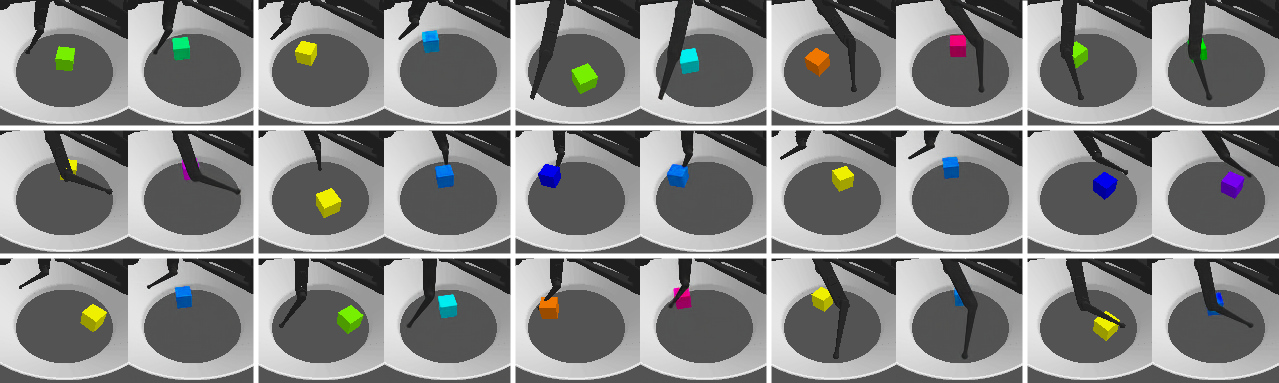}
    \caption[Each pair shows an OOD2 input (left) and reconstruction (right) by a VAE with perfectly disentangled factors of variation. The VAE was trained \emph{without} input noise.]{Each pair shows an OOD2 input (left) and reconstruction (right) by a VAE with perfectly disentangled factors of variation. The VAE was trained \emph{without} input noise. Since the cube colors are out of distribution with respect to the encoder's training distribution, the mapping from an input image to a latent representation is not well defined. The encoder is especially likely to fail if the input color is not ``similar enough'' to a color in the training distribution (e.g., when the cube is yellow). Unsurprisingly, the generative model always reconstructs colors that were present in the training set.}
    \label{fig:OOD2-reconstructions-disentangled-noiseless}
\end{figure}

First, as discussed above, it is crucial to define \emph{in what sense} we are measuring generalization: If the representation function is OOD, the resulting representations will not be disentangled---in fact, they may not even be faithful to the data. For example, if the cube color in the input image never appeared in the VAE's training distribution, it is arguably unreasonable to expect the encoder to correctly interpret the unseen color. 
Indeed, in our experiments, encoders are typically unreliable out of distribution, as they sometimes fail at inferring even in-distribution factors. For example, when the color is OOD, the cube rotation is very often inferred incorrectly (see \cref{fig:OOD2-reconstructions-disentangled-noiseless}). This appears to hold regardless of the degree of disentanglement, which suggests that any disentanglement capabilities are lost when the data distribution shifts.\looseness=-1

Second, when the downstream model is an MLP, the degree of entanglement does not seem to matter at all. This is not surprising, since an MLP can be expected to disentangle factors of variation that are entangled in a representation. However, we observe a surprising phenomenon in the OOD1 case (i.e., when the encoder is \emph{not} OOD): when the representation function \emph{perfectly disentangles} the factors, the downstream models reliably attain a very low error on the task, while in other cases we observe a significant variance (see \cref{fig:evaulations_transfer_ood1}, left). On the other hand, disentanglement is not correlated with the OOD2 generalization of a downstream MLP, i.e., when the encoder is OOD (see \cref{fig:evaulations_transfer_ood2}).

In summary, in our experiments \emph{disentanglement appears to matter only for OOD1 generalization, and what matters is mostly whether the representation is fully disentangled or not}. A possible explanation is that, when the prediction target is a (relatively simple) non-linear function of a single input feature, the optimization easily and reliably converges to the optimal solution where all other input features are ignored, including those that might be OOD at test time.

\begin{figure}
    \centering
    \includegraphics[width=\linewidth]{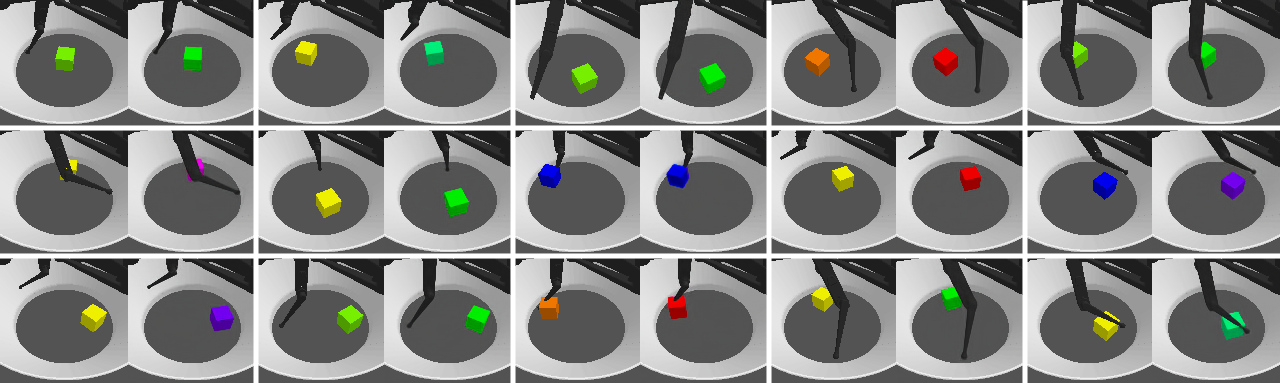}
    \caption[Each pair shows an OOD2 input (left) and reconstruction (right) by a VAE with perfectly disentangled factors of variation. The VAE was trained \emph{with} input noise.]{Each pair shows an OOD2 input (left) and reconstruction (right) by a VAE with perfectly disentangled factors of variation. The VAE was trained \emph{with} input noise. Unsurprisingly, the generative model always reconstructs colors that were present in the training set. However, unlike in the noiseless case in \cref{fig:OOD2-reconstructions-disentangled-noiseless}, the other factors of variation are inferred relatively well.}
    \label{fig:OOD2-reconstructions-disentangled-noisy}
\end{figure}

\begin{figure}[bt]
    \centering
    \begin{subfigure}[b]{\textwidth}
        \centering
        \includegraphics[width=\linewidth]{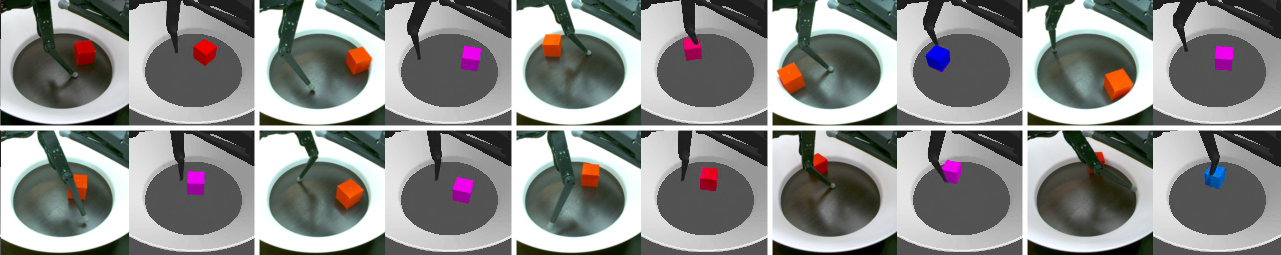}
        \caption{Trained without noise.}
    \end{subfigure}
    
    \vspace{1.5em}
    
    \begin{subfigure}[b]{\textwidth}
        \centering
        \includegraphics[width=\linewidth]{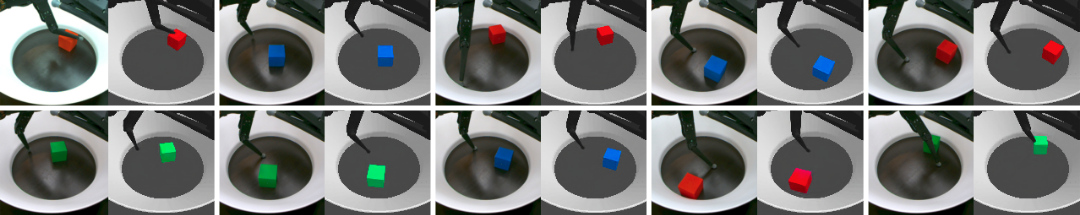}
        \caption{Trained with noise.}
    \end{subfigure}
    
    \caption{Each pair shows a real-world image input (left) and its reconstruction by trained VAEs (right). This is an OOD2 scenario. The model trained with input noise (b) infers the ground-truth factors of variation significantly more accurately than the model trained without noise (a).}
    \label{fig:OOD2-real-reconstructions-disentangled-both}
\end{figure}

Another interesting observation is that adding random noise to the input of the representation function during training significantly improves its generalization, both in simulation (OOD cube colors) and to the real world. This is not entirely surprising, as noisy inputs have been shown theoretically and empirically to improve the robustness of neural networks. However, the extent of the improvement is undoubtedly significant, as seen qualitatively by comparing \cref{fig:OOD2-reconstructions-disentangled-noiseless} and \cref{fig:OOD2-reconstructions-disentangled-noisy}.
Note that this falls into the OOD2 generalization category, and is particularly useful in practice in sim-to-real scenarios---\cref{fig:OOD2-real-reconstructions-disentangled-both} shows examples of the effectiveness of input noise for zero-shot sim-to-real generalization.
Quantitative results are presented in \cref{sec:transfer} (\cref{fig:evaulations_transfer_w_noise}).\looseness=-1

\section{Study on robotic tasks}

In Paper~\paperTwo (\cref{chapter:iclr2022}), we investigate the role of pretrained representations on the performance and generalization of downstream reinforcement learning agents in the robotic setup introduced earlier, both in simulation and in the real world. 
To the best of our knowledge, this is the first systematic and extensive account of the OOD generalization of downstream reinforcement learning agents in robotics, and how this generalization is affected by characteristics of the upstream pretrained representation functions.
In this section, we briefly outline the experimental setup for this study. We then discuss a few key insights in \cref{sec:contribs_results_RL}, leaving a more extended exposition for \cref{chapter:iclr2022}.

In this study, we evaluate distribution shifts within a framework that is analogous to the one in Paper~\paperOne. However, the experimental study in this robotic setting is significantly more complex and demanding:
\begin{itemize}
    \item The downstream task is considerably more challenging: While predicting one factor of variation involves training a downstream model on 10k samples for a few thousand steps, we train our reinforcement learning policies for 3M steps for pushing.
    
    \item While in the simple prediction task we can precompute the representations of the entire training and test sets, in the reaching and pushing tasks we must encode each input image at runtime. This is an additional source of computational cost, besides the mere fact that the task is harder.
    
    \item Since the objective function in reinforcement learning is typically based on some form of expected cumulative reward, its gradients with respect to the model parameters are not computable in closed form and have to be estimated \cite{mohamed2020monte}.
    Gradients in reinforcement learning are thus generally harder to estimate than in supervised learning tasks such as factor prediction, where the training loss is directly differentiable with respect to all parameters.
    This, together with the fact that the tasks are harder in the first place, require us to use many (10 for pushing, 20 for reaching) random seeds for downstream RL training. This further increases the already high overall computation time by one order of magnitude.
\end{itemize}
Thus, we opt to reduce the computational burden by limiting the number of pretrained representations, and use only a subset of the models trained for Paper~\paperOne.

We train reinforcement learning policies with SAC \cite{haarnoja2018soft} to either reach the cube in the arena or push it to a given target location.
For reaching, we measure success by the fractional progress made by the tip of the robot finger from its initial position to the 
surface
of the cube. 
For pushing, we measure the fractional volumetric overlap between the cube and the goal (which is defined as a cube of the same size).

We select a subset of the representation functions from Paper~\paperOne and train 11,520 downstream policies in total.
For each encoder, we use 20 random seeds for training downstream policies on reaching, and 10 seeds on pushing.
Since disentanglement is one of the central themes of this study, we also explore the effect of L1 regularization on the input, in an attempt to encourage the downstream policies to disregard nuisance factors and therefore be more robust to distribution shifts.

The input to the downstream policies is the concatenation of: (1) the input representation $r(\xb)$, where $r$ is the frozen, pretrained representation function, (2) the ground-truth angles and velocities of the robot joints, and (3) the target position and orientation of the cube, in the pushing task. The cube's position and orientation are thus the only pieces of information that must be contained in $r(\xb)$, since all other relevant quantities are available in the ground-truth state observation. Despite this, solving these tasks proved to be difficult---in fact, they can be challenging even from complete ground-truth information \cite{ahmed2020causalworld}.

\section{Key findings on robotic tasks}\label{sec:contribs_results_RL}

One of the motivations for introducing the annotated robotics dataset in Paper~\paperOne was to scale up disentangled representation learning to more realistic settings and eventually evaluate its usefulness on more relevant downstream tasks than factor prediction on toy datasets, such as reinforcement learning on a robot. 
In Paper~\paperTwo, we deliver on this promise. However, since the results on disentanglement are mostly negative, we expand our study to investigate more in general how the generalization of downstream policies relates to a variety of metrics that can be computed on the representations before training the policies.

We start by analyzing the effect of disentanglement on the generalization of the trained policies. In Paper~\paperOne, we observed that disentanglement is not beneficial in the OOD2 scenario, but it can be helpful for OOD1 generalization (i.e., when the encoder is in distribution) only when the representations are perfectly disentangled.
In Paper~\paperTwo, on the other hand, we find the role of disentanglement to be negligible: it appears not to be beneficial even when the encoder is kept in distribution and even if it perfectly disentangles the factors of variation.

As in Paper~\paperOne, we test the encoder's robustness (OOD2 generalization) by evaluating the policies on unseen cube colors in simulation, as well as on the real robot. Here we also test on an unseen shape (a sphere) both in simulation and in the real world. 
In simulation, where it is feasible to evaluate a large number of policies on hundreds of episodes each, we observe that training the representation functions with input noise significantly improves OOD2 generalization, which is in line with the conclusion we reach on the simpler prediction setting of Paper~\paperOne. 
In particular, some of the policies trained in simulation generalize surprisingly well zero-shot to the real robot, without any fine-tuning or domain randomization during training.
Crucially, the best policies in the real-world setting tend to be the best ones in the OOD2 setting in simulation, for example with unseen cube colors. Therefore, \emph{we can use a policy's OOD2 performance in simulation to predict how it will perform on the real robot}.

Perhaps the most valuable takeaway from this study, however, is that we can interpret the out-of-distribution performance on the simple downstream tasks from Paper~\paperOne as \emph{generalization scores}, to be added to the collection of representation metrics.
Thus, if we are interested in a specific type of generalization for downstream policies---e.g., corresponding to a specific training distribution and test-time distribution shift---we can simply replicate a similar train/test scenario on a simple prediction task and expect the OOD performance on such a task to be predictive of the policies' OOD performance. We can use the toy task as a \emph{proxy task} to predict the performance of downstream models on a \emph{target task}.
Notably, the distribution shift need not be exactly the same in the proxy and target tasks: for example, the results on substantially different OOD2 shifts---OOD cube colors in simulation, OOD shape in simulation, and sim-to-real shift---appear to be correlated.

Let us consider a concrete example to highlight the relevance of these results.
Assume we have trained a large number of representation functions to be used as vision backbones for downstream reinforcement learning. We train downstream policies in simulation and deploy them on a real robot. It is reasonable to assume that we should train many policies for each representation function, varying hyperparameters and random seed. 
Fortunately, our study suggest that we can leverage simple proxy tasks to drastically reduce the number of policies that need to be trained: First, we train factor prediction models for all representation functions and evaluate them on images from the real robot. Then, we select the encoders with the best OOD performance and run a hyperparameter sweep for the policies using only this subset of encoders as upstream models.
Note that, since the OOD2 performance of the policies in simulation and in the real world are also correlated, we could then further reduce the cost of deploying multiple policies to the real robot by filtering out the ones that do not perform particularly well on the simulated OOD2 setting.
However, the most impactful advantage is arguably the possibility to pre-select representations that are more likely to lead to robust downstream policies.

Finally, 
although not discussed in the paper, it would not be surprising if these results held beyond reinforcement learning in robotics. In fact, we could hypothesize that the correlations between the performances on proxy and target tasks may primarily depend on how similar the downstream models and tasks are.

Regarding the \emph{similarity of downstream models}, in Paper~\paperTwo we indeed observe significant correlations only when the downstream model for factor prediction is an MLP, like the neural networks in the RL agents. Note that this is probably a soft constraint: First, each RL agent actually consists of multiple MLPs (the policy, value, and Q networks), therefore a direct comparison is not straightforward. Second, in Paper~\paperOne we observe that the behaviors of MLPs with up to 3 hidden layers are similar to each other, but different from gradient boosted trees, random forests, and $k$-nearest neighbors.\looseness=-1

The \emph{similarity of the proxy and target downstream tasks} can probably be expressed in terms of the mutual information between the prediction targets of the two tasks.
Assume that the data is defined by a set of ground-truth (generative) factors $\GGG$, and that solving the proxy task $T_p$ and the target task $T_t$ requires information about two subsets $\GGG_p,\GGG_t \subset \GGG$ of factors.\footnote{For simplicity we could refer to the factors by some arbitrary indices such that $\GGG \subset \mathbb{N}$.} We might then observe that models for $T_p$ that are downstream of a representation function $r$ tend to exhibit some properties, e.g., good in-distribution performance or robustness to a specific kind of distribution shift. Intuitively, if $\GGG_p$ and $\GGG_t$ are ``similar'', we may expect the same properties to hold to some extent for models that are trained downstream of $r$ to solve task $T_t$.\footnote{Denoting by $\yb_p$ and $\yb_t$ the prediction targets of the proxy and target tasks, respectively, we could attempt to formalize this notion using the mutual information $I(\yb_p;\,\yb_t)$. In reinforcement learning, for example, the prediction target might be the reward (or expected discounted return) for a state--action pair or the action probabilities of an optimal stochastic policy.}
Note that, if the tasks are entirely unrelated (i.e., $\GGG_p \cap \GGG_t = \varnothing$), some properties may still be consistent: for example, if $r$ is robust to distribution shifts, downstream models trained on $T_p$ and $T_t$ might both exhibit good OOD2 performance; and if $r$ is \emph{not} robust, both downstream models will probably not perform well in an OOD2 setting.
Although these relationships are not straightforward to characterize, we found some empirical evidence that partially supports this hypothesis:
The OOD1 performance of the policies is particularly correlated with the OOD1 accuracy when predicting the \emph{factors that are not included in the ground-truth state}, i.e., those that necessarily have to be inferred from the learned representation (see \cref{fig:sim_world_OOD_rank_correlations_reaching_no_reg}), while the correlation with the prediction performance of other factors is significantly milder. A similar result holds for OOD2 scenarios.

\chapter{Study on object-centric representations}\label{chapter:contribs_objects}

In this chapter, we motivate and outline the design of the study in Paper~\paperThree, and discuss its main results. Further details can be found in the paper (\cref{chapter:objects}) and in the corresponding supplementary material (\cref{chapter:appendix:objects}).
The most relevant background for this chapter is in \cref{sec:background - representation learning,sec:background - objects}.

\section{Introduction}

Compositional generalization is widely recognized as a fundamental issue in deep learning \cite{lake2017building,greff2020binding}. 
The object-centric learning literature is mainly concerned with proposing new methods, while \emph{assuming} that learning about objects may (at least partially) solve the issue of generalization.
In Paper~\paperThree \cite{dittadi2021generalization}, we investigate this claim via a systematic empirical study on unsupervised object-centric learning, with a focus on image data.\looseness=-1

\subsection{Overview of the study design}
We start by formulating three hypotheses about the unsupervised learning of object-centric representations, roughly following assumptions made in previous work but largely left implicit. In summary, an encoder that separates objects should:
\begin{enumerate}
    \item Separately represent each object's properties in a complete and accurate manner, and therefore be useful for arbitrary downstream tasks.
    
    \item Be robust to distribution shifts affecting a single object per image, in the sense that the representations of all other objects should still be reliable, even if the affected object has out-of-distribution properties.
    
    \item Be robust to distribution shifts globally affecting the input, such as introducing occlusions, cropping the input image, or increasing the number of objects beyond the maximum number observed in the training distribution.
\end{enumerate}

To investigate these hypotheses, we design a systematic experimental study in which we train and evaluate object-centric learning models on multi-object datasets that are annotated with segmentation masks and object properties.

\paragraph{Models.}
We focus on slot-based models, a popular approach for unsupervised object-centric learning (see \cref{sec:background - slot-based-models}), but we include models covering different approaches to object separation: instance slots (Slot Attention), sequential slots (MONet and GENESIS), and spatial slots (SPACE).
Moreover, we train standard variational autoencoders (\cref{sec:background - vaes}) in an attempt to gauge the downstream usefulness of non-object-centric representations (see \cref{sec:objects-vae-discussion}).

\begin{figure}
    \centering
    \includegraphics[width=0.8\textwidth]{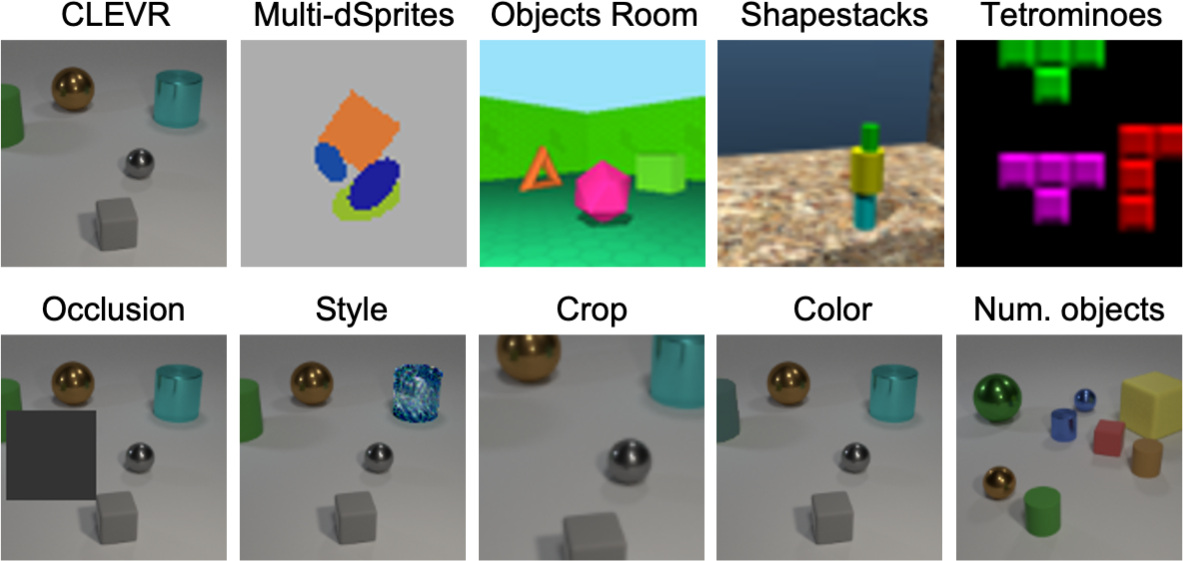}
    \caption{Top: examples from the five datasets considered in the study. Bottom: Example of distribution shifts applied to CLEVR. Figure from Paper~\paperThree \cite{dittadi2021generalization}.}
    \label{fig:objects_dataset_and_shifts_copy}
\end{figure}

\paragraph{Datasets.} 
We train and evaluate the models on five multi-object datasets that have been used in the literature as benchmarks for object-centric models. All of them have segmentation masks; four of them have annotations for all object properties. In \cref{fig:objects_dataset_and_shifts_copy} (top row), we show examples from the datasets (see \cref{fig:dataset_overview} in \cref{app:datasets} for more details).

\paragraph{Evaluation.}
We use segmentation masks to directly assess how a trained model separates objects.
The object property annotations allow us to evaluate representations via a downstream task that consists in predicting the ground-truth factors of variation of all objects.
This task is similar to downstream prediction tasks that are common in the disentanglement literature, including Papers~\paperOne and~\paperTwo. The general idea is that, if this information can be retrieved efficiently from the representation, it stands to reason that any arbitrary downstream task that depends on these underlying factors should also be solvable. 

\paragraph{Probing generalization.}
Finally, to assess how object separation and usefulness of the representations are affected by distribution shifts (Hypotheses 2 and 3), we evaluate segmentation accuracy and downstream task performance after introducing these shifts.
In \cref{fig:objects_dataset_and_shifts_copy} (bottom row), we show examples of distribution shifts considered in our study (see \cref{fig:transform/transforms_datasets} in \cref{app:evaluation_shifts} for a complete overview).

\subsection{On the comparison with non-object-centric models}\label{sec:objects-vae-discussion}

In \cref{chapter:contribs_disentanglement} (see Papers~\paperOne and~\paperTwo in \cref{chapter:iclr2021,chapter:iclr2022}) we have discussed that disentanglement may not necessarily always be useful for downstream tasks. Similarly, in Paper~\paperThree, we investigate the extent to which object properties can be accurately predicted from the ``flat'' distributed representations of VAEs, as opposed to object-centric representations.
To this end, we train VAEs with a latent space size that is comparable with the global representation size of slot-based models, and attempt a direct comparison on downstream task performance.

While it is not straightforward to gauge the fairness of this comparison, there are a few crucial points that should be discussed.
The first point concerns the size of the models.
We performed a light hyperparameter search (mostly on model depth) to obtain models that would at least reconstruct the input relatively well. Since some of these datasets are relatively complex---although this may not be clear from their visual appearance---the final VAEs employed in our study are relatively large. (Interestingly, Multi-dSprites turned out to be the most challenging dataset in this hyperparameter search, possibly due to the visual complexity and heavy occlusions.)
This may raise questions regarding the fairness of the comparison. On the other hand, we should note that: (i) the decoder now faces the challenge of generating the entire image at once, while typically in slot-based models a simpler decoder is applied independently to each slot to generate a single object in the scene; (ii) the object-centric models considered in this study are already rather diverse in terms of size 
and the largest ones are even comparable to our VAEs in terms of parameter count.\looseness=-1

A second observation is that the downstream task typically considered in object-centric learning is in some sense meant for object-centric models.
In fact, in a multi-object setting, predicting the properties of all objects is effectively a \emph{set prediction} task. In a slot-based object-centric representation where all slots have a common format, one can apply a simple downstream model to each slot and then match the predictions to the ground-truth labels to minimize the total loss. In other words, we can directly exploit the set-like structure of the representations. Conversely, in the VAE setting, a single vector represents the entire scene, so the same setup is not directly applicable. 

A possible approach to overcome this issue, as suggested by \citet{greff2019multi}, is to use the whole-scene representation to predict the properties of all objects at once, such that the objects are sorted lexicographically according to their properties. Potential issues with this approach are:
\begin{enumerate}
    \item The model's input and output size are substantially larger than those of the downstream model used for object-centric representations, and there is no weight sharing.
    \item The model must learn to sort objects, which could be relatively challenging.
    \item On the other hand, since the target is sorted according to object properties, there is a bias that can be exploited by the model to predict object properties better than by random chance.
    \item When testing such a model with more objects than seen during training, we are in effect probing the generalization capabilities (extrapolation, in some sense) of the model. Conversely, in the slot-based case, matching more slots to more objects has nothing to do with the downstream model, as it is simply performed by a combinatorial optimization algorithm (e.g., the Hungarian algorithm). 
\end{enumerate}
Another possible solution is to have the model output a flat vector containing predicted properties of all objects in no particular order, and then match the predictions to ground-truth object properties using the loss function, as we do in the slot-based case. This has similar issues and biases as the method discussed above.

In summary, when using set prediction as a downstream task, comparing object-centric and distributed representations is a challenging problem. Although we attempt to adapt this task to distributed representations in a relatively simple manner in order to minimize potential confounding effects, the comparison is arguably still not entirely fair.

\subsection{Library}
A side product of this study is a PyTorch-based \cite{paszke2019pytorch} Python library for training and evaluating object-centric learning methods. 
Since the five chosen datasets are not necessarily straightforward to use in general (e.g., the ones from the DeepMind dataset suite \cite{multiobjectdatasets19} require TensorFlow \cite{abadi2016tensorflow}), we repackaged all datasets into a common, general-purpose format.
We adapted the official PyTorch implementations of GENESIS and SPACE to our framework, and re-implemented MONet and the Slot Attention autoencoder.
Note that, since none of the models was originally tested on all the datasets considered here, some hyperparameter sweeps were necessary for some model--dataset combinations.
After training a model, the library allows for automatic evaluation of segmentation metrics and reconstruction error, as well as downstream property prediction with various matching strategies and downstream models. All these evaluations can be performed out of the box on new models or datasets, as long as they implement the standard interface defined in our library.

\section{Main results and discussion}
\label{sec:contribs_objects_discussion}

In this section, we summarize and discuss the key findings of our study.

\summary{MSE correlates with ARI (which is the most useful segmentation metric)}
We find strong correlations between the ARI and the MSE across all five datasets: given a set of models trained on the same dataset, those that reconstruct the input more accurately also tend to separate objects better according to the ARI. Therefore, the MSE can be a useful proxy metric to select high-ARI models, when ground-truth validation masks are not available.
Other segmentation metrics (SC and mSC) agree with the ARI to a varying extent depending on the dataset, and their correlation with the MSE is milder and inconsistent across datasets (see \cref{fig:indistrib_correlation/metrics_metrics}). However, as discussed below, we found that the ARI appears to be a generally more useful segmentation metric than the others considered in our study.

\summary{Model selection for downstream performance: ARI, or MSE if no masks}
Segmentation may not be the ultimate goal, and we might be interested in learning object-centric representations for downstream tasks (see \cref{sec:bgr - task-agnostic representations}). This is related to our first hypothesis above: object representations should contain useful information for downstream tasks, and information about different objects should be stored separately. 
Indeed, in our experiments, we observe that good performance on object property prediction can be achieved by downstream models that receive learned object-centric representations as input.
Interestingly, we find the ARI to be the only segmentation metric that consistently has a strong correlation with downstream performance across all datasets, object properties, and downstream models. This points to the ARI as a valuable metric for model selection when ground-truth segmentation masks are available for validation.
The MSE is also significantly correlated with downstream performance---which is unsurprising, considering its correlation with the ARI---but to a lesser extent and less consistently than the ARI. Although less effective than the ARI, the reconstruction error can therefore be useful for model selection when masks are not available.

\summary{ARI is robust even with textures; MSE is useless}
Note, however, that in more realistic scenarios there may be irrelevant details in the image (possibly on the objects themselves) that are in effect nuisances with respect to the downstream tasks of interest (see the discussion on nuisance factors in \cref{sec:background - representation learning - importance of representation,sec:properties learned representations - content}). In such cases, accurate reconstruction might be practically unrelated to correct object separation and downstream usefulness of the representations.
In \citet{papa2022inductive}, we confirm this hypothesis. We test different variations of MONet and Slot Attention on datasets with complex textures added to the objects via neural style transfer, and study the relationship between reconstruction quality, segmentation accuracy, and downstream performance on the same tasks considered in Paper~\paperThree. We find that the ARI is still consistently correlated with downstream performance, while the MSE is not. It follows that, unsurprisingly, reconstructing detailed textures is largely irrelevant for downstream tasks that focus on higher-level object properties, while segmentation quality remains crucial. 
In conclusion, the ARI may be a valuable and relatively robust proxy metric for the downstream usefulness of object-centric representations, even when objects have complex textures. Conversely, the reconstruction MSE is not necessarily informative when objects have a more complex appearance such as in the presence of diverse textures \cite{papa2022inductive}.

\summary{Downstream performance of OC vs. non-OC}
Next, we are interested in comparing the downstream performance attainable from object-centric representations and from ``flat'', distributed ones.
As discussed in \cref{sec:objects-vae-discussion}, making a fair comparison is challenging, since the set prediction task considered here is particularly well-suited for set-like representations in the first place.
In order to make a sensible assessment of distributed representations, we set up simple experiments where the downstream predictions are initialized to a random vector of object properties for all objects in the scene, and are optimized with the standard training procedure for downstream tasks for distributed representations. We find that the baseline performance for distributed representations is often remarkably higher than the standard ``random guess'' baseline we would use for object-centric representations (although in some cases it is approximately equal). 
Despite this advantage, downstream predictors from object-centric representations tend to outperform those from distributed representations. Although the task may arguably be more difficult for a single downstream MLP that has to predict all objects at once (see \cref{sec:objects-vae-discussion}), we also observe that the performance typically does not improve significantly when increasing model size up to three hidden layers.

\summary{Generalization with single-object shifts (hyp. 2)}
Our second hypothesis is that, when one object is out of distribution, the representations of other objects should be robust, i.e., they should still faithfully represent those objects' properties. 
First, we observe that the models tend to be able to segment the scene correctly in this single-OOD-object scenario. 
Then, we focus on the representations of the objects in the scene, and use the downstream prediction performance to assess how they are affected by distribution shifts.
Our results, presented in \cref{sec:results/hyp2}, indicate that:
\begin{itemize}
    \item Property prediction for out-of-distribution objects deteriorates. Moreover, the representations of the out-of-distribution objects are not reliable, since even retraining the downstream model after the distribution shift has occurred does not significantly improve performance.
    \item The downstream model can correctly predict the properties of the in-distribution objects, as if the distribution shift did not occur at all---this suggests that the representations of the in-distribution objects are largely unaffected by one object being out of distribution.
\end{itemize}

\begin{figure}
    \centering
    \includegraphics[height=0.37\textwidth]{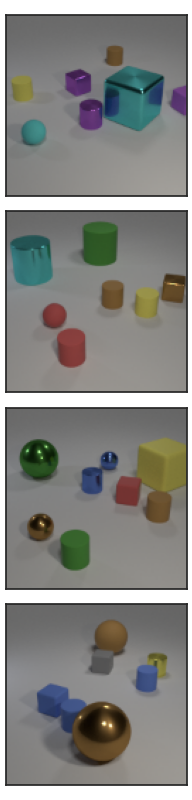}
    \hspace{2pt}
    \includegraphics[height=0.37\textwidth]{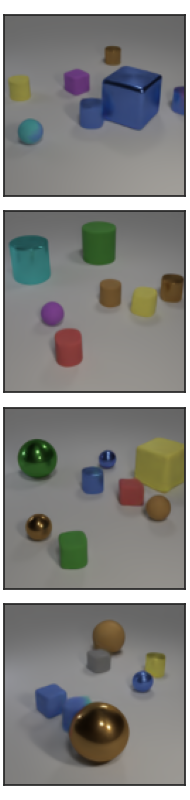}
    \hspace{-6pt}
    \includegraphics[height=0.37\textwidth]{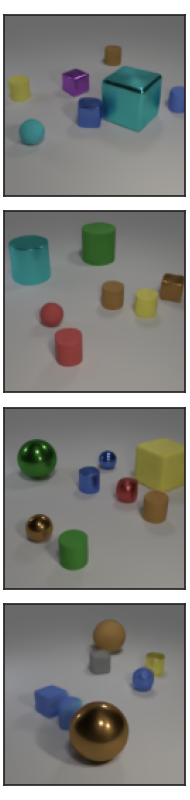}
    \hspace{-6pt}
    \includegraphics[height=0.37\textwidth]{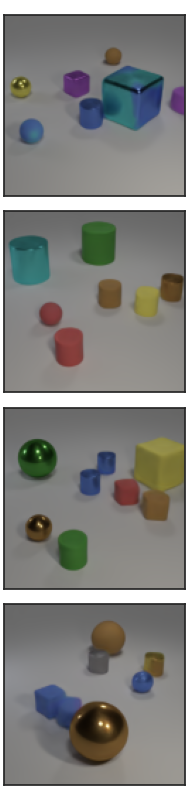}
    \hspace{-6pt}
    \includegraphics[height=0.37\textwidth]{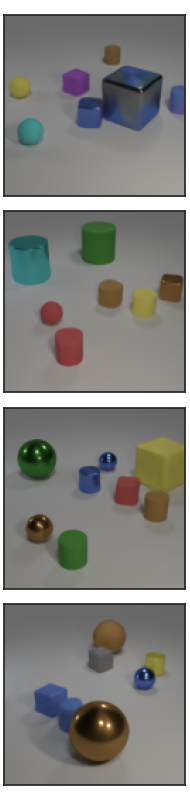}
    \hspace{-6pt}
    \includegraphics[height=0.37\textwidth]{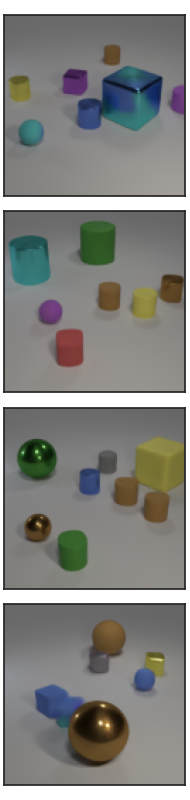}
    \hspace{-6pt}
    \includegraphics[height=0.37\textwidth]{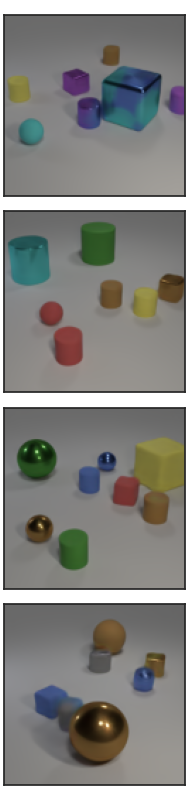}
    \hspace{-6pt}
    \includegraphics[height=0.37\textwidth]{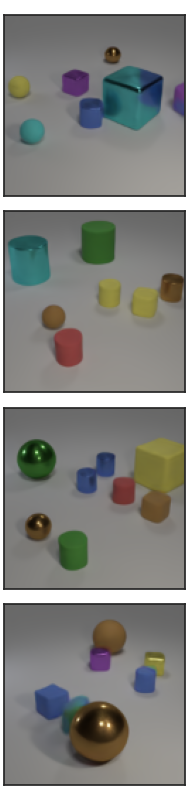}
    \hspace{-6pt}
    \includegraphics[height=0.37\textwidth]{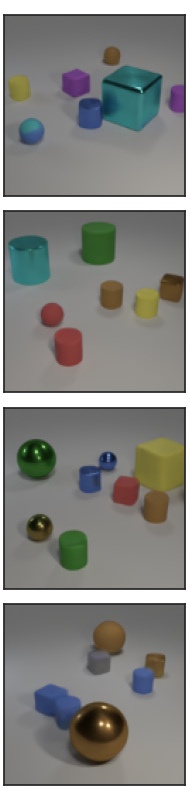}
    \hspace{-6pt}
    \includegraphics[height=0.37\textwidth]{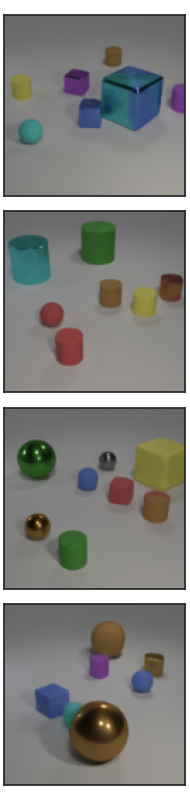}
    \hspace{-6pt}
    \includegraphics[height=0.37\textwidth]{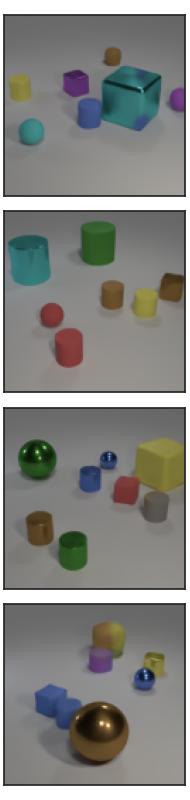}
    \caption[Input and reconstructions of 4 randomly selected CLEVR images with more than 6 objects by VAEs trained with up to 6 objects.]{Input and reconstructions of 4 randomly selected CLEVR images with more than 6 objects by VAEs trained with up to 6 objects. Leftmost column: input image. Other columns: reconstructions by the 10 VAEs in our study that were trained on CLEVR6. Note that some of the variability comes from sampling $\zb$ rather than using the posterior mean (which is typically used as representation in VAEs), but this does not explain that a few object properties are sometimes inferred incorrectly. On the other hand, it is worth noting that all VAEs seem to be able to produce coherent samples with more objects than observed during training.}
    \label{fig:baselines_all_numobjects_qualitative}
\end{figure}

\summary{Generalization with global shifts (hyp. 3)}
Our third hypothesis is that object-centric models are robust even when a distribution shifts affects global properties of the scene.
First, we look for a potential degradation of segmentation performance at test time when the data undergoes a distribution shift of this type.
Unlike for single-object shifts, for global shifts we observe varying results depending on the specific shift. For example, cropping and enlarging the scene often harms segmentation quality, 
while occlusions appear to have a minor effect.
Interestingly, when the number of objects increases at test time (in CLEVR), we see a relatively small drop in the ARI for object-centric models, and the reconstruction error increases less significantly than for the vanilla VAE models. This is presumably due to the strong inductive bias of object-centric models towards treating objects separately. However, what may seem surprising is that VAEs trained with up to 6 objects can generate rather coherent samples that contain more. This suggests that they might be capable of representing more objects than observable in the training distribution. In fact, given an image with more objects they seem to be able to reconstruct the correct number of objects, as shown in \cref{fig:baselines_all_numobjects_qualitative}. On the other hand, some of these objects have incorrect properties, suggesting that this extrapolation behaviour---where object-centric models have a clear advantage by design---is limited.
We present a quantitative analysis in Paper~\paperThree (\cref{chapter:objects}), and \cref{app:results} includes additional qualitative and quantitative results.

Regarding the informativeness and usefulness of the object representations under global distribution shifts, our findings are mostly negative: in all datasets and for all models, the prediction performance deteriorates significantly for most object properties (although to varying degrees), and cannot be recovered even by adjusting the downstream model post hoc on the OOD data.

\summary{Discussion on global shifts}
A point that merits discussion is that, while the definition of single-object shifts is clear---one object is OOD an the others are ID---it is not as straightforward to characterize \emph{global} distribution shifts.
\emph{Cropping} is essentially equivalent to an object-wise distribution shift that enlarges all objects. On the other hand, if we accept this interpretation, we also have to view this distribution shift as affecting the number of objects in the scene, their relative spatial arrangement (their centers are farther apart), and the fact that, on average, more objects than usual may be partially out of the field of view. Finally, note that cropping may be more simply seen as a shift in a global property of the scene such as camera view---from this point of view, it is more easily interpreted as global shift.
\emph{Occlusion}, another global distribution shift in our categorization, could also be viewed as adding a new object that is very different from any previously seen object, while leaving other objects untouched. From this perspective, it is not particularly surprising that object-centric models seem to exhibit at least partial robustness to this shift.
Finally, the \emph{number of objects} can easily be interpreted as a global property of the scene. However, considering one object at a time, there is in fact no distribution shift at all. This partly explains why some object-centric models tend to be relatively robust to this distribution shift. On the other hand, most of these models still have to compute visual features and, most importantly, some form of attention mask, from the entire image: this, following the argument on OOD2 generalization from \cref{chapter:contribs_disentanglement} and Papers~\paperOne and~\paperTwo, may explain the minor drop in performance.

\summary{We always do OOD2}
In fact, \emph{all} distribution shifts considered in this study bring the encoders out of distribution, and therefore correspond to the OOD2 setting in Papers~\paperOne and~\paperTwo. However, the fact that both segmentation and representation quality are affected to a widely varying degree suggests that there might be inductive biases in the models that make representation functions more robust to some shift types than to others. In particular, object-centric models do not appear to be significantly affected when only one object is out of distribution. In future work, it would be relevant to investigate this further by comparing the generalization of object-centric models and other models in a fairer way, and by testing a broader range of distribution shifts.

\summary{OOD1?}
On a related note, it would be interesting to consider the OOD1 scenario from Papers~\paperOne and~\paperTwo: given a class of representation functions trained on a large enough data distribution (such that they are never out of distribution at test time), do downstream models generalize better if the representations are object-centric? For example, we could learn representation functions on the full CLEVR dataset, and then train downstream models on a subset of CLEVR that contains few objects, or on another subset that does not contain red objects. We would then be able to investigate which type of representation shows more potential for the OOD generalization of downstream models.
Arguably, this is in fact the flavor of generalization that should be enabled by structured, or even causal, representations of the data. What good is a representation function that perfectly inverts the data generating process if we use it out of distribution, where it no longer works reliably?

\summary{Other shifts, e.g. correlations}
Another type of distribution shift we are not considering in this study is a shift in correlations between factors of variation.
One example is a correlation between factors within each object, as we have explored in \citet{trauble2021disentangled} for disentanglement learning. In the multi-object setting, a similar correlation has been investigated by \citet{greff2019multi}, who test IODINE on held-out images from CLEVR containing green spheres. It would be interesting to test the robustness of object-centric models to this shift, and study whether per-object disentanglement plays a role (see the generalization results in \citet{trauble2021disentangled}).
Another example that specifically applies to the multi-object setting is object co-occurrence. E.g., if all images either have zero or two red spheres, will an object-centric model represent two red spheres in the same slot? Should it not? Will it be able to generalize to images that only show one red sphere? This scenario, which includes the special case of object--part hierarchies, also brings up the question of how the notion of object should be defined \cite{greff2020binding}. 

\summary{Evaluation}
Beside distribution shifts, further considerations should be made on the different ways models and representations could be evaluated.
While a clear advantage of object-centric representations is the possibility of performing separate interventions on single objects,\footnote{%
This includes, for example, performing latent traversals on one slot at a time, or replacing two slots with a convex combination of the two to smoothly interpolate between them. 
}
this is in fact not feasible in a sensible, coherent manner in many existing object-centric models (e.g., MONet blends generated objects into a single image using alpha masks that are inferred directly from the input).
In addition, as discussed in \cref{sec:objects-vae-discussion}, although factor prediction is a reasonable proxy task, testing learned representations on more complex and practically relevant downstream tasks (e.g., relational question answering or reinforcement learning) would improve our understanding of models trained on multi-object data.

\summary{Simpler models might be interesting---yet another reason for better evaluations}
We have also observed that distributed representations (in our case, obtained by training standard VAEs) might in fact be more useful than expected, but we could not draw definitive conclusions because of the fundamental incompatibility of these representations with the set prediction task in our study.
Moreover, we found object-centric models to be rather sensitive to hyperparameters and random seeds, probably because they are structured models with discrete components. When the optimization goes wrong, and a model does not segment the scene in a meaningful way, the suitability for downstream tasks may decrease drastically. By contrast, although simpler models like VAEs lack some of the beneficial inductive biases of object-centric models, they are generally easier to optimize and require less hyperparameter tuning. 
Especially in light of the recent progress made by focusing on scaling and engineering rather than methodology, an open question is whether these simpler models may indeed be sufficient, given enough data and large enough models.
For these reasons, a more thorough comparison of object-centric and distributed representations is needed.
New downstream tasks should ideally make this comparison fairer and sounder, and further inform future research efforts.\looseness=-1

%%%%%%%%%%%%%%%%%%%%%%%%%%%%%%%%%%%%%%%%%%%%%%%%%%%%%%%%%%%%%%%%%%%%%%%%%%%%%%%%%%%%%%%%%%

\chapter{Paper~\paperOne: On the Transfer of Disentangled Representations in Realistic Settings}
\label{chapter:iclr2021}
\chaptermark{Paper \paperOne}

\pubinfo{%
    Andrea Dittadi\eqcontrib, \and 
    Frederik Tr{\"a}uble\eqcontrib, \and
    Francesco Locatello, \and
    Manuel W{\"u}thrich, \and
    Vaibhav Agrawal, \and
    Ole Winther, \and
    Stefan Bauer, \and
    Bernhard Sch{\"o}lkopf.\\[0.15em]
    \eqcontribfn {\small Equal contribution.}
}{%
    Published in \emph{International Conference on Learning Representations}, 2021.
}{%
    Learning meaningful representations that disentangle the underlying structure of the data generating process is considered to be of key importance in machine learning. While disentangled representations were found to be useful for diverse tasks such as abstract reasoning and fair classification, their scalability and real-world impact remain questionable.
    We introduce a new high-resolution dataset with 1M simulated images and over 1,800 annotated real-world images of the same setup. In contrast to previous work, this new dataset exhibits correlations, a complex underlying structure, and allows to evaluate transfer to unseen simulated and real-world settings where the encoder i) remains in distribution or ii) is out of distribution.
    We propose new architectures in order to scale disentangled representation learning to realistic high-resolution settings and conduct a large-scale empirical study of disentangled representations on this dataset. We observe that disentanglement is a good predictor for out-of-distribution (OOD) task performance.
}

\iffull
\section{Introduction}

\begin{wrapfigure}[12]{r}{0.5\linewidth}
\vspace{-10pt}
\includegraphics[height=0.2\textheight]{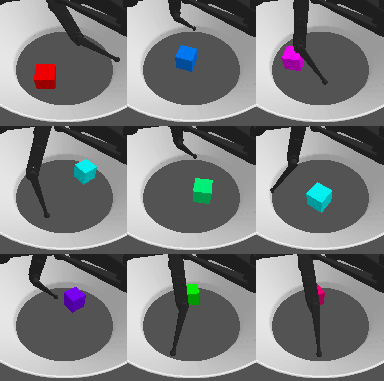}
\hfill
\includegraphics[height=0.2\textheight]{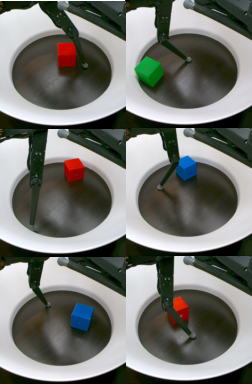}
\caption{Images from the simulated dataset (left) and the real-world setup (right).}
\label{fig:dataset}
\end{wrapfigure}
Disentangled representations hold the promise of generalization to unseen scenarios \citep{higgins2017darla}, increased interpretability \citep{adel2018discovering, higgins2018scan} and faster learning on downstream tasks \citep{locatello2019fairness,van2019disentangled}. However, most of the focus in learning disentangled representations has been on small synthetic datasets whose ground truth factors exhibit perfect independence by design. More realistic settings remain largely unexplored. We hypothesize that this is because real-world scenarios present several challenges that have not been extensively studied to date.
Important challenges are scaling (much higher resolution in observations and factors), occlusions, and correlation between factors. Consider, for instance, a robotic arm moving a cube: Here, the robot arm can occlude parts of the cube, and its end-effector position exhibits correlations with the cube's position and orientation, which might be problematic for common disentanglement learners \citep{trauble2021disentangled}. Another difficulty is that we typically have only limited access to ground truth labels in the real world, which requires robust frameworks for model selection when no or only weak labels are available.

The goal of this work is to provide a path towards disentangled representation learning in realistic settings. First, we argue that this requires a new dataset that captures the challenges mentioned above. We propose a dataset consisting of simulated observations from a scene where a robotic arm interacts with a cube in a stage (see \cref{fig:dataset}). This setting exhibits correlations and occlusions that are typical in real-world robotics.
Second, we show how to scale the architecture of disentanglement methods to perform well on this dataset. Third, we extensively analyze the usefulness of disentangled representations in terms of out-of-distribution downstream generalization, both in terms of held-out factors of variation and sim-to-real transfer. In fact, our dataset is based on the TriFinger robot from \citet{wuthrich2020trifinger}, which can be built to test the deployment of models in the real world. While the analysis in this paper focuses on the transfer and generalization of predictive models, we hope that our dataset may serve as a benchmark to explore the usefulness of disentangled representations in real-world control tasks.

The contributions of this paper can be summarized as follows:
\begin{itemize}

    \item We propose a new dataset for disentangled representation learning, containing 1M simulated high-resolution images from a robotic setup, with seven partly correlated factors of variation. Additionally, we provide a dataset of over 1,800 annotated images from the corresponding real-world setup that can be used for challenging sim-to-real transfer tasks.  These datasets are made publicly available.\footnote{\url{http://people.tuebingen.mpg.de/ei-datasets/iclr_transfer_paper/robot_finger_datasets.tar} (6.18 GB)}\looseness=-1
    
    \item We propose a new neural architecture to successfully scale VAE-based disentanglement learning approaches to complex datasets.
    
    \item We conduct a large-scale empirical study on generalization to various transfer scenarios on this challenging dataset. We train 1,080 models using state-of-the-art disentanglement methods and discover that disentanglement is a good predictor for out-of-distribution (OOD) performance of downstream tasks.
\end{itemize}

%===============================================================================

\section{Related Work}

\label{sec:related_work}
\paragraph{Disentanglement methods.} Most state-of-the-art methods for disentangled representation learning are based on the framework of variational autoencoders (VAEs) \citep{kingma2013auto,rezende2014stochastic}.
A (high-dimensional) observation $\xb$ is assumed to be generated according to the latent variable model $p_\theta(\xb|\zb)p(\zb)$ where the latent variables $\zb$ have a fixed prior $p(\zb)$.
The generative model $p_\theta(\xb|\zb)$ and the approximate posterior distribution $q_\phi(\zb|\xb)$ are typically parameterized by neural networks, which are optimized by maximizing the evidence lower bound (ELBO):
\begin{equation}
\mathcal{L}_{VAE} = \mathbb{E}_{q_\phi(\zb|\xb)} [ \log p_\theta (\xb | \zb) ]
- D_\mathrm{KL}( q_\phi(\zb|\xb) \| p (\zb) ) \leq  \log p(\xb)
\end{equation}
As the above objective does not enforce any structure on the latent space except for some similarity to $p (\zb)$, different regularization strategies have been proposed, along with evaluation metrics to gauge the disentanglement of the learned representations \citep{higgins2016beta, kim2018disentangling, burgess2018understanding, kumar2017variational, chen2018isolating, eastwood2018framework}. Recently, \citet[Theorem 1]{locatello2018challenging} showed that the purely unsupervised learning of disentangled representations is impossible. This limitation can be overcome without the need for explicitly labeled data by introducing weak labels \citep{locatello2020weakly, shu2019weakly}. Ideas related to disentangling the factors of variation date back to the non-linear ICA literature \citep{comon1994independent, hyvarinen1999nonlinear, bach2002kernel, jutten2003advances, hyvarinen2016unsupervised, hyvarinen2018nonlinear, gresele2019incomplete}. Recent work combines non-linear ICA with disentanglement \citep{khemakhem2020variational, sorrenson2020disentanglement, klindt2020towards}.

\paragraph{Evaluating disentangled representations.}
The \emph{BetaVAE}~\citep{higgins2016beta} and \emph{FactorVAE}~\citep{kim2018disentangling} scores measure disentanglement by performing an intervention on the factors of variation and predicting which factor was intervened on. The \emph{Mutual Information Gap (MIG)}~\citep{chen2018isolating}, \emph{Modularity}~\citep{ridgeway2018learning}, \emph{DCI Disentanglement}~\citep{eastwood2018framework} and \emph{SAP} scores~\citep{kumar2017variational} are based on matrices relating factors of variation and codes (e.g. pairwise mutual information, feature importance and predictability).

\paragraph{Datasets for disentanglement learning.} \emph{dSprites} \citep{higgins2016beta}, which consists of binary low-resolution 2D images of basic shapes, is one of the most commonly used synthetic datasets for disentanglement learning. \emph{Color-dSprites}, \emph{Noisy-dSprites}, and \emph{Scream-dSprites} are slightly more challenging variants of dSprites. The \emph{SmallNORB} dataset contains toy images rendered under different lighting conditions, elevations and azimuths \citep{lecun2004learning}. \emph{Cars3D} \citep{reed2015deep} exhibits different car models from \citet{fidler20123d} under different camera viewpoints. \emph{3dshapes} is a popular dataset of simple shapes in a 3D scene \citep{kim2018disentangling}. Finally, \citet{gondal2019transfer} proposed \emph{MPI3D}, containing images of physical 3D objects with seven factors of variation, such as object color, shape, size and position available in a simulated, simulated and highly realistic rendered simulated variant. Except MPI3D which has over 1M images, the size of the other datasets is limited with only $17,568$ to $737,280$ images. All of the above datasets exhibit perfect independence of all factors, the number of possible states is on the order of 1M or less, and due to their static setting they do not allow for dynamic downstream tasks such as reinforcement learning. In addition, except for {SmallNORB}, the image resolution is limited to 64x64 and there are no occlusions.

\paragraph{Other related work.}

\citet{locatello2020weakly} probed the out-of-distribution generalization of downstream tasks trained on disentangled representations. However, these representations are trained on the entire dataset. 
Generalization and transfer performance especially for representation learning has likewise been studied in \citet{dayan1993improving, muandet2013domain, heinze2017conditional,  rojas2018invariant, suter2019robustly, li2018deep, arjovsky2019invariant, krueger2020out, gowal2020achieving}.
For the role of disentanglement in causal representation learning we refer to the recent overview by \citet{scholkopf2021toward}.
\citet{trauble2021disentangled} systematically investigated the effects of correlations between factors of variation on disentangled representation learners.
Transfer of learned disentangled representations from simulation to the real world has been recently investigated by \citet{gondal2019transfer} on the MPI3D dataset, and previously by \citet{higgins2017darla} in the context of reinforcement learning.
Sim-to-real transfer is of major interest in the robotic learning community, because of limited data and supervision in the real world \citep{tobin2017domain,rusu2017sim, peng2018sim,james2019sim,yan2020close, andrychowicz2020learning}.

%===============================================================================

\section{Scaling Disentangled Representations to Complex Scenarios}
\label{sec:scaling}

\begin{table}
    \centering
    \caption{Factors of variation in the proposed dataset. Values are linearly spaced in the specified intervals. Joint angles are in radians, cube positions in meters.}
    \begin{tabular}{ll}
        \toprule
        \textbf{FoV} & \textbf{Values} \\
        \midrule
        Upper joint & 30 values in $[-0.65, +0.65]$ \\
        Middle joint & 30 values in $[-0.5, +0.5]$ \\
        Lower joint & 30 values in $[-0.8, +0.8]$ \\
        Cube position x & 30 values in $[-0.11, +0.11]$ \\
        Cube position y & 30 values in $[-0.11, +0.11]$ \\
        Cube rotation & 10 values in $[0^{\circ}, 81^{\circ}]$ \\
        Cube color hue & 12 values in $[0^{\circ}, 330^{\circ}]$ \\
        \bottomrule
    \end{tabular}
    \label{tab:dataset_fov}
\end{table}

\paragraph{A new challenging dataset.} Simulated images in our dataset are derived from the trifinger robot platform introduced by \citet{wuthrich2020trifinger}. The motivation for choosing this setting is that (1) it is challenging due to occlusions, correlations, and other difficulties encountered in robotic settings, (2) it requires modeling of fine details such as tip links at high resolutions, and (3) it corresponds to a robotic setup, so that learned representations can be used for control and reinforcement learning in simulation and in the real world. The scene comprises a robot finger with three joints that can be controlled to manipulate a cube in a bowl-shaped stage. \cref{fig:dataset} shows examples of scenes from our dataset.
The data is generated from 7 different factors of variation (FoV) listed in \cref{tab:dataset_fov}.
Unlike in previous datasets, not all FoVs are independent: The end-effector (the tip of the finger) can collide with the floor or the cube, resulting in infeasible combinations of the factors (see \cref{sec:app_dataset_corr}). We argue that such correlations are a key feature in real-world data that is not present in existing datasets. The high FoV resolution results in approximately 1.52 billion feasible states, but the dataset itself only contains one million of them (approximately 0.065\% of all possible FoV combinations), realistically rendered into $128 \times 128$ images.
Additionally, we recorded an annotated dataset under the same conditions in the real-world setup: we acquired 1,809 camera images from the same viewpoint and recorded the labels of the 7 underlying factors of variation. This dataset can be used for out-of-distribution evaluations, few-shot learning, and testing other sim-to-real aspects.

\paragraph{Model architecture.}
When scaling disentangled representation learning to more complex datasets, such as the one proposed here, one of the main bottlenecks in current VAE-based approaches is the flexibility of the encoder and decoder networks. In particular, using the architecture from \citet{locatello2018challenging}, none of the models we trained correctly captured all factors of variation or yielded high-quality reconstructions. While the increased image resolution already presents a challenge, the main practical issue in our new dataset is the level of detail that needs to be modeled. In particular, we identified the cube rotation and the lower joint position to be the factors of variation that were the hardest to capture. This is likely because these factors only produce relatively small changes in the image and hence the reconstruction error.\looseness=-1

To overcome these issues, we propose a deeper and wider neural architecture than those commonly used in the disentangled representation learning literature, where the encoder and decoder typically have 4 convolutional and 2 fully-connected layers.
Our encoder consists of a convolutional layer, 10 residual blocks, and 2 fully-connected layers. Some residual blocks are followed by 1x1 convolutions that change the number of channels, or by average pooling that downsamples the tensors by a factor of 2 along the spatial dimensions. Each residual block consists of two 3x3 convolutions with a leaky ReLU nonlinearity, and a learnable scalar gating mechanism \citep{bachlechner2020rezero}. Overall, the encoder has 23 convolutional layers and 2 fully connected layers. The decoder mirrors this architecture, with average pooling replaced by bilinear interpolation for upsampling. The total number of parameters is approximately 16.3M.
See \cref{sec:implementation_details} for further implementation details.

\paragraph{Experimental setup.} We perform a large-scale empirical study on the simulated dataset introduced above by training 1,080 $\beta$-VAE models.\footnote{Training these models requires approximately 2.8 GPU years on NVIDIA Tesla V100 PCIe.}
For further experimental details we refer the reader to \cref{sec:implementation_details}.
The hyperparameter sweep is defined as follows:\looseness=-1
\begin{itemize}
    \item We train the models using either unsupervised learning or weakly supervised learning \citep{locatello2020weakly}. In the weakly supervised case, a model is trained with pairs of images that differ in $k$ factors of variation. Here we fix $k=1$ as it was shown to lead to higher disentanglement by \citet{locatello2020weakly}. The dataset therefore consists of 500k pairs of images that differ in only one FoV.
    \item We vary the parameter $\beta$ in $\{1, 2, 4\}$, and use linear deterministic warm-up \citep{bowman2015generating,sonderby2016ladder} over the first $\{0, 10000, 50000\}$ training steps.
    \item The latent space dimensionality is in $\{10, 25, 50\}$.
    \item Half of the models are trained with additive noise in the input image. This choice is motivated by the fact that adding noise to the input of neural networks has been shown to be beneficial for out-of-distribution generalization \citep{sietsma1991creating,bishop1995training}.
    \item Each of the 108 resulting configurations is trained with 10 random seeds.
\end{itemize}

\begin{figure}
\centering
    \includegraphics[width=\linewidth]{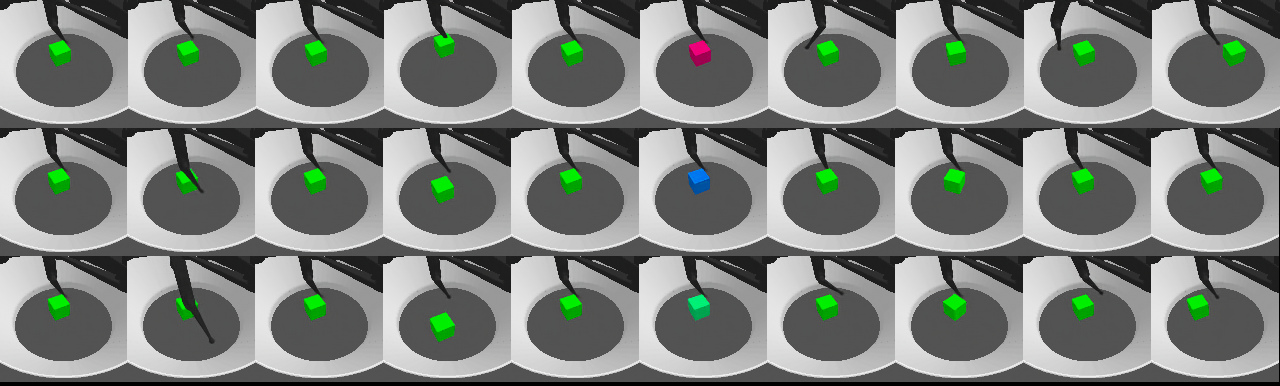}
\caption{Latent traversals of a trained model that perfectly disentangles the dataset's FoVs. In each column, all latent variables but one are fixed.}
\label{fig:data_and_traversals}
\end{figure}

\paragraph{Can we scale up disentanglement learning?}
Most of the trained VAEs in our empirical study fully capture all the elements of a scene, correctly model heavy occlusions, and generate detailed, high-quality samples and reconstructions (see \cref{sec:app_additional_modelviz}).
From visual inspections such as the latent traversals in \cref{fig:data_and_traversals}, we observe that many trained models fully disentangle the ground-truth factors of variation. This, however, appears to only be possible in the weakly supervised scenario. The fact that models trained without supervision learn entangled representations is in line with the impossibility result for the unsupervised learning of disentangled representations from \citet{locatello2018challenging}. Latent traversals from a selection of models with different degrees of disentanglement are presented in \cref{sec:app_additional_traversals}.
Interestingly, the high-disentanglement models seem to correct for correlations and interpolate infeasible states, i.e. the fingertip traverses through the cube or the floor.

\textbf{Summary:} The proposed architecture can scale disentanglement learning to more realistic settings, but a form of weak supervision is necessary to achieve high disentanglement.\looseness=-1

\paragraph{How useful are common disentanglement metrics in realistic scenarios?}

\begin{figure}
\centering
\begin{minipage}[c]{0.55\linewidth}
\centering
    \includegraphics[width=\textwidth]{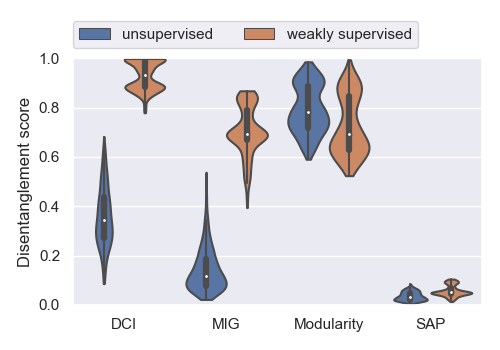}
\end{minipage}
\hspace{20pt}
\begin{minipage}[c]{0.3\linewidth}
\centering
    \includegraphics[width=\linewidth]{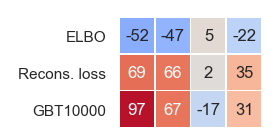}

    \includegraphics[width=\linewidth]{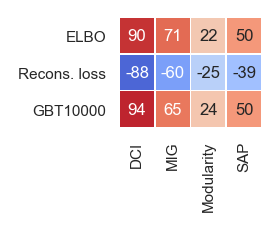}
\end{minipage}
\caption{\textbf{Left}: Disentanglement metrics aggregating all hyperparameters except for supervision type. \textbf{Right}: Spearman Rank correlations of the disentanglement metrics with the ELBO, the reconstruction loss, and the test error of a GBT classifier trained on 10,000 labelled data points. The upper rank correlations correspond to the unsupervised models and the lower ones to the weakly supervised models.
}
\label{fig:disentanglemet_metrics}
\end{figure}

The violin plot in \cref{fig:disentanglemet_metrics} (left) shows that DCI and MIG measure high disentanglement under weak supervision and lower disentanglement in the unsupervised setting. This is consistent with our qualitative conclusion from visual inspection of the models (\cref{sec:app_additional_traversals}) and with the aforementioned impossibility result.
Many of the models trained with weak supervision exhibit a very high DCI score (29\% of them have $>$99\% DCI, some of them up to 99.89\%).
SAP and Modularity appear to be ineffective at capturing disentanglement in this setting, as also observed by \citet{locatello2018challenging}. %We thus conclude that they are not useful for model selection in this context.
Finally, note that the BetaVAE and FactorVAE metrics are not straightforward to be evaluated on datasets that do not contain all possible combinations of factor values.
According to \cref{fig:disentanglemet_metrics} (right), DCI and MIG strongly correlate with test accuracy of GBT classifiers predicting the FoVs. In the weakly supervised setting, these metrics are strongly correlated with the ELBO (positively) and with the reconstruction loss (negatively). We illustrate these relationships in more detail in \cref{sec:app_scatter_plots}. Such correlations were also observed by \citet{locatello2020weakly} on significantly less complex datasets, and can be exploited for unsupervised model selection: these unsupervised metrics can be used as proxies for disentanglement metrics, which would require fully labeled data.

\textbf{Summary:} DCI and MIG appear to be useful disentanglement metrics in realistic scenarios, whereas other metrics seem to fall short of capturing disentanglement or can be difficult to compute. When using weak supervision, we can select disentangled models with unsupervised metrics.

%===============================================================================

\section{Framework for the Evaluation of OOD Generalization}
\label{sec:framework}

Previous work has focused on evaluating the usefulness of disentangled representations for various downstream tasks, such as predicting ground truth factors of variation, fair classification, and abstract reasoning. Here we propose a new framework for evaluating the out-of-distribution (OOD) generalization properties of representations. 
More specifically, we consider a downstream task---in our case, regression of ground truth factors---trained on a learned representation of the data, and evaluate the performance on a held-out test set. 
While the test set typically follows the same distribution as the training set (in-distribution generalization), we also consider test sets that follow a different distribution (out-of-distribution generalization). 
Our goal is to investigate to what extent, if at all, downstream tasks trained on disentangled representations exhibit a higher degree of OOD generalization than those trained on entangled representations.

Let $\dataset$ denote the training set for disentangled representation learning. To investigate OOD generalization, we train downstream regression models on a subset $\dataset_1 \subset \dataset$ to predict ground truth factor values from the learned representation computed by the encoder. 
We independently train one predictor per factor. 
We then test the regression models on a set $\dataset_2$ that differs distributionally from the training set $\dataset_1$, as it either contains images corresponding to held-out values of a chosen FoV (e.g. unseen object colors), or it consists of real-world images.
We now differentiate between two scenarios: (1) $\dataset_2 \subset \dataset$, i.e. the OOD test set is a subset of the dataset for representation learning; (2) $\dataset$ and $\dataset_2$ are disjoint and distributionally different. These two scenarios will be denoted by \emph{OOD1} and \emph{OOD2}, respectively.
For example, consider the case in which distributional shifts are based on one FoV: the color of the object. Then, we could define these datasets such that images in $\dataset$ always contain a red or blue object, and those in $\dataset_1 \subset \dataset$ always contain a red object. In the OOD1 scenario, images in $\dataset_2$ would always contain a blue object, whereas in the OOD2 case they would always contain an object that is neither red nor blue.

The regression models considered here are Gradient Boosted Trees (GBT), random forests, and MLPs with $\{1,2,3\}$ hidden layers. Since random forests exhibit a similar behavior to GBTs, and all MLPs yield similar results to each other, we choose GBTs and the 2-layer MLP as representative models and only report results for those.
To quantify prediction quality, we normalize the ground truth factor values to the range $[0,1]$, and compute the mean absolute error (MAE). Since the values are normalized, we can define our transfer metric as the average of the MAE over all factors (except for the FoV that is OOD).

%===============================================================================

\section{Benefits and Transfer of Structured Representations}
\label{sec:transfer}

\paragraph{Experimental setup.} We evaluate the transfer metric introduced in \cref{sec:framework} across all 1,080 trained models. To compute this metric, we train regression models to predict the ground truth factors of variation, and test them under distributional shift. 
We consider distributional shifts in terms of cube color or sim-to-real, and we do not evaluate downstream prediction of cube color.
We report scores for two different regression models: a Gradient Boosted Tree (GBT) and an MLP with 2 hidden layers of size 256.
In \cref{sec:implementation_details} we provide details on the datasets used in this section.

In the OOD1 setting, we have $\dataset_2 \subset \dataset$, hence the encoder is in-distribution: we are testing the predictor on representations of images that were in the training set of the representation learning algorithm. Therefore, we expect the representations to be meaningful. We consider three scenarios:
\begin{itemize}
    \item OOD1-A: The regression models are trained on 1 cube color (red) and evaluated on the remaining 7 colors.
    \item OOD1-B: The regression models are trained on 4 cube colors with high hue in the HSV space, and evaluated on 4 cube colors with low hue (extrapolation).
    \item OOD1-C: The regression models are again trained and evaluated on 4 cube colors, but the training and evaluation colors are alternating along the hue dimension (interpolation).
\end{itemize}
In the more challenging setting where even the encoder is out-of-distribution (OOD2, with $\dataset_2 \cap D = \varnothing$), we train the regression models on a subset of the training set $\dataset$ that includes all 8 cube colors, and we consider the two following scenarios:
\begin{itemize}
    \item OOD2-A: The regression models are evaluated on simulated data, on 4 cube colors that are out of the encoder's training distribution.
    \item OOD2-B: The regression models are evaluated on real-world images of the robotic setup, without any adaptation or fine-tuning.
\end{itemize}

\begin{figure}
    \begin{minipage}[c]{\linewidth}
        \centering
        \includegraphics[height=0.27\textheight]{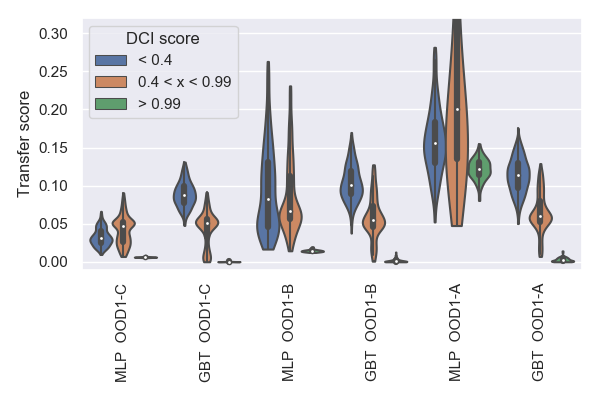}
        \hfill
        \includegraphics[height=0.24\textheight]{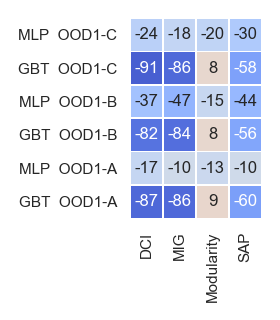}
    \end{minipage}
\caption{Higher disentanglement corresponds to better generalization across all OOD1 scenarios, as seen from the transfer scores (left). The transfer score is computed as the mean absolute prediction error of ground truth factor values (lower is better). This correlation is particularly evident in the GBT case, whereas MLPs appear to exhibit better OOD1 transfer with very high disentanglement only. These results are mirrored in the Spearman rank correlations between transfer scores and disentanglement metrics (right).}
\label{fig:evaulations_transfer_ood1}
\end{figure}

\paragraph{Is disentanglement correlated with OOD1 generalization?}
In \cref{fig:evaulations_transfer_ood1} we consistently observe a negative correlation between disentanglement and transfer error across all OOD1 settings. The correlation is mild when using MLPs, strong when using GBTs. This difference is expected, as GBTs have an axis-alignment bias whereas MLPs can---given enough data and capacity---disentangle an entangled representation more easily.
Our results therefore suggest that \textbf{highly disentangled representations are useful for generalizing out-of-distribution as long as the encoder remains in-distribution}.
This is in line with the correlation found by \citet{locatello2018challenging} between disentanglement and the GBT10000 metric. There, however, GBTs are tested on the same distribution as the training distribution, while here we test them under distributional shift.
Given that the computation of disentanglement scores requires labels, this is of little benefit in the unsupervised setting. However, it can be exploited in the weakly supervised setting, where disentanglement was shown to correlate with ELBO and reconstruction loss (\cref{sec:scaling}). Therefore, model selection for representations that transfer well in these scenarios is feasible based on the ELBO or reconstruction loss, when weak supervision is available.
Note that, in absolute terms, the OOD generalization error with encoder in-distribution (OOD1) is very low in the high-disentanglement case (the only exception being the MLP in the OOD1-C case, with the 1-7 color split, which seems to overfit). This suggests that disentangled representations can be useful in downstream tasks even when transferring out of the training distribution.

\textbf{Summary:} Disentanglement seems to be positively correlated with OOD generalization of downstream tasks, provided that the encoder remains in-distribution (OOD1).
Since in the weakly supervised case disentanglement correlates with the ELBO and the reconstruction loss, model selection can be performed using these metrics as proxies for disentanglement. These metrics have the advantage that they can be computed without labels, unlike disentanglement metrics.

\begin{figure}
    \begin{minipage}[c]{\linewidth}
        \centering
        \includegraphics[height=0.28\textheight]{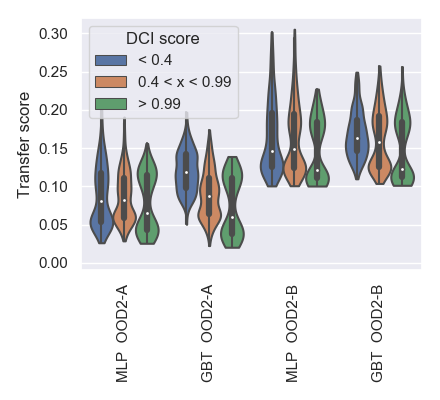}
        \hspace{1cm}
        \raisebox{0.24\height}{\includegraphics[height=0.19\textheight]{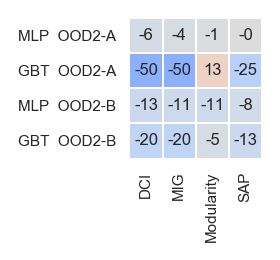}}
    \end{minipage}
\caption{Disentanglement affects generalization across the OOD2 scenarios only minimally as seen from transfer scores (left) and corresponding rank correlations with disentanglement metrics (right).}
\label{fig:evaulations_transfer_ood2}
\end{figure}
\begin{figure}
\vspace{15pt}
    \begin{minipage}[c]{\linewidth}
        \centering
        \includegraphics[height=0.29\textheight]{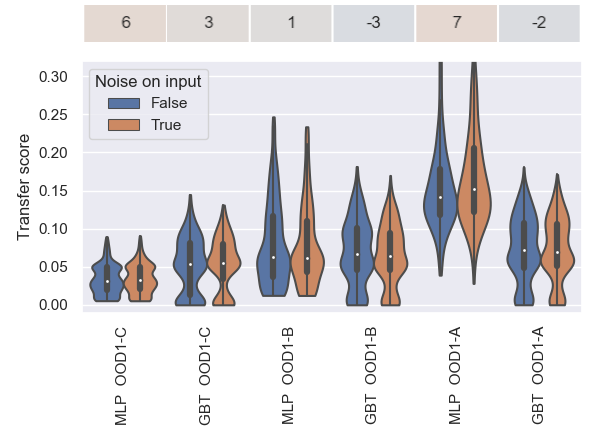}
        \includegraphics[height=0.29\textheight]{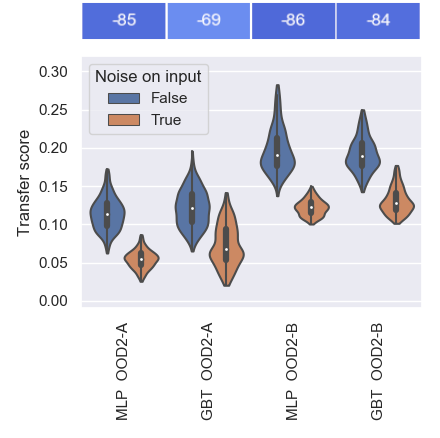}
    \end{minipage}
\caption{Noise improves generalization across the OOD2 scenarios and less so for the OOD1 scenarios as seen from the transfer scores. Top row: Spearman rank correlation coefficients between transfer metrics and presence of noise in the input.}
\label{fig:evaulations_transfer_w_noise}
\end{figure}

\paragraph{Is disentanglement correlated with OOD2 generalization?}
As we can observe in \cref{fig:evaulations_transfer_ood2}, the negative correlation between disentanglement and GBT transfer error is weaker when the encoder is out of distribution (OOD2). Nonetheless, we observe a non-negligible correlation for GBTs in the OOD2-A case, where we investigate out-of-distribution generalization along one FoV, with observations in $\dataset_2$ still generated from the same simulator.
In the OOD2-B setting, where the observations are taken from cameras in the corresponding real-world setting, the correlation between disentanglement and transfer performance appears to be minor at best.
This scenario can be considered a variant of zero-shot sim-to-real generalization.

\textbf{Summary:} Disentanglement has a minor effect on out-of-distribution generalization outside of the training distribution of the encoder (OOD2).

\paragraph{What else matters for OOD2 generalization?}
Results in \cref{fig:evaulations_transfer_w_noise} suggest that adding Gaussian noise to the input during training as described in \cref{sec:scaling} leads to significantly better OOD2 generalization, and has no effect on OOD1 generalization.
Adding noise to the input of neural networks is known to lead to better generalization \citep{sietsma1991creating,bishop1995training}. This is in agreement with our results, since OOD1 generalization does not require generalization of the encoder, while OOD2 does.
Interestingly, closer inspection reveals that the contribution of different factors of variation to the generalization error can vary widely.
See \cref{sec:app_additional_ood} for further
\begin{wrapfigure}[13]{r}{0.34\columnwidth}
    \centering
    \vspace{-5pt}
    \includegraphics[width=0.27\columnwidth]{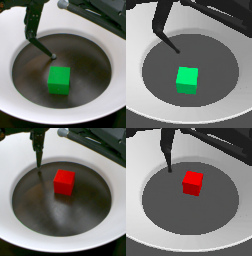}
    \caption{Zero-shot transfer of our models trained in simulation to real images. Left: input; right: reconstruction.}
    \label{fig:recostructions_real_closeup}
\end{wrapfigure}
details. In particular, with noisy input, the position of the cube is predicted accurately even in real-world images ($<$5\% mean absolute error on each axis). This is promising for robotics applications, where the true state of the joints is observable but inference of the cube position relies on object tracking methods.
\cref{fig:recostructions_real_closeup} shows an example of real-world inputs and reconstructions of their simulated equivalents.

\textbf{Summary:} Adding input noise during training appears to be significantly beneficial for OOD2 generalization, while having no effect when the encoder is kept in its training distribution (OOD1).

%===============================================================================

\section{Conclusion}
\label{sec:conclusion}

Despite the growing importance of the field and the potential societal impact in the medical
domain \citep{chartsias2018factorised} and fair decision making \citep{locatello2019fairness},  state-of-the-art approaches for learning disentangled representations have so far only been systematically evaluated on synthetic toy datasets. 
Here we introduced a new high-resolution dataset with 1M simulated images and over 1,800 annotated real-world images of the same setup. 
This dataset exhibits a number of challenges and features which are not present in previous datasets: it contains correlations between factors, occlusions, a complex underlying structure, and it allows for evaluation of transfer to unseen simulated and real-world settings. We proposed a new VAE architecture to scale disentangled representation learning to this realistic setting and conducted a large-scale empirical study of disentangled representations on this dataset.
We discovered that disentanglement is a good predictor of OOD generalization of downstream tasks and showed that, in the context of weak supervision, model selection for good OOD performance can be based on the ELBO or the reconstruction loss, which are accessible without explicit labels.
Our setting allows for studying a wide variety of interesting downstream tasks in the future, such as reinforcement learning or learning a dynamics model of the environment. Finally, we believe that in the future it will be important to take further steps in the direction of this paper by considering settings with even more complex structures and stronger correlations between factors.

\subsubsection*{Acknowledgements}
The authors thank Shruti Joshi and Felix Widmaier for their useful comments on the simulated setup, Anirudh Goyal for helpful discussions and comments, and CIFAR for the support. We thank the International Max
Planck Research School for Intelligent Systems (IMPRS-IS)
for supporting Frederik Tr{\"a}uble.

\fi

\chapter{Paper \paperTwo: The Role of Pretrained Representations for the {OOD} Generalization of Reinforcement Learning Agents}
\label{chapter:iclr2022}
\chaptermark{Paper \paperTwo}

\vspace{-20pt}
\pubinfo{%
    Andrea Dittadi\eqcontrib, \and 
    Frederik Tr{\"a}uble\eqcontrib, \and
    Manuel W{\"u}thrich, \and
    Felix Widmaier, \and
    Peter Gehler, \and
    Ole Winther, \and
    Francesco Locatello, \and
    Olivier Bachem, \and
    Bernhard Sch{\"o}lkopf,\and
    Stefan Bauer.\\[0.15em]
    \eqcontribfn {\small Equal contribution (the order was chosen at random and may be changed).}
}{%
    Published in \emph{International Conference on Learning Representations}, 2022.
}{%
    Building sample-efficient agents that generalize out-of-distribution (OOD) in real-world settings remains a fundamental unsolved problem on the path towards achieving higher-level cognition. 
    One particularly promising approach is to begin with low-dimensional, pretrained representations of our world, which should facilitate efficient downstream learning and  generalization.
    By training 240 representations and over 10,000 reinforcement learning (RL) policies on a simulated robotic setup, we evaluate to what extent different properties of pretrained VAE-based representations affect the OOD generalization of downstream agents. 
    We observe that many agents are surprisingly robust to realistic distribution shifts, including the challenging sim-to-real case. In addition, we find that the generalization performance of a simple downstream proxy task reliably predicts the generalization performance of our RL agents under a wide range of OOD settings. Such proxy tasks can thus be used to select pretrained representations that will lead to agents that generalize.
}

\iffull
\section{Introduction}
\label{sec:intro}

Robust out-of-distribution (OOD) generalization is one of the key open challenges in machine learning. This is particularly relevant for the deployment of ML models to the real world, where we need systems that generalize beyond the i.i.d.\ (independent and identically distributed) data setting \cite{scholkopf2021toward, djolonga2021robustness, koh2020wilds, barbu2019objectnet, azulay2019deep, roy2018effects, gulrajani2020search, hendrycks2019robustness, michaelis2019benchmarking, funkbenchmark}. One instance of such models are agents that learn by interacting with a training environment and we would like them to generalize to other environments with different statistics~\cite{zhang2018study, pfister2019learning, cobbe2019quantifying, ahmed2020causalworld, ke2021systematic}.
Consider the example of a robot with the task of moving a cube to a target position: Such an agent can easily fail as soon as some aspects of the environment differ from the training setup, e.g.\ the shape, color, and other object properties, or when transferring from simulation to real world. 

Humans do not suffer from these pitfalls when transferring learned skills beyond a narrow training domain, presumably because they represent visual sensory data in a concise and useful manner~\cite{marr1982vision,gordon1996s,lake2017building,anand2019unsupervised, spelke1990principles}.
Therefore, a particularly promising path is to base predictions and decisions on similar low-dimensional representations of our world~\citep{bengio2013representation,kaiser2019model,finn2016deep,barreto2017successor,dittadi2021planning,stooke2020decoupling,vinyals2019grandmaster}. The learned representation should facilitate efficient downstream learning~\cite{eslami2018neural, anand2019unsupervised,stooke2020decoupling,van2019disentangled} and exhibit better generalization~\cite{zhang2020learning,srinivas2020curl}.
Learning such a representation from scratch for every downstream task and every new variation would be inefficient. If we learned to juggle three balls, we should be able to generalize to oranges or apples without learning again from scratch. We could even do it with cherimoyas, a fruit that we might have never seen before. We can effectively reuse our generic representation of the world.

We thus consider deep learning agents trained from pretrained representations and ask the following questions: 
To what extent do they generalize under distribution shifts similar to those mentioned above? 
Do they generalize in different ways or to different degrees depending on the type of distribution shift, including sim-to-real?
Can we predict the OOD generalization of downstream agents from properties of the pretrained representations? 

\iffalse
\begin{wraptable}[13]{r}{5.3cm}
    \centering
    {\small
    \begin{tabular}{ll}
        \toprule
        \textbf{Metric class} & \textbf{Metrics}\\
        \midrule
        \textbf{Unsupervised} & ELBO, rec. loss\\
        \addlinespace[0.17cm]
        \textbf{FoV prediction} & MLP, GBT\\
        \addlinespace[0.17cm]
        \multirow{3}{*}{\textbf{Generalization}} & GS-OOD1,\\
         & GS-OOD2-sim,\\
         &  GS-OOD2-real\\
        \addlinespace[0.17cm]
        \multirow{2}{*}{\textbf{Disentanglement}} & DCI, MIG, SAP,\\
        & Modularity\\
        \bottomrule
    \end{tabular}
    }
    \caption{Overview of the 11 representation metrics discussed in this work.Respective acronyms and definitions discussed in section 2 and 4. \label{tab:metrics_table}}
\end{wraptable}
\fi
%
To answer the questions above, we need our experimental setting to be realistic, diverse, and challenging, but also controlled enough for the conclusions to be sound. 
We therefore base our study on the robot platform introduced by \citet{wuthrich2020trifinger}. The scene comprises a robot finger with three joints that can be controlled to manipulate a cube in a bowl-shaped stage. 
\citet{dittadi2021transfer} conveniently introduced a dataset of simulated and real-world images of this setup with ground-truth labels, which can be used to pretrain and evaluate representations. 
To train downstream agents, we adapted the simulated reinforcement learning benchmark CausalWorld from \citet{ahmed2020causalworld} that was developed for this platform.
Building upon these works, we design our experimental study as follows (see \cref{fig:experimental_setup}):
First, we pretrain representations from static simulated images of the setup and evaluate a collection of representation metrics. Following prior work \cite{watter2015embed,van2016stable,ghadirzadeh2017deep,nair2018visual,ha2018recurrent,eslami2018neural}, we focus on autoencoder-based representations. 
Then, we train downstream agents from this fixed representation on a set of environments. 
Finally, we investigate the zero-shot generalization of these agents to new environments that are out of the training distribution, including the real robot.

The goal of this work is to provide the first systematic and extensive account of the OOD generalization of downstream RL agents in a robotic setup, and how this is affected by characteristics of the upstream pretrained representations.
We summarize our contributions as follows:
\begin{itemize}
 \item We train 240 representations and 11,520 downstream policies,\footnote{Training the representations required approximately 0.62 GPU years on NVIDIA Tesla V100. Training and evaluating the downstream policies required about 86.8 CPU years on Intel Platinum 8175M.} and systematically investigate their performance under a diverse range of distribution shifts.\footnote{Additional results and videos are provided at \url{https://sites.google.com/view/ood-rl}.}
 \item We extensively analyze the relationship between the generalization of our RL agents and a substantial set of representation metrics. % (see \cref{tab:metrics_table}).
 \item Notably, we find that a specific representation metric that measures the generalization of a simple downstream proxy task reliably predicts the generalization of downstream RL agents under the broad spectrum of OOD settings considered here. This metric can thus be used to select pretrained representations that will lead to more robust downstream policies.
 \item In the most challenging of our OOD scenarios, we deploy a subset of the trained policies to the corresponding real-world robotic platform, and observe surprising zero-shot sim-to-real generalization without any fine-tuning or domain randomization.
 \end{itemize}

\begin{figure}
    \centering
    \includegraphics[width=0.93\linewidth]{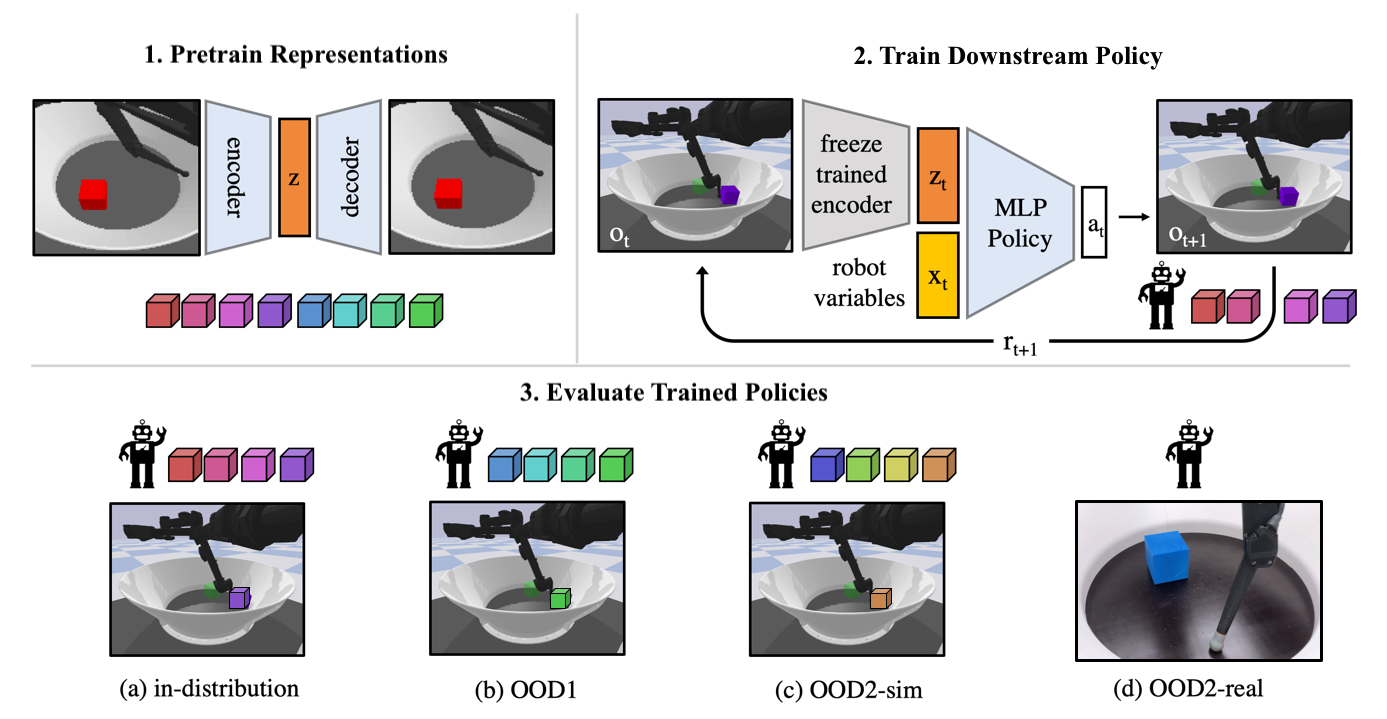}
    \caption[Overview of our experimental setup for investigating out-of-distribution generalization in downstream tasks.]{\textbf{Overview of our experimental setup for investigating out-of-distribution generalization in downstream tasks.} (1)~We train 240 $\beta$-VAEs on the robotic dataset from \citet{dittadi2021transfer}. (2)~We then train downstream policies to solve \textit{object reaching} or \textit{pushing}, using multiple random RL seeds per VAE. The input to a policy consists of the output of a pretrained encoder and additional task-related observable variables. Crucially, the policy is only trained on a subset of the cube colors from the pretraining dataset. (3)~Finally, we evaluate these policies on their respective tasks in four different scenarios: (a)~in-distribution, i.e. with cube colors used in policy training; (b)~OOD1, i.e. with cube colours previously seen by the encoder but OOD for the policy; (c)~OOD2-sim, having cube colours also OOD to the encoder; (d)~sim-to-real zero-shot on the real-world setup.}
    \label{fig:experimental_setup}
\end{figure}

\section{Background}
\label{sec:background}
In this section, we provide relevant background on the methods for representation learning and reinforcement learning, and on the robotic setup to evaluate out-of-distribution generalization.

\paragraph{Variational autoencoders.}
VAEs \cite{kingma2013auto,rezende2014stochastic} are a framework for optimizing a latent variable model $\ptheta(\xb) = \int_\zb \ptheta(\xb \given \zb) p(\zb) d\zb$ with parameters $\theta$, typically with a fixed prior $p(\zb)=\mathcal{N}(\zb; \mathbf{0}, \mathbf{I})$, using amortized stochastic variational inference. A variational distribution $\qphi(\zb \given \xb)$ with parameters $\phi$ approximates the intractable posterior $\ptheta(\zb \given \xb)$. The approximate posterior and generative model, typically called encoder and decoder and parameterized by neural networks, are jointly optimized by maximizing a lower bound to the log likelihood (the ELBO):
\begin{align}
    \log \ptheta(\xb) 
    &\geq \E_{\qphi(\zb \given \xb)}  \left[ \log \ptheta(\xb \given \zb) \right] - \kl\left( \qphi(\zb \given \xb) \| p(\zb)\right)
    = \mathcal{L}^{ELBO}_{\theta,\phi} (\xb) \ .
\end{align}
In $\beta$-VAEs, the KL term is modulated by a factor $\beta$ to enforce a more structured latent space \cite{higgins2016beta,burgess2018understanding}.
While VAEs are typically trained without supervision, we also employ a form of weak supervision \cite{locatello2020weakly} that encourages disentanglement.

\paragraph{Reinforcement learning.}
A Reinforcement Learning (RL) problem is typically modeled as a Partially Observable Markov Decision Process (POMDP) defined as a tuple $(S, A, T, R, \Omega, O, \gamma, \rho_{0}, H)$ with states $s \in S$, actions $a \in A$ and observations $o \in \Omega$ determined by the state and action of the environment $O(o|s,a)$. $T(s_{t+1}|s_{t},a_{t})$ is the transition probability distribution function, $R(s_t,a_t)$ is the reward function, $\gamma$ is the discount factor, $\rho_{0}(s)$ is the initial state distribution at the beginning of each episode, and $H$ is the time horizon per episode.
The objective in RL is to learn a policy $\pi: S \times A \rightarrow [0, 1]$, typically parameterized by a neural network, that maximizes the total discounted expected reward $J(\pi)=\E \big[ \sum_{t=0}^{H}\gamma^t R(s_t, a_t) \big]$.
There is a broad range of model-free learning algorithms to find $\pi^*$ by policy gradient optimization or by learning value functions while trading off exploration and exploitation \cite{haarnoja2018soft, schulman2017proximal, sutton1999policy, schulman2015high, schulman2015trust, silver2014deterministic, fujimoto2018addressing}. Here, we optimize the objective above with \emph{Soft Actor Critic} (SAC), an off-policy method that simultaneously maximizes the expected reward and the entropy $H(\pi(\cdot|s_t))$, and is widely used in control tasks due to its sample efficiency \cite{haarnoja2018soft}.

\paragraph{A robotic setup to evaluate out-of-distribution generalization.} 
Our study is based on a real robot platform where a robotic finger with three joints manipulates a cube in a bowl-shaped stage \cite{wuthrich2020trifinger}. We pretrain representations on a labeled dataset introduced by \citet{dittadi2021transfer} which consists of simulated and real-world images of this setup. This dataset has 7 underlying factors of variation (FoV): angles of the three joints, and position (x and y), orientation, and color of the cube. 
Some of these factors are correlated \cite{dittadi2021transfer}, which may be problematic for representation learners, especially in the context of disentanglement \cite{trauble2021disentangled,chen2021boxhead}.
After training the representations, we train downstream agents and evaluate their generalization on an adapted version of the simulated CausalWorld benchmark \cite{ahmed2020causalworld} that was developed for the same setup. Finally, we test sim-to-real generalization on the real robot.

Our experimental setup, illustrated in \cref{fig:experimental_setup}, allows us to systematically investigate a broad range of out-of-distribution scenarios in a controlled way.
We pretrain our representations from this simulated dataset that covers 8 distinct cube colors. We then train an agent from this fixed representation on a subset of the cube colors, and evaluate it (1) on the same colors (this is the typical scenario in RL), (2) on the held-out cube colors that are still known to the encoder, or (3) OOD w.r.t. the encoder's training distribution, e.g. on novel colors and shapes or on the real world.

We closely follow the framework for measuring OOD generalization proposed by \citet{dittadi2021transfer}. In this framework, a representation is initially learned on a training set $\dataset$, and a simple downstream model is trained on a subset $\dataset_1 \subset \dataset$ to predict the ground-truth factors from the learned representation. 
Generalization is then evaluated by testing the downstream model on a set $\dataset_2$ that differs distributionally from $\dataset_1$, e.g. containing images corresponding to held-out values of a chosen factor of variation (FoV).
\citet{dittadi2021transfer} consider two flavors of OOD generalization depending on the choice of $\dataset_2$:
First, the case when $\dataset_2 \subset \dataset$, i.e. the OOD test set is a subset of the dataset for representation learning. This is denoted by \textbf{OOD1} and corresponds to the scenario~(2) from the previous paragraph. In the other scenario, referred to as \textbf{OOD2}, $\dataset$ and $\dataset_2$ are disjoint and distributionally different. This even stronger OOD shift corresponds to case~(3) above.
The generalization score for $\dataset_2$ is then measured by the (normalized) mean absolute prediction error across all FoVs except for the one that is OOD. Following \citet{dittadi2021transfer}, we use a simple 2-layer Multi-Layer Perceptron (MLP) for downstream factor prediction, we train one MLP for each FoV, and report the \textit{negative} error.
This simple and cheap generalization metric could serve as a convenient proxy for the generalization of more expensive downstream tasks.
We refer to these generalization scores as GS-OOD1, GS-OOD2-sim, and GS-OOD2-real depending on the scenario.

The focus of \citet{dittadi2021transfer} was to scale VAE-based approaches to more realistic scenarios and study the generalization of these simple downstream tasks, with a particular emphasis on disentanglement. Building upon their contributions, we can leverage the broader potential of this robotic setup with many more OOD2 scenarios to study our research questions: To what extent can agents generalize under distribution shift? Do they generalize in different ways depending on the type of shift (including sim-to-real)? Can we predict the OOD generalization of downstream agents from properties of the pretrained representations such as the GS metrics from \citet{dittadi2021transfer}?

\section{Study design}
\label{sec:study_design}
\paragraph{Robotic setup.} Our setup is based on TriFinger \cite{wuthrich2020trifinger} and consists of a robotic finger with three joints that can be controlled to manipulate an object (e.g. a cube) in a bowl-shaped stage.
The agent receives a camera observation consistent with the images in \citet{dittadi2021transfer} and outputs a three-dimensional action.  
During training, which always happens in simulation, the agent only observes a cube of four possible colors, randomly sampled at every episode (see \cref{fig:experimental_setup}, step 2).\looseness=-1

\paragraph{Distribution shifts.} After training, we evaluate these agents in 7 environments: (1) the training environment, which is the typical setting in RL, (2) the OOD1 setting with cube colors that are OOD for the agent but still in-distribution for the encoder, (3) the more challenging OOD2-sim setting where the colors are also OOD for the encoder, (4-6) the OOD2 settings where the object colors are as in the 3 previous settings but the cube is replaced by a sphere (a previously unseen shape), (7) the OOD2-real setting, where we evaluate zero-shot sim-to-real transfer on the real robotic platform.\looseness=-1

\paragraph{Tasks.}
We begin our study with the \textit{object reaching} downstream control task, where the agent has to reach an object placed at an \textit{arbitrary} random position in the arena. This is significantly more challenging than directly predicting the ground-truth factors, as the agent has to learn to reach the cube by acting on the joints, with a scalar reward as the only learning signal.
Consequently, the compute required to learn this task is about 1,000 times greater than in the simple factor prediction case.
We additionally include in our study a \textit{pushing} task which consists of pushing an object to a goal position that is sampled at each episode. Learning this task takes one order of magnitude more compute than \textit{object reaching}, likely due to the complex rigid-body dynamics and object interactions. To the best of our knowledge, this is the most challenging manipulation task that is currently feasible on our setup. \citet{ahmed2020causalworld} report solving a similar pushing task, but require the full ground-truth state to be observable.

\paragraph{Training the RL agents.}
The inputs at time $t$ are the camera observation $o_t$ and a vector of observable variables $x_t$ containing the joint angles and velocities, as well as the target object position in \textit{pushing}. We then feed the camera observation $o_t$ into an encoder $e$ that was pretrained on the dataset in \citet{dittadi2021transfer}. The result is concatenated with $x_t$, yielding a state vector $s_t = [x_t, e(o_t)]$.
We then use SAC to train the policy with $s_t$ as input. The policy, value, and Q networks are implemented as MLPs with 2 hidden layers of size 256. When training the policies, we keep the encoder frozen.\looseness=-1

\paragraph{Model sweep.}
To shed light on the research questions outlined in the previous sections, we perform a large-scale study in which we train 240 representation models and 11,520 downstream policies, as described below. See \cref{app:implementation_details} for further implementation details.
\begin{itemize}
    \item We train 120 $\beta$-VAEs \cite{higgins2016beta} and 120 Ada-GVAEs \cite{locatello2020weakly} with a subset of the hyperparameter configurations and neural architecture from \citet{dittadi2021transfer}. Specifically, we consider $\beta \in \{1, 2, 4\}$, $\beta$ annealing over $\{0, 50000\}$ steps, with and without input noise, and 10 random seeds per configuration. The latent space size is fixed to $10$ following prior work \cite{kim2018disentangling,chen2018isolating,locatello2020weakly,trauble2021disentangled}.\looseness=-1
    
    \item For \textit{object reaching}, we train 20 downstream policies (varying random seed) for each of the 240 VAEs. The resulting 4,800 policies are trained for 400k steps (approximately 2,400 episodes).
    
    \item Since \textit{pushing} takes substantially longer to train, we limit the number of policies trained on this task: We choose a subset of 96 VAEs corresponding to only 4 seeds, and then use 10 seeds per representation. The resulting 960 policies are trained for 3M steps (about 9,000 episodes).
    
    \item Finally, for both tasks we also investigate the role of regularization on the policy. More specifically, we repeat the two training sweeps from above (5,760 policies), with the difference that now the policies are trained with L1 regularization on the first layer. 
    
\end{itemize}

\paragraph{Limitations of our study.}
Although we aim to provide a sound and extensive empirical study, such studies are inevitably computationally demanding. Thus, we found it necessary to make certain design choices. For each of these choices, we attempted to follow common practice, in order to maintain our study as relevant, general, and useful as possible. 
One such decision is that of focusing on autoencoder-based representations. To answer our questions on the effect of upstream representations on the generalization of downstream policies, we need a diverse range of representations. 
How these representations are obtained is not directly relevant to answer our research question. 
Following \citet{dittadi2021transfer}, we chose to focus on $\beta$-VAE and Ada-GVAE models, as they were shown to provide a broad set of representations, including fully disentangled ones. Although we conjecture that other classes of representation learning algorithms should generally reveal similar trends as those found in our study, this is undoubtedly an interesting extension.
As for the RL algorithm used in this work, SAC is known to be a particularly sample-efficient model-free RL method that is a popular choice in robotics \cite{haarnoja2018learning, kiran2021deep, singh2019end}.
Extensive results on pushing from ground-truth features on the same setup in \citet{ahmed2020causalworld} indicate that methods like TD3 \cite{fujimoto2018addressing} or PPO \cite{schulman2017proximal} perform very similarly to SAC under the same reward structure and observation space. Thus, we expect the results of our study to hold beyond SAC.
Another interesting direction is the study of additional regularization schemes on the policy network, an aspect that is often overlooked in RL. 
We expect the potential insights from extending the study along these axes to not justify the additional compute costs and corresponding carbon footprint. However, with improving efficiency and decreasing costs, we believe these could become worthwhile validation experiments in the future.

\section{Results}
\label{sec:results}
We discuss our results in three parts: In \cref{subsec:results_training_and_indistrib_performance}, we present the training results of our large-scale sweep, and how policy regularization and different properties of the pretrained representations affect in-distribution reward. 
\Cref{subsec:results_ood_generalization_simulation} gives an extensive account of which metrics of the pretrained representations predict OOD generalization of the agents in simulated environments. Finally, in \cref{subsec:results_ood_generalization_real_robot} we perform a similar evaluation on the real robot, in a zero-shot sim-to-real scenario.

\begin{figure}
    \centering
    \begin{minipage}{0.37\textwidth}
        \centering
        \includegraphics[width=\linewidth]{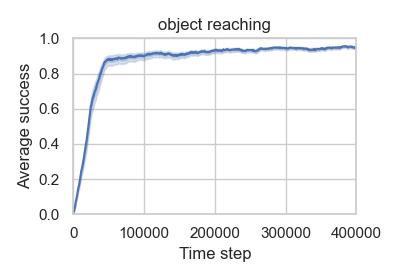}
        
        \includegraphics[width=\linewidth]{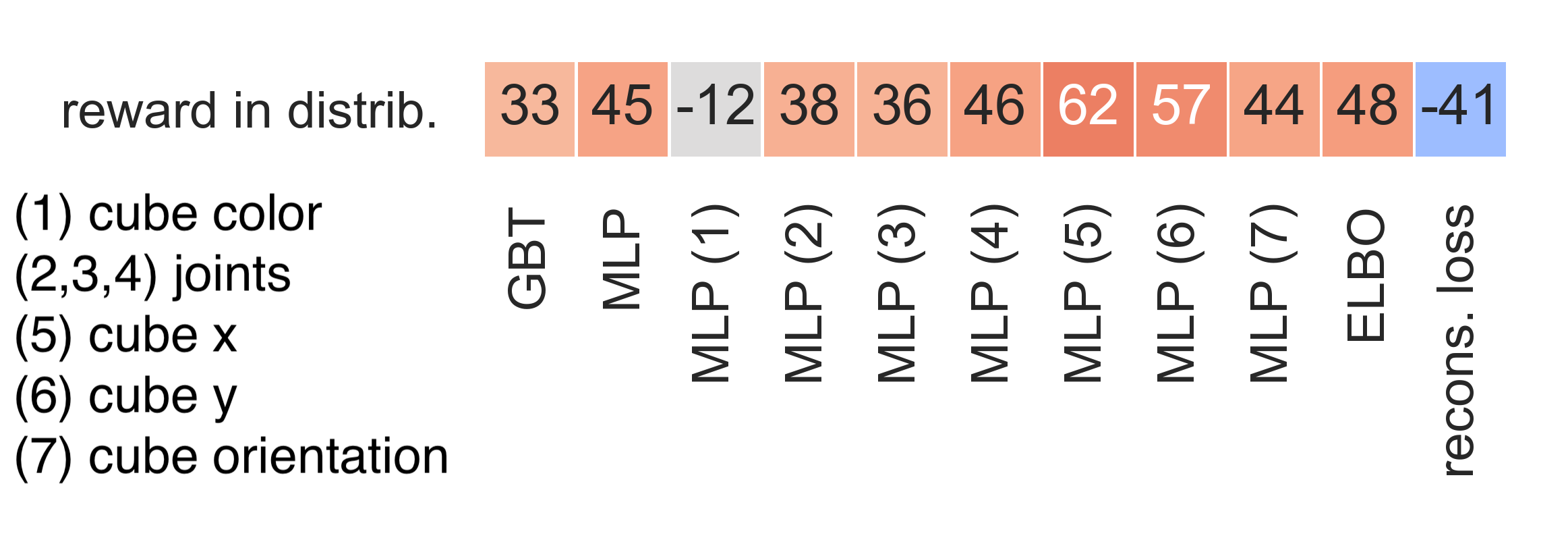}
    \end{minipage}
    \qquad
    \begin{minipage}{0.37\textwidth}
        \centering
        \includegraphics[width=\linewidth]{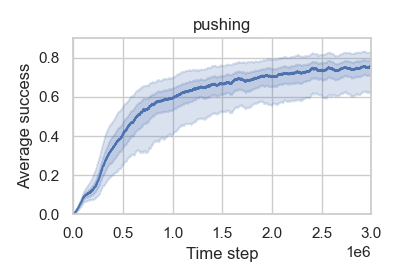}
        
        \includegraphics[width=\linewidth]{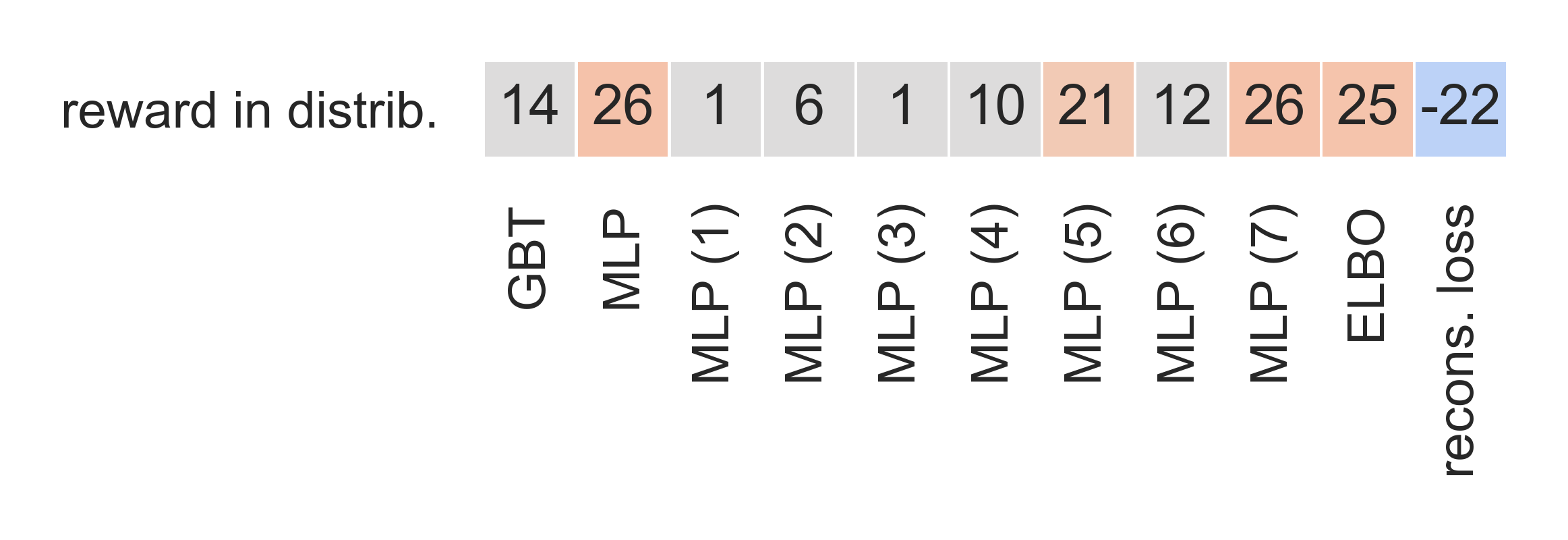}
    \end{minipage}
    \caption{Top: Average training success, aggregated over \emph{all} policies from the sweep (median, quartiles, 5th/95th percentiles). Bottom: Rank correlations between representation metrics and in-distribution reward (evaluated when the policies are fully trained), in the case without regularization. Correlations are color-coded in red (positive) or blue (negative) when statistically significant (p<0.05), otherwise they are gray.
    }
    \label{fig:results_on_training_env}
\end{figure}

\subsection{Results in the training environment}
\label{subsec:results_training_and_indistrib_performance}
\cref{fig:results_on_training_env} shows the training curves of all policies for \textit{object reaching} and \textit{pushing} in terms of the task-specific success metric.
Here we use success metrics for interpretability, as their range is always $[0,1]$.
In \textit{object reaching}, the success metric indicates progress from the initial end effector position to the optimal distance from the center of the cube. It is 0 if the final distance is not smaller than the initial distance, and 1 if the end effector is touching the center of a face of the cube.
In \textit{pushing}, the success metric is defined as the volumetric overlap of the cube with the goal cube, and the task can be visually considered solved with a score around 80\%.

From the training curves we can conclude that both tasks can be consistently solved from pixels using pretrained representations. In particular, all policies on \textit{object reaching} attain almost perfect scores.
Unsurprisingly, the more complex \textit{pushing} task requires significantly more training, and the variance across policies is larger. Nonetheless, almost all policies learn to solve the task satisfactorily.

%\begin{wrapfigure}[9]{r}{5.1cm}
%    \centering
%    \vspace{-17pt}
%    \includegraphics[width=\linewidth]{figures/iclr2022/fig3}
%    % \vspace*{-5mm}
%    \caption{Correlations between training (in distrib.) and OOD rewards (p<0.05).}
%    \label{fig:OOD_correlations_reward_trasfer}
%\end{wrapfigure}
%
To investigate the effect of representations on the training reward, we now compute its Spearman rank correlations with various supervised and unsupervised metrics of the representations (\cref{fig:results_on_training_env} bottom). By training reward here we mean the average reward of a fully trained policy over 200 episodes in the training environment (see \cref{app:implementation_details}).
On \textit{object reaching}, the final reward correlates with the ELBO and the reconstruction loss. 
A simple supervised metric to evaluate a representation is how well a small downstream model can predict the ground-truth factors of variation. Following \citet{dittadi2021transfer}, we use the MLP10000 and GBT10000 metrics (simply MLP and GBT in the following), where MLPs and Gradient Boosted Trees (GBTs) are trained to predict the FoVs from 10,000 samples. The training reward correlates with these metrics as well, especially with the MLP accuracy. This is not entirely surprising: if an MLP can predict the FoVs from the representations, our policies using the same architecture could in principle retrieve the FoVs relevant for the task.
Interestingly, the correlation with the overall MLP metric mostly stems from the cube pose FoVs, i.e. those that are not included in the ground-truth state $x_t$.
These results suggest that these metrics can be used to select good representations for downstream RL.
On the more challenging task of \textit{pushing}, the correlations are milder but most of them are still statistically significant.

\paragraph{Summary.} Both tasks can be consistently solved from pixels using pretrained representations. Unsupervised (ELBO, reconstruction loss) and supervised (ground-truth factor prediction) in-distribution metrics of the representations are correlated with reward in the training environment.

\subsection{Out-of-distribution generalization in simulation}
\label{subsec:results_ood_generalization_simulation}

\paragraph{In- and out-of-distribution rewards.}
After training, the in-distribution reward correlates with OOD1 performance on both tasks (especially with regularization), but not with OOD2 performance (see \cref{fig:OOD_correlations_reward_trasfer}).
Moreover, rewards in OOD1 and OOD2 environments are moderately correlated across tasks and regularization settings.
\begin{figure}
    \centering
    \includegraphics[width=0.38\linewidth]{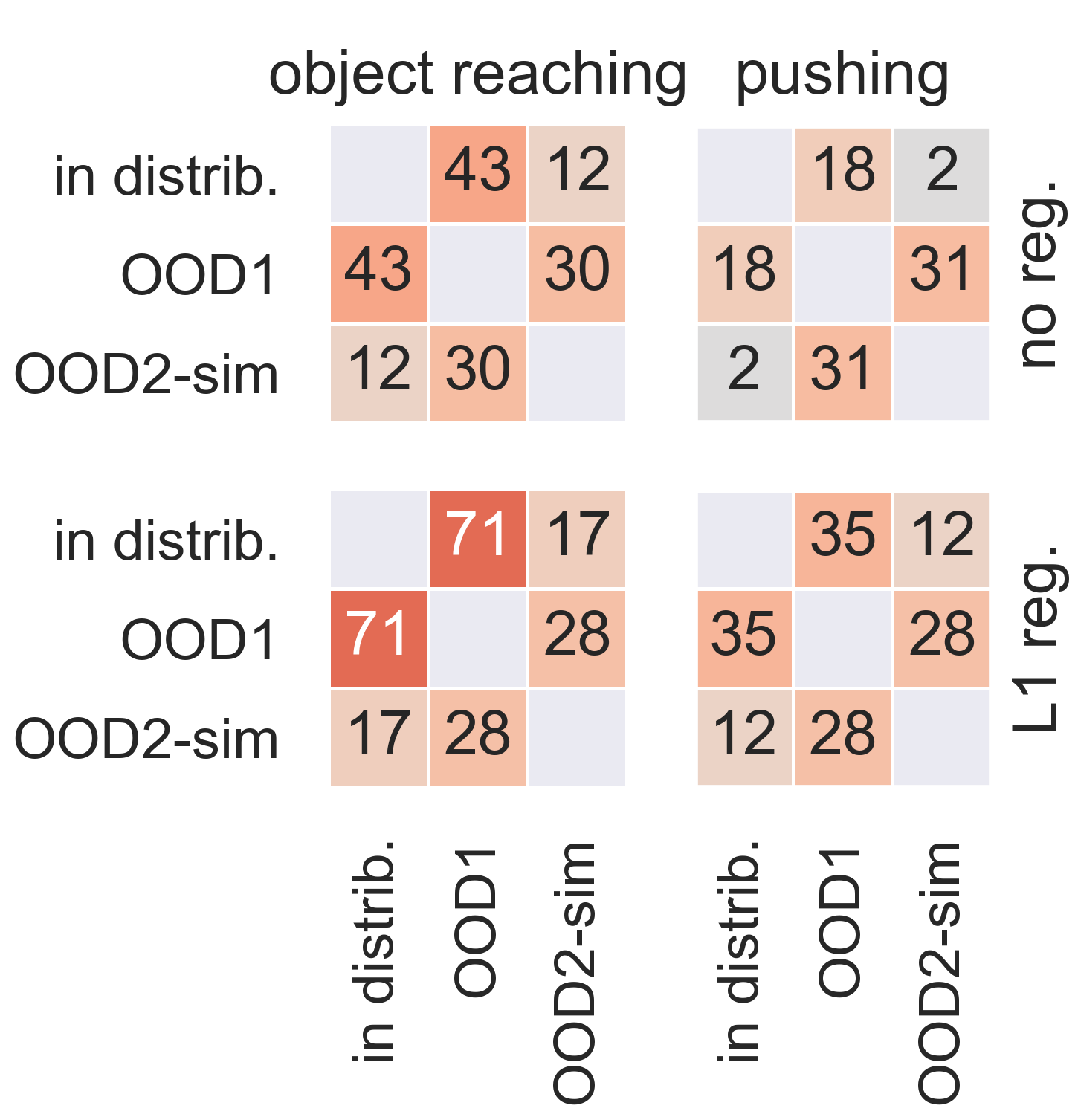}
    \caption{Correlations between training (in distrib.) and OOD rewards (p<0.05).}
    \label{fig:OOD_correlations_reward_trasfer}
\end{figure}

\paragraph{Unsupervised metrics and informativeness.}
In \cref{fig:sim_world_OOD_rank_correlations_reaching_no_reg} (left) we assess the relation between OOD reward and in-distribution metrics (ELBO, reconstruction loss, MLP, and GBT). 
Both ELBO and reconstruction loss exhibit a correlation with OOD1 reward, but not with OOD2 reward. These unsupervised metrics can thus be useful for selecting representations that will lead to more robust downstream RL tasks, as long as the encoder is in-distribution.
While the GBT score is not correlated with reward under distribution shift, we observe a significant correlation between OOD1 reward and the MLP score, which measures downstream factor prediction accuracy of an MLP with the same architecture as the one parameterizing the policies. As in \cref{subsec:results_training_and_indistrib_performance}, we further investigate the source of this correlation, and find it in the pose parameters of the cube. 
Correlations in the OOD2 setting are much weaker, thus we conclude that these metrics do not appear helpful for model selection in this case. 
Our results on \textit{pushing} confirm these conclusions although correlations are generally weaker, presumably due to the more complicated nature of this task. An extensive discussion is provided in \cref{app:additional_results_ood_simulation}.

\begin{figure}
    \begin{minipage}[t]{0.335\linewidth}
    \vspace{0pt}
    \includegraphics[width=\linewidth]{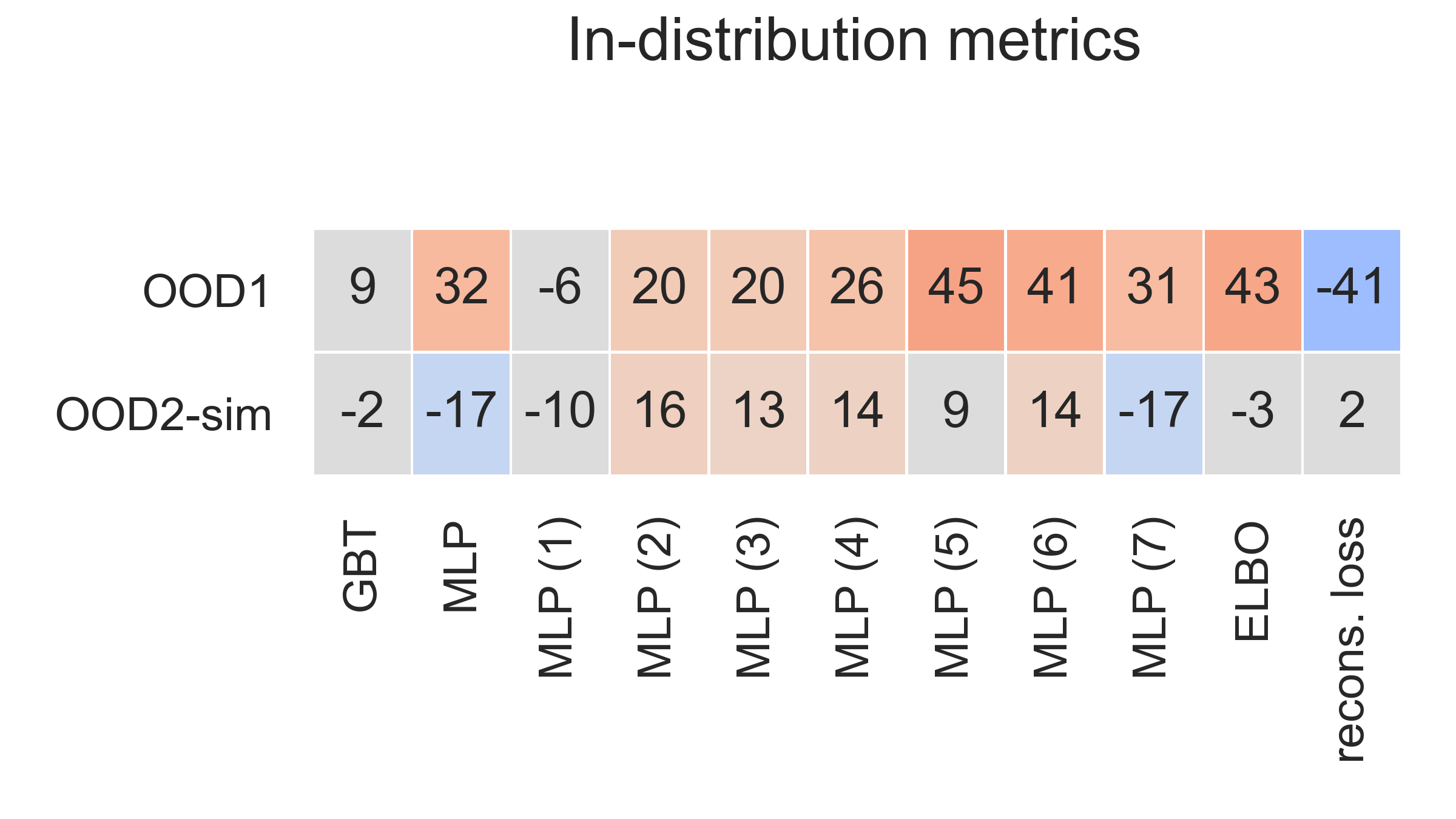}
    \end{minipage}
    \hfill
    \begin{minipage}[t]{0.65\linewidth}
    \vspace{0pt}
    \includegraphics[width=\linewidth]{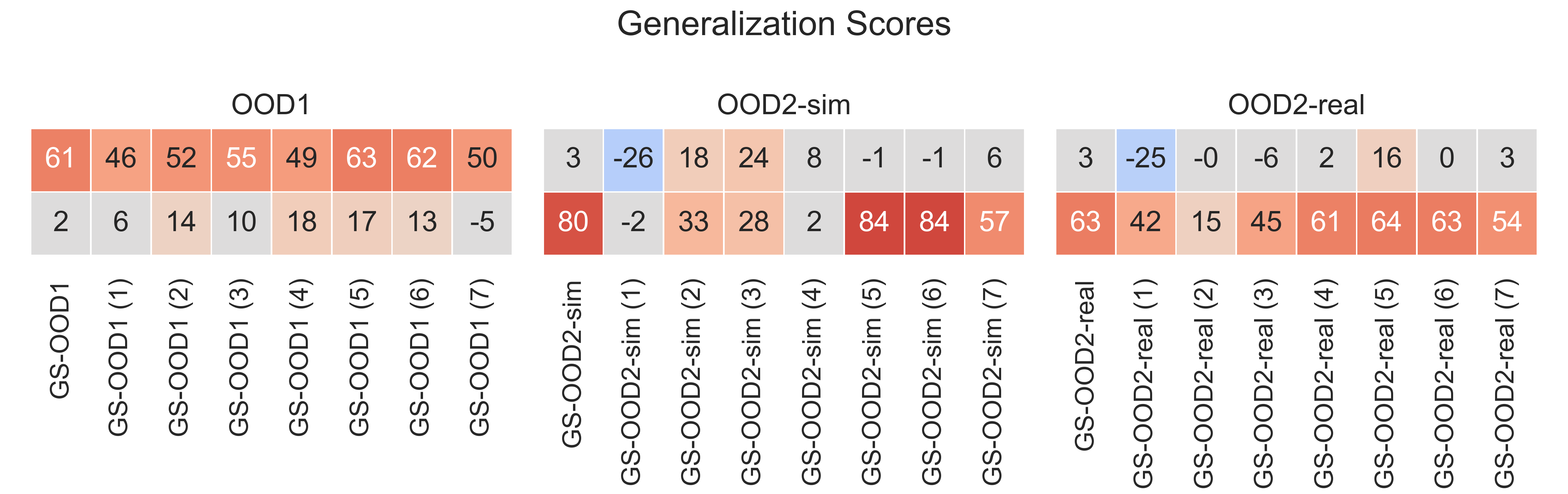}
    \end{minipage}
    \caption{Rank correlations of representation properties with OOD1 and OOD2 reward on \textit{object reaching} without regularization.
    Numbering when splitting metrics by FoV: (1) cube color; (2--4) joint angles; (5--7) cube position and rotation. Correlations are color-coded as described in \cref{fig:results_on_training_env}.}
    \label{fig:sim_world_OOD_rank_correlations_reaching_no_reg}
\end{figure}

\paragraph{Correlations with generalization scores.}
Here we analyze the link between generalization in RL and the generalization scores (GS) discussed in \cref{sec:background}, which measure the generalization of downstream FoV predictors \textit{out of distribution}, as opposed to the MLP and GBT metrics considered above. For both OOD scenarios, the distribution shifts underlying these GS scores are the same as the ones in the RL tasks in simulation.
We summarize our findings in \cref{fig:sim_world_OOD_rank_correlations_reaching_no_reg} (right) on the \textit{object reaching} task. 
Reward in the OOD1 setting is significantly correlated with the GS-OOD1 metric of the pretrained representation.
We observe an even stronger correlation between the reward in the simulated OOD2 setting and the corresponding GS-OOD2-sim and GS-OOD2-real scores. On a per-factor level, we see that the source of the observed correlations primarily stems from the generalization scores w.r.t.\,the pose parameters of the cube. 
The OOD generalization metrics can therefore be used as proxies for the corresponding form of generalization in downstream RL tasks.
This has practical implications for the training of RL downstream policies which are generally known to be brittle to distribution shifts, as we can measure a representation's generalization score from a few labeled images. This allows for selecting representations that yield more robust downstream policies.

\paragraph{Disentangled representations.}
\label{subsubsec:disentanglement_and_RL_reward}
Disentanglement has been shown to be helpful for downstream performance and OOD1 generalization even with MLPs \cite{dittadi2021transfer}. However, in \textit{object reaching}, we only observe a weak correlation with some disentanglement metrics (\cref{fig:disentanglement_and generalization_reaching}). In agreement with \cite{dittadi2021transfer}, disentanglement does not correlate with OOD2 generalization. 
The same study observed that disentanglement correlates with the informativeness of a representation. To understand if these weak correlations originate from this common confounder, we investigate whether they persist after adjusting for MLP FoV prediction accuracy. Given two representations with similar MLP accuracy, does the more disentangled one exhibit better OOD1 generalization? To measure this we predict success from the MLP accuracy using kNN (k=5) \cite{locatello2019fairness} and compute the residual reward by subtracting the amount of reward explained by the MLP metric. \cref{fig:disentanglement_and generalization_reaching} shows that this resolves the remaining correlations with disentanglement. Thus, for the RL downstream tasks considered here, disentanglement per se does not seem to be useful for OOD generalization. 
We present similar results on \textit{pushing} in \cref{app:additional_results_ood_simulation}.\looseness=-1

\begin{figure}
    \centering
    \begin{minipage}{0.22\linewidth}
    \includegraphics[width=\linewidth]{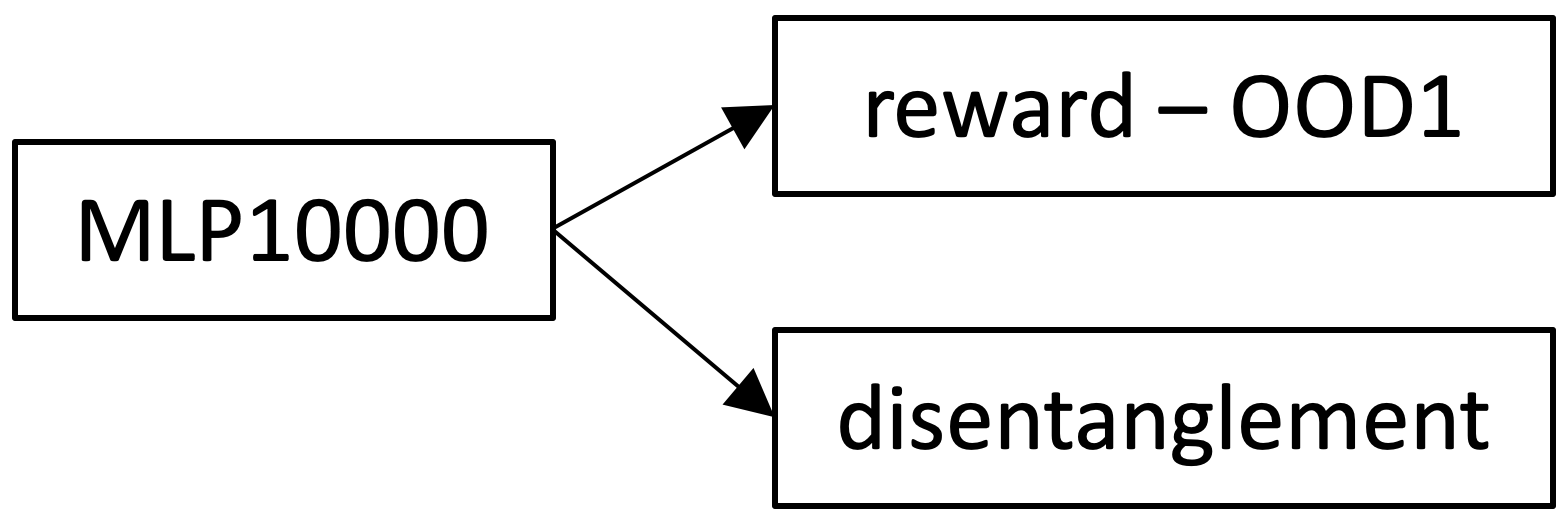}
    
    \includegraphics[width=\linewidth]{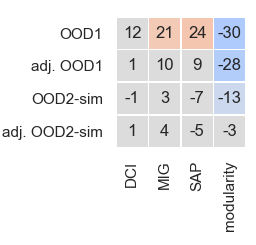}
    \end{minipage}
    \hfill
    \begin{minipage}{0.7\linewidth}
    \includegraphics[width=\linewidth]{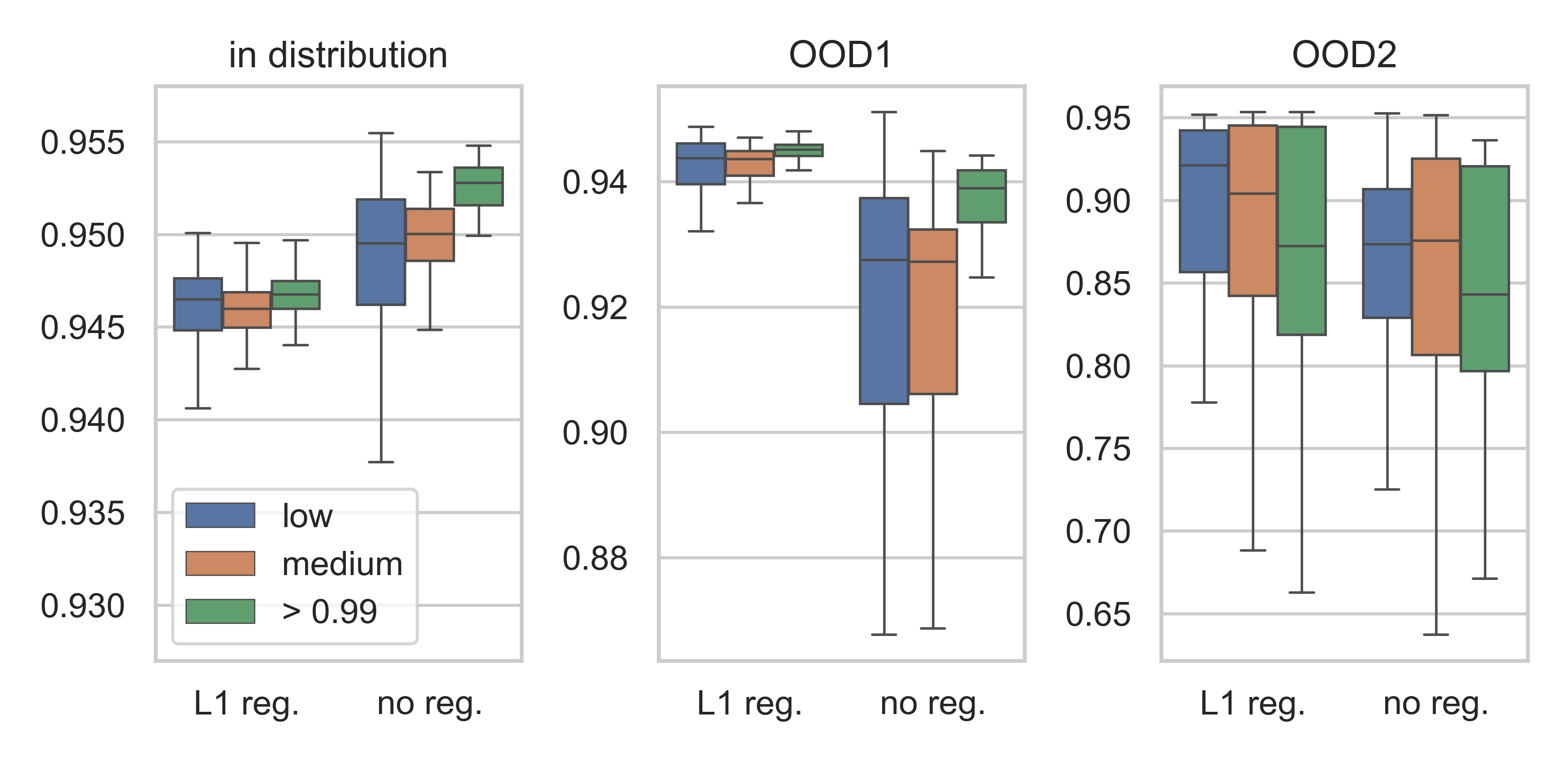}
    \end{minipage}
    \caption[{Box plots}: fractional success on \textit{object reaching} split according to low (blue), medium-high (orange), and almost perfect (green) disentanglement. {Correlation matrix} (left): although we observe a mild correlation between some disentanglement metrics and OOD1 (but not OOD2) generalization, this does not hold when adjusting for representation informativeness.]{\textbf{Box plots}: fractional success on \textit{object reaching} split according to low (blue), medium-high (orange), and almost perfect (green) disentanglement. 
    L1 regularization in the first layer of the MLP policy has a positive effect on OOD1 and OOD2 generalization with minimal sacrifice in terms of training reward (see scale).
    \textbf{Correlation matrix} (left): although we observe a mild correlation between some disentanglement metrics and OOD1 (but not OOD2) generalization, this does not hold when adjusting for representation informativeness. 
    Correlations are color-coded as described in \cref{fig:results_on_training_env}. We use disentanglement metrics from \citet{eastwood2018framework, chen2018isolating, kumar2017variational, ridgeway2018learning}.}
    \label{fig:disentanglement_and generalization_reaching}
\end{figure}

\paragraph{Policy regularization and observation noise.}
It might seem unsurprising that disentanglement is not useful for generalization in RL, as MLP policies do not have any explicit inductive bias to exploit it. Thus, we attempt to introduce such inductive bias by repeating all experiments with L1 regularization on the first layer of the policy. Although regularization improves OOD1 and OOD2 generalization in general (see box plots in \cref{fig:disentanglement_and generalization_reaching}), we observe no clear link with disentanglement. Furthermore, in accordance with \citet{dittadi2021transfer}, we find that observation noise when training representations is beneficial for OOD2 generalization. See \cref{app:additional_results_ood_simulation} for a detailed discussion.

\paragraph{Stronger OOD shifts: evaluating on a novel shape.}
On \textit{object reaching}, we also test generalization w.r.t. a novel shape by replacing the cube with a sphere. This corresponds to a strong OOD2-type shift, since shape was never varied when training the representations. 
Surprisingly, the policies appear to be robust to the novel shape. In fact, when the sphere has the same colors that the cube had during policy training, \emph{all} policies get closer than 5 cm to the sphere on average, with a mean success metric of 95\%. On sphere colors from the OOD1 split, more than 98.5\% move the finger closer than this threshold, and on the strongest distribution shift (OOD2-sim colors, and cube replaced by sphere) almost 70\% surpass that threshold with an average success metric above 80\%.

\paragraph{Summary.}
(1) In- and out-of-distribution rewards are correlated, as long as the representation remains in its training distribution (OOD1). (2) Similarly, in-distribution representation metrics (both unsupervised and supervised) predict OOD1 reward, but are not reliable when the representation is OOD (OOD2). (3) Disentanglement does not correlate with generalization in our experiments, while (4) input noise when training representations is beneficial for OOD2 generalization. (5) Most notably, the {GS metrics}, which measure generalization under distribution shifts, are {significantly correlated} with RL performance under similar distribution shifts. We thus recommend using these convenient proxy metrics for selecting representations that will yield robust downstream policies.

\subsection{Deploying policies to the real world}
\label{subsec:results_ood_generalization_real_robot}
We now evaluate a large subset of the agents on the real robot without fine-tuning, quantify their zero-shot sim-to-real generalization, and find metrics that correlate with real-world performance.

\begin{figure}
    \centering
    \includegraphics[width=0.38\linewidth]{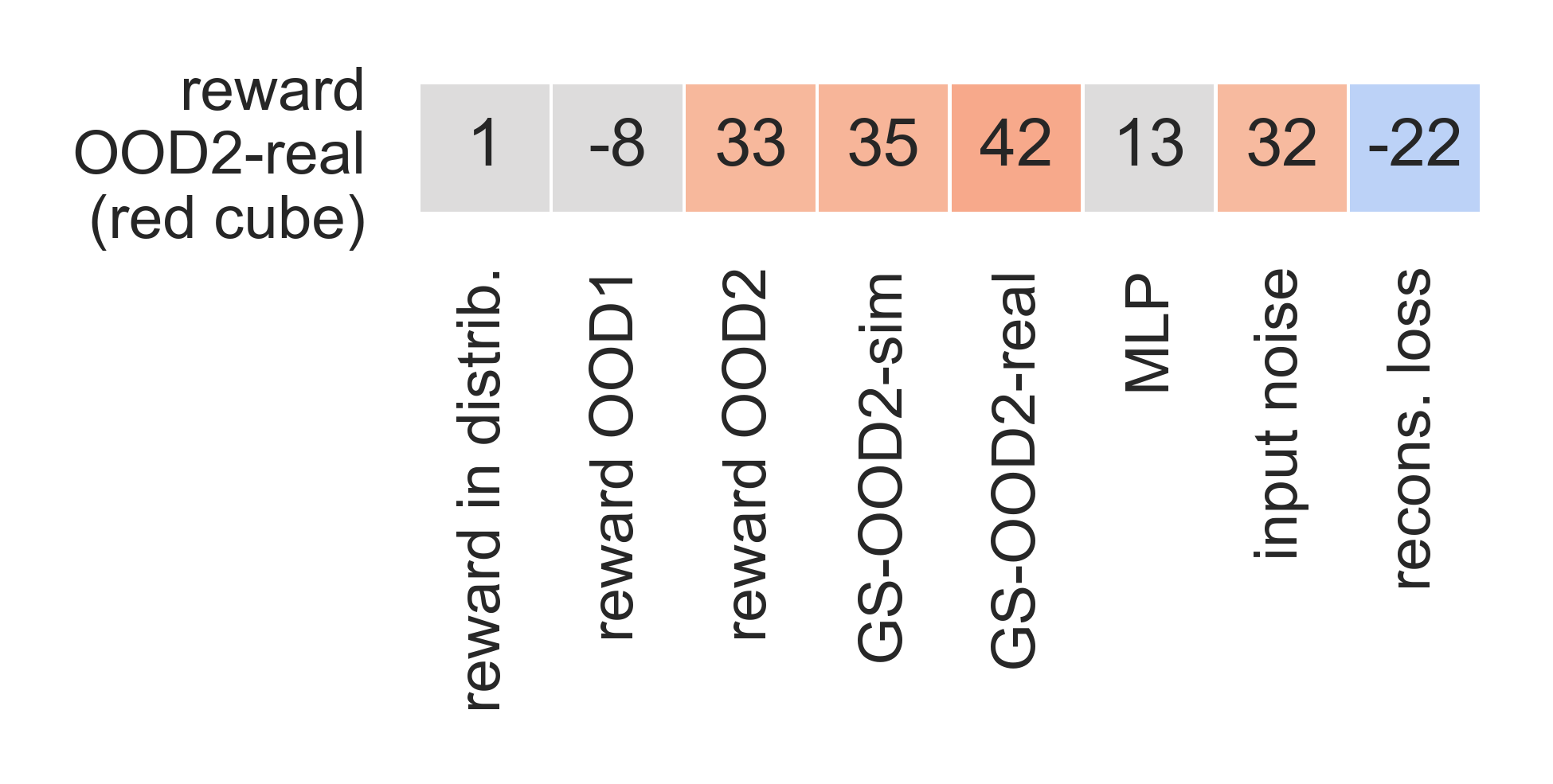}
    \quad
    \includegraphics[width=0.22\linewidth]{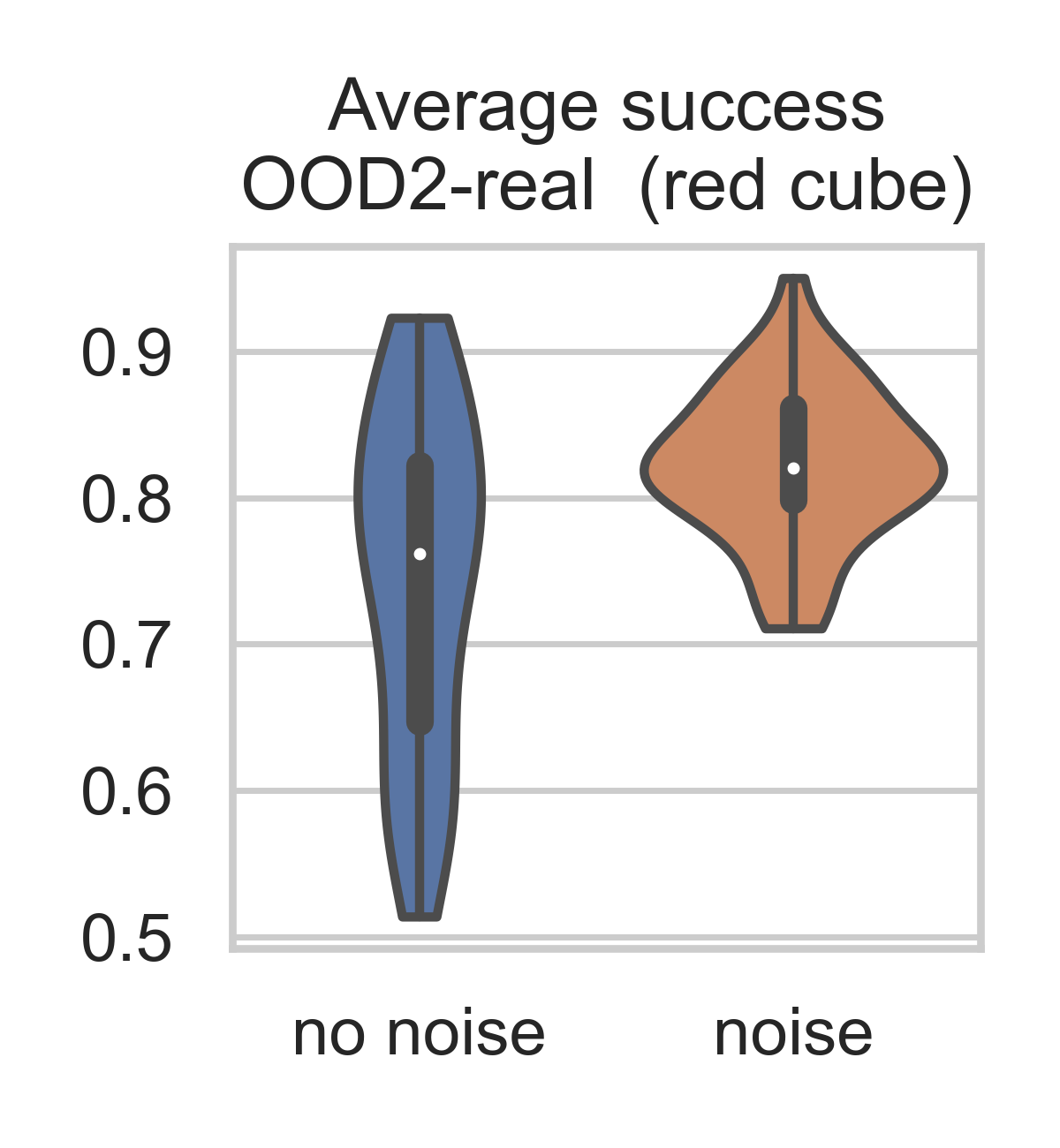}
    \quad
    \includegraphics[width=0.26\linewidth]{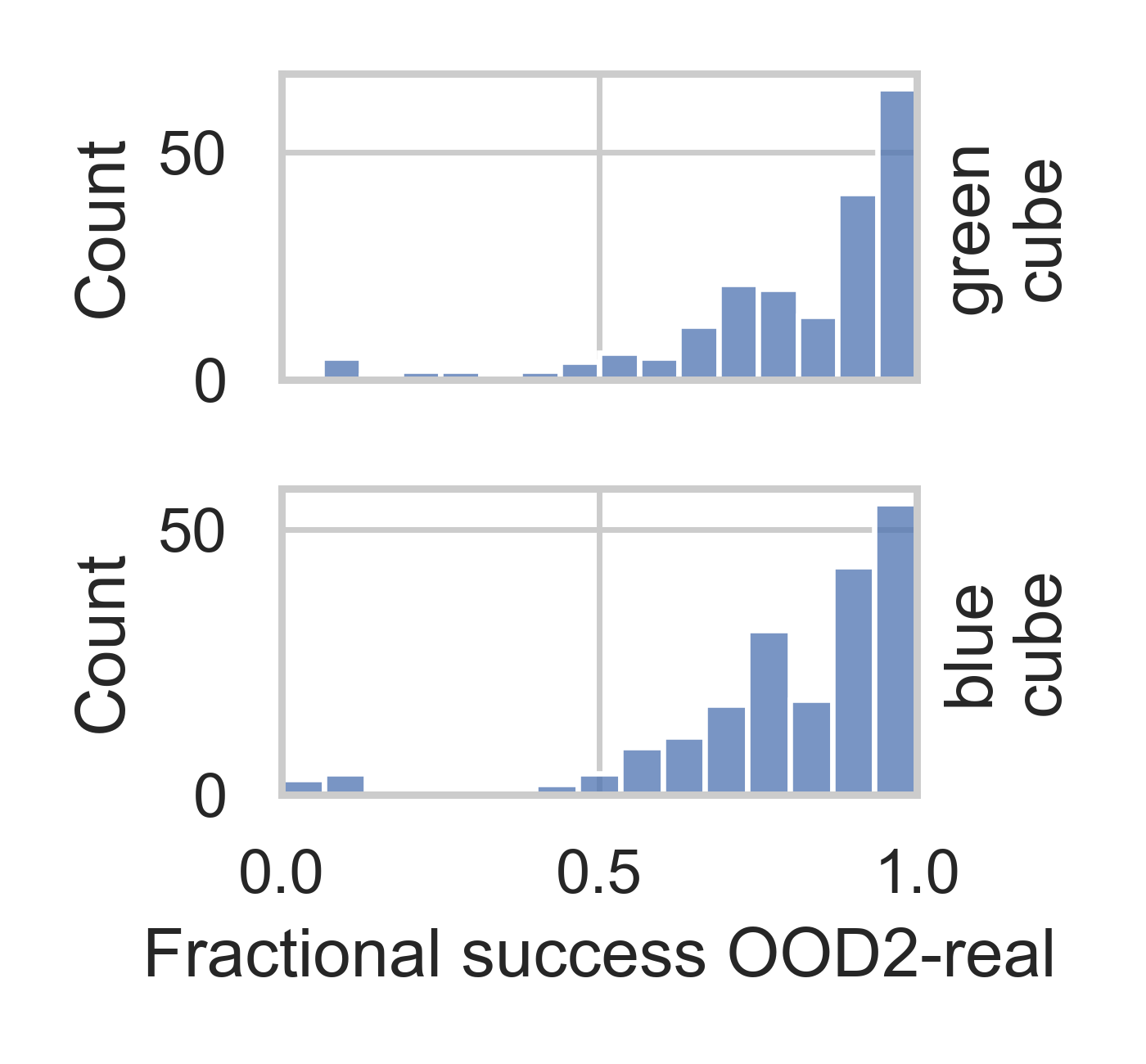}
    \caption[{Zero-shot sim-to-real} on {object reaching} on over 2,000 episodes.]{\textbf{Zero-shot sim-to-real} on \textit{object reaching} on over 2,000 episodes. \textbf{Left:} Rank-correlations on the real platform with a red cube (color-coded as described in \cref{fig:results_on_training_env}). \textbf{Middle}: Training encoders with additive noise improves sim-to-real generalization. \textbf{Right}: Histogram of fractional success in the more challenging OOD2-real-\{green,blue\} scenario from 50 policies across 4 different goal positions.}
    \label{fig:OOD2_evaluations_reaching}
\end{figure}

\paragraph{Reaching.}
We choose 960 policies trained in simulation, based on 96 representations and 10 random seeds, and evaluate them on two (randomly chosen, but far apart) goal positions using a red cube. 
While a red cube was in the training distribution, we consider this to be OOD2 because real-world images represent a strong distribution shift for the encoder~\cite{dittadi2021transfer, djolonga2021robustness}.
Although sim-to-real in robotics is considered to be very challenging without domain randomization or fine-tuning \cite{tobin2017domain,finn2017generalizing,rusu2017sim}, many of our policies obtain a high fractional success without resorting to these methods.
In addition, in \cref{fig:OOD2_evaluations_reaching} (left) we observe significant correlations between zero-shot real-world performance and some of the previously discussed metrics. 
First, there is a positive correlation with the OOD2-sim reward: Policies that generalize to unseen cube colors in simulation also generalize to the real world.
Second, repre\-sen\-tations with high GS-OOD2-sim and (especially) GS-OOD2-real scores are promising candidates for sim-to-real transfer.
Third, if no labels are available, the weaker correlation with the reconstruction loss on the simulated images can be exploited for representation selection. Finally, as observed by \citet{dittadi2021transfer} for simple downstream tasks, input noise while learning representations is beneficial for sim-to-real generalization (\cref{fig:OOD2_evaluations_reaching}, middle).\looseness=-1

\begin{figure}
    \centering
    \includegraphics[width=0.75\linewidth]{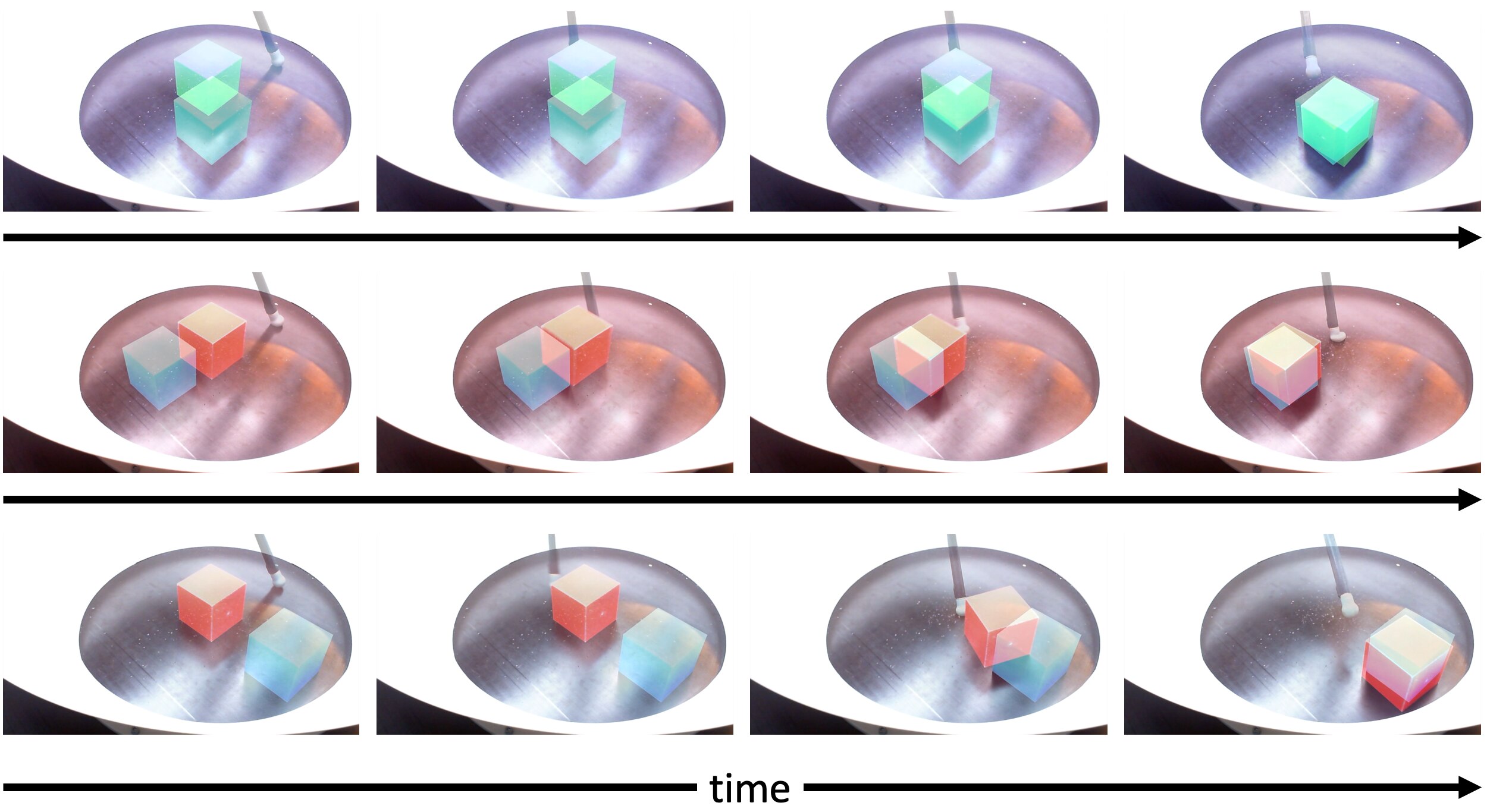}
    \caption{We select pushing policies with high GS-OOD2-real score. When deployed on the real robot without fine-tuning, they succeed in pushing the cube to a specified goal position (transparent blue cube).
    \label{fig:pushing_real_robot_frames}
    }
\end{figure}

Based on these findings, we select 50 policies with a high GS-OOD2-real score, and evaluate them on the real world with a green and a blue cube, which is an even stronger OOD2 distribution shift.
In \cref{fig:OOD2_evaluations_reaching} (right), where metrics are averaged over 4 cube positions per policy, we observe that most policies can still solve the task: approximately 80\% of them position the finger less than 5 cm from the cube. 
Lastly, we repeat the evaluations on the green sphere that we previously performed in simulation, and observe that many policies successfully reach this completely novel object. 
See \cref{app:additional_results_ood_real_world} and the project website for additional results and videos of deployed policies.

\paragraph{Pushing.}
We now test whether our real-world findings on \textit{object reaching} also hold for \textit{pushing}. We again select policies with a high GS-OOD2-real score and encoders trained with input noise. We record episodes on diverse goal positions and cube colors to support our finding that pushing policies in simulation can generalize to the real robot. In \cref{fig:pushing_real_robot_frames}, we show three representative episodes with successful task completions and refer to the project site for video recordings and further episodes.

\paragraph{Summary.} 
Policies trained in simulation can solve the task on the real robot without domain randomization or fine-tuning. Reconstruction loss, encoder robustness, and OOD2 reward in simulation are all good predictors of real-world performance. For real-world applications, we recommend using GS-OOD2-sim or GS-OOD2-real for model selection, and training the encoder with noise.

\section{Other related work}

A key unsolved challenge in RL is the brittleness of agents to distribution shifts in the environment, even if the underlying structure is largely unchanged~\cite{cobbe2019quantifying,ahmed2020causalworld}.
This is related to studies on representation learning and generalization in downstream tasks \citep{gondal2019transfer,steenbrugge2018improving,dittadi2021generalization,esmaeili2019structured,chaabouni2020compositionality}, as well as domain generalization (see \citet{wang2021generalizing} for an overview).
More specifically for RL, \citet{higgins2017darla} focus on domain adaptation and zero-shot transfer in DeepMind Lab and MuJoCo environments, and claim disentanglement improves robustness. To obtain better transfer capabilities, \citet{asadi2020learning} argue for discretizing the state space in continuous control domains by clustering states where the optimal policy is similar.
\citet{kulkarni2015deep} propose geometric object representations by means of keypoints or image-space coordinates and \citet{wulfmeier2021representation} investigate the effect of different representations on the learning and exploration of different robotics tasks.
Transfer becomes especially challenging from the simulation to the real world, a phenomenon often referred to as the sim-to-real gap. This is particularly crucial in RL, as real-world training is expensive, requires sample-efficient methods, and is sometimes unfeasible if the reward structure requires accurate ground truth labels \cite{dulac2019challenges, kormushev2013reinforcement}. 
This issue is typically tackled with large-scale domain randomization in simulation \cite{akkaya2019solving, james2019sim}.

\section{Conclusion}
Robust out-of-distribution (OOD) generalization is still one of the key open challenges in machine learning.
We attempted to answer central questions on the generalization of reinforcement learning agents in a robotics context, and how this is affected by pretrained representations.
We presented a large-scale empirical study in which we trained over 10,000 downstream agents given pretrained representations, and extensively tested them under a variety of distribution shifts, including sim-to-real.
We observed agents that generalize OOD, and found that some properties of the pretrained representations can be useful to predict which agents will generalize better.
We believe this work brings us one step closer to understanding the generalization abilities of learning systems, and we hope that it encourages many further important studies in this direction.

\section*{Acknowledgements}
We would like to thank Anirudh Goyal, Georg Martius, Nasim Rahaman, Vaibhav Agrawal, Max Horn, and the Causality group at the MPI for useful discussions and feedback. We thank the International Max Planck Research School for Intelligent Systems (IMPRS-IS) for supporting FT. Part of the experiments were generously supported with compute credits by Amazon Web Services. 

\fi

\chapter{Paper~\paperThree: Generalization and Robustness Implications in Object-Centric Learning}
\label{chapter:objects}
\chaptermark{Paper \paperThree}

\pubinfo{%
    Andrea Dittadi, \and 
    Samuele Papa, \and
    Michele De Vita, \and
    Bernhard Sch{\"o}lkopf, \and
    Ole Winther, \and
    Francesco Locatello.
}{%
    Published in \emph{International Conference on Machine Learning}, 2022.
}{%
    The idea behind object-centric representation learning is that natural scenes can better be modeled as compositions of objects and their relations as opposed to distributed representations. This inductive bias can be injected into neural networks to potentially improve systematic generalization and performance of downstream tasks in scenes with multiple objects.
    In this paper, we train state-of-the-art unsupervised models on five common multi-object datasets and evaluate segmentation metrics and downstream object property prediction.
    In addition, we study generalization and robustness by investigating the settings where either a single object is out of distribution---e.g., having an unseen color, texture, or shape---or global properties of the scene are altered---e.g., by occlusions, cropping, or increasing the number of objects.
    From our experimental study, we find object-centric representations to be useful for downstream tasks and generally robust to most distribution shifts affecting objects. However, when the distribution shift affects the input in a less structured manner, robustness in terms of segmentation and downstream task performance may vary significantly across models and distribution shifts.
}

\iffull
\section{Introduction}

In object-centric representation learning, we make the assumption that visual scenes are composed of multiple entities or objects that interact with each other, and exploit this compositional property as inductive bias for neural networks. Informally, the goal is to find transformations $r$ of the data $\xb$ into a \emph{set} of vector representations $r(\xb) = \{\zb_k\}$ each corresponding to an individual object, without supervision~\citep{eslami2016air, greff2017neural,kosiorek2018sequential,crawford2019spatially, greff2019multi, burgess2019monet, engelcke2020genesis, lin2020space, locatello2020object, chen2020learning,weis2020unmasking,mnih2014recurrent,gregor2015draw,yuan2019generative}.
Relying on this inductive bias, object-centric representations are conjectured to be more robust than distributed representations, and to enable the systematic generalization typical of symbolic systems while retaining the expressiveness of connectionist approaches \cite{bengio2013representation,lake2017building,greff2020binding,scholkopf2021toward}.
Grounding for these claims comes mostly from cognitive psychology and neuroscience \cite{spelke1990principles, teglas2011pure, wagemans2015oxford}. E.g., infants learn about the physical properties of objects as entities that behave consistently over time \cite{baillargeon1985object,spelke2007core} and are able to re-apply their knowledge to new scenarios involving previously unseen objects \cite{dehaene2020how}. Similarly, in complex machine learning tasks like physical modelling and reinforcement learning, it is common to train from the internal representation of a simulator \cite{battaglia2016interaction, sanchez-gonzalez2020learning} or of a game engine \cite{vinyals2019grandmaster, openai2019dota} rather than from raw pixels, as more abstract representations facilitate learning.
Finally, learning to represent objects separately is a crucial step towards learning causal models of the data from high-dimensional observations, as objects can be interpreted as causal variables that can be manipulated independently \cite{scholkopf2021toward}. Such causal models are believed to be crucial for human-level generalization \cite{pearl2009causality,peters2017elements}, but traditional causality research assumes causal variables to be given rather than learned \cite{scholkopf2019causality}.

As object-centric learning developed recently as a subfield of representation learning, we identify three key hypotheses and design systematic experiments to test them. 
(1)~\emph{The unsupervised learning of objects as pretraining task is useful for downstream tasks.} Besides learning to separate objects without supervision, current approaches are expected to separately represent information about each object's properties, so that the representations can be useful for arbitrary downstream tasks. 
(2)~\emph{In object-centric models, distribution shifts affecting a single object do not affect the representations of other objects.}
If objects are to be represented independently of each other to act as compositional building blocks for higher-level cognition \cite{greff2020binding}, changes to one object in the input should not affect the representation of the unchanged objects. This should hold even if the change leads to an object being out of distribution (OOD).
(3)~\emph{Object-centric models are generally robust to distribution shifts, even if they affect global properties of the scene.} 
Even if the whole scene is OOD---e.g., if it contains more objects than in the training set---object-centric approaches should be robust thanks to their inductive bias.

In this paper, we systematically investigate these three concrete hypotheses by re-implementing popular unsupervised object discovery approaches and testing them on five multi-object datasets.\footnote{Training and evaluating all the models for the main study requires approximately 1.44 GPU years on NVIDIA V100.} 
We find that:
(1)~Object-centric models achieve good downstream performance on property prediction tasks. We also observe a strong correlation between segmentation metrics, reconstruction error, and downstream property prediction performance, suggesting potential model selection strategies. 
(2)~If a single object is out of distribution, the overall segmentation performance is not strongly impacted. Remarkably, the downstream prediction of in-distribution (ID) objects is mostly unaffected. 
(3)~Under more global distribution shifts, the ability to separate objects depends significantly on the model and shift at hand, and downstream performance may be severely affected.

As an additional contribution, we provide a library\footnote{\url{https://github.com/addtt/object-centric-library}}
for benchmarking object-centric representation learning, which can be extended with more datasets, methods, and evaluation tasks. We hope this will foster further progress in the learning and evaluation of object-centric representations.

%%%%%%%%%%%%%%%%%%%%%%%%%%%%%%%%%%%%%%%%%%%%%%%%%%%%%%%%%%%%%%%%%%%%%%
\section{Study design and hypotheses}\label{sec:objects_study_design}

\textbf{Problem definition:} 
Vanilla deep learning architectures learn distributed representations that do not capture the compositional properties of natural scenes---see, e.g., the ``superposition catastrophe''~\cite{von1986thinking,bowers2014neural,greff2020binding}. Even in disentangled representation learning~\cite{higgins2016beta, kim2018disentangling, chen2018isolating, ridgeway2018learning, kumar2017variational, eastwood2018framework}, factors of variations are encoded in a vector representation that is the output of a standard CNN encoder. This introduces an unnatural ordering of the objects in the scene and fails to capture its compositional structure in terms of objects.
Formally defining objects is challenging \citep{greff2020binding} and there is no consensus even outside of machine learning \citep{smith1996origin,green2019theory}. \citet{greff2020binding} put forth three properties for object-centric representations: \emph{separation}, i.e., object features in the set of vectors $r(\xb)$ do not interact with each other, and each object is individually captured in a single element of $r(\xb)$; \emph{common format}, i.e., each element of $r(\xb)$ shares the same representational format; and \emph{disentanglement}, i.e., each element of $r(\xb)$ is represented in a disentangled format that exposes the factors of variation. 
In this paper, we consider representations $r(\xb)$ that are sets of vectors with each element sharing the representational format. 
We take a pragmatic perspective and focus on two clear desiderata for object-centric approaches:\looseness=-1

\textbf{Desideratum 1: } \emph{Object embodiment.}
The representation should contain information about the object's location and its embodiment in the scene. 
As we focus on unsupervised object discovery, this translates to segmentation masks. This is related to separation and common format, as the decoder is applied to the elements of $r(\xb)$ with shared parameters.

\textbf{Desideratum 2:} \emph{Informativeness of the representation.} Instead of learning disentangled representations of objects, which is challenging even in single-object scenarios \citep{locatello2018challenging}, we want the representation to contain useful information for downstream tasks, not necessarily in a disentangled format. We define objects through their properties as annotated in the datasets we consider, and predict these properties from the representations. Note that this may not be the only way to define objects (e.g., defining faces and edges as objects and deducing shapes as composition of those). The fact that existing models learn informative representations is our first hypothesis (see below).

\textbf{Design principle:} These desiderata offer well-defined quantitative evaluations for object-centric approaches and we want to understand the implications of learning such representations. To this end, we train four different state-of-the-art methods on five datasets, taking hyperparameter configurations from the respective publications and adapting them to improve performance when necessary. Assuming these models succeeded in learning an object-centric representation, we investigate the following hypotheses.\looseness=-1

\textbf{Hypothesis 1:} \emph{The unsupervised learning of objects as pretraining task is useful for downstream tasks.}  Existing empirical evaluations largely focus on our \emph{Desideratum 1} and evaluate the performance at test time in terms of segmentation metrics. The hope, however, is that the representation would be useful for other downstream tasks besides segmentation (\emph{Desideratum 2}). We test this hypothesis by training small downstream prediction models on the frozen object-representations with shared parameters to predict the object properties. We match the predictions to the ground-truth properties with the Hungarian algorithm \cite{kuhn1955hungarian} following \citet{locatello2020object}.\looseness=-1

\textbf{Hypothesis 2:} 
\emph{In object-centric models, distribution shifts affecting a single object do not affect the representations of other objects.}
A change in the properties of one object in the input should not affect the representation of the other objects. Even OOD objects with previously unseen properties should be segmented correctly by a network that learned the notion of objects~\citep{greff2020binding,scholkopf2021toward}.
We test this hypothesis by (1)~evaluating the segmentation of the scene after the distribution shift, and (2)~training downstream models to predict object properties, and evaluating them on representations extracted from scenes with one OOD object. More specifically, we test changes in the shape, color, or texture of one object.\looseness=-1

\textbf{Hypothesis 3:} \emph{Object-centric models are generally robust to distribution shifts, even if they affect global properties of the scene.} Early evidence~\citep{romijnders2021representation} points to the conjecture that learning object-centric representations biases the network towards learning more robust representations of the overall scene. Intuitively, the notion of objects is an additional inductive bias for the network to exploit to maintain accurate predictions if simple global properties of the scene are altered. We test this hypothesis by training downstream models to predict object properties, and evaluating them on representations of scenes with OOD global properties. In this case, we test robustness by cropping, introducing occlusions, and increasing the number of objects.

\begin{figure}
\centering
\includegraphics[width=0.8\linewidth]{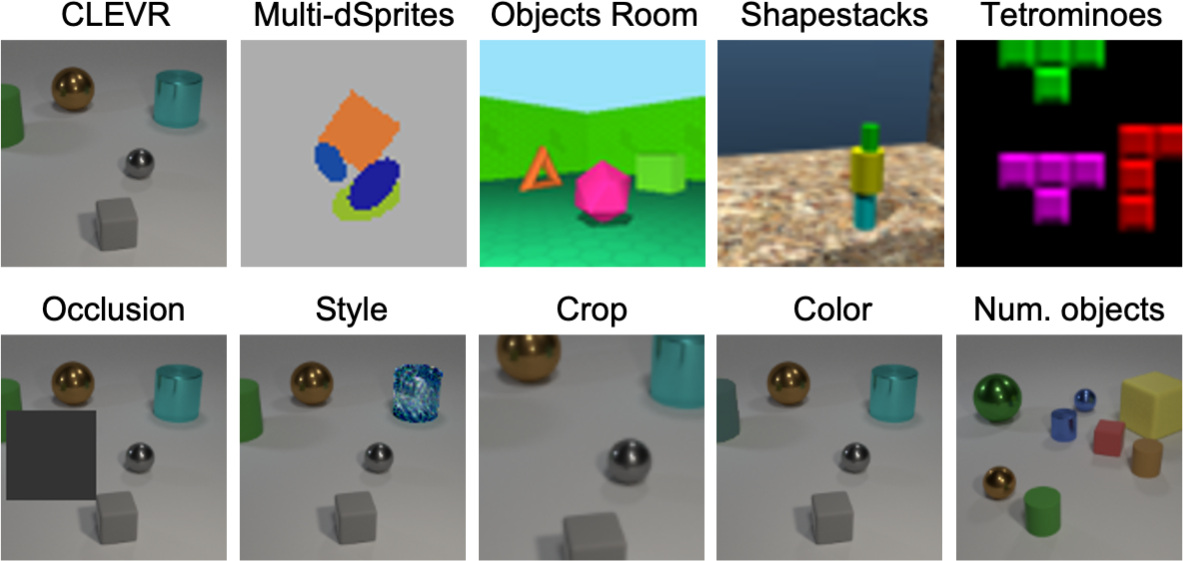}
\caption{\textbf{Top:} examples from the five datasets in this study. \textbf{Bottom:} distribution shifts in CLEVR.}
\label{fig:datasets}
\end{figure}

%%%%%%%%%%%%%%%%%%%%%%%%%%%%%%%%%%%%%%%%%%%%%%%%%%%%%%%%%%%%%%%%%%%%%%
\section{Experimental setup}
\label{sec:experimental_setup}

Here we provide an overview of our experimental setup. After introducing the relevant models and datasets, we outline the evaluation protocols for segmentation accuracy (Desideratum~1) and downstream task performance (Desideratum~2). Then, we discuss the distribution shifts that we use to test robustness---the aforementioned evaluations are repeated once again under these distribution shifts.
We conclude with a discussion on the limitations of this study.

\textbf{Models and datasets.}
We implement four state-of-the-art object-centric models---MONet \cite{burgess2019monet}, GENESIS~\cite{engelcke2020genesis}, Slot Attention~\cite{locatello2020object}, and SPACE~\cite{lin2020space}---as well as vanilla variational autoencoders (VAEs) \citep{kingma2013auto,rezende2014stochastic} as baselines for distributed representations. 
We use one VAE variant with a broadcast decoder \cite{watters2019spatial} and one with a regular convolutional decoder.
See \cref{app:models} for an overview of the models with implementation details.
We then collect five popular multi-object datasets: \emph{Multi-dSprites}, \emph{Objects Room}, and \emph{Tetrominoes} from DeepMind's Multi-Object Datasets collection~\cite{multiobjectdatasets19}, \emph{CLEVR}~\cite{johnson2017clevr}, and \emph{Shapestacks}~\cite{groth2018shapestacks}. The datasets are shown in \cref{fig:datasets} (top row) and described in detail in \cref{app:datasets}.
For each dataset, we define train, validation, and test splits. The test splits, which always contain at least $\num{2000}$ images, are exclusively used for evaluation.
We train each model on all datasets, using 10 random seeds for object-centric models and 5 for each VAE variant, resulting in 250 models in total.

\textbf{Metrics.} We evaluate the segmentation accuracy of object-centric models with the Adjusted Rand Index (ARI)~\cite{hubertComparingPartitions1985}, Segmentation Covering (SC)~\cite{arbelaez2010contour}, and mean Segmentation Covering (mSC)~\cite{engelcke2020genesis}. For all models, we additionally evaluate reconstruction quality via the mean squared error (MSE).
\cref{app:evaluation_metrics} includes detailed definitions of these metrics.

\textbf{Downstream property prediction.} % actually the next two pars are also about downstream
We evaluate object-centric representations by training downstream models to predict ground-truth object properties from the representations. More specifically, exploiting the fact that object slots share a common representational format, a single downstream model $f$ can be used to predict the properties of each object independently: for each slot representation $\zb_k$ we predict a vector of object properties $\ybpred_k = f(\zb_k)$.
As in previous work on object property prediction~\cite{locatello2020object}, each model simultaneously predicts all properties of an object. For learning, we use the cross-entropy loss for categorical properties and MSE for numerical properties, and denote by $\ell(\ybpred_k, \yb_m)$ the overall loss for a single object, where $\yb_m$ are its ground-truth properties. Here $k \in \{1,\ldots,K\}$ and $m \in \{1,\ldots,M\}$ with $K$ the number of slots and $M$ the number of objects.
In order to optimize the downstream models, the vector $\ybpred_k$ (the properties predicted from the $k$th representational slot) needs to be matched to the ground-truth properties $\yb_m$ of the $m$th object. This is done by computing a $M \times K$ matrix of \emph{matching losses} for each slot--object pair, and then solving the assignment problem using the Hungarian algorithm~\cite{kuhn1955hungarian} to minimize the total matching loss, which is the sum of $\min(M, K)$ terms from the loss matrix.
As matching loss we use either the negative cosine similarity between predicted and ground-truth masks (as in \citet{greff2019multi}), or the downstream loss $\ell(\ybpred_k, \yb_m)$ itself (as in \citet{locatello2020object}). In the following, we will refer to these strategies as \emph{mask matching} and \emph{loss matching}, respectively.
For property prediction, we use 4 different downstream models: a linear model, and MLPs with up to 3 hidden layers of size 256 each.
Given a pretrained object-centric model, we train each downstream model on the representations of $\num{10000}$ images. The downstream models are then tested on $\num{2000}$ held-out images from the test set, which may exhibit distribution shifts as discussed below.
Further details on this evaluation are provided in \cref{app:evaluation_downstream}

\textbf{Evaluating distributed representations.}
Since in non-slot-based models, such as classical VAEs, the representations of the single objects are not readily available, matching representations to objects for downstream property prediction is not trivial.
Although this is an inherent limitation of distributed representations, we are nevertheless interested in evaluating their usefulness.
Using the matching framework presented above, we require the downstream model $f$ to output the predicted properties of \emph{all} objects, and then match these with the true object properties to evaluate prediction quality. 
Our downstream model in this case will thus take as input the entire representation $\zb = r(\xb)$ (which is now a single vector rather than a set of vectors) and output the predictions for all objects together as a vector $f(\zb)$. Finally, we split $f(\zb)$ into $K$ vectors $\{\ybpred_k\}_{k=1}^K$, where $K$ loosely corresponds to the number of slots in object-centric models. At this point, we can compute the loss $\ell(\ybpred_k, \yb_m)$ for each pair, as usual.
We now consider two matching strategies: As before, \emph{loss matching} simply defines the matching loss of a slot--object pair as the prediction loss itself. In the \emph{deterministic matching} strategy, following \citet{greff2019multi}, we lexicographically sort objects according to a canonical order of object properties. 
Calling $\pi$ the permutation that defines this sorting, the $k$th slot is deterministically matched with the $m$th object, where $m = \pi^{-1}(k)$.

\textbf{Baselines.}
To correctly assess performance on downstream tasks, it is fundamental to compare with sensible baselines. Here we consider as baseline the best performance that can be achieved by a downstream model that outputs a constant vector that does not depend on the image. When predicting properties independently for each object (in slot-based models), the optimal solution is to predict the mean of continuous properties and the majority class for categorical ones. When using deterministic matching in the distributed case, the downstream model can exploit the predefined total order to predict more accurately than random guessing even without using information from the input (this effect is non-negligible only for the properties that are most significant in the order). Finally, in a few cases, loss matching for distributed representations can be significantly better than deterministic matching.\footnote{Intuitively, a (constant) diverse set of uninformed predictions $\{\ybpred_k\}$ might be sufficient for the matching algorithm to find suitable enough objects for most predictions.} 
For simplicity, for both matching strategies in the distributed case, we directly learn a vector $\ybpred$ by gradient descent to minimize the prediction loss. As this depends on random initialization and optimization dynamics, we repeat this for 10 random seeds and report error bars in the plots.

\textbf{Distribution shifts.}
We test the robustness of the learned representations under two classes of distribution shifts: one where one object goes OOD, and one where global properties of the scene are changed. All such distribution shifts occur at test time, i.e., the unsupervised models are always trained on the original datasets.
To evaluate generalization to distribution shifts affecting a \textit{single object}, we systematically induce changes in the color, shape, and texture of objects.
To change color, we apply a random color shift to one random object in the scene, using the available masks (we do not do this in Multi-dSprites, as the training distribution covers the entire RGB color space).
To test robustness to unseen textures, we apply neural style transfer~\cite{gatys_neural_2016} to one random object in each scene, using \emph{The Great Wave off Kanagawa} as style image.
When either a new color or a new texture is introduced, prediction of material (in CLEVR only) and color is not performed.
To introduce a new shape, we select images from Multi-dSprites that have at most 4 objects (in general, they have up to 5), and add a randomly colored triangle, in a random position, at a random depth in the object stack. In this case, shape prediction does not apply.
Finally, to test robustness to \textit{global} changes in the scene, we change the number of objects (in CLEVR only), introduce occlusions (a gray square at a random location), or crop images at the center and restore their original size via bilinear interpolation.
See \cref{fig:datasets} for examples, and \cref{app:evaluation_shifts} for further details.\looseness=-1

\textbf{Limitations of this study.}
While we aim to conduct a sound and informative experimental study to answer the research questions from \cref{sec:experimental_setup}, inevitably there are limitations regarding datasets, models, and evaluations.
Although the datasets considered here vary significantly in complexity and visual properties, they all consist of synthetic images where object properties are independent of each other and independent between objects.
Regarding object-centric models, we only focus on autoencoder-based approaches that model a scene as a mixture of components.
As official implementations are not always available, and none of the methods in this work has been applied to all the datasets considered here, we re-implement these methods and choose hyperparameters following a best-effort approach.
Finally, we only consider the downstream task of object property prediction, and assess generalization using only a few representative single-object and global distribution shifts.

%%%%%%%%%%%%%%%%%%%%%%%%%%%%%%%%%%%%%%%%%%%%%%%%%%%%%%%%%%%%%%%%%%%%%%
\section{Results}

In this section, we highlight our findings with plots that are representative of our main results. The full experimental results are presented in \cref{app:results}. In \cref{sec:results/metrics} we focus on the different evaluation metrics and the performance we obtained re-training the methods considered in this study. We then focus on our three hypotheses in \cref{sec:results/hyp1,sec:results/hyp2,sec:results/hyp3}.

\subsection{Learning and evaluating object discovery}
\label{sec:results/metrics}

\begin{figure}
\centering
\includegraphics[width=0.8\linewidth]{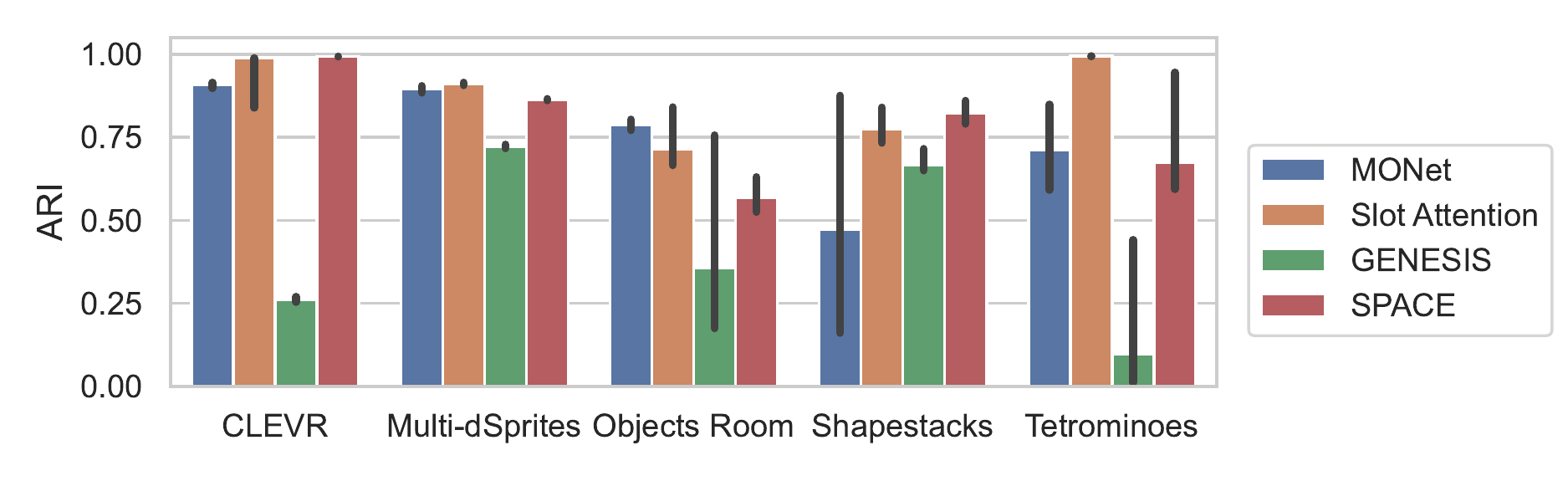}
\caption{ARI of all models and datasets on $\num{2000}$ test images. Medians and 95\% confidence intervals with 10 seeds.}
\label{fig:main_text/indistrib/metrics/box_plots}
\end{figure}

Since all methods included in our study were originally evaluated only on a subset of the datasets and metrics considered here, we first test how well these models perform.

\cref{fig:main_text/indistrib/metrics/box_plots} shows the segmentation performance of the models in terms of ARI across models, datasets, and random seeds. \cref{fig:indistrib/metrics/box_plots} in \cref{app:results} provides an overview of the reconstruction MSE and all segmentation metrics. Although these results are in line with published work, we observe substantial differences in the ranking between models depending on the metric. This indicates that, in practice, these metrics are not equivalent for measuring object discovery.

\begin{figure}
\centering
\includegraphics[width=0.9\linewidth]{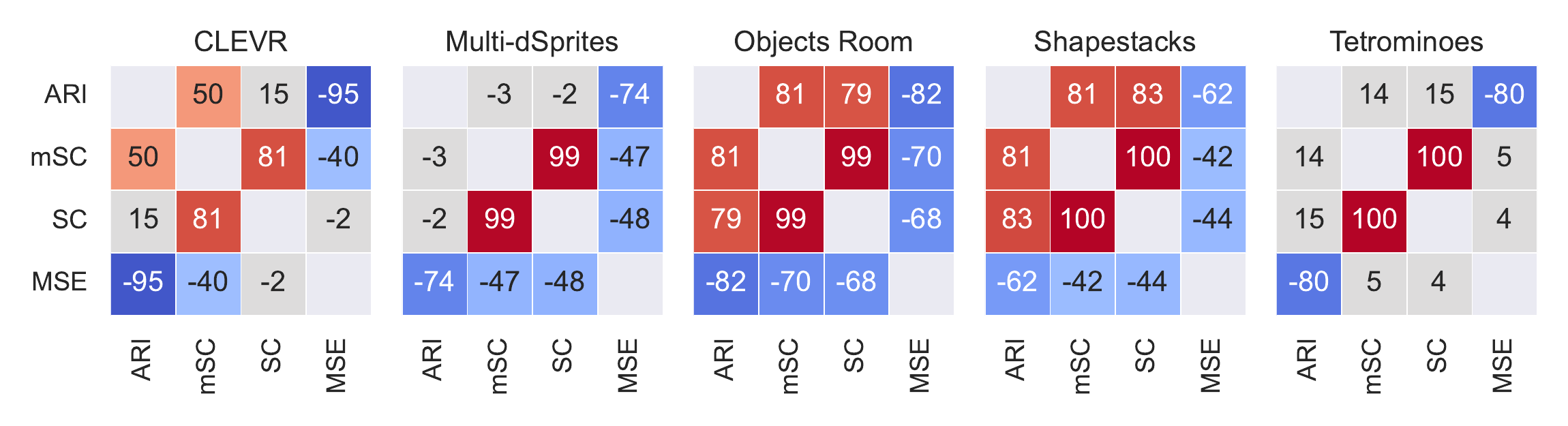}
\caption{Spearman rank correlations between evaluation metrics across models and random seeds (color-coded only when p\textless 0.05).}
\label{fig:indistrib_correlation/metrics_metrics}
\end{figure}

This is confirmed in \cref{fig:indistrib_correlation/metrics_metrics}, which shows rank correlations between metrics on different datasets (aggregating over different models).
We also observe a strong negative correlation between ARI and MSE across models and datasets, suggesting that models that learn to more accurately reconstruct the input tend to better segment objects according to the ARI score.
This trend is less consistent for the other segmentation metrics, as MSE significantly correlates with mSC in only three datasets (Multi-dSprites, Objects Room, and CLEVR), and with SC in two (Multi-dSprites and Objects Room).
SC and mSC measure very similar segmentation notions and therefore are significantly correlated in all datasets, although to a varying extent. However, they correlate with ARI only on two and three datasets, respectively (the same datasets where they correlate with the MSE).

\textbf{Summary:} We observe strong differences in performance and ranking between the models depending on the evaluation metric. In the tested datasets, we find that the ARI, which requires ground-truth segmentation masks to compute, correlates particularly well with the MSE, which is unsupervised and provides training signal.

\subsection{Usefulness for downstream tasks (Hypothesis 1)}
\label{sec:results/hyp1}
To test Hypothesis 1, we first evaluate whether frozen object-centric representations can be used to train downstream models measuring {Desideratum 2} from \cref{sec:objects_study_design}. As discussed in \cref{sec:experimental_setup}, this type of downstream task requires matching the true object properties with the predictions of the downstream model. In the following, we will only present results obtained with \emph{loss matching}, and show results for other matching strategies in \cref{app:results}.

\begin{figure*}
\centering
\includegraphics[width=\linewidth]{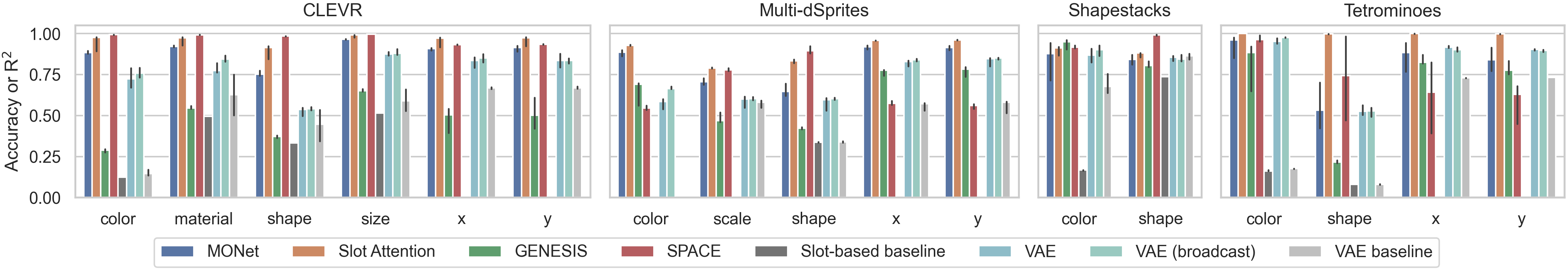}
\caption[Comparison of downstream property prediction performance for object-centric (slot-based) and distributed (VAE) representations, using an MLP with one hidden layer as downstream model.]{Comparison of downstream property prediction performance for object-centric (slot-based) and distributed (VAE) representations, using an MLP with one hidden layer as downstream model. The metric is accuracy for categorical properties or \rsq for numerical ones. The baselines in gray indicate the best performance that can be achieved by a model that outputs a constant vector that does not depend on the input. The bars show medians and 95\% confidence intervals with 10 random seeds.}
\label{fig:indistrib/downstream/barplots_loss_linear_main}
\end{figure*}

\cref{fig:indistrib/downstream/barplots_loss_linear_main} shows downstream prediction performance on all datasets and models, when the downstream model is a single-layer MLP.
Although results vary across datasets and models, accurate prediction of object properties seems to be possible in most of the scenarios considered here.
\cref{fig:indistrib/downstream/box_plots} in \cref{app:results} shows similar results when using a linear model or MLPs with up to 3 hidden layers.

In \cref{fig:indistrib/downstream/barplots_loss_linear_main}, we also compare the downstream prediction performance from object-centric and distributed representations. We observe that VAE representations tend to achieve lower scores in downstream prediction, although not always by a large margin. In particular, color and size in CLEVR and color in Tetrominoes are predicted relatively well, and significantly better than the baseline. On the other hand, in many cases where VAE representations perform well, they have in fact a considerable advantage if we take the baselines into account (scale in Multi-dSprites, color in Shapestacks, x~and~y in CLEVR, Multi-dSprites, and Tetrominoes). Moreover, performance from distributed representations often does not improve significantly when using a larger downstream model (see \cref{fig:indistrib/downstream/barplots_compare_downstream_models_loss}).
In conclusion, although the two classes of representations are difficult to compare on this task, these results suggest that the quantities of interest are present in the VAE representations, but they appear to be less explicit and less easily usable.

\begin{figure}
\centering
\includegraphics[width=0.75\linewidth]{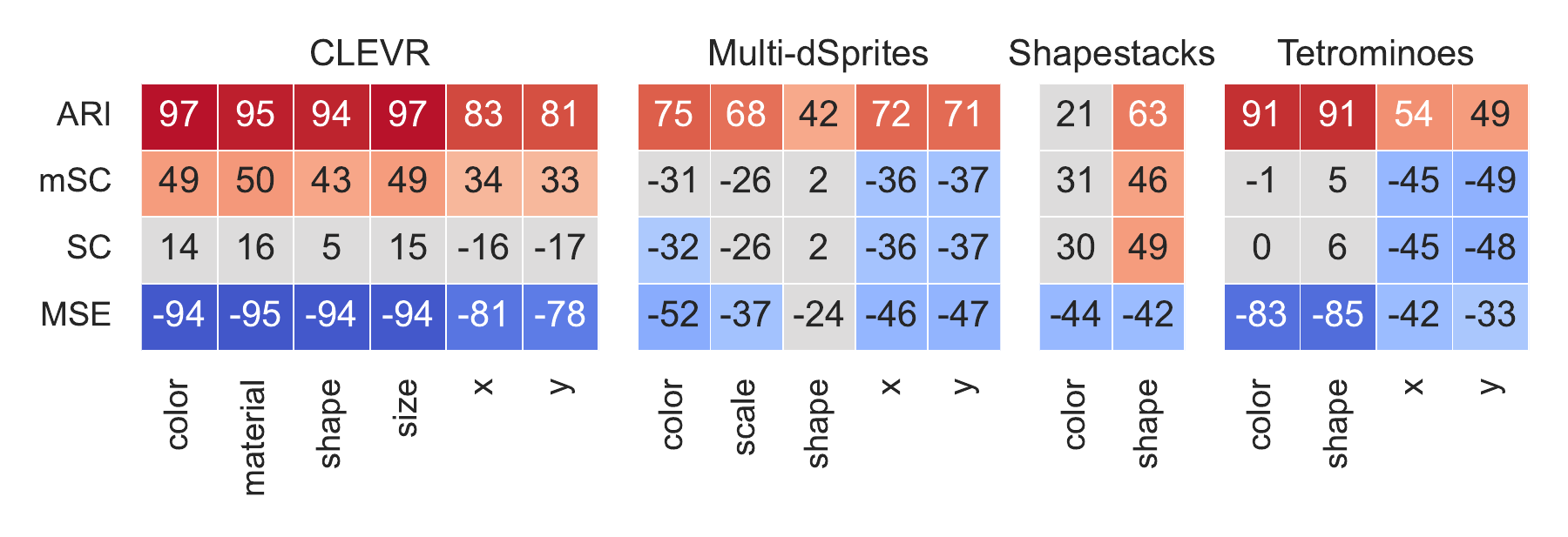}
\caption{Spearman rank correlations between evaluation metrics and downstream performance with an MLP. The correlations are color-coded only when p\textless 0.05.}
\label{fig:indistrib_correlations/metrics_downstream}
\end{figure}

Finally, we investigate the relationship between downstream performance and evaluation metrics.
\cref{fig:indistrib_correlations/metrics_downstream} shows the Spearman rank correlation of the segmentation and reconstruction metrics with the test performance of downstream predictors.
For all datasets and 
object 
properties, downstream performance is strongly correlated with the ARI.
On the other hand, SC and mSC exhibit inconsistent trends across datasets.
% Object-centric 
Models that correctly separate objects according to the ARI are therefore useful for downstream object property prediction, confirming Hypothesis 1.
Downstream prediction performance is also significantly correlated with the reconstruction MSE in all datasets. This is not particularly surprising, since the representation of a model that cannot properly reconstruct the input might not contain the information necessary for property prediction. However, the correlation is generally stronger with the ARI than with the MSE, suggesting that having a notion of objects is more important for downstream tasks than reconstruction accuracy. This is consistent with the findings by \citet{papa2022inductive}, where the ARI still correlates strongly with downstream performance when objects have complex textures, while the MSE does not. When segmentation masks are available for validation, ARI should therefore be the preferred metric to select useful representations for downstream tasks.
\cref{fig:indistrib/downstream/corr_metrics_downstream_all} in \cref{app:results} shows analogous results for mask matching and for the three other downstream models---these results are broadly similar, except that correlations with ARI tend to be stronger when using mask matching (perhaps unsurprisingly) or larger downstream models.

\textbf{Summary:}
Models that accurately segment objects allow for good downstream
prediction 
performance.
Despite often having an advantage, distributed representations generally perform worse,
but not always significantly: the information is present but less easily accessible. 
The ARI is consistently correlated with downstream performance, and is therefore useful for model selection when masks are available. The MSE can be a practical unsupervised alternative on these datasets, but it may be less robust on complex textures.

%%%%%%%%%%%%%%%%%%%%%%%%%%%%%%%%%%%%%%%%%%%%%%%%%%%%%%%%%%%%%%%%%%%%%%
\subsection{Generalization with one OOD object (Hypothesis 2)}
\label{sec:results/hyp2}
To test Hypothesis 2, we construct settings where a single object is OOD and the others are ID. We change the object style with neural style transfer, change the color of one object at random (only in CLEVR, Tetrominoes, and Shapestacks), or introduce a new shape (only in Multi-dSprites).
The unsupervised models are always trained on the original datasets. Then we train downstream models to predict the object properties from the learned representations. 
We consider two scenarios for this task: (1)~train the predictors on the original datasets and test them on the variants with a modified object, (2)~train and test the predictors on each variant.
In both cases, we test the predictors on representations that might be inaccurate, because the representation function (encoder) is OOD. However, since in case (2) the downstream model is \emph{trained} under distribution shift, this experiment quantifies the extent to which the representation can still be used by a downstream task that is allowed to adapt to the shift---although the representation might no longer represent objects faithfully, it could still contain useful information.

\begin{figure*}
\centering
\includegraphics[width=\linewidth]{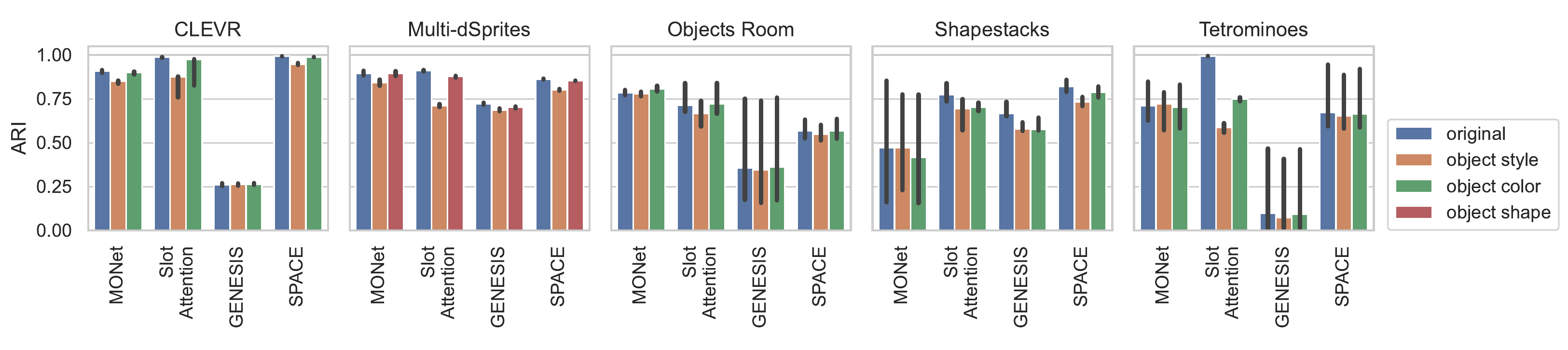}
\caption{Effect of single-object distribution shifts on the ARI. Medians and 95\% confidence intervals with 10 random seeds.}
\label{fig:main_text/ood/metrics/generalization_compositional/box_plots}
\end{figure*}

For {Desideratum 1}, we observe in \cref{fig:main_text/ood/metrics/generalization_compositional/box_plots} that the models are generally robust to distribution shifts affecting a single object.
Introducing a new color or a new shape typically does not affect segmentation quality (but note Slot Attention on Tetrominoes), while changing the texture of an object via neural style transfer leads to a moderate drop in ARI in some cases.
In \cref{fig:ood/metrics/generalization_compositional/box_plots} (\cref{app:results}) we observe that SC and mSC show a compatible but less pronounced trend, while the MSE more closely mirrors the ARI. 
We conclude that the encoder is still partially able to separate objects when one object undergoes a distribution shift at test time.

\begin{figure}
\centering
\includegraphics[width=0.7\linewidth]{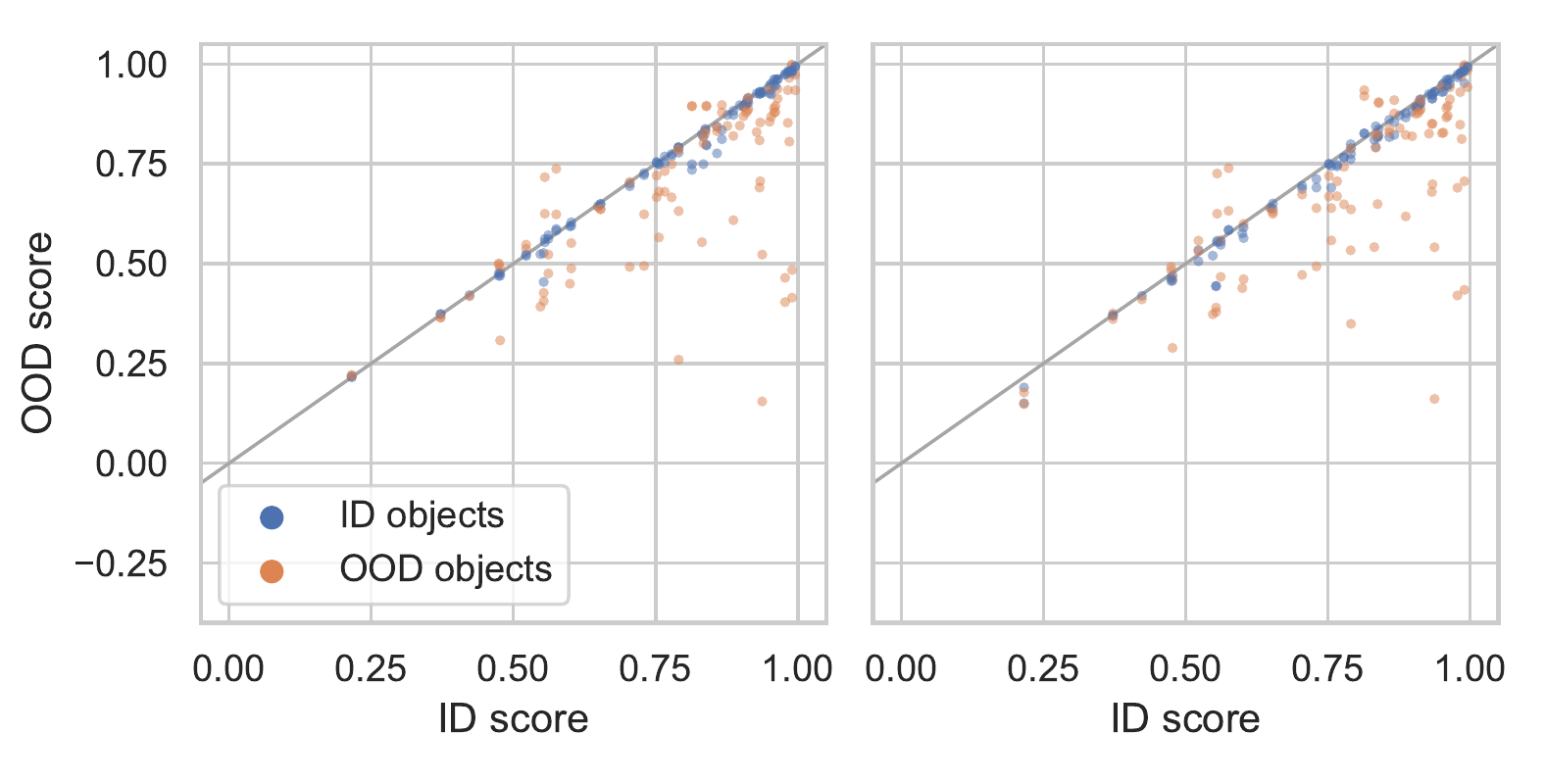}
\caption[{ID vs OOD downstream performance} with single-object distribution shifts. All datasets, models, and object properties are shown.]{\textbf{ID vs OOD downstream performance} with single-object distribution shifts. All datasets, models, and object properties are shown. Metrics: accuracy for categorical attributes, \rsq for numerical attributes. The downstream model (an MLP with one hidden layer) is tested zero-shot out-of-distribution (left) or retrained after the distribution shift has occurred (right).}
\label{fig:main_text/ood/downstream/generalization_compositional/scatter}
\end{figure}

For {Desideratum 2}, we observe in \cref{fig:main_text/ood/downstream/generalization_compositional/scatter} (left) that property prediction performance for objects that underwent distribution shifts (color, shape, or texture) is often significantly worse than in the original dataset, whereas the prediction of ID objects is largely unaffected.
This is in agreement with Hypothesis 2: changes to one object do not affect the representation of other objects, even when these objects are OOD.
Extensive results, including further splits and all downstream models, are shown in \cref{fig:ood/downstream/slotted/match_loss/object/retrain_False} in \cref{app:results}.
On the right plot in \cref{fig:main_text/ood/downstream/generalization_compositional/scatter}, we observe that retraining the downstream models after the distribution shifts does not lead to significant improvements. This suggests that the shifts introduced here negatively affect not only the downstream model, but also the representation itself.
This result also holds with different downstream models and with mask matching (see \cref{fig:ood/downstream/slotted/match_loss/object/retrain_True,fig:ood/downstream/slotted/match_mask/object/retrain_True}). 
While in principle we observe a similar trend for VAEs (see e.g. \cref{fig:ood/downstream/vae/match_loss/object/retrain_False,fig:ood/downstream/vae/match_loss/object/retrain_True} in \cref{app:results}), their performance is often too close to the respective baseline (\cref{fig:indistrib/downstream/barplots_loss_linear_main}) for a definitive conclusion to be drawn.

\textbf{Summary:}
The models are generally robust to distribution shifts affecting a single object.
Downstream prediction is largely unaffected for ID objects, but may be severely affected for OOD objects.
Finally, there seems to be no clear benefit in retraining downstream models after the shifts, indicating that the deteriorated representations cannot easily be adjusted \emph{post hoc}.

%%%%%%%%%%%%%%%%%%%%%%%%%%%%%%%%%%%%%%%%%%%%%%%%%%%%%%%%%%%%%%%%%%%%%%
\subsection{Robustness to global shifts (Hypothesis 3)}
\label{sec:results/hyp3}
Finally, we investigate the robustness of object-centric models to transformations changing the global properties of a scene at test time.
Here, we consider variants of the datasets with occlusions, cropping, or more objects (only on CLEVR).
We train downstream predictors on the original datasets and report their test performance on the dataset variants with global shifts. As before, we also report results of downstream models retrained on the OOD datasets.

\begin{figure*}
\centering
\includegraphics[width=\linewidth]{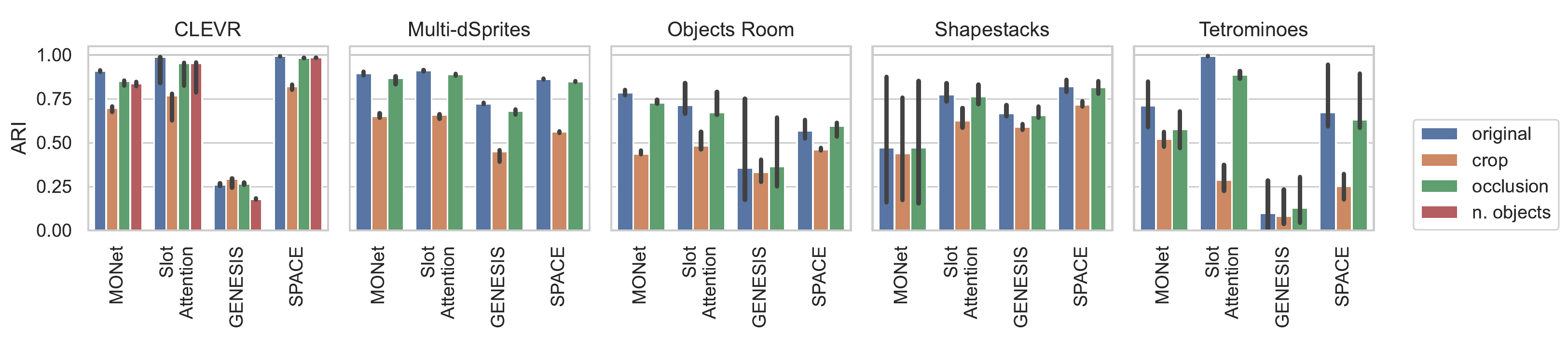}
\caption{Effect of distribution shifts on global scene properties on the ARI. Medians and 95\% confidence intervals with 10 seeds.}
\label{fig:figures/ood/metrics/only_ARI_ood_global/barplots_ARI}
\end{figure*}

For {Desideratum 1}, \cref{fig:figures/ood/metrics/only_ARI_ood_global/barplots_ARI} shows that segmentation quality is generally only marginally affected by occlusion, but cropping often leads to a significant degradation. In CLEVR, the effect on the ARI of increasing the number of objects is comparable to the effect of occlusions, which suggests that learning about objects is useful for this type of systematic generalization.
These trends persist when considering SC and mSC, but appear less pronounced and less consistent across datasets (see \cref{fig:ood/metrics/generalization_global/box_plots} in \cref{app:results} for detailed results).
As might be expected, when the number of objects is increased in CLEVR, the MSE increases more conspicuously for VAEs than for object-centric models (\cref{fig:ood/metrics/generalization_global/box_plots}, bottom left), likely due to their explicit modeling of objects. However, \cref{fig:ood_visualizations_clevr} shows that VAEs may, in fact, generalize relatively well to an unseen number of objects, although not nearly as well as some object-centric models.
\looseness=-1

\begin{figure}
\centering
\includegraphics[width=0.7\linewidth]{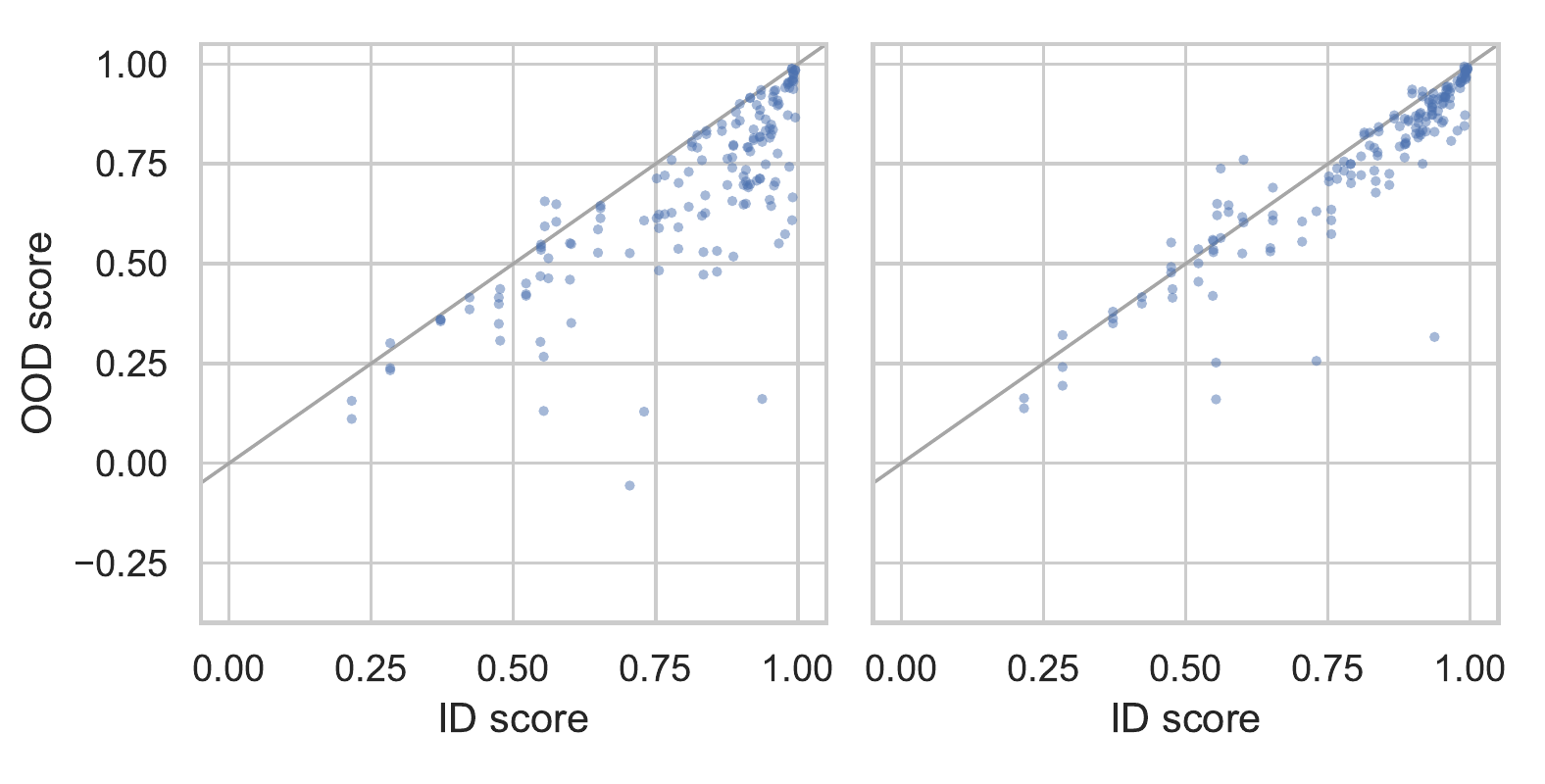}
\caption[{ID vs OOD downstream performance} with global distribution shifts. All datasets, models, and object properties are shown.]{\textbf{ID vs OOD downstream performance} with global distribution shifts. All datasets, models, and object properties are shown. Metrics: accuracy for categorical attributes, \rsq for numerical attributes. The downstream model (an MLP with one hidden layer) is tested zero-shot out-of-distribution (left) or retrained after the distribution shift has occurred (right).}
\label{fig:main_text/ood/downstream/generalization_global/scatter}
\end{figure}

For {Desideratum 2}, we train a downstream model on the original dataset and test it under global distribution shifts. These shifts generally have a negative effect on downstream property prediction (\cref{fig:main_text/ood/downstream/generalization_global/scatter}, left), although this is comparable to the effect on OOD objects when only one object is OOD. This is in agreement with the observation made in \cref{sec:results/hyp2} that these shifts negatively affect the representation, which is no longer accurate because the encoder is OOD (cf. the ``OOD2'' scenario in \citet{dittadi2021transfer}).
When retraining the downstream models on the OOD datasets while keeping the representation frozen, the performance improves slightly but does not reach the corresponding results on the training distribution (\cref{fig:main_text/ood/downstream/generalization_global/scatter}, right), as in \cref{sec:results/hyp2}. 
These observations also hold for different downstream models and with mask matching (\cref{fig:ood/downstream/slotted/match_loss/global/retrain_False,fig:ood/downstream/slotted/match_mask/global/retrain_False,fig:ood/downstream/slotted/match_loss/global/retrain_True,fig:ood/downstream/slotted/match_mask/global/retrain_True}), as well as for distributed representations (see, e.g., \cref{fig:ood/downstream/vae/match_loss/global/retrain_False/scatter}) although with similar caveats as in \cref{sec:results/hyp2}.

\textbf{Summary:}
The impact of global distribution shifts on the segmentation capability of object-centric models depends on the chosen shift; e.g., cropping consistently has a significant effect. Moreover, the usefulness for downstream tasks decreases substantially in many cases, and the performance of downstream prediction models cannot be satisfactorily recovered by retraining them.

%%%%%%%%%%%%%%%%%%%%%%%%%%%%%%%%%%%%%%%%%%%%%%%%%%%%%%%%%%%%%%%%%%%%%%
\section{Other related work}

Recent years have seen a number of systematic studies on disentangled representations~\citep{locatello2018challenging,locatello2019fairness,van2019disentangled,trauble2021disentangled}, some of which focusing on their effect on generalization~\citep{gondal2019transfer,dittadi2021transfer,montero2021the,pmlr-v89-esmaeili19a,trauble2022role}.
In the context of object-centric learning, \citet{engelcke2020reconstruction} investigate their reconstruction bottlenecks to understand how these models can separate objects from the input in an unsupervised manner. In contrast, we specifically test some key implications of learning object-centric representations.

Slot-based object-centric models can be classified according to their approach to separating the objects at a representational level~\cite{greff2020binding}.
In models that use \emph{instance slots}~\cite{greff_tagger_2016,greff2017neural,van2018relational,greff2019multi,locatello2020object,lowe2020learning, huang2020better,le2011learning,goyal2019recurrent,van2020investigating,NEURIPS2019_32bbf7b2,yang2020learning,kipf2019contrastive,racah2020slot,kipf2021conditional}, each slot is used to represent a different part of the input. This introduces a routing problem, because all slots are identical but they cannot all represent the same object, so a mechanism needs to be introduced to allow slots to communicate with each other.
In models based on \emph{sequential slots}~\cite{eslami2016air,stelzner2019supair,kosiorek2018sequential,stove,burgess2019monet,engelcke2020genesis,engelcke2021genesisv2}, the representational slots are computed in a sequential fashion, which solves the routing problem and allows to dynamically change the number of slots, but introduces dependencies between slots.
In models based on \emph{spatial slots}~\cite{nash2017multi,crawford2019spatially,jiang2020scalor,crawford2020exploiting,lin2020space,deng2021generative,lin2020improving,dittadi2019lavae}, a spatial coordinate is associated with each slot, introducing a dependency between slot and spatial location. 
In this work, we focus on four scene-mixture models as representative examples of approaches based on instance slots (Slot Attention), sequential slots (MONet and GENESIS), and spatial slots (SPACE).\looseness=-1

%%%%%%%%%%%%%%%%%%%%%%%%%%%%%%%%%%%%%%%%%%%%%%%%%%%%%%%%%%%%%%%%%%%%%%
\section{Conclusions}

In this paper, we identify three key hypotheses in object-centric representation learning: learning about objects is useful for downstream tasks, it facilitates strong generalization, and it improves overall robustness to distribution shifts.
To investigate these hypotheses,
we re-implement and systematically evaluate four state-of-the-art unsupervised object-centric learners on a suite of five common multi-object datasets.
We find that object-centric representations are generally useful for downstream object property prediction, and downstream performance is strongly correlated with segmentation quality and reconstruction error.
Regarding generalization, we observe that when a single object undergoes distribution shifts the overall segmentation quality and downstream performance for in-distribution objects is largely unaffected.
Finally, we find that object-centric models can still relatively robustly separate objects even under global distribution shifts. However, this may depend on the specific shift, and downstream performance appears to be more severely affected.

An interesting avenue for future work is to continue our systematic investigation of object-centric learning on more complex data with diverse textures, as well as a wide range of more challenging downstream tasks. 
Furthermore, it would be interesting to compare object-centric and non-object-centric models more fairly: while learning about objects offers clear advantages, the full potential of distributed representations in this context is still not entirely clear, particularly when scaling up datasets and models.
Finally, while we limit our study to unsupervised object discovery, it would be relevant to consider methods that leverage some form of supervision when learning about objects.
We believe our benchmarking library will facilitate progress along these and related lines of research.

\section*{Acknowledgements}
We would like to thank Thomas Brox, Dominik Janzing, Sergio Hernan Garrido Mejia, Thomas Kipf, and Frederik Träuble for useful comments and discussions, and the anonymous reviewers for valuable feedback.

\fi

\chapter{Conclusion}\label{chapter:conclusion}

\section{Summary}

The central theme of this thesis is learning representations that reflect the data's underlying structure---e.g., by disentangling the ground-truth generative factors of variation or by separately representing objects in a modular fashion---with little or no supervision.
In particular, we focused on the potential usefulness of these representations for learning downstream tasks, where models such as classifiers or reinforcement learning agents are trained downstream of pretrained representation functions.
Moreover, learning structured representations holds the promise of better systematic generalization, which is a significant issue in modern deep learning. In this dissertation, we investigated to what extent this may help in practice.

In the first two papers, we noted that disentanglement has been shown to be beneficial for a variety of purposes, but thorough quantitative studies have so far largely focused on (1) toy datasets and (2) simple contrived downstream tasks.
In Paper~\paperOne, we took a step towards a more practically relevant scenario: robotic manipulation. We proposed a new dataset and showed that fully disentangled representations can be learned in this more complex setting by using weak supervision and more expressive neural architectures.
We then investigated the role of disentanglement for generalization in simple downstream tasks that consist in predicting the ground-truth factors of variation.
In Paper~\paperTwo, we extended this work to challenging robotic tasks, and studied the relationship between properties of the pretrained representations, the generalization of simple downstream tasks, and the generalization of downstream reinforcement learning agents.\looseness=-1

We found that the the quality of the representations is generally affected by distribution shifts in the data, largely due to a lack of robustness of the learned representation function. We called this scenario ``OOD2'' and showed that training with input noise is a simple but effective strategy to improve encoder robustness. This is beneficial both for simple downstream prediction tasks and for more complex robotic manipulation, and allows for zero-shot sim-to-real transfer in both cases. Crucially, we did not find evidence of other representation properties being particularly helpful for this generalization scenario.

In contrast, when the representation function is in distribution but the downstream model is out of distribution, we are effectively \emph{testing the OOD generalization of the downstream model itself} because, in this case, the representations can be assumed to be accurate and meaningful. With this test scenario, which we called ``OOD1'', we want to answer the following question: \emph{are there properties of the pretrained representations that lead downstream models to be generally robust to systematic distribution shifts?}
A positive answer comes from Paper~\paperOne, where we observed that representation functions that \emph{perfectly} separate the true factors of variation tend to lead to robust downstream models for factor prediction (as long as the encoder remains in distribution). On the other hand, when the representations are not fully disentangled, their degree of entanglement does not seem to affect downstream generalization.

In the more challenging robotic tasks in Paper~\paperTwo, disentanglement does not seem to be beneficial for downstream policies. However, we found that, given a pretrained representation function, downstream factor predictors and reinforcement learning agents generalize in similar ways. This can be very convenient in practice: to obtain a downstream model that is robust to a specific class of distribution shifts, we can pre-select promising upstream representation functions using a simpler proxy task where we mirror the desired generalization scenarios of the target downstream task.

Finally, in Paper~\paperThree, we considered images with multiple objects and studied the implications of learning object-centric representations, with a particular focus on generalization.
We observed that object-centric models that successfully separate objects learn useful representations for simple downstream set-prediction tasks. Regarding generalization, we found that object-centric models are particularly robust to distribution shifts that are in some sense related to the compositional structure of an image---e.g., when the number of objects increases, or when a single object is out of distribution---while the picture appears less clear when the shifts affect the data in a less structured manner.

\section{Discussion}

In this dissertation, we have studied representations that explicitly encode some of the structure found in the data. Disentangled representations (at least in their traditional formulation) separate factors of variation; object-centric representations, which are strictly related to the former, focus on separating objects and are more suitable for multi-object settings.
A typical argument in favor of explicitly representing structure in this manner is that it should facilitate systematic generalization, thereby narrowing the gap with human-level intelligence.
However, based on the findings and discussions in this dissertation, a few remarks are in order.

First, while it may generally be true that encoding structure is beneficial for generalization, the role of disentanglement appears to depend on both the downstream task and the downstream model.
A relevant direction for future work includes validating our results on a broader range of representation learning methods, downstream tasks (e.g., abstract reasoning), and downstream models (e.g., different reinforcement learning algorithms). 
    
Second, in the multi-object setting, unstructured models (variational autoencoders in our case) seem to generalize better than expected, although object-centric approaches still have an edge due to their explicit inductive biases. 
Since quantitative comparisons between these two classes of models are not entirely fair in our setup, future work should attempt to fill this gap in order to clarify more precisely when and how structured models are more useful than unstructured ones.

Finally, the argument that structured representations benefit generalization is not necessarily valid when the representation function itself is out of distribution (the OOD2 setting), since in that case it may not even be encoding information correctly. In fact, we have shown in a variety of settings---both disentanglement and object-centric learning; both in property prediction and in robotic tasks; and under various distribution shifts---that the main bottleneck for OOD2 generalization is the robustness of the encoder, rather than the format of the representations it computes.
On the other hand, our study on object-centric learning in Paper~\paperThree suggests there may be some useful inductive biases in object-centric models that make them relatively robust to some distribution shifts (e.g., one OOD object). An interesting avenue for future work is to systematically investigate these biases.
As an orthogonal but related direction, it would be valuable to study how different choices in the architecture, objective, and optimization affect the learning of useful and modular object representations, in order to discover inductive biases that could enable scaling object-centric learning to real data with minimal supervision.

%%%%%%%%%%%%%%%%%%%%%%%%%%%%%%%%%%%%%%%%%%%%%%%%%%%%%%%%%%%%%%%%%%%%%%%%%%%%%%%%%%%%%

\appendixpage\appendix
\pagestyle{appruled} % Comment to remove the word "Appendix" from the header
\iffull
\chapter{Supplementary material for Chapter~\ref{chapter:iclr2021}}

\section{Implementation Details}\label{sec:implementation_details}

\paragraph{Training.}
We train the $\beta$-VAEs by maximizing the following objective function:
\begin{equation*}
\mathcal{L}^{\beta}_{VAE} = \mathbb{E}_{q_\phi(\bm{z}|\bm{x})} [ \log p_\theta (\bm{x} | \bm{z}) ] 
- \beta D_\mathrm{KL}( q_\phi(\bm{z}|\bm{x}) \| p (\bm{z}) ) \leq  \log p(\bm{x})
\end{equation*}
with $\beta > 0$ using the Adam optimizer \citep{kingma2014adam} with default parameters. We use a batch size of 64 and train for 400k steps. The learning rate is initialized to 1e-4 and halved at 150k and 300k training steps. We clip the global gradient norm to $1.0$ before each weight update.
Following \citet{locatello2018challenging}, we use a Gaussian encoder with an isotropic Gaussian prior for the latent variable, and a Bernoulli decoder.
Our implementation of weakly supervised learning is based on Ada-GVAE \citep{locatello2020weakly}, but uses a symmetrized KL divergence:
\begin{equation*}
    \tilde{D}_{\mathrm{KL}}(p , q) = \frac{1}{2} D_{\mathrm{KL}}(p \| q) + \frac{1}{2} D_{\mathrm{KL}}(q \| p) 
\end{equation*}
to infer which latent dimensions should be aggregated.

The noise added to the encoder's input consists of two independent components, both iid Gaussian with zero mean: one is independent for each subpixel (RGB) and has standard deviation $0.03$, the other is a $8 \times 8$ pixel-wise (greyscale) noise with standard deviation $0.15$, bilinearly upsampled by a factor of 16. The latter has been designed (by visual inspection) to roughly mimic observation noise in the real images due to complex lighting conditions.

\paragraph{Neural architecture.}
Architectural details are provided in \cref{tab:architecture_encoder_decoder,tab:architecture_residual}, and \cref{fig:architecture} provides a high-level overview. In preliminary experiments, we observed that batch normalization, layer normalization, and dropout did not significantly affect performance in terms of ELBO, model samples, and disentanglement scores, both in the unsupervised and weakly supervised settings. On the other hand, layer normalization before the posterior parameterization (last layer of the encoder) appeared to be beneficial for stability in early training.
While using an architecture based on residual blocks leads to fast convergence, in practice we observed that it may be challenging to keep the gradients in check at the beginning of training.\footnote{This instability may also be exacerbated in probabilistic models by the sampling step in latent space, where a large log variance causes the decoder input to take very large values. Intuitively, this might be a reason why layer normalization before latent space appears to be beneficial for training stability.} 
In order to solve this issue, we resorted to a simple scalar gating mechanism in the residual blocks \citep{bachlechner2020rezero} such that each residual block is initialized to the identity.

\paragraph{Datasets and OOD evaluation.} 
Because we evaluate OOD generalization in terms of cube color hue (except in the sim2real case), we first sampled 8 color hues at random from the 12 specified in \cref{tab:dataset_fov}. The chosen hues are: $[0^{\circ}, 120^{\circ}, 150^{\circ}, 180^{\circ},$ $210^{\circ}, 270^{\circ}, 300^{\circ}, 330^{\circ}]$. Then, the dataset $D$ used for training VAEs is generated by randomly sampling values for the factors of variation from \cref{tab:dataset_fov}, with the color hue restricted to the above-mentioned values. This makes OOD2 evaluation possible, specifically OOD2-A where the learned predictors are tested on representations extracted from images with held-out values of the cube hue.

For evaluation of out-of-distribution generalization, we train the downstream predictors on a subset $D_1 \subset D$ of the representation training set. The downstream training set $D_1$ is sampled at random from $D$ but only contains a (not necessarily proper) subset of the 8 cube colors. This subset contains 1 color in the OOD1-A case, 4 colors in OOD1-B and OOD1-C, all 8 colors in OOD2 (in this case $D_1$ is simply a random subset of $D$).
Then we test the downstream predictors on a set $D_2$ distributionally different from $D_1$ in terms of cube color (all OOD1 scenarios as well as OOD2-A) or sim2real (OOD2-B). In the OOD1 case, $D_2$ is also a subset of $D$ and is generated the same way. In each OOD1 case, the test set $D_2$ is paired with its corresponding $D_1$ that was used to train the downstream predictors. $D_2$ contains all colors in $D$ minus those in $D_1$. In the OOD2-A case, $D_2$ is a separate dataset containing 5k simulated images like those in $D$, except that these only contain the 4 colors that were left out from the VAE training set $D$ (hue in $[30^{\circ}, 60^{\circ}, 90^{\circ}, 240^{\circ}]$). In the OOD2-B case, the set $D_2$ is the dataset of real images.
Following previous work (e.g. the GBT10000 metric in \citet{locatello2018challenging}), the training set $D_1$ and test set $D_2$ for downstream tasks contain 10k and 5k images, respectively, except in the OOD2-B case, where the size is limited by the size of the real dataset.

\let\oldtimes\times
\renewcommand{\times}{\!\oldtimes\!}

\begin{table}
\centering
\caption[Encoder and decoder architectures.]{Encoder (left) and decoder (right) architectures. The latent space dimensionality is denoted by $d$, and $K=3$ indicates the number of image channels. Last line in the encoder architecture: the fully connected layer parameterizing the log variance of the approximate posterior distributions of the latent variables has custom initialization. The weights are initialized with $1/10$ standard deviation than the default value, and the biases are initialized to $-1$ instead of $0$. Empirically, this together with (learnable) LayerNorm was beneficial for training stability at the beginning of training.}
\label{tab:architecture_encoder_decoder}
\begin{minipage}[t]{.48\linewidth}
    \strut\vspace*{-\baselineskip}\newline%
    \centering
    
    \scalebox{0.9}{
    \begin{tabular}{ll}
        \toprule
        \multicolumn{2}{c}{\textbf{Encoder}}\\
        \toprule
        \textbf{Operation} & \textbf{Output Shape}\\
        \midrule
        Input & $128 \times 128 \times K$ \\
        Conv 5x5, stride 2, 64 ch. & $64 \times 64 \times 64$\\
        LeakyReLU(0.02) & ---\\
        2x ResidualBlock(64) & ---\\
        Conv 1x1, 128 channels & $64 \times 64 \times 128$\\
        AveragePool(2)  & $32 \times 32 \times 128$\\
        2x ResidualBlock(128) &  ---\\
        AveragePool(2) &  $16 \times 16 \times 128$\\
        2x ResidualBlock(128) & ---\\
        Conv 1x1, 256 channels & $16 \times 16 \times 256$\\
        AveragePool(2)  & $8 \times 8 \times 256$\\
        2x ResidualBlock(256) & ---\\
        AveragePool(2)  & $4 \times 4 \times 256$\\
        2x ResidualBlock(256)  & ---\\
        Flatten & $4096$\\
        LeakyReLU(0.02) & ---\\
        FC(512) & $512$\\
        LeakyReLU(0.02) & ---\\
        LayerNorm & ---\\
        2x FC($d$) & $2d$\\
        \bottomrule
    \end{tabular}
    }
\end{minipage}
\hfill
\begin{minipage}[t]{.48\linewidth}
    \strut\vspace*{-\baselineskip}\newline%
    \centering
    
    \scalebox{0.9}{
    \begin{tabular}{ll}
        \toprule
        \multicolumn{2}{c}{\textbf{Decoder}}\\
        \toprule
        \textbf{Operation} & \textbf{Output Shape}\\
        \midrule
        Input & $d$ \\
        FC(512) & $512$\\
        LeakyReLU(0.02) & ---\\
        FC(4096) & $4096$\\
        Reshape & $4 \times 4 \times 256$\\
        2x ResidualBlock(256)  & ---\\
        BilinearInterpolation(2) & $8 \times 8 \times 256$\\
        2x ResidualBlock(256)  & ---\\
        Conv 1x1, 128 channels & $8 \times 8 \times 128$\\
        BilinearInterpolation(2) & $16 \times 16 \times 128$\\
        2x ResidualBlock(128)  & ---\\
        BilinearInterpolation(2) & $32 \times 32 \times 128$\\
        2x ResidualBlock(128)  & ---\\
        Conv 1x1, 64 channels & $32 \times 32 \times 64$\\
        BilinearInterpolation(2) & $64 \times 64 \times 64$\\
        2x ResidualBlock(64)  & ---\\
        BilinearInterpolation(2) & $128 \times 128 \times 64$\\
        LeakyReLU(0.02) & ---\\
        Conv 5x5, $K$ channels & $128 \times 128 \times K$\\
        \bottomrule
    \end{tabular}
    }
\end{minipage}
\end{table}

\begin{table}
    \centering
    \caption[Architecture of a residual block.]{Architecture of a residual block. The scalar gate is implemented by multiplying the tensor by a learnable scalar parameter before adding it to the block input. Initializing the residual block to the identity by setting this parameter to zero has been originally proposed by \citet{bachlechner2020rezero}. The tensor shape is constant throughout the residual block.}
    \label{tab:architecture_residual}
    \scalebox{0.9}{
    \begin{tabular}{l}
        \toprule
        \textbf{Residual Block}\\
        \midrule
        Input: shape $H \times W \times C$ \\
        LeakyReLU(0.02) \\
        Conv 3x3, $C$ channels \\
        LeakyReLU(0.02) \\
        Conv 3x3, $C$ channels \\
        Scalar gate \\
        Sum with input\\
        \bottomrule
    \end{tabular}
    }
\end{table}

\renewcommand{\times}{\oldtimes}

\begin{figure}
    \centering
    \includegraphics[width=\linewidth]{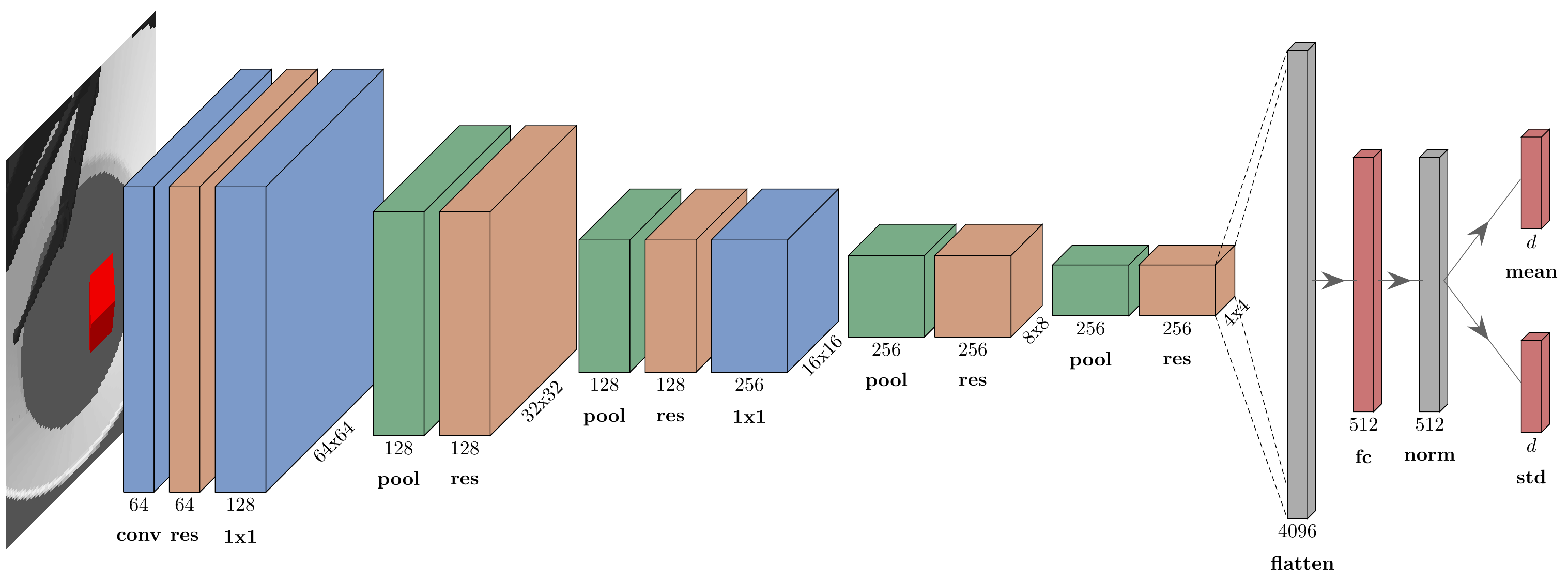}
    
    \vspace{0.8cm}
    \includegraphics[width=\linewidth]{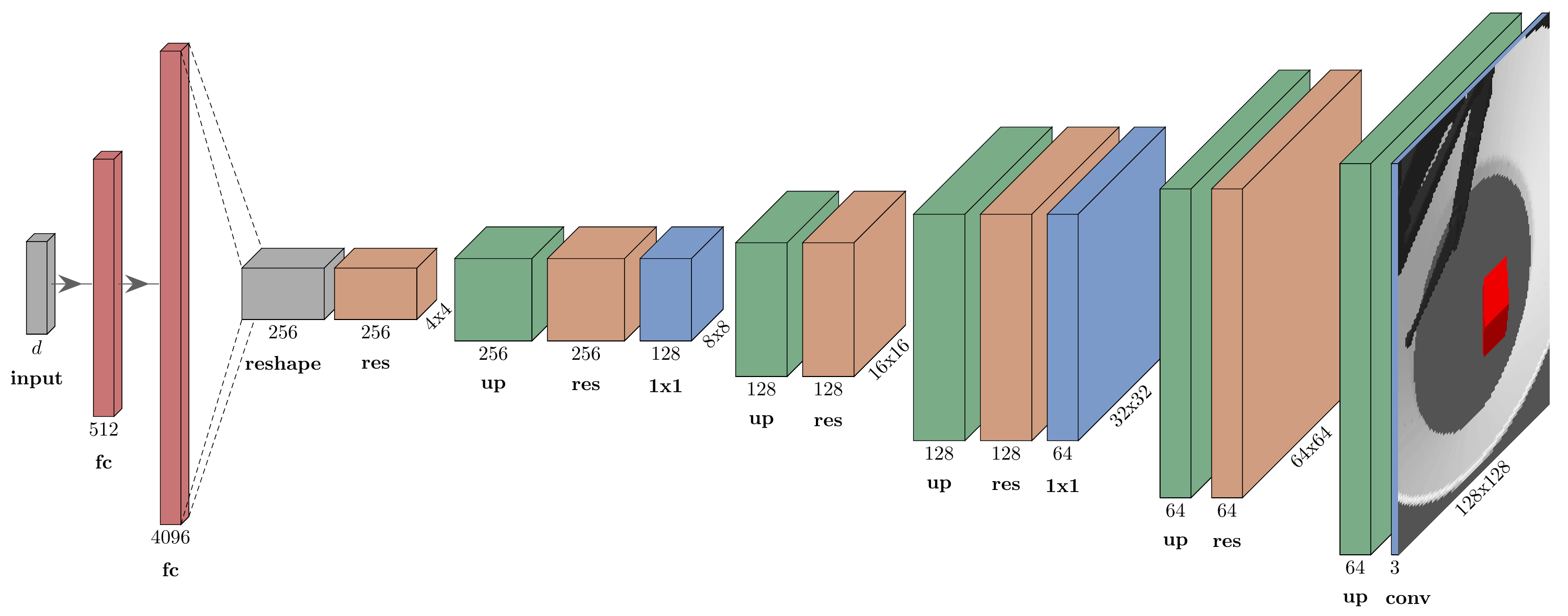}
    \caption[Schemes of the encoder and decoder architectures.]{Schemes of the encoder (top) and decoder (bottom) architectures. In both schemes, information flows left to right. Blue blocks represent convolutional layers: those labeled ``conv'' have 5x5 kernels and stride 2, while those labeled ``1x1'' have 1x1 kernels. Each orange block represents a pair of residual blocks (implementation details of a residual block are provided in \cref{tab:architecture_residual}). Green blocks in the encoder represent average pooling with stride 2, and those in the decoder denote bilinear upsampling by a factor of 2. Red blocks represent fully-connected layers. The block labeled ``norm'' indicates layer normalization. Dashed lines denote tensor reshaping.}
    \label{fig:architecture}
\end{figure}

\clearpage

\section{Additional Results}\label{sec:additional_results}

\subsection{Dataset Correlations}\label{sec:app_dataset_corr}

\begin{figure}[h]
    \centering
    \includegraphics[width=0.56\linewidth]{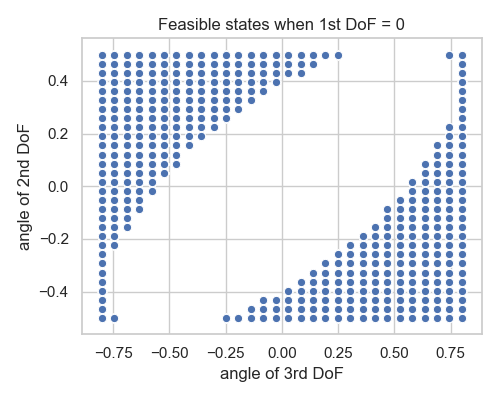}
    \caption{Feasible states of the 2nd and 3rd DoF when the angle of the 1st DoF is $0$. Angles are in radians.}
    \label{fig:joints_scatter}
\end{figure}

\vspace{1cm}
\begin{figure}[h]
    \centering
    \includegraphics[width=0.56\linewidth]{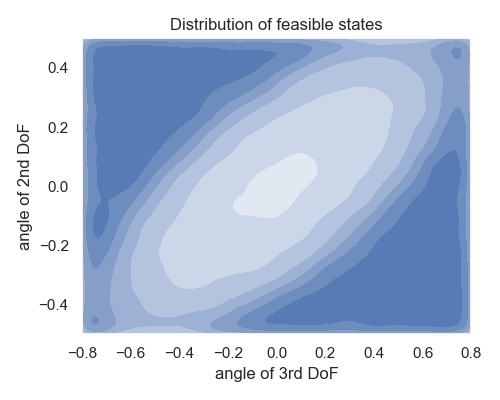}
    \caption{Density of feasible states of 2nd and 3rd DoF over the whole training dataset. Darker shades of blue indicate regions of higher density. Angles are in radians.}
    \label{fig:joints_kde}
\end{figure}

\clearpage

\subsection{Samples and Reconstructions}\label{sec:app_additional_modelviz}

\begin{figure}[h]
\centering
    \includegraphics[width=0.45\linewidth]{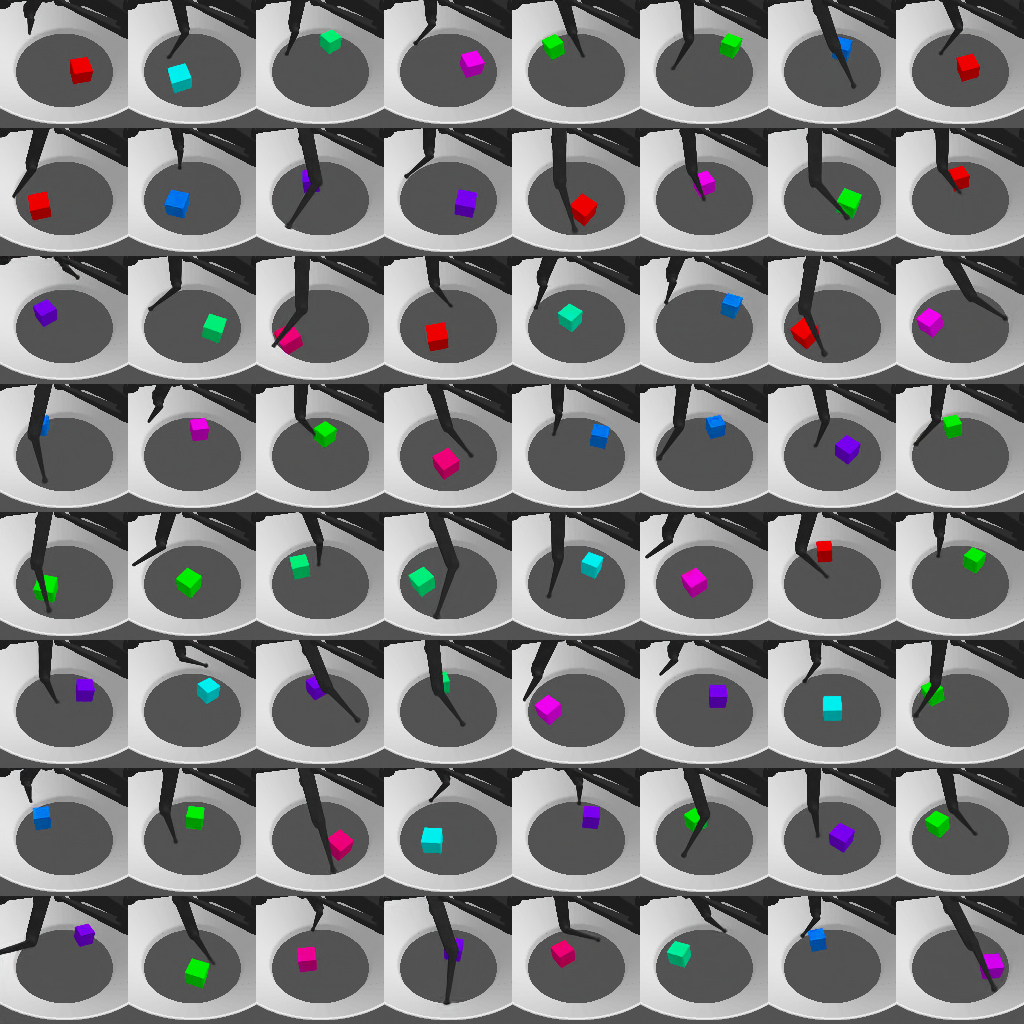}
\caption{Samples generated by a trained model (selected based on the ELBO).}
\label{fig:model_samples}
\end{figure}
\begin{figure}[h]
\centering
    \includegraphics[width=0.95\linewidth]{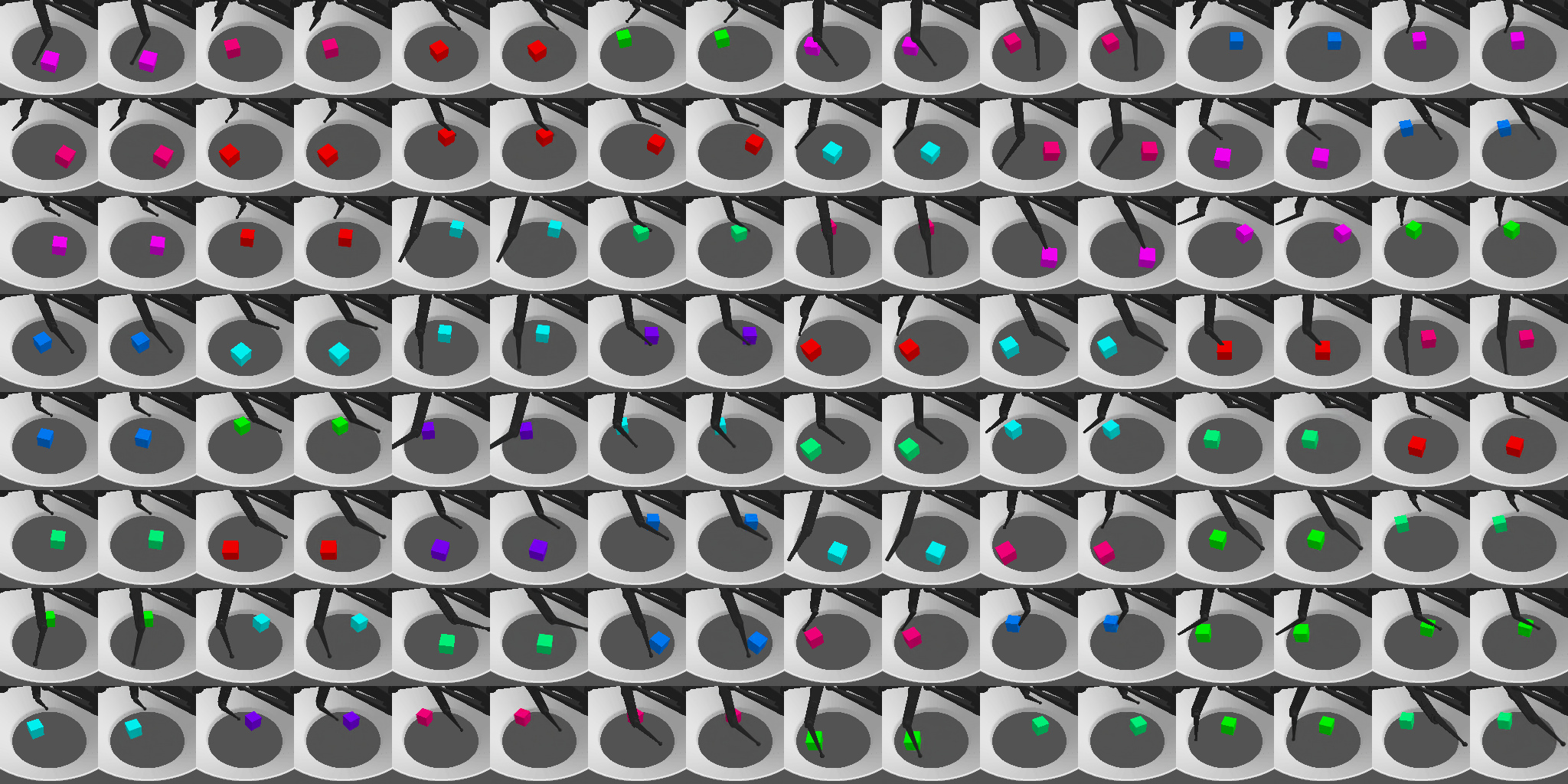}
\caption{Input reconstructions by a trained model. This model was selected based on the ELBO. Image inputs are on odd columns, reconstructions on even columns.}
\label{fig:model_reconstructions}
\end{figure}

\clearpage

\subsection{Latent Traversals}\label{sec:app_additional_traversals}

\begin{figure}[h]
  \centering
  \includegraphics[height=0.25\textheight]{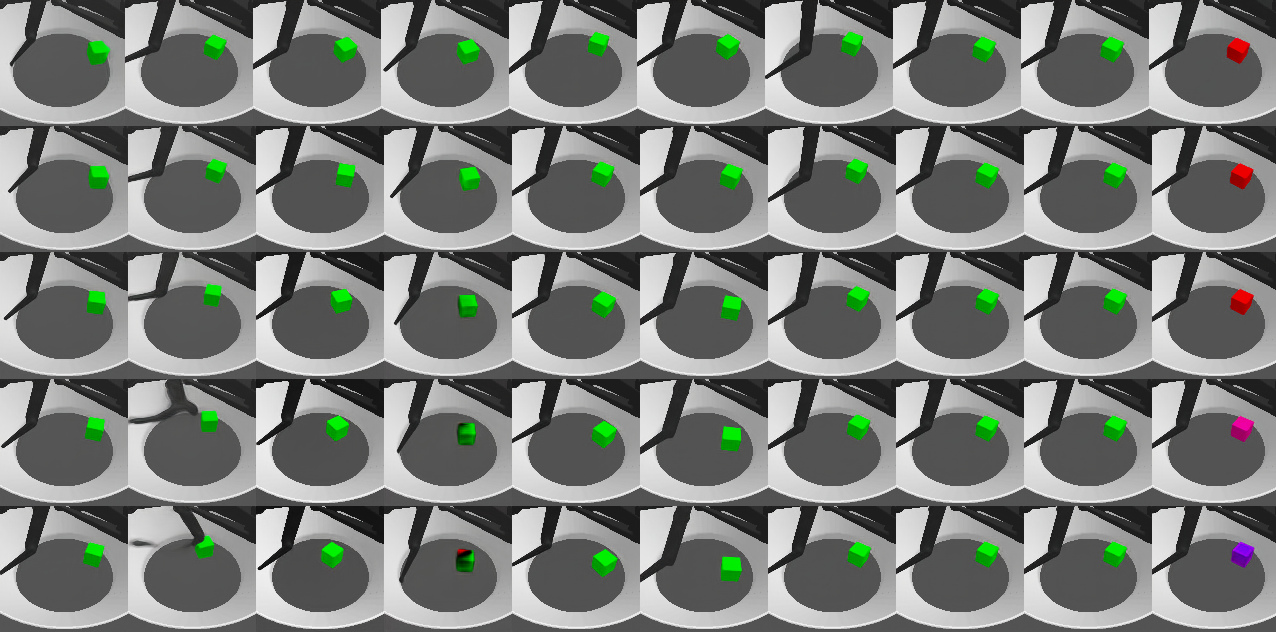}
  \vspace{10pt}
  
  \includegraphics[height=0.25\textheight]{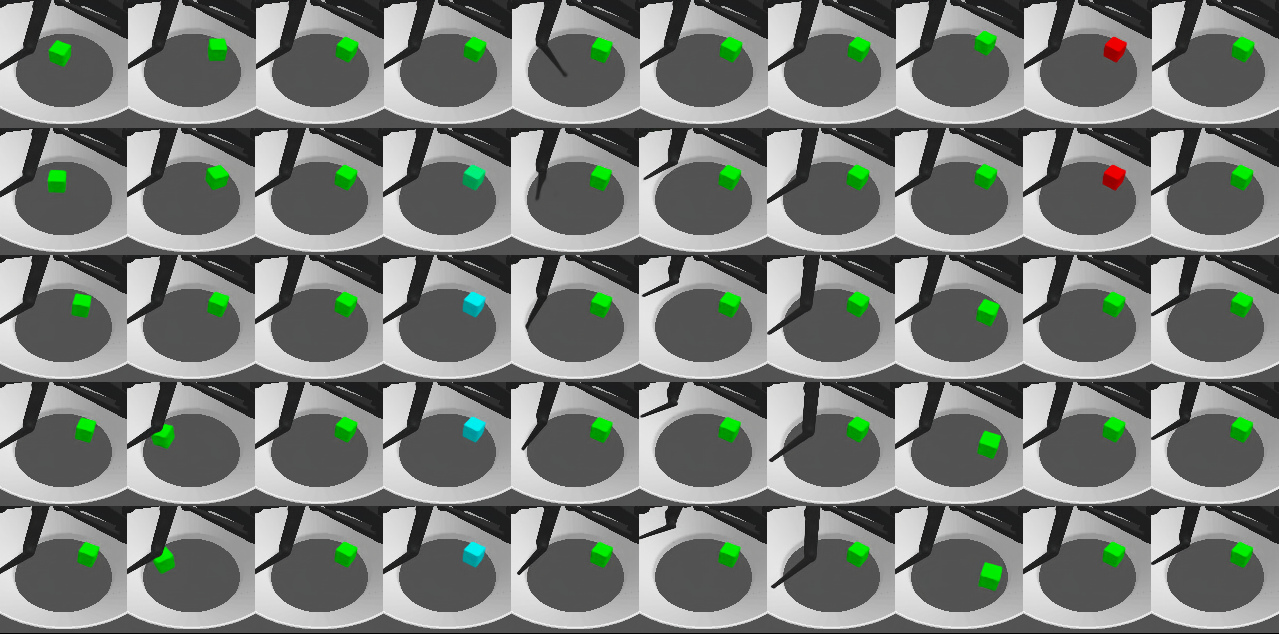}
  \vspace{10pt}
  
  \includegraphics[height=0.25\textheight]{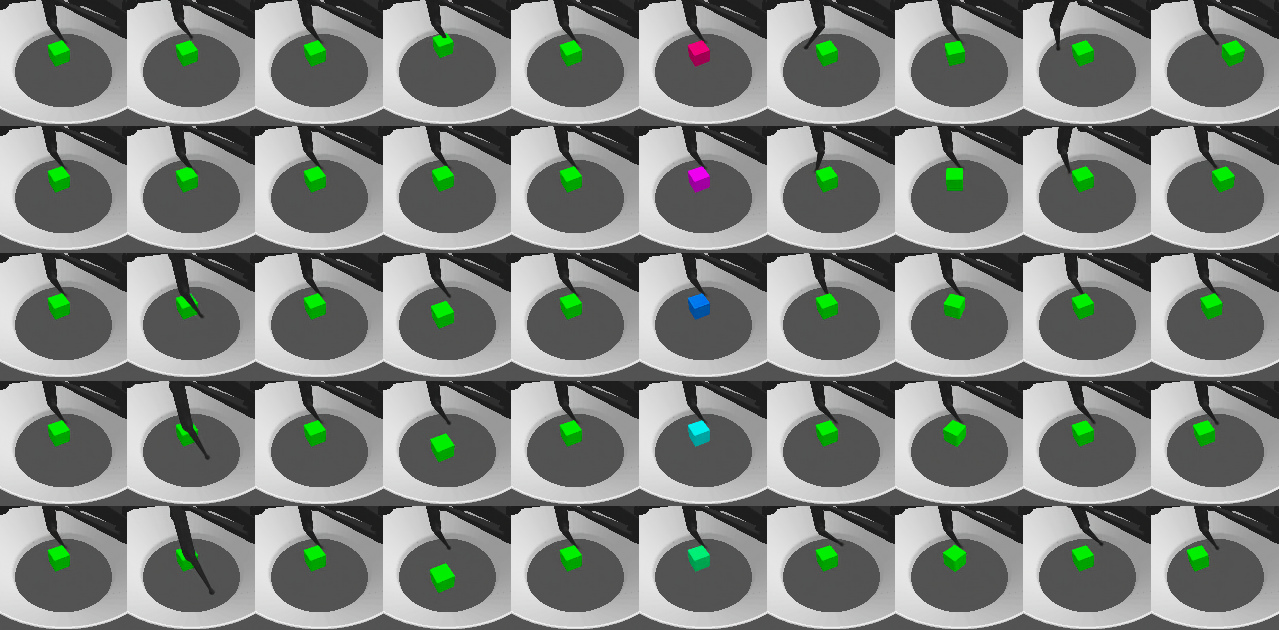}
  
  \caption{From top to bottom: latent traversals for a model with low ($0.15$), medium ($0.5$), and high ($1.0$) DCI score.}
  \label{fig:latent_traversals_appendix}
\end{figure}

\clearpage

\subsection{Unsupervised Metrics and Disentanglement}\label{sec:app_scatter_plots}

\begin{figure}[h]
\centering
    \includegraphics[width=0.48\linewidth]{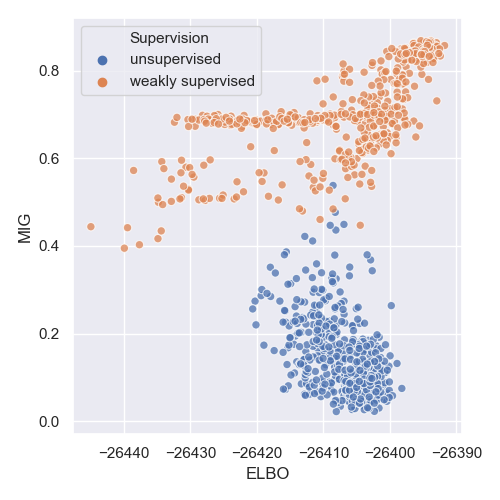}\hfill
    \includegraphics[width=0.48\linewidth]{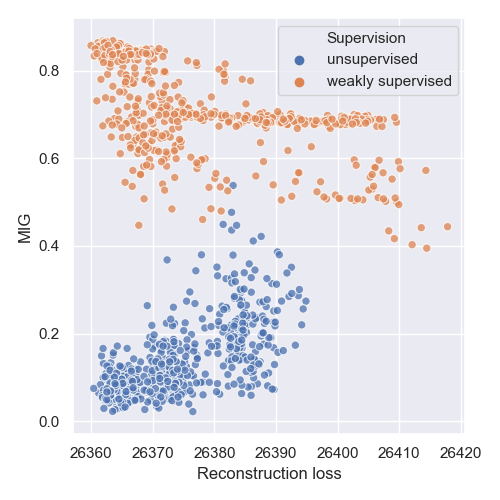}
    
    \vspace{1cm}
    \includegraphics[width=0.48\linewidth]{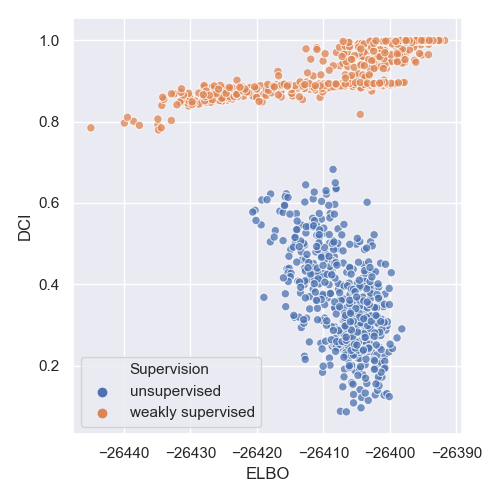}\hfill
    \includegraphics[width=0.48\linewidth]{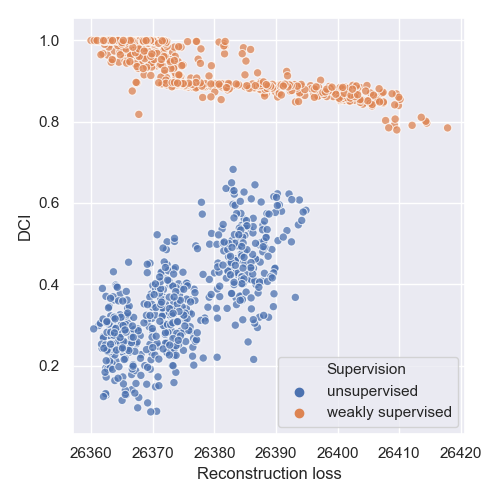}
\caption{Scatter plots of unsupervised metrics (left: ELBO; right: reconstruction loss) vs disentanglement (top: MIG; bottom: DCI) for 1,080 trained models, color-coded according to supervision. Each point represents a trained model.}
\label{fig:scatter_metrics}
\end{figure}

\clearpage

\subsection{Out-of-Distribution Transfer}\label{sec:app_additional_ood}

\begin{figure}[h]
\centering
\begin{minipage}[c]{\linewidth}
\centering
    \includegraphics[width=0.9\linewidth]{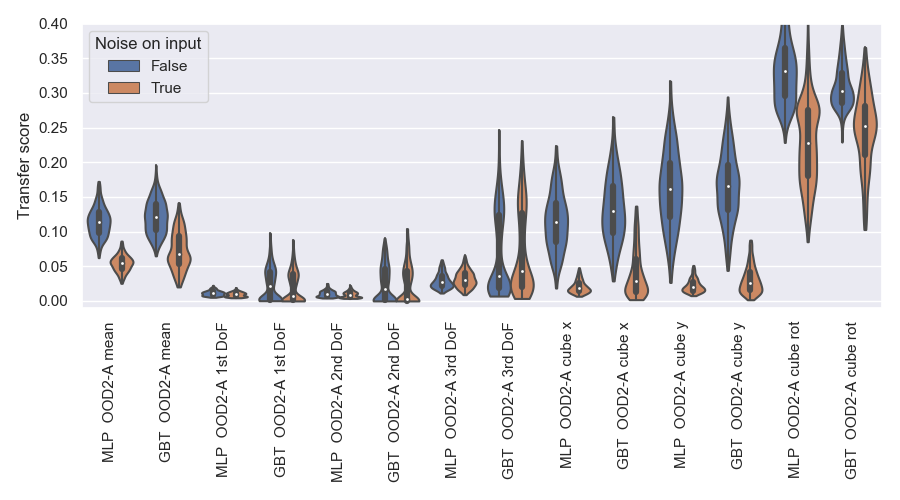}
    
    \includegraphics[width=0.9\linewidth]{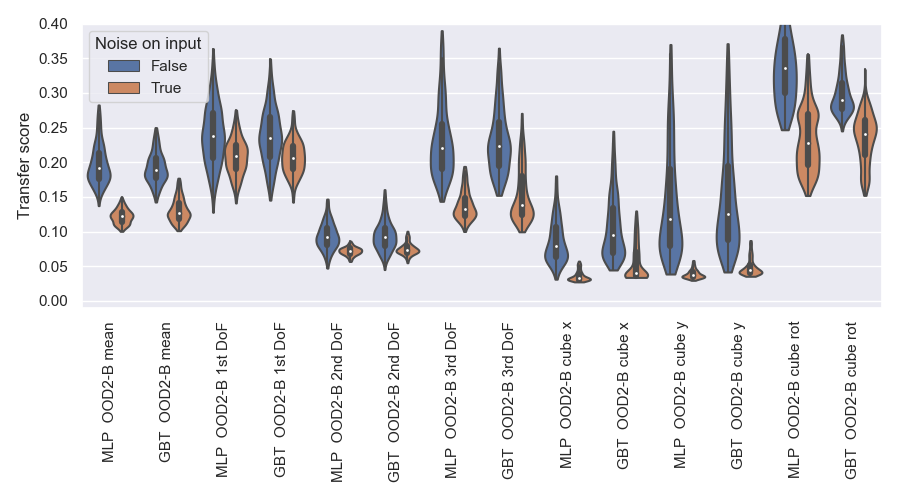}
\end{minipage}
\caption[Transfer metric in OOD2-A and OOD2-B settings, decomposed according to the factor of variation and presence of input noise.]{Transfer metric in OOD2-A (top) and OOD2-B (bottom) settings, decomposed according to the factor of variation and presence of input noise. When noise is added to the input during training, the inferred cube position error is relatively low (the scores are the mean absolute error, and they are normalized to $[0, 1]$). This is particularly useful in the OOD2-B setting (real world) where the joint state is anyway considered known, while object position has to be inferred with tracking methods.}
\label{fig:violin_transfer_factors_ood2}
\end{figure}

\clearpage

\subsection{Out-of-Distribution Reconstructions}\label{sec:app_additional_ood_recons}

\begin{figure}[h]
  \centering
  \includegraphics[height=0.25\textheight]{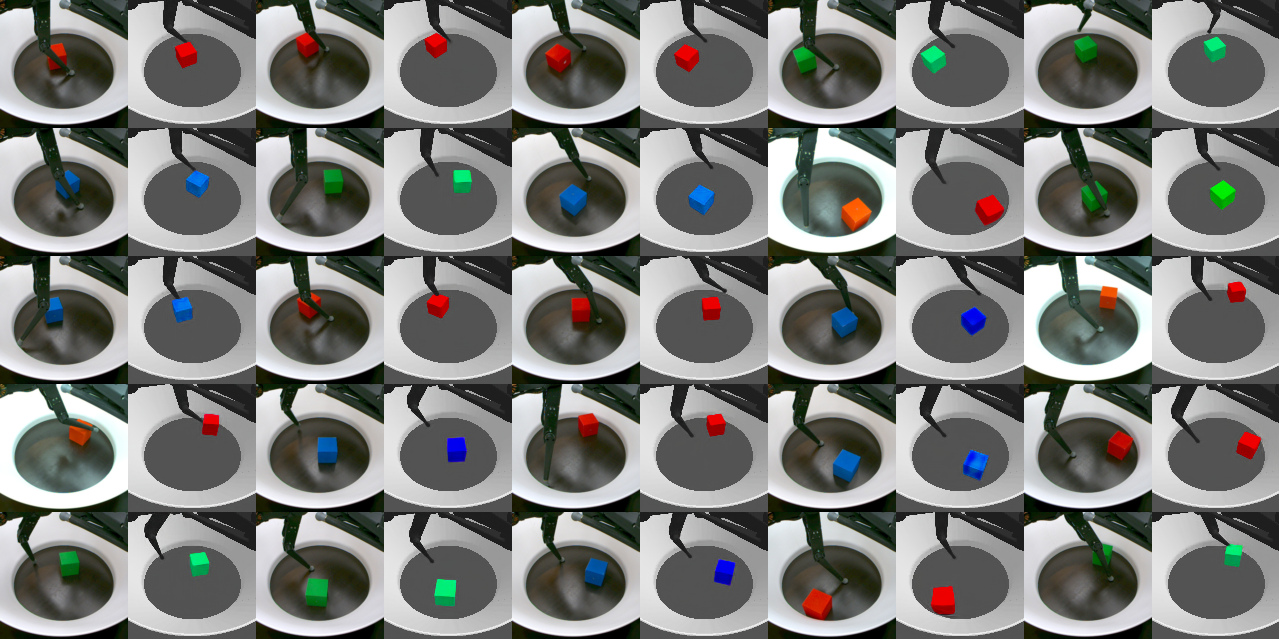}
  \vspace{10pt}
  
  \includegraphics[height=0.25\textheight]{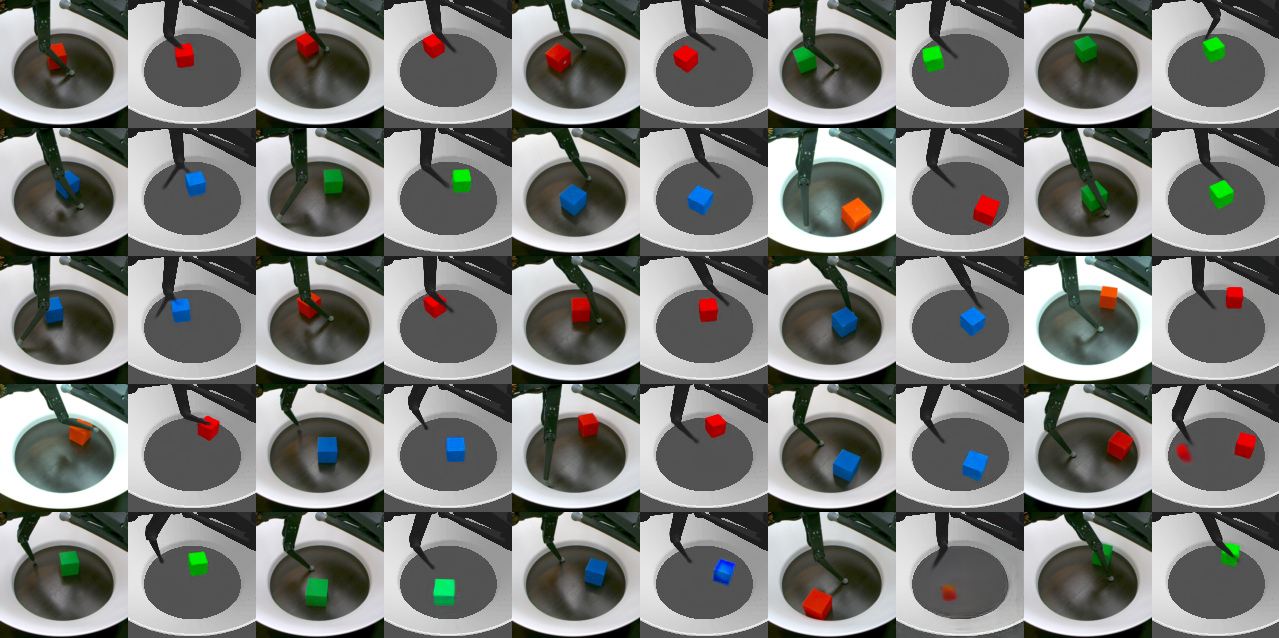}
  \vspace{10pt}
  
  \includegraphics[height=0.25\textheight]{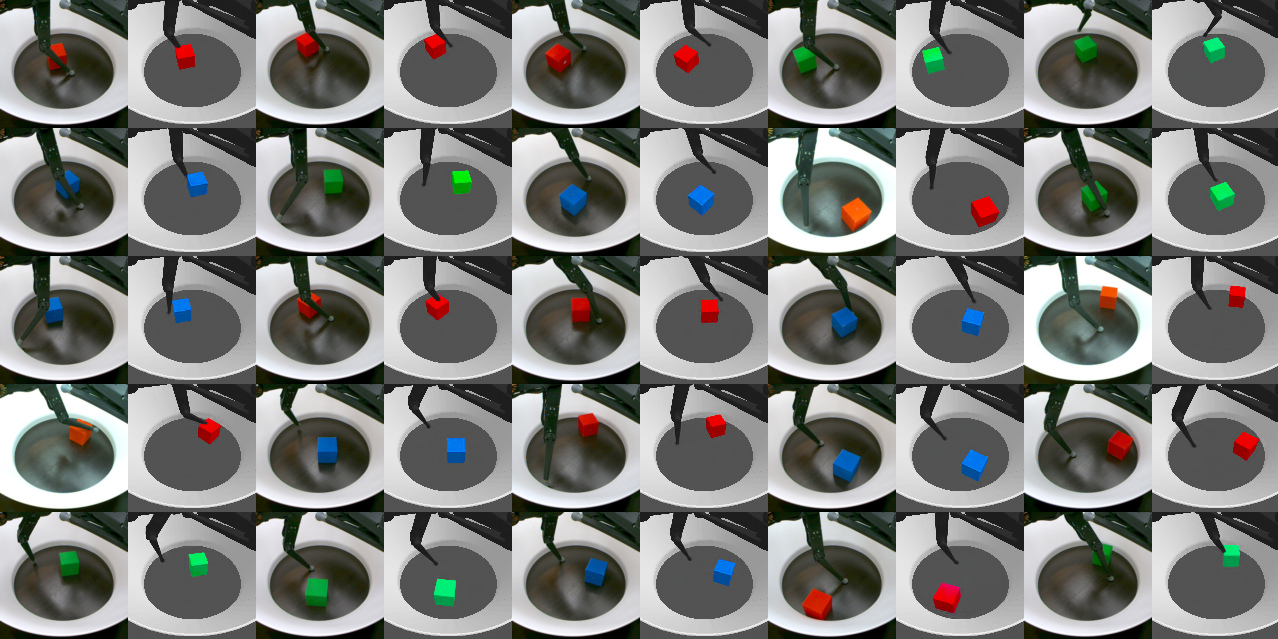}
  
  \caption{From top to bottom: reconstructions of real-world images (OOD2-B) for a model with low ($0.15$), medium ($0.5$), and high ($1.0$) DCI score.}
  \label{fig:real_reconstructions_appendix}
\end{figure}

\begin{figure}
  \centering
  \includegraphics[height=0.25\textheight]{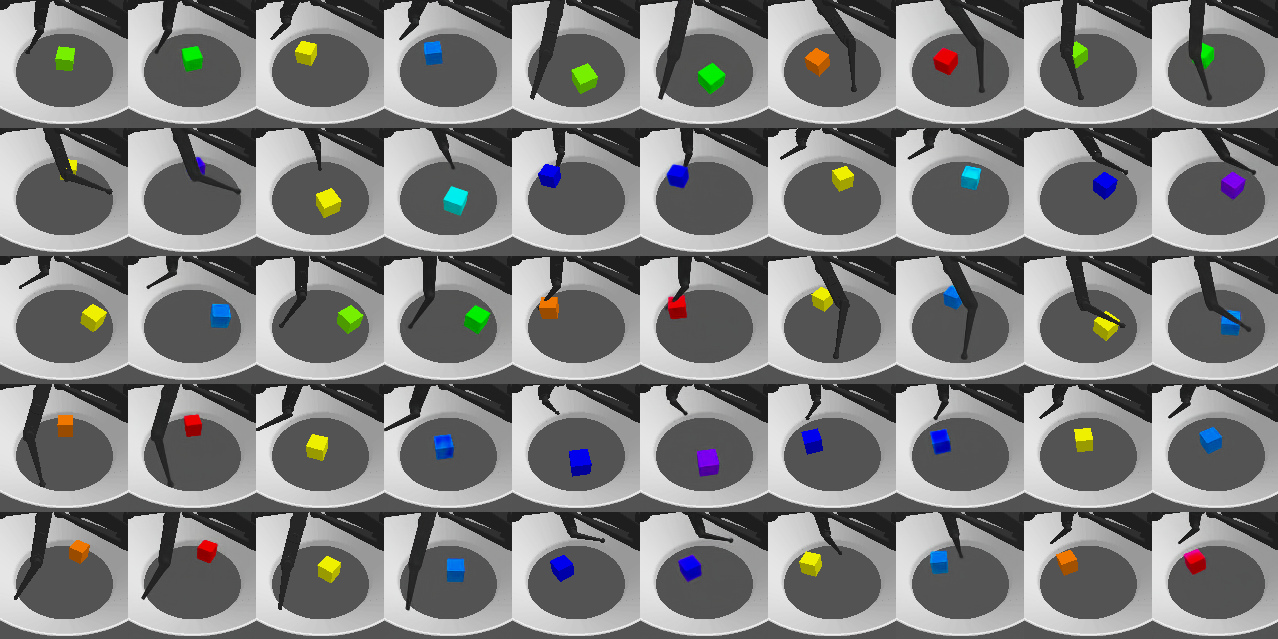}
  \vspace{10pt}
  
  \includegraphics[height=0.25\textheight]{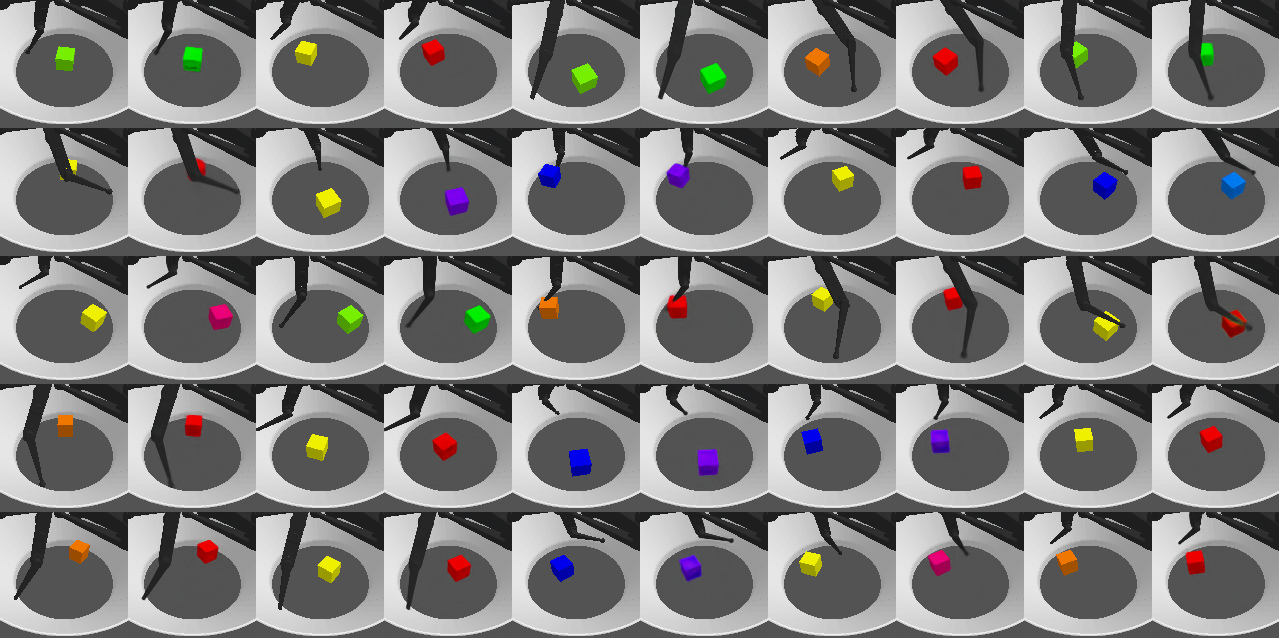}
  \vspace{10pt}
  
  \includegraphics[height=0.25\textheight]{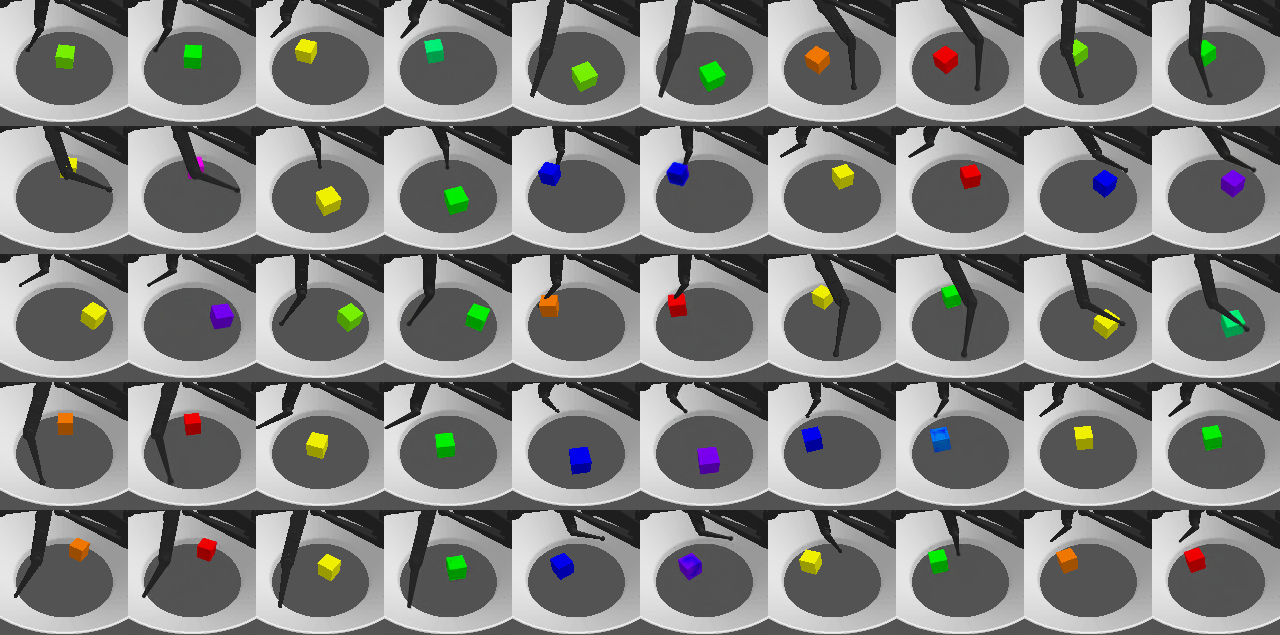}
  
  \caption{Reconstructions of simulated images with out-of-distribution encoder (OOD2-A) for a model with low ($0.15$), medium ($0.5$), and high ($1.0$) DCI score.}
  \label{fig:ood2a_reconstructions_appendix}
\end{figure}

\chapter{Supplementary material for Chapter~\ref{chapter:iclr2022}}  % cref gives problems with hyperref

\section{Implementation details}
\label{app:implementation_details}

\paragraph{Task definitions and reward structure. }
We derive both tasks, \textit{object reaching} and \textit{pushing}, from the CausalWorld environments introduced by \citet{ahmed2020causalworld}. We pretrain representations on the dataset introduced by \citet{dittadi2021transfer}, and allow only one finger to move in our RL experiments. We introduce the \textit{object reaching} environment that involves an unmovable cube. 
We used reward structures similar to those in \citet{ahmed2020causalworld}:
\begin{itemize}
    \item \textit{object reaching}: $ r_t = -750 \left[ d(g_t,e_t) - d(g_{t-1},e_{t-1}) \right]$
    \item \textit{pushing}: $ r_t = -750 \left[ d(o_t,e_t) - d(o_{t-1},e_{t-1}) \right] -250 \left[ d(o_t,g_t) - d(o_{t-1},g_{t-1}) \right] + \rho_t$ 
\end{itemize}
where $t$ denotes the time step, $\rho_t\in [0,1]$ is the fractional overlap with the goal cube at time $t$, $e_{t} \in \mathbf{R}^3$ is the end-effector position, $o_{t} \in \mathbf{R}^3$ the cube position, $g_{t} \in \mathbf{R}^3$ the goal position, and $d(\cdot,\cdot)$ denotes the Euclidean distance. The cube in \textit{object reaching} is fixed, i.e. $o_{t} = g_{t}$ for all $t$. The time limit is 2 seconds in \textit{object reaching} and 4 seconds in \textit{pushing}.

\paragraph{Success metrics.}
Besides the accumulated reward along episodes, that is determined by the reward function, we also report two reward-independent normalized success definitions for better interpretability:
In \textit{object reaching}, the success metric indicates progress from the initial end effector position to the optimal distance from the center of the cube. It is 0 if the final distance is greater than or equal to the initial distance, and 1 if the end effector is touching the center of a face of the cube. In \textit{pushing}, the success metric is defined as the volumetric overlap of the cube with the goal cube, and the task can be visually considered solved with a score around 80\%. We observed that accumulated reward and success are very strongly correlated, thus allowing us to use one or the other for measuring performance.

\paragraph{Training and evaluation details.}

During training, the goal position is randomly sampled at every episode. Similarly, the object color is sampled from the 4 specified training colors from $\dataset_1$ that correspond to the OOD1-B split from \citet{dittadi2021transfer}.

For each policy evaluation (in-distribution and out-of-distribution variants), we average the accumulated reward and final success over 200 episodes with randomly sampled cube positions and the respective object color in both tasks.

\paragraph{SAC implementation.}
Our implementation of SAC builds on \texttt{stable-baselines}, a Python package introduced by~\citet{stable-baselines}. We use the same SAC hyperparameters used for pushing in \citet{ahmed2020causalworld}. When using L1 regularization, we add to the loss function the L1 norm of the first layers of all MLPs, scaled by a coefficient $\alpha$. We gradually increase regularization by linearly annealing $\alpha$ from 0 to $5\cdot10^{-7}$ over 200,000 time steps in \textit{object reaching}, and from 0 to $6\cdot10^{-8}$ over 3,000,000 time steps in \textit{pushing}.

\section{Additional results}
\label{app:additional_results}

\subsection{Training environment}
\label{app:additional_results_train_environment}

\begin{figure}
    \centering
    \includegraphics[width=\linewidth]{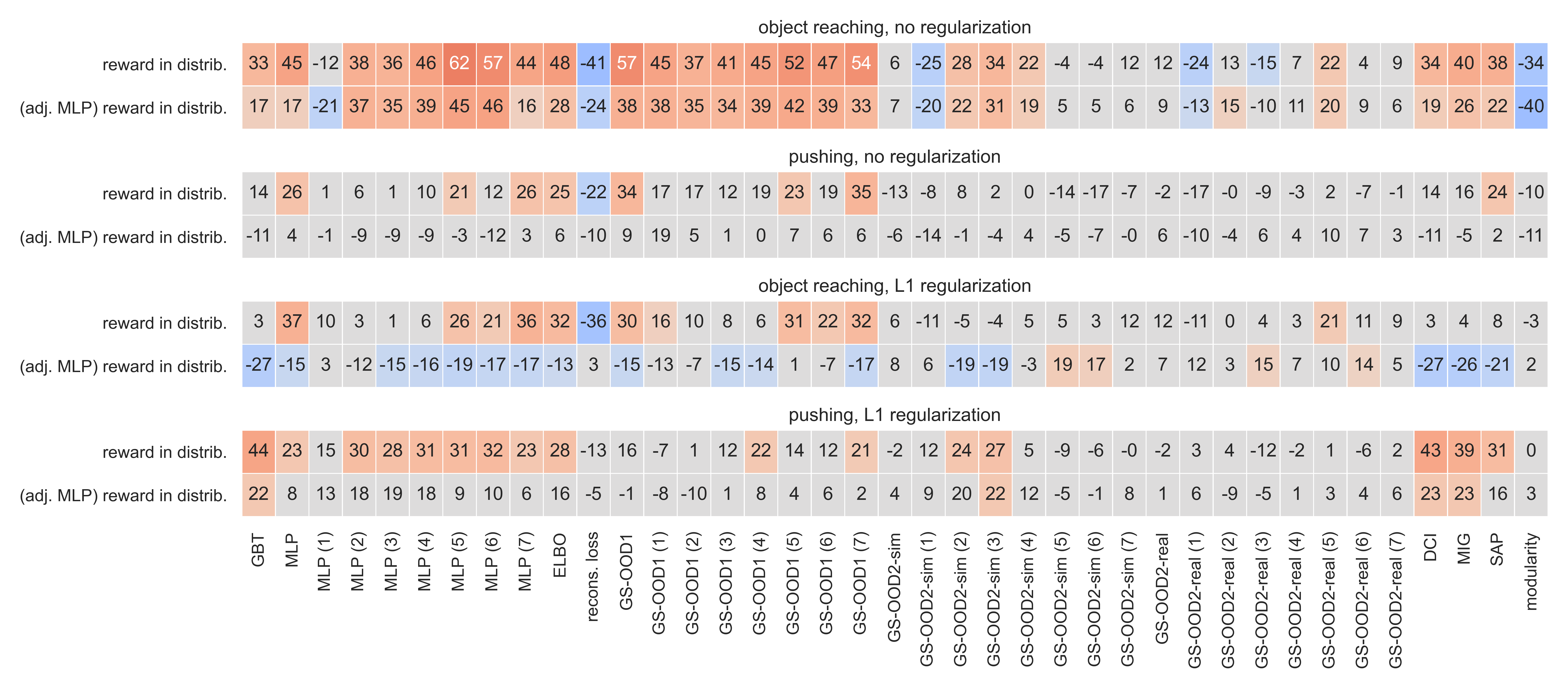}
    \caption{Rank correlations between metrics and in-distribution reward, with and without adjusting for informativeness. Correlations are color-coded as described in \cref{fig:results_on_training_env}.}
    \label{fig:appendix_correlations_indistribution}
\end{figure}

\cref{fig:results_on_training_env} in \cref{subsec:results_training_and_indistrib_performance} shows correlations of unsupervised and supervised metrics with in-distribution reward for \textit{object reaching} and \textit{pushing}, only in the case without regularization. In \cref{fig:appendix_correlations_indistribution} we also show these results in the case with regularization, as well as when adjusting for MLP informativeness.

\subsection{Out-of-distribution generalization in simulation}
\label{app:additional_results_ood_simulation}

In \cref{subsec:results_ood_generalization_simulation} we discussed rank-correlations of OOD rewards with unsupervised, informativeness and generalization scores on \textit{object reaching} without regularization. In \cref{fig:appendix_correlations_ood}  we also show these results for the case with regularization and on \textit{pushing}, as well as when adjusting for MLP informativeness.
Without regularization, we observe on \textit{pushing} very similar correlations along all metrics as we observed on  \textit{object reaching}, confirming our conclusions on this more complex task. When using regularization, rank correlations are very similar across both tasks. Interestingly, the correlation between GS-OOD2 scores and OOD2 generalization of the policy is even stronger when using L1 regularization. In contrast to our observations without regularization, we find that the correlation between GS-OOD1 and OOD1 generalization of the policy vanishes when adjusting for the MLP metric.

\begin{figure}
    \centering
    \includegraphics[width=\linewidth]{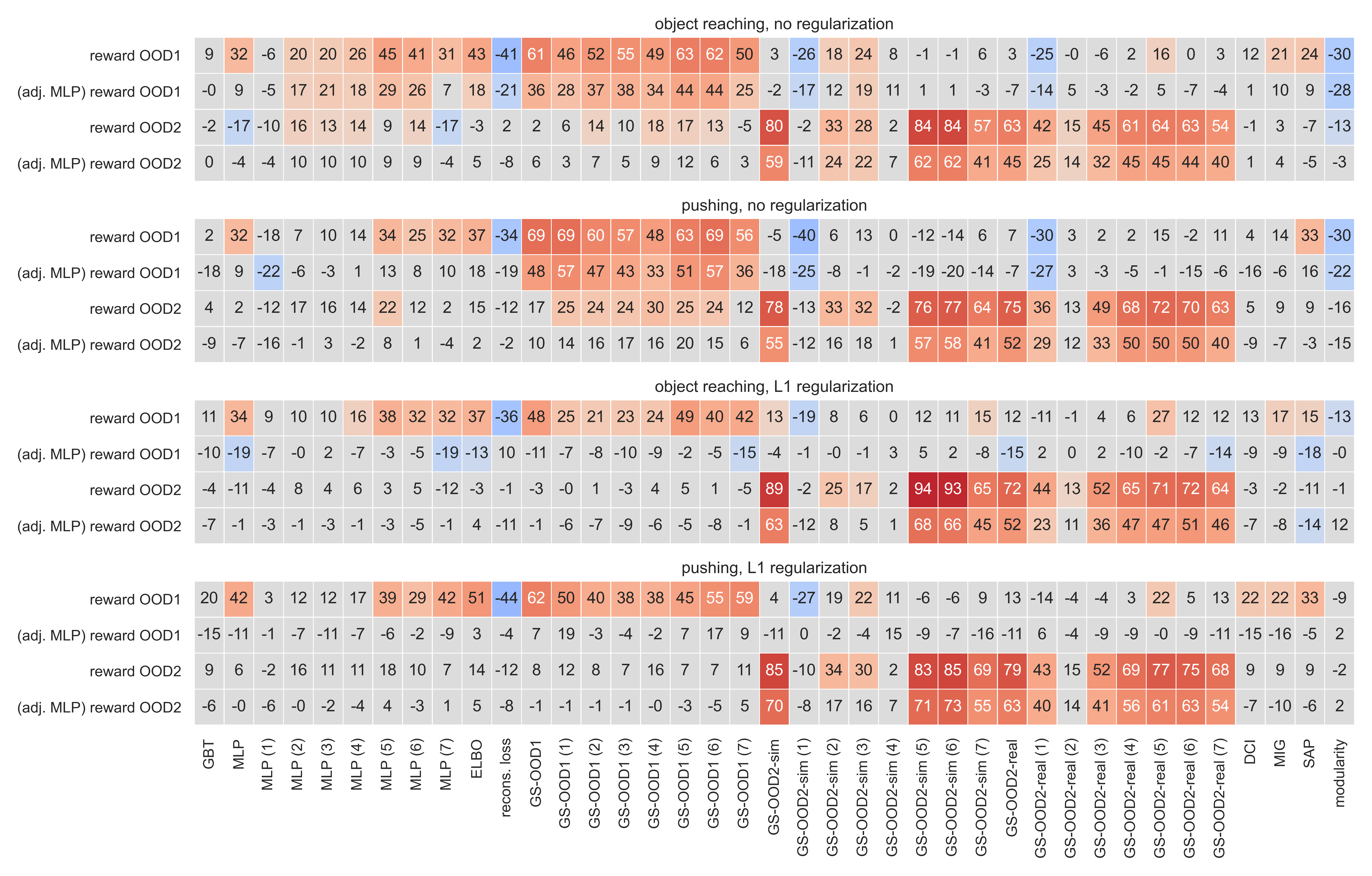}
    \caption{Rank correlations between metrics and OOD reward, with and without adjusting for informativeness. Correlations are color-coded as described in \cref{fig:results_on_training_env}.}
    \label{fig:appendix_correlations_ood}
\end{figure}

\subsubsection{Disentangled representations}
\begin{figure}
    \centering
    \includegraphics[width=0.79\linewidth]{figures/iclr2022/fig5_box_success_reg}
    
    \includegraphics[width=0.79\linewidth]{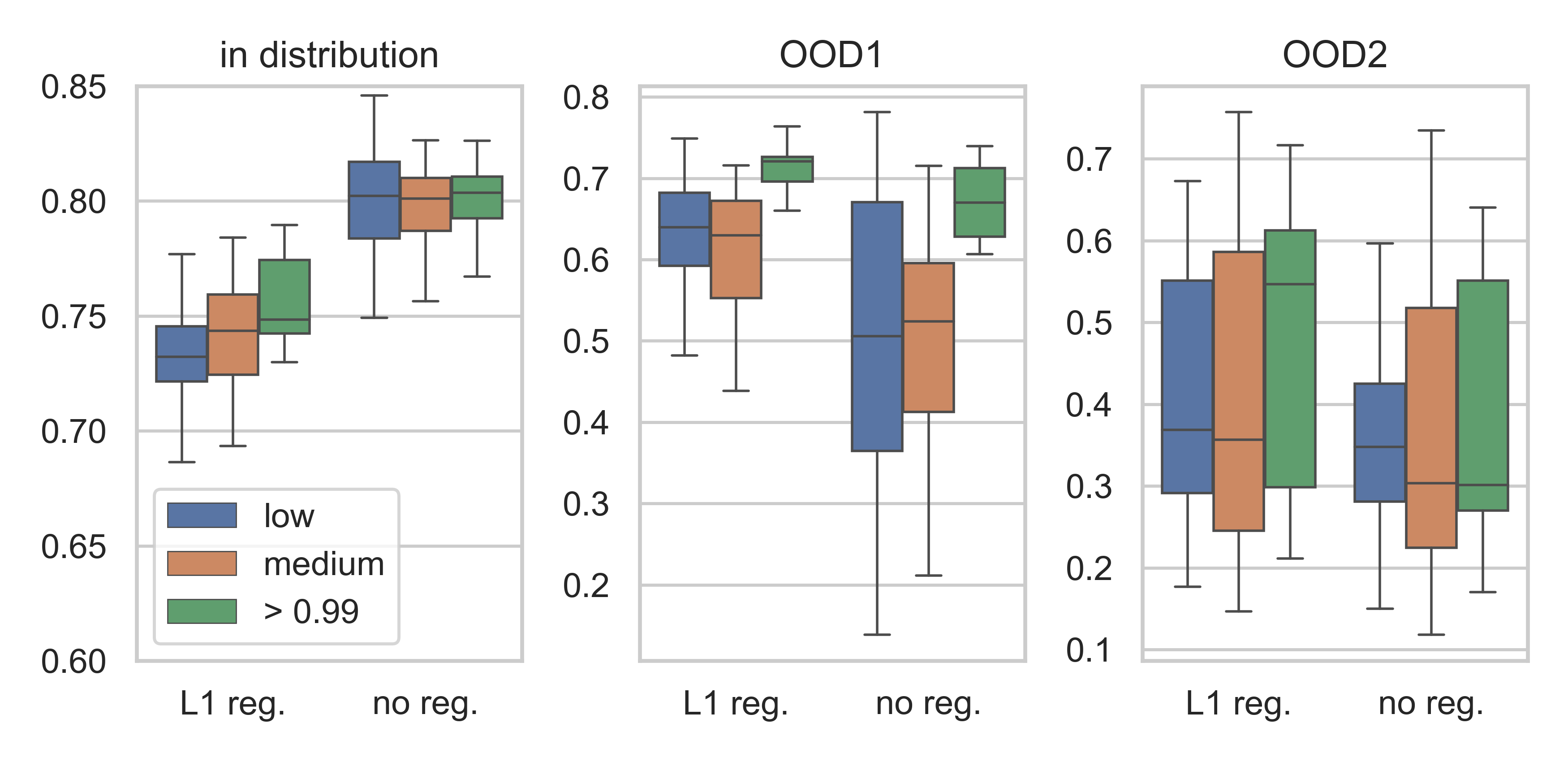}
    \caption{Fractional success on \textit{object reaching} (\textbf{top}) and \textit{pushing} (\textbf{bottom}), split according to low (blue), medium-high (orange), and almost perfect (green) disentanglement. Results for \textit{object reaching} are also reported in \cref{fig:disentanglement_and generalization_reaching} in \cref{subsec:results_ood_generalization_simulation}.}
    \label{fig:disentanglement_box_plots}
\end{figure}
As discussed in \cref{subsec:results_ood_generalization_simulation} for \textit{object reaching} without regularization, we observe in \cref{fig:appendix_correlations_ood} a weak correlation between some disentanglement metrics and OOD1 reward, which however vanishes when adjusting for MLP informativeness.
In agreement with \citet{dittadi2021transfer}, we observe no significant correlation between disentanglement and OOD2 generalization, for both tasks, with and without regularization.
From \cref{fig:disentanglement_box_plots} we see that in some cases, especially without regularization, a very high DCI score seems to lead to better performance on average. However, this behavior is not significant (within error bars), as opposed to the results shown in simpler downstream tasks by \citet{dittadi2021transfer}. Furthermore, this trend is likely due to representation informativeness, since the correlations with disentanglement disappear when adjusting for the MLP score, as discussed above.

\subsubsection{Regularization}
As seen in \cref{fig:disentanglement_box_plots}, regularization generally has a positive effect on OOD1 and OOD2 generalization, which is particularly prominent in the OOD1 setting. On the other hand, it leads to lower training rewards both in \textit{object reaching} and in \textit{pushing}. In the latter, the performance drop is particularly significant, while in \textit{object reaching} it is negligible.

\subsubsection{Sample efficiency}

In addition to the analysis reported in the main paper, we investigate how representation properties affect sample efficiency. Specifically, we store checkpoints of our policies at $t\in\{20\text{k},50\text{k},100\text{k}, 400\text{k}\}$ for \textit{object reaching} and $t\in\{200\text{k},500\text{k},1\text{M},3\text{M}\}$ for \textit{pushing}. We then evaluate policies at these time step on the same three environments as before: (1) on the cube colors from training; (2) on the OOD1 cube colors; and (3) on the OOD2-sim cube colors. Results are summarized in \cref{fig:appendix_correlation_sample_efficiency_reach} for \textit{object reaching} and \cref{fig:appendix_correlation_sample_efficiency_push} for \textit{pushing}.

\begin{figure}
    \centering
    \includegraphics[width=\linewidth]{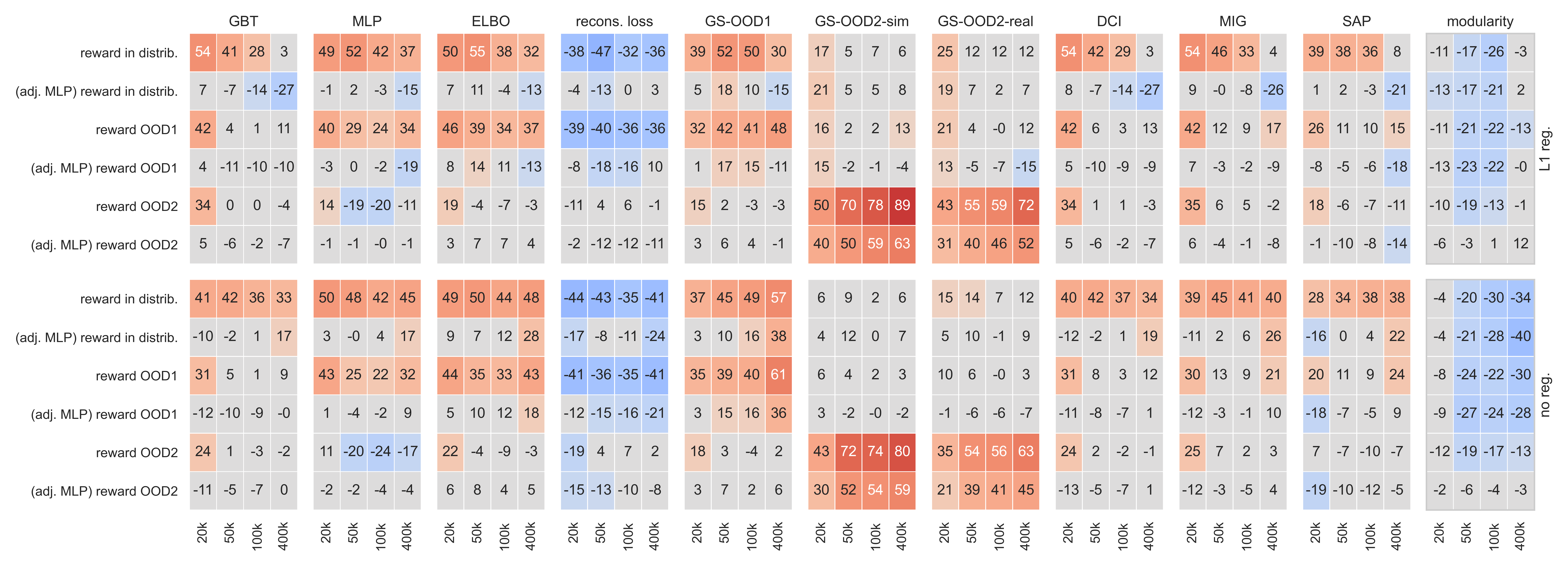}
    \caption{Sample efficiency analysis for \textit{object reaching}. Rank correlations of rewards with relevant metrics along multiple time steps. Correlations are color-coded as described in \cref{fig:results_on_training_env}.}
    \label{fig:appendix_correlation_sample_efficiency_reach}
\end{figure}

\begin{figure}
    \centering
    \includegraphics[width=\linewidth]{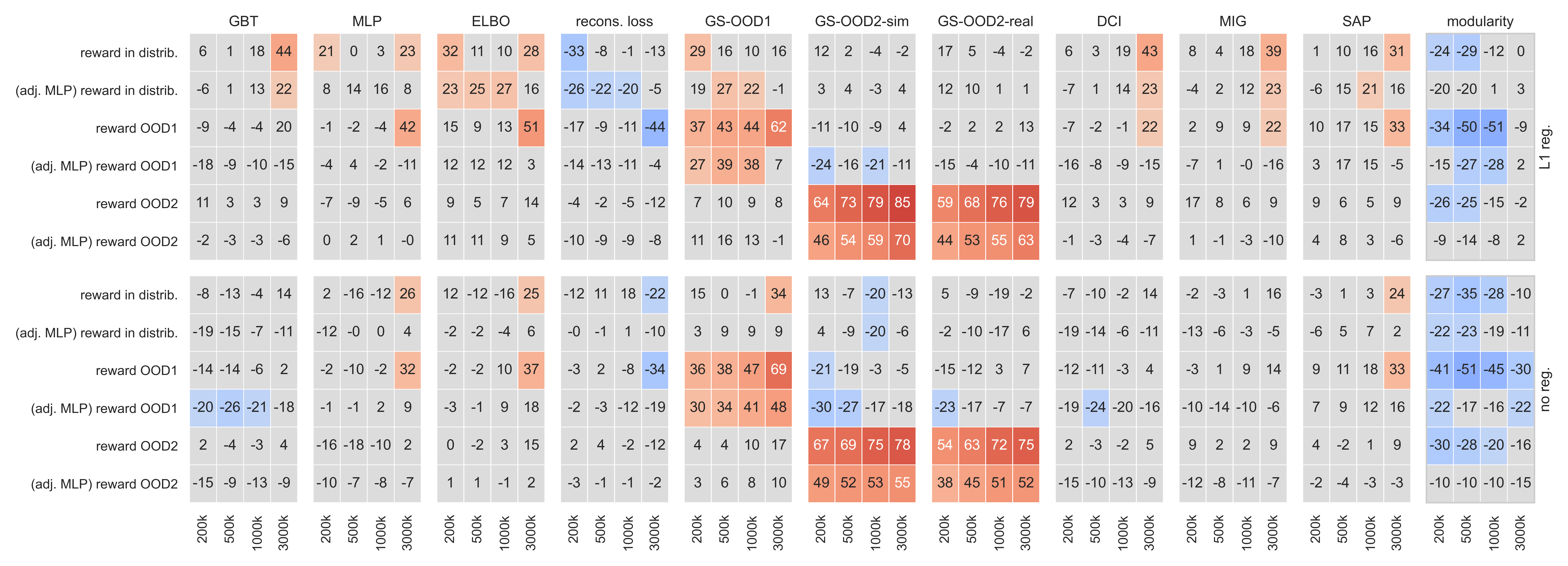}
    \caption{Sample efficiency analysis for \textit{pushing}. Rank correlations of rewards with relevant metrics along multiple time steps. Correlations are color-coded as described in \cref{fig:results_on_training_env}.}
    \label{fig:appendix_correlation_sample_efficiency_push}
\end{figure}

On \textit{object reaching} (\cref{fig:appendix_correlation_sample_efficiency_reach}), we observe very similar trends with and without regularization: Unsupervised metrics (ELBO and reconstruction loss) display a correlation with the training reward, as do the supervised informativeness metrics (GBT and MLP). This is strongest on early timesteps, meaning these scores could be important for sample efficiency. Similarly, we observe a correlation with the disentanglement scores DCI, MIG and SAP. With the help of the additional evaluation of rewards adjusted for MLP informativeness, we can attribute this correlation again to this common confounder. 
Crucially, we see that the generalization scores (GS) are correlated with generalization of the corresponding policies under OOD1 and OOD2 shifts for all recorded time steps, confirming the results in the main text.

On \textit{pushing} (\cref{fig:appendix_correlation_sample_efficiency_push}), many correlations at early checkpoints are significantly reduced, especially with regularization. This behavior might be due to the more complicated nature of the task, which involves learning to reach the cube first, and then push it to the goal. Correlations are primarily seen towards the end of training, with similar spurious correlations with disentanglement as elaborated above. Importantly, correlations between generalization scores (GS) and policy generalization under the same distribution shifts remain strong and statistically significant, corroborating the analysis in the main text.

\subsubsection{Generalization to a novel shape}
As mentioned in \cref{subsec:results_ood_generalization_simulation}, on the \textit{object reaching} task, we also test generalization w.r.t. a novel object shape by replacing the cube with an unmovable sphere. This corresponds to a strong OOD2-type shift, since shape was never varied when training the representations. 
We then evaluate a subset of 960 trained policies as before, with the same color splits. Surprisingly, the policies appear to handle the novel shape as we see from the histograms in \cref{fig:reaching_spheres_real_robot_analysis} in terms of success and final distance. In fact, when the sphere has the same colors that the cube had during policy training, \emph{all} policies get closer than 5 cm to the sphere on average, with a mean success metric of about 95\%. On sphere colors from the OOD1 split, more than 98.5\% move the finger closer than this threshold, and on the strongest distribution shift (OOD2-sim colors and cube replaced by sphere) almost 70\% surpass that threshold with an average success metric above~80\%.

\begin{figure}
    \centering
    \includegraphics[width=\linewidth]{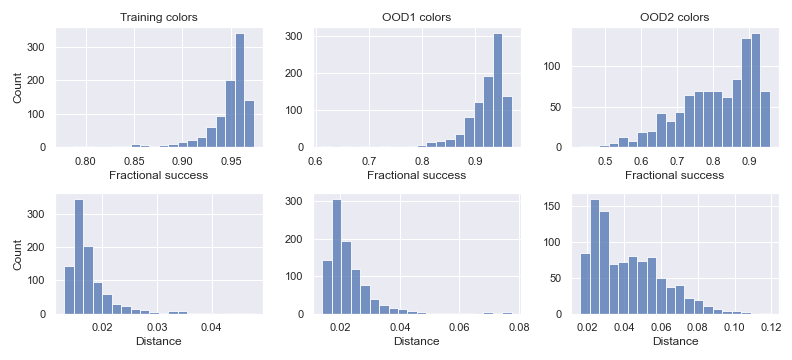}
    \caption{Testing policies for \textit{object reaching} under the same in-distribution, OOD1, and OOD2 evaluation protocols regarding object color in simulation, but replacing the cube with a sphere, which was never used in training.}
    \label{fig:reaching_spheres_real_robot_analysis}
\end{figure}

\subsection{Deploying policies to the real world}
In \cref{fig:reaching_spheres_real_robot_frames} we show three representative episodes of testing a reaching policy on the real robot for the strong OOD shift with a novel sphere object shape instead of the cube from training. We present the respective videos in the project page. There we also present videos of additional real-world episodes on pushing and reaching cubes of different colors.

\label{app:additional_results_ood_real_world}
\begin{figure}
    \centering
    \includegraphics[width=0.85\linewidth]{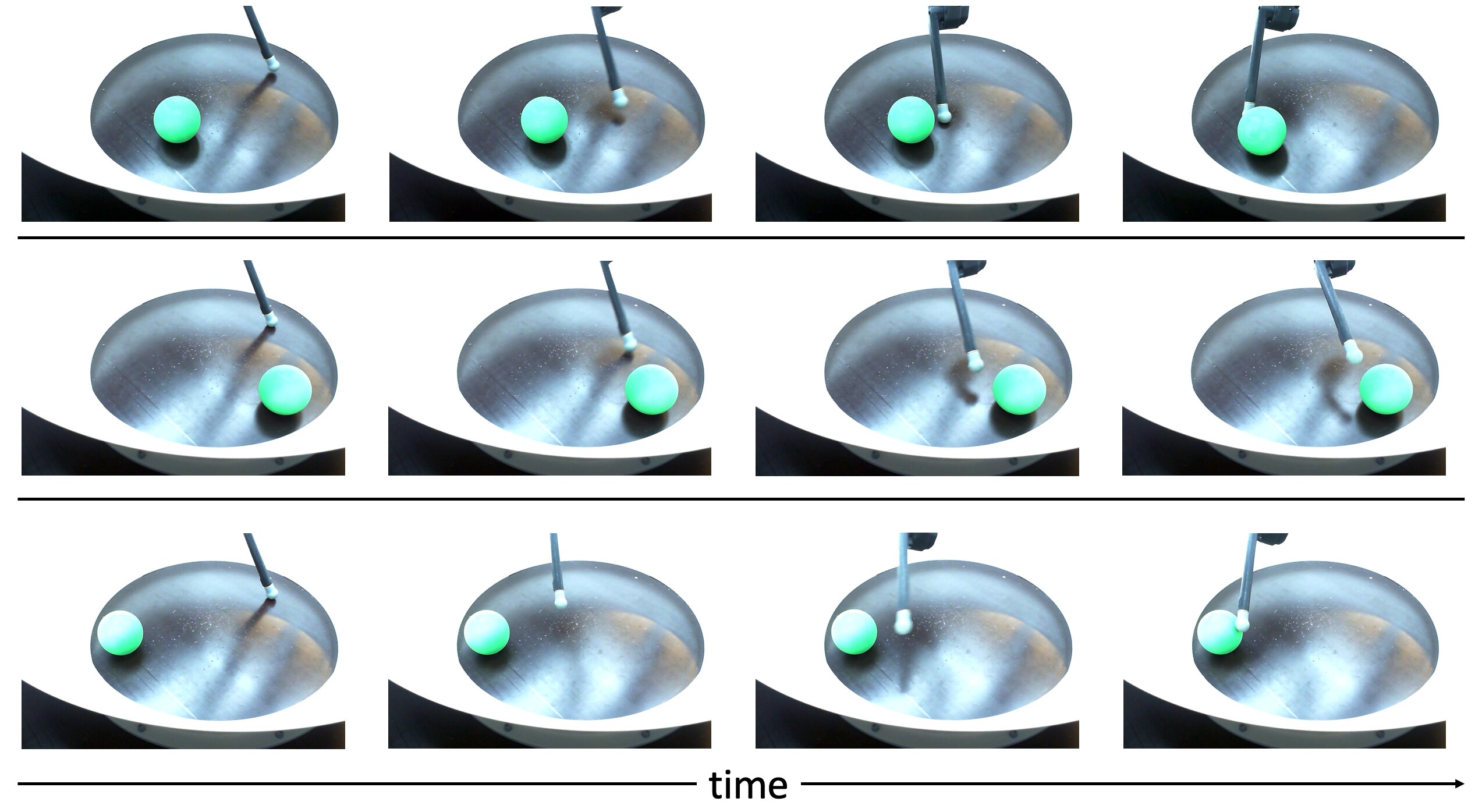}
    \caption{Transferring policies for \textit{object reaching} to the real robot setup without any fine-tuning on a green sphere (unseen shape \emph{and} color). Correlations are color-coded as described in \cref{fig:results_on_training_env}.}
    \label{fig:reaching_spheres_real_robot_frames}
\end{figure}

\chapter{Supplementary material for Chapter~\ref{chapter:objects}}  % cref gives problems with hyperref
\label{chapter:appendix:objects}

\section{Models}
\label{app:models}

In this section, we give an informal overview of the models included in this study and provide details on the implementation and hyperparameter choices.

\subsection{Overview of the models}

\paragraph{MONet.} In MONet~\cite{burgess2019monet}, attention masks are computed by a recurrent segmentation network that takes as input the image and the current \emph{scope}, which is the still unexplained portion of the image. For each slot, a variational autoencoder (the \emph{component VAE}) encodes the full image and the current attention mask, and then decodes the latent representation to an image reconstruction and mask. The reconstructed images are combined using the \emph{attention} masks (\emph{not} the masks decoded by the component VAE) into the final reconstructed image. 
The reconstruction loss is the negative log-likelihood of a spatial Gaussian mixture model (GMM) with one component per slot, where each pixel is modeled independently. The overall training loss is a (weighted) sum of the reconstruction loss, the KL divergence of the component VAEs, and an additional mask reconstruction loss for the component VAEs.

\paragraph{GENESIS.} Similarly to MONet, GENESIS~\cite{engelcke2020genesis} models each image as a spatial GMM. The spatial dependencies between components are modeled by an autoregressive prior distribution over the latent variables that encode the mixing probabilities.
From the image, an encoder and a recurrent network are used to compute the latent variables that are then decoded into the mixing probabilities. The mixing probabilities are pixel-wise and can be seen as attention masks for the image. Each of these is concatenated with the original image and used as input to the component VAE, which finds latent representations and reconstructs each scene component. These are combined using the mixing probabilities to obtain the reconstruction of the image.
While in MONet the attention masks are computed by a deterministic segmentation network, GENESIS defines an autoregressive prior on latent codes that are decoded into attention masks. GENESIS is therefore a proper probabilistic generative model, and it is trained by maximizing a modification of the ELBO introduced by \citet{rezende2018taming}, which adaptively trades off the likelihood and KL terms in the ELBO.

\paragraph{Slot Attention.} As our focus is on the object discovery task, we use the autoencoder model proposed in the Slot Attention paper~\cite{locatello2020object}.
The encoder consists of a CNN followed by the Slot Attention module, which maps the feature map to a set of {slots} through an iterative refinement process. 
At each iteration, dot-product attention is computed with the input vectors as keys and the current slot vectors as queries. The attention weights are then normalized over the slots, introducing competition between the slots to explain the input. Each slot is then updated using a GRU that takes as inputs the current slot vectors and the normalized attention vectors. After the refinement steps, the slot vectors are decoded into the appearance and mask of each object, which are then combined to reconstruct the entire image. The model is optimized by minimizing the MSE reconstruction loss.
While MONet and GENESIS use sequential slots to represent objects, Slot Attention employs instance slots.

\paragraph{SPACE.}
Spatially Parallel Attention and Component Extraction (SPACE) \cite{lin2020space} combines the approaches of scene-mixture models and spatial attention models. The foreground objects are segregated using bounding boxes computed through a parallel spatial attention process. The parallelism allows for a larger number of bounding boxes to be processed compared to previous related approaches. The background elements are instead modeled by a mixture of components. The use of bounding boxes for the foreground objects could lead to under- or over-segmentation if the size of the bounding box is not tuned appropriately. An additional boundary loss tries to address the over-segmentation issue by penalizing splitting objects across bounding boxes.

\paragraph{VAE baselines.}
We train variational autoencoders (VAEs)~\cite{kingma2013auto,rezende2014stochastic} as baselines that learn distributed representations. Following \citet{greff2019multi}, we use two different decoder architectures: one consisting of an MLP followed by transposed convolutions, and one where the MLP is replaced by a broadcast decoder~\cite{watters2019spatial}.
The VAEs are trained by maximizing the usual variational lower bound (ELBO).

\subsection{Implementation details}

\begin{table}
    \centering
    \caption{Datasets used for quantitative and/or qualitative evaluation in the publications corresponding to the four object-centric models considered in this study. Here we train and evaluate all models on all datasets.}
    \label{tab:app_models_x_datasets}
    \resizebox{\columnwidth}{!}{%
    \begin{tabular}{lccccc}
    \toprule
                   & CLEVR      & Multi-dSprites & Objects Room & Shapestacks & Tetrominoes\\
   \midrule
    MONet          & \checkmark & \checkmark$^*$ & \checkmark   &             &          \\
    Slot Attention & \checkmark & \checkmark\phantom{$^*$} &              &             &\checkmark\\
    GENESIS        &            & \checkmark$^*$ & \checkmark   & \checkmark  &          \\
    SPACE          &            &                &              &             &          \\
    \midrule 
    \multicolumn{6}{c}{\small $^*$These publications use a variant of Multi-dSprites with colored background as opposed to grayscale.}
    \\ \bottomrule
    \end{tabular}%
    }
    
    % \vspace{8pt}
    %
    % \begin{minipage}{\linewidth}
    %     \centering
    %     \small
    %
    %     $^*$These publications use a variant of Multi-dSprites with colored background as opposed to grayscale.
    %
    % \end{minipage}
\end{table}

We implement our library in PyTorch \cite{paszke2019pytorch}. All models are either re-implemented or adapted from available code, and quantitative results from the literature are reproduced, when available.
As shown in \cref{tab:app_models_x_datasets}, all methods included in our study were originally evaluated only on a subset of the datasets considered in our study. Thus, the recommended hyperparameters for a given model are likely to be suboptimal in the datasets on which such model was not evaluated. When a model performed particularly bad on a dataset, we attempted to find better hyperparameter values for the sake of the soundness of our study.
We provide implementation and training details for each model below.

\paragraph{MONet.}

We re-implement MONet following the implementation details in \citet{burgess2019monet}. In order to make this model work satisfactorily on Shapestacks and Tetrominoes---the two datasets where MONet was not originally tested---we ran a grid search over hyperparameters on both datasets, as follows:
\begin{itemize}
    \item Optimizer: Adam or RMSprop, both with default PyTorch parameters.
    \item $\beta \in \{0.1, 0.5\}$.
    \item Learning rate in \{3e-5, 1e-4\}.
    \item $(\sigma_{\mathrm{bg}}, \sigma_{\mathrm{fg}}) \in \{(0.06, 0.1), (0.12, 0.18), (0.2, 0.24), (0.25, 0.3), (0.3, 0.36)\}$.
\end{itemize}
A summary of the final hyperparameter choices is shown in \cref{tab:app_models_monet_parameters}.

\begin{table}
\centering
\caption{Overview of the main hyperparameter values for MONet. When dataset-specific values are not given, the defaults are used.}
\label{tab:app_models_monet_parameters}
\resizebox{\columnwidth}{!}{%
\begin{tabular}{@{}lcccc@{}}
\toprule
\multicolumn{1}{c}{\multirow{2}{*}{\textbf{Hyperparameter}}} & \multirow{2}{*}{\textbf{Default value}} & \multicolumn{3}{c}{\textbf{Dataset-specific values}}                         \\ \cmidrule(lr){3-5} 
\multicolumn{1}{c}{}                                &                                & \textbf{CLEVR}                & \textbf{Shapestacks}          & \textbf{Tetrominoes}          \\ \midrule
Optimizer                                           & Adam                           & RMSprop              & RMSprop              & ---                  \\
Learning rate                                       & 1e-4                   & 3e-5         & ---                  & ---                  \\
Batch size                                          & 64                             & 32                   & ---                  & ---                  \\
Training steps                                      & 500k                    & ---                  & ---                  & ---                  \\
$\sigma_\mathrm{bg}$                                & 0.06                           & ---                  & 0.2                  & 0.3                  \\
$\sigma_\mathrm{fg}$                                & 0.1                            & ---                  & 0.24                 & 0.36                 \\
$\beta$                                             & 0.5                            & ---                  & 0.1                  & ---                  \\
$\gamma$                                            & 0.5                            & ---                  & ---                  & ---                  \\
Latent space size                                   & 16                             & ---                  & ---                  & ---                  \\
U-Net blocks                                        & 5                              & 6                    & ---                  & 4                    \\ \bottomrule
\end{tabular}%
}
\end{table}

\paragraph{Slot Attention.}
We re-implement the Slot Attention autoencoder based on the official TensorFlow implementation and the corresponding publication \cite{locatello2020object}. 
We mostly use the recommended hyperparameter values and learning rate schedule.
On Objects Room and Shapestacks, we use the same parameters as for Multi-dSprites, which has the same resolution.
On CLEVR, we make a few changes to accommodate the larger image size. For the decoder, we follow the approach in \citet{locatello2020object} and use the broadcast decoder from a broadcasted shape of $8\times 8$ rather than $128\times 128$, and use four times a stride of 2 in the decoder.
For the encoder, we follow the set prediction architecture in \citet{locatello2020object} and use two strides of 2 in the encoder.
Finally, we use a batch size of 32 rather than 64.

\paragraph{GENESIS.}

We re-implement GENESIS based on the official implementation and the corresponding publication \cite{engelcke2020genesis}, and use the recommended hyperparameter values.
On Objects Room, we use the same hyperparameters as described in the paper for Multi-dSprites and Shapestacks, which have the same resolution. 
On CLEVR, which has $128 \times 128$ images, we use an additional stride of $2$ in the convolutional layer at the middle of both encoder and decoder (the output padding in the decoder is adjusted accordingly). In this case we also reduce the batch size from 64 to 32.
On Tetrominoes ($32\times 32$ images), we change the first stride in the encoder and the last stride in the decoder from 2 to 1.

\paragraph{SPACE.}

We adapt the official PyTorch implementation of SPACE to integrate it in our library. While in \citet{lin2020space} the authors train SPACE for 160k steps, here we train it for 200k. Since SPACE was not tested on any of the five datasets considered here (see \cref{tab:app_models_x_datasets}), we perform a hyperparameter sweep for all datasets. For each dataset, we run a random search over hyperparameters by training 100 models for 100k steps. \cref{tab:app_models_space_parameters} shows the random search definition, the hyperparameter values used for each dataset, and how they differ from those used in the original publication for the \emph{3D-Rooms} dataset (although we omit some hyperparameters that we leave unchanged).

\begin{table}
\centering
\caption[Hyperparameters for SPACE experiments.]{Hyperparameters for SPACE experiments. Here we show: the hyperparameters recommended by \citet{lin2020space} for the 3D-Rooms dataset on the official code repository; the hyperparameter space considered for our random search; the chosen default values across datasets; the dataset-specific values for CLEVR and Tetrominoes, which override the defaults. We omit some of the hyperparameters that we left unchanged from \citet{lin2020space}.}
\label{tab:app_models_space_parameters}
\resizebox{\columnwidth}{!}{%
\begin{tabular}{lccccc}
\toprule
\multirow{2}{*}{\textbf{Hyperparameter}} &
  \multirow{2}{*}{\begin{tabular}[c]{@{}c@{}}\textbf{Original}\\ \textbf{(3D-Rooms)}\end{tabular}} &
  \multirow{2}{*}{\textbf{Sweep values}} &
  \multirow{2}{*}{\textbf{Default value}} &
  \multicolumn{2}{c}{\textbf{Dataset-specific values}} \\ \cmidrule(lr){5-6} 
\multicolumn{1}{c}{}   &         &                              &         & \textbf{CLEVR} & \textbf{Tetrominoes} \\ \midrule
FG optimizer           & RMSprop & RMSprop                      & RMSprop & ---   & ---         \\
FG learning rate       & 1e-5    & \{3e-6, 1e-5, 3e-5, 1e-4\}   & 3e-5    & 1e-4  & 1e-4        \\
BG optimizer           & Adam    & Adam                         & Adam    & ---   & ---         \\
BG learning rate       & 1e-3    & 1e-3                         & 1e-3    & ---   & ---         \\
Batch size             & 12      & $\{16, 32\}$                 & 32      & ---   & ---         \\
$\sigma_\mathrm{bg}$   & 0.15    & $\{0.05, 0.15, 0.35\}$       & 0.15    & 0.05  & ---         \\
$\sigma_\mathrm{fg}$   & 0.15    & $\{0.02, 0.05, 0.15, 0.35\}$ & 0.15    & 0.05  & ---         \\
$G$ (FG grid size)     & 8       & $\{4, 8\}$                   & 8       & ---   & 4           \\
$K$ (BG n. of slots)   & 5       & $\{1, 5\}$                   & 5       & ---   & ---         \\
Boundary loss off step & 100k    & \{20k, 100k\}                & 20k     & ---   & 100k        \\
$\tau$ anneal end step & 20k     & \{20k, 50k\}                 & 50k     & 20k   & ---         \\
\begin{tabular}[c]{@{}l@{}}Mean of $p(\zb_\mathrm{pres})$ \\ \ \ (start/end values)\end{tabular} &
  (0.1, 0.01) &
  \{(0.1, 0.01), (0.5, 0.05)\} &
  (0.5, 0.05) &
  (0.1, 0.01) &
  (0.1, 0.01) \\
\begin{tabular}[c]{@{}l@{}}Mean of $p(\zb_\mathrm{scale})$ \\ \ \ (start/end values)\end{tabular} &
  $(-1, -2)$ &
  $\{(-1, -2), (0, -1)\}$ &
  $(0, -1)$ &
  --- &
  --- \\ \bottomrule
\end{tabular}%
}
\end{table}

\paragraph{VAEs.}

The architecture details for the VAEs are presented in \cref{tab:bsaeline_VAE_encoder_structure,tab:baseline_VAE_decoder_MLP,tab:baseline_VAE_decoder_broadcast}. These are used for Shapestacks, Multi-dSprites, and Objects Room. For CLEVR, an additional ResidualBlock with 64 channels and a AvgPool2D layer is added at the end of the stack of ResidualBlocks, to downsample the image one more time. This is mirrored in the decoder, where a  ResidualBlock with 256 channels and a (bilinear) Interpolation layer is added at the beginning of the stack of ResidualBlocks. The same happens in the broadcast decoder case. For Tetrominoes, the number of layers is the same, but the last AvgPool2D layer is removed from the encoder and the first Interpolation layer is removed from the decoder, to have one less downsampling and upsampling, respectively. 
The latent space size is chosen to be 64 times the number of slots that would be used when training an object-centric model on the same dataset. Note that the default number of slots varies depending on the dataset, as shown in \cref{tab:params_datasets}.\footnote{Here we consider the default for MONet, Slot Attention, and GENESIS, and we disregard SPACE. Although SPACE has a much larger number of slots, this is not comparable with the other models because of the grid-based spatial attention mechanism.}

\begin{table}
\centering
\caption{Structure of the encoder for both the vanilla and broadcast VAE, excluding the final linear layer that parameterizes $\mu$ and $\log \sigma^2$ of the approximate posterior.}
\begin{tabular}{lcc}
\toprule
\textbf{Encoder}           &          &                             \\ \toprule
\emph{Type}              & \emph{Size/Ch.} & \emph{Notes}                     \\ \midrule
Input: $\xb$   & $3$        &                             \\
Conv $5 \times 5$ &          & Stride $2$, Padding $2$     \\
LeakyReLU        &          &                             \\
          &          &                             \\
Residual Block    & $64$       & $2$ Conv layers               \\
Residual Block    & $64$       & $2$ Conv layers               \\
Conv $1 \times 1$ & $128$      &                             \\
AvgPool2D       &          & Kernel size $2$, Stride $2$ \\
          &          &                             \\
Residual Block    & $128$      & $2$ Conv layers               \\
Residual Block    & $128$      & $2$ Conv layers               \\
AvgPool2D       &          & Kernel size $2$, Stride $2$ \\
          &          &                             \\
Residual Block    & $128$      & $2$ Conv layers               \\
Residual Block    & $128$      & $2$ Conv layers               \\
Conv $1 \times 1$ & $256$      &                             \\
          &          &                             \\
Residual Block    & $256$      & $2$ Conv layers               \\
Residual Block    & $256$      & $2$ Conv layers               \\
          &          &                             \\
Flatten           &          &                             \\
LeakyReLU        &          &                             \\
Linear            & $512$      &                             \\
LeakyReLU        &          &                             \\
LayerNorm        &          &                             \\ \bottomrule
\end{tabular}
\label{tab:bsaeline_VAE_encoder_structure}
\end{table}

\begin{table}
\centering
\caption{Structure of the decoder for the vanilla VAE.}
\begin{tabular}{lcc}
\toprule
\textbf{Vanilla Decoder}   &                        &                         \\ \toprule
\emph{Type}            & \emph{Size/Ch.}               & \emph{Notes}                 \\ \midrule
Input: $\zb$   & $64 \times$ num. slots &                         \\
LeakyReLU        &                        &                         \\
Linear            & $512$                  &                         \\
LeakyReLU        &                        &                         \\
Unflatten         &                        &                         \\
          &          &                             \\
Residual Block    & $256$                  & $2$ Conv layers         \\
Residual Block    & $256$                  & $2$ Conv layers         \\
Conv $1 \times 1$ & $128$                  &                         \\
Interpolation     &                        & Scale $2$               \\
          &          &                             \\
Residual Block    & $128$                  & $2$ Conv layers         \\
Residual Block    & $128$                  & $2$ Conv layers         \\
Interpolation     &                        & Scale $2$               \\
          &          &                             \\
Residual Block    & $128$                  & $2$ Conv layers         \\
Residual Block    & $128$                  & $2$ Conv layers         \\
Conv $1 \times 1$ & $64$                   &                         \\
Interpolation     &                        & Scale $2$               \\
          &          &                             \\
Residual Block    & $64$                   & $2$ Conv layers         \\
Residual Block    & $64$                   & $2$ Conv layers         \\
Interpolation     &                        & Scale $2$               \\
          &          &                             \\
LeakyReLU        &                        &                         \\
Conv $5 \times 5$ & Image channels         & Stride $1$, Padding $2$\\\bottomrule
\end{tabular}
\label{tab:baseline_VAE_decoder_MLP}
\end{table}

\begin{table}
\centering
\caption[Structure of the decoder for the broadcast VAE.]{Structure of the decoder for the broadcast VAE. One less Interpolation is required, because the final image size for this architecture is $64$ and the broadcasting is to a feature map of size~$8$. }
\begin{tabular}{lll}
\toprule
\textbf{Broadcast Decoder} &                               &                         \\ \toprule
\emph{Type}              & \emph{Size/Ch.}             & \emph{Notes}        \\ \midrule
Input: $\zb$            & $64 \times$ num. slots        &                         \\
Broadcast                  & $64 \times$ num. slots $ + 2$ & Broadcast dim. $8$      \\
          &          &                             \\
Residual Block             & $256$                         & $2$ Conv layers         \\
Residual Block             & $256$                         & $2$ Conv layers         \\
Conv $1 \times 1$          & $128$                         &                         \\
          &          &                             \\
Residual Block             & $128$                         & $2$ Conv layers         \\
Residual Block             & $128$                         & $2$ Conv layers         \\
Interpolation              &                               & Scale $2$               \\
          &          &                             \\
Residual Block             & $128$                         & $2$ Conv layers         \\
Residual Block             & $128$                         & $2$ Conv layers         \\
Conv $1 \times 1$          & $64$                          &                         \\
Interpolation              &                               & Scale $2$               \\
          &          &                             \\
Residual Block             & $64$                          & $2$ Conv layers         \\
Residual Block             & $64$                          & $2$ Conv layers         \\
          &          &                             \\
LeakyReLU                 &                               &                         \\
Conv $5 \times 5$          & Image channels                & Stride $1$, Padding $2$ \\ \bottomrule
\end{tabular}
\label{tab:baseline_VAE_decoder_broadcast}
\end{table}

%%%%%%%%%%%%%%%%%%%%%%%%%%%%%%%%%%%%%%%%%%%%%%%%%%%%%%%%%%%%%%%%%%%%%%%%%%%%%%%%%%%%%%%%%%%%%%%%

\clearpage

\section{Datasets}
\label{app:datasets}

We collected 5 existing multi-object datasets and converted them into a common format. 
\emph{Multi-dSprites}, \emph{Objects Room} and \emph{Tetrominoes} are from DeepMind's Multi-Object Datasets collection, under the Apache 2.0 license~\cite{multiobjectdatasets19}. \emph{CLEVR} was originally proposed by \citet{johnson2017clevr}, with segmentation masks introduced by \citet{multiobjectdatasets19}. \emph{Shapestacks} was proposed by \citet{groth2018shapestacks} under the GPL 3.0 license.
Details on these datasets are provided in the following subsections. See \cref{fig:dataset_overview} for sample images and ground-truth segmentation masks for these datasets.
In \cref{tab:params_datasets}, we report dataset splits, number of foreground and background objects, and number of slots used when training object-centric models.

\subsection{CLEVR}

This dataset consists of $128 \times 128$ images of 3D scenes with up to 10 objects, possibly occluding each other.
Objects can have different colors (8 in total), materials (rubber or metal), shapes (sphere, cylinder, cube), sizes (small or large), x and y positions, and rotations.
Objects can be occluded by others. On average, 6.2 objects are visible.
As in previous work~\cite{greff2019multi,locatello2020object}, we learn object-centric representations on the CLEVR6 variant, which contains at most 6 objects.
There are $\num{100000}$ samples in the full dataset, and $\num{53483}$ in the CLEVR6 variant (at most 6 objects).
The CLEVR dataset has been cropped and resized according to the procedure detailed originally by~\citet{burgess2019monet}.

Each object is annotated with the following properties:
\begin{itemize}[parsep=3pt]
 \item \code{color} (categorical): 8 colors:
        \begin{itemize}[parsep=3pt]
            \item Red. RGB:\code{[173, 35, 35]}
            \item Cyan. RGB:\code{[41, 208, 208]}
            \item Green. RGB:\code{[29, 105, 20]}
            \item Blue. RGB:\code{[42, 75, 215]}
            \item Brown. RGB:\code{[129, 74, 25]}
            \item Gray. RGB:\code{[87, 87, 87]}
            \item Purple. RGB:\code{[129, 38, 192]}
            \item Yellow. RGB:\code{[255, 238, 51]}
        \end{itemize}
 \item \code{material} (categorical): The material of the object: rubber or metal.
 \item \code{shape} (categorical): The shape of the object: sphere, cylinder or cube.
 \item \code{size} (categorical): The size of the object: small or large.
 \item \code{x} (numerical): The x coordinate in 3D space.
 \item \code{y} (numerical): The y coordinate in 3D space.
\end{itemize}

\subsection{Multi-dSprites}
This dataset is based on the \emph{dSprites} dataset~\cite{dsprites17}. Following previous work~\cite{greff2019multi,locatello2020object}, we use the Multi-dSprites variant with colored sprites on a grayscale background. Each scene has 2--5 objects with random shapes (ellipse, square, heart), sizes (6 discrete values in $[0.5, 1]$), x and y position, orientation, and color (randomly sampled in HSV space). Objects can occlude each other. The intensity of the uniform grayscale background is randomly sampled in each image. Images have size $64 \times 64$.

Each object is annotated with the following properties:
\begin{itemize}[parsep=3pt]
    \item \code{color} (numerical): 3-dimensional RGB color vector.
    \item \code{scale} (numerical): Scaling of the object, 6 uniformly spaced values between 0.5 and 1.
    \item \code{shape} (categorical): The shape type of the object (ellipse, heart, square).
    \item \code{x} (numerical): Horizontal position between 0 and 1.
    \item \code{y} (numerical): Vertical position between 0 and 1.
\end{itemize}

\begin{table}
\centering
\caption{Dataset splits, number of foreground and background objects, and number of slots used when training object-centric models.}
\label{tab:params_datasets}
\resizebox{\columnwidth}{!}{%
\begin{tabular}{lcccccc}
\toprule
\textbf{Dataset Name} & \textbf{Train} & \textbf{Validation} & \textbf{Test} & \textbf{Background} &  \textbf{Foreground} & \textbf{Slots}  \\
                      &  \textbf{Size} & \textbf{Size}       & \textbf{Size} & \textbf{Objects}       & \textbf{Objects}        &    \\ 
\midrule
{CLEVR6}         & 49483 & 2000 & 2000 & 1 & 3--6 & 7$^*$ \\
{Multi-dSprites} & 90000 & 5000 & 5000 & 1 & 2--5  & 6$^*$ \\
{Objects Room}   & 90000 & 5000 & 5000 & 4 & 1--3 & 7$^*$\\ 
{Shapestacks}    & 90000 & 5000 & 5000 & 1 & 2--6 & 7$^*$\\
{Tetrominoes}    & 90000 & 5000 & 5000 & 1 & 3 & 4$^\dagger$\\
\midrule
\multicolumn{7}{l}{\small
$^*$In SPACE we use 69 slots: 5 background slots, and a grid of $8\times 8$ foreground slots.
}\\
\multicolumn{7}{l}{\small
$^\dagger$In SPACE we use 21 slots: 5 background slots, and a grid of $4\times 4$ foreground slots.
}\\
\bottomrule
\end{tabular}%
}

\end{table}

\begin{figure}
    \def\datasetleftspacing{\qquad\qquad\qquad\qquad}
    \def\datasetvspace{\vspace{6pt}}
    
    \newlength{\datasetheight}
    \setlength{\datasetheight}{3cm}
    
    \datasetleftspacing \includegraphics[height=\datasetheight]{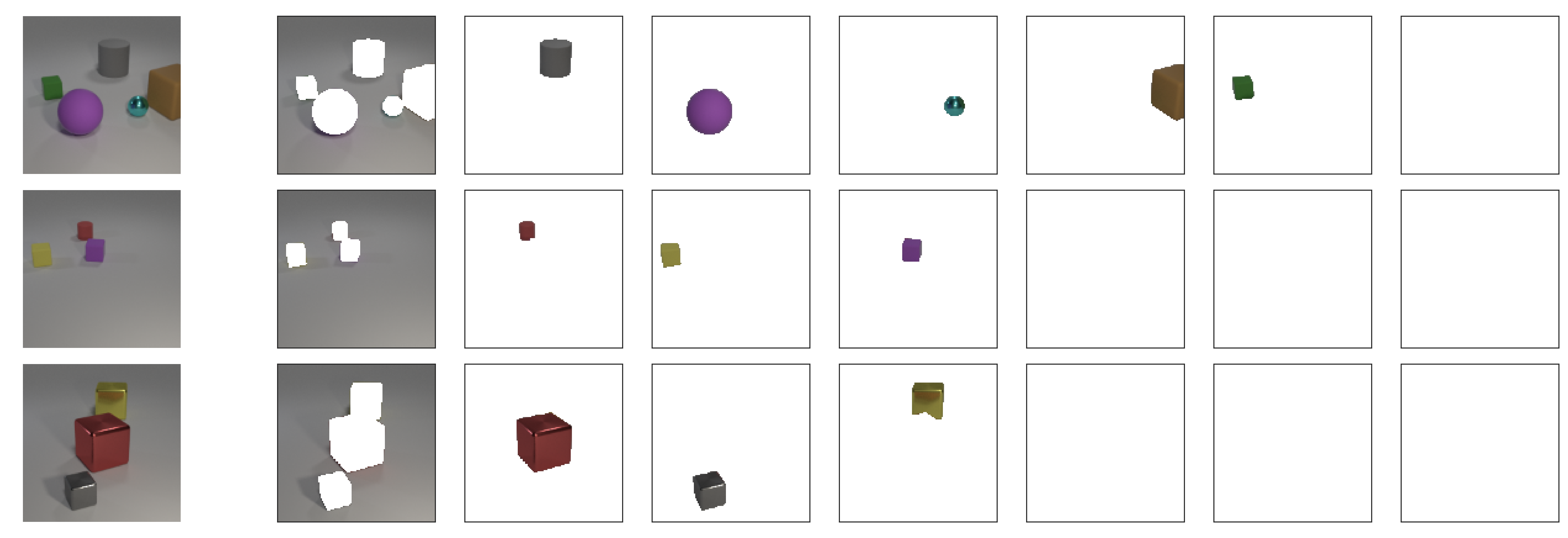}
    
    \datasetvspace
    
    \datasetleftspacing
    \includegraphics[height=\datasetheight]{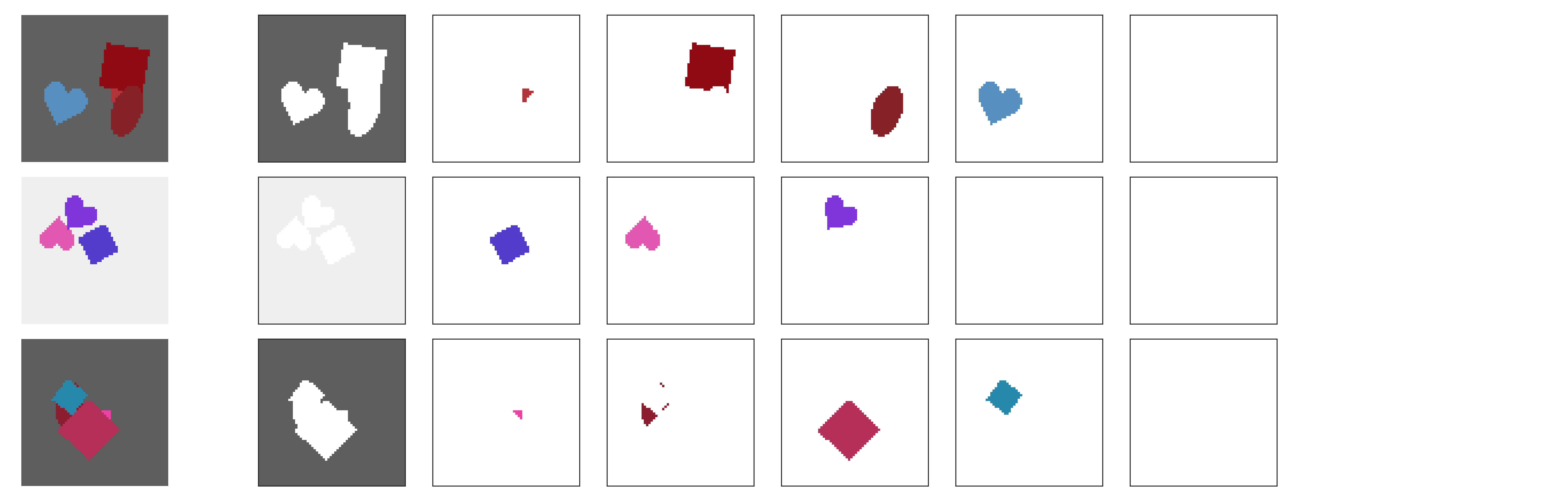}
    
    \datasetvspace
    
    \datasetleftspacing \includegraphics[height=\datasetheight]{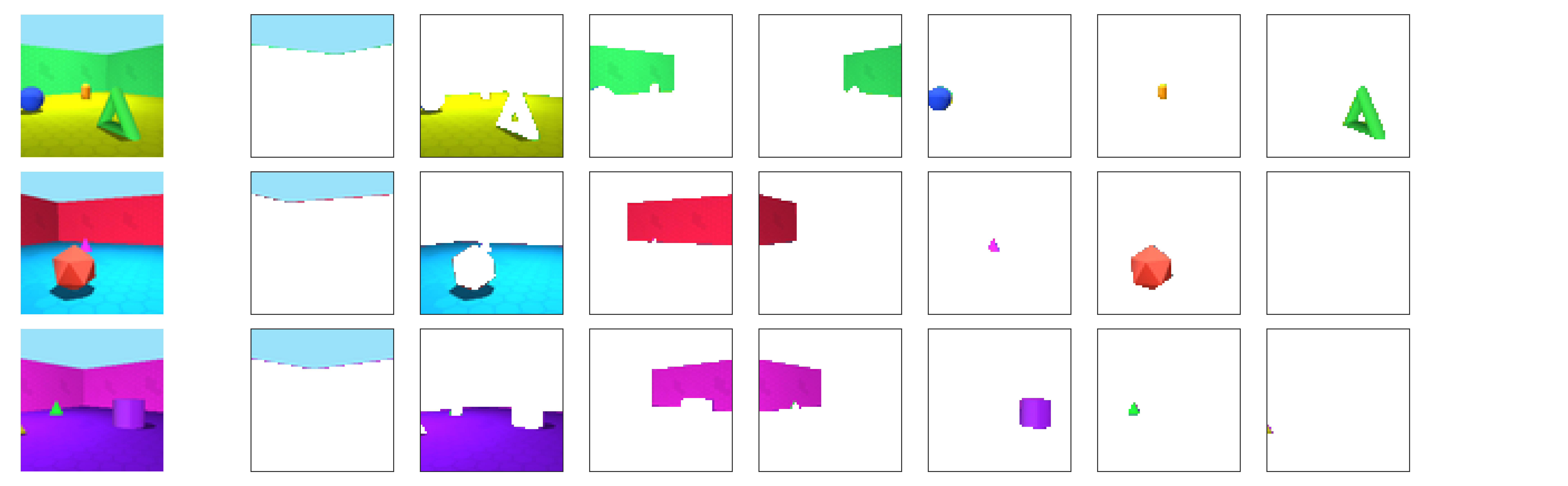}
    
    \datasetvspace
    
    \datasetleftspacing \includegraphics[height=\datasetheight]{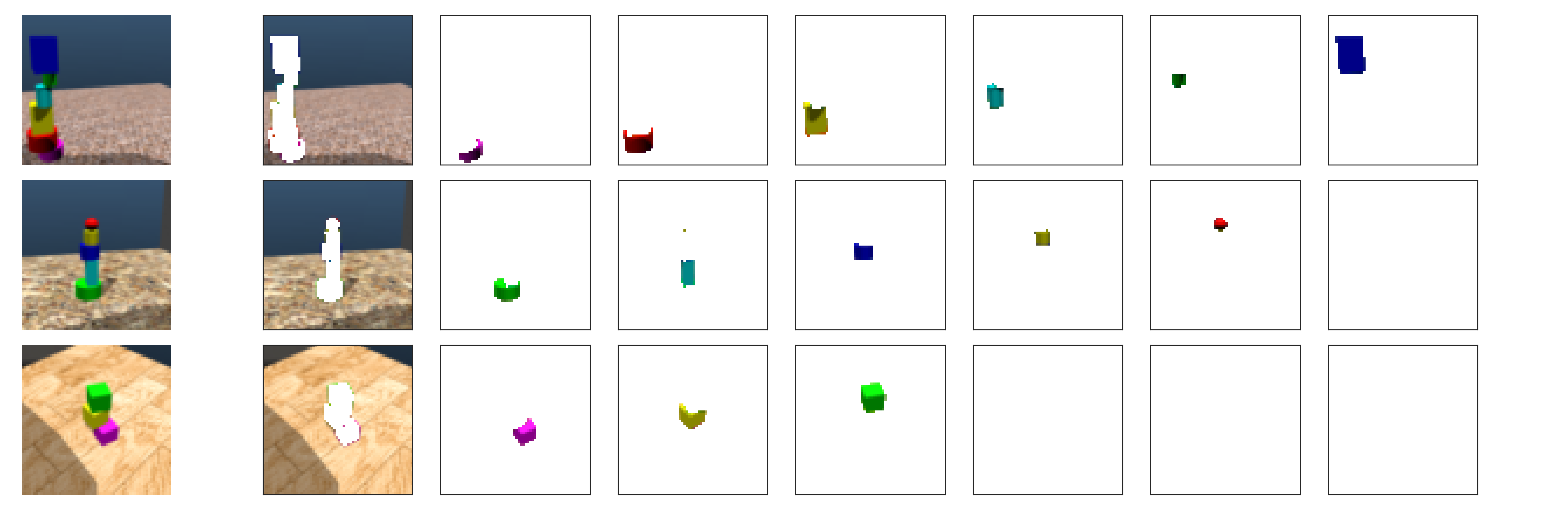}
    
    \datasetvspace
    
    \datasetleftspacing \includegraphics[height=\datasetheight]{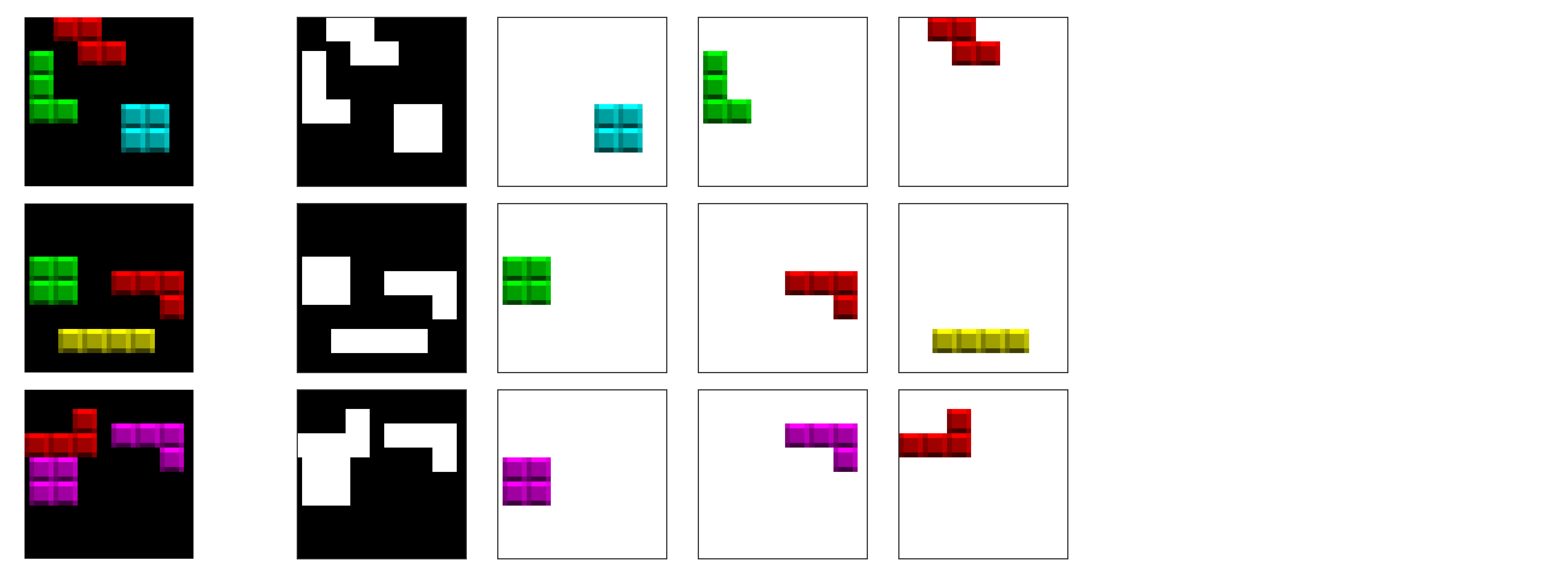}
    
    \datasetvspace

	\caption{Examples of images from the datasets considered in this work. The leftmost column represents the original image, the other columns show all the objects in the scene according to the ground-truth segmentation masks. Top to bottom: CLEVR6, Multi-dSprites, Objects Room, Shapestacks, Tetrominoes.}
	\label{fig:dataset_overview}
\end{figure}

\subsection{Objects Room}

This dataset was originally introduced by \citet{eslami2018neural} and consists of $64\times 64$ images of 3D scenes with up to three objects.
Since this dataset includes masks but no labels for the object properties, we can use it only to evaluate segmentation performance.

\subsection{Shapestacks}

This dataset consists of $64\times 64$ images of 3D scenes where objects are stacked to form a tower. Each scene is available under different camera views. Object properties are shape (cube, cylinder, sphere), color (6 possible values), size (numerical) and ordinal position in the stack.

%
% There are $311\ 200$ samples in the dataset.
%
Each object is annotated with the following properties:
\begin{itemize}[parsep=3pt]
 \item \code{shape} (categorical): shape of the object: cylinder, sphere or cuboid.
 \item \code{color} (categorical): 6 colors:
            \begin{itemize}[parsep=3pt]
                 \item Blue. RGB:\code{[0, 0, 255]}
                 \item Green. RGB:\code{[0, 255, 0]}
                 \item Cyan. RGB:\code{[0, 255, 255]}
                 \item Red. RGB:\code{[255, 0, 0]}
                 \item Purple. RGB:\code{[255, 0, 255]}
                 \item Yellow. RGB:\code{[255, 255, 0]}
            \end{itemize}
\end{itemize} 

\subsection{Tetrominoes}
This dataset consists of $32 \times 32$ images (cropped from the original $35 \times 35$ for simplicity) of 3D-textured tetris pieces placed on a black background. There are always 3 objects in a scene, and no occlusions. Objects have different shapes (19 in total), colors (6 fully saturated colors), x and y position.

Each object is annotated with the following properties:
\begin{itemize}[parsep=3pt]
 \item \code{shape} (categorical): 19 shapes:
        \begin{itemize}[parsep=3pt]
         \item Horizontal I piece.
         \item Vertical I piece.
         \item L piece pointing downward.
         \item J piece pointing upward.
         \item L piece pointing upward.
         \item J piece pointing downward.
         \item L piece pointing left.
         \item J piece pointing left.
         \item J piece pointing right.
         \item L piece pointing right.
         \item Horizontal Z piece.
         \item Horizontal S piece.
         \item Vertical Z piece.
         \item Vertical S piece.
         \item T piece pointing upward.
         \item T piece pointing downward.
         \item T piece pointing left.
         \item T piece pointing right.
         \item O piece.
        \end{itemize}
 
 \item \code{color} (categorical): 6 colors:
        \begin{itemize}[parsep=3pt]
         \item Blue. RGB:\code{[0, 0, 255]}
         \item Green. RGB:\code{[0, 255, 0]}
         \item Cyan. RGB:\code{[0, 255, 255]}
         \item Red. RGB:\code{[255, 0, 0]}
         \item Purple. RGB:\code{[255, 0, 255]}
         \item Yellow. RGB:\code{[255, 255, 0]}
        \end{itemize}
 \item \code{x} (numerical): Horizontal position.
 \item \code{y} (numerical): Vertical position.
\end{itemize}

%\clearpage

\section{Evaluations}
\label{app:evaluation}

In this section, we discuss in more detail the chosen reconstruction and segmentation metrics (\cref{app:evaluation_metrics}), provide implementation details on the downstream property prediction task (\cref{app:evaluation_downstream}), and more closely examine the distribution shifts considered in this study (\cref{app:evaluation_shifts}).

\subsection{Reconstruction and segmentation metrics}\label{app:evaluation_metrics}

\paragraph{Mean reconstruction error.}
Since all models in this study are autoencoders, we can use the reconstruction error to 
This is potentially an informative metric as it should roughly indicate the amount and accuracy of information captured by the models and present in the representations.
All models include some form of reconstruction term in their losses, but they may take different forms. We then choose to evaluate the reconstruction error with the mean squared error (MSE), defined for an image $\xb$ and its reconstruction $\hat{\xb}$ as follows:
\begin{equation}
    \mathrm{MSE} \left( \xb, \hat{\xb} \right) = \| \xb - \hat{\xb} \|_2^2 = \frac{1}{D} \sum_{i=1}^D (x_i - \hat{x}_i)^2
\end{equation}
where for simplicity we assume a vector representation of $\xb$ and $\hat{\xb}$, both with dimension $D$ equal to the number of pixels times the number of color channels.

\paragraph{Adjusted Rand Index (ARI).}

The Adjusted Rand Index (ARI) \cite{hubertComparingPartitions1985} measures the similarity between two partitions of a set (or clusterings). Interpreting segmentation as clustering of pixels, the ARI can be used to measure the degree of similarity between two sets of segmentation masks. Segmentation accuracy is then assessed by comparing ground-truth and predicted masks.
The expected value of the ARI on random clustering is 0, and the maximum value is 1 (identical clusterings up to label permutation).
As in prior work \cite{burgess2019monet,engelcke2020genesis,locatello2020object}, we only consider the ground-truth masks of foreground objects when computing the ARI.
Below, we define the Rand Index and the Adjusted Rand Index in more detail.

The Rand Index is a symmetric measure of the similarity between two partitions of a set~\citep{rand1971objective,hubertComparingPartitions1985,wagner2007comparing}. It is inspired by traditional classification metrics that compare the number of correctly and incorrectly classified elements. 
The Rand Index is defined as follows:
Let $S$ be a set of $n$ elements, and let $A = \{A_1, \dots, A_{n_A}\}$ and $B = \{B_1, \dots, B_{n_B}\}$ be partitions of $S$. Furthermore, let us introduce the following quantities:%
\begin{itemize}
	\item $m_{11}$: number of pairs of elements that are in the same subset in both $A$ and $B$,
	\item $m_{00}$: number of pairs of elements that are in different subsets in both $A$ and $B$,
	\item $m_{10}$: number of pairs of elements that are in the same subset in $A$ and in different subsets in $B$,
	\item $m_{01}$: number of pairs of elements that are in different subsets in $A$ and in the same subset in $B$.
\end{itemize}
The Rand Index is then given by:
\begin{equation}
	\mathrm{RI} (A,B) = \cfrac{m_{11}+m_{00}}{m_{11}+m_{00}+m_{10}+m_{01}} = \cfrac{2(m_{11}+m_{00})}{n(n-1)}
\end{equation}
and quantifies the number of elements that have been correctly classified over the total number of elements. 

The Rand Index ranges from 0 (no pair classified in the same way under $A$ and $B$) to 1 ($A$ and $B$ are identical up to a permutation). However, the result is strongly dependent on the number of clusters and on the number of elements in each cluster. If we fix $n_A$, $n_B$, and the proportion of elements in each subset of the two partitions, then the Rand Index will increase as $n$ increases, and even converge to 1 in some cases \cite{fowlkes1983method}. The expected value of a random clustering also depends on the number of clusters and on the number of elements $n$.

The Adjusted Rand Index (ARI) \cite{hubertComparingPartitions1985} addresses this issue by normalizing the Rand Index such that, with a random clustering, the metric will be 0 in expectation.
Given the same conditions as above, let $n_{i,j} = |A_i \cap B_j|$, $a_i = |A_i|$, and $b_i = |B_i|$, with $i=1,\ldots,n_A$ and $i=1,\ldots,n_B$. The ARI is then defined as:
\begin{equation}
	\mathrm{ARI}(A,B)  = \cfrac{\sum_{i,j} \binom{n_{i,j}}{2} - \cfrac{\sum_{i} \binom{a_i}{2}\sum_{j} \binom{b_j}{2}}{\binom{n}{2}}}{\frac{1}{2}\left[\sum_i\binom{a_i}{2}+\sum_j\binom{b_j}{2}\right]-\cfrac{\sum_i\binom{a_i}{2}\sum_j\binom{b_j}{2}}{\binom{n}{2}}}
\end{equation}
which is 0 in expectation for random clusterings, and 1 for perfectly matching partitions (up to a permutation). Note that the ARI can be negative.

\paragraph{Segmentation covering metrics.}
Segmentation Covering (SC) \cite{arbelaez2010contour} uses the intersection over union (IOU) between pairs of segmentation masks from the sets $A$ and $B$. How the segmentation masks are matched depends on whether we are considering the covering of $B$ by $A$ (denoted by $A \rightarrow B$) or vice versa ($B \rightarrow A$).
We use the slightly modified definition by \citet{engelcke2020genesis}:
\begin{equation}
    \mathrm{SC}(A\rightarrow B) = \frac{1}{\sum_{R_B \in B} \left|R_B\right|} \sum_{R_B \in B} \left|R_B\right| \max_{R_A \in A} \operatorname{\textsc{iou}}(R_A, R_B)\ ,
    \label{eq:sc}
\end{equation}
where $|R|$ denotes the number of pixels belonging to mask $R$, and the intersection over union is defined as:
\begin{equation}
    \operatorname{\textsc{iou}}(R_A, R_B) = \frac{\left|R_A \cap R_B\right|}{\left|R_A \cup R_B\right|}\ .
    \label{eq:iou}
\end{equation}

While standard (weighted) segmentation covering weights the IOU by the size of the ground truth mask, mean (or unweighted) segmentation covering (mSC) \cite{engelcke2020genesis} gives the same importance to masks of different size:
\begin{equation}
    \mathrm{mSC}(A\rightarrow B) = \frac{1}{|B|} \sum_{R_B \in B}  \max_{R_A \in A} \operatorname{\textsc{iou}}(R_A, R_B)\ ,
    \label{eq:msc}
\end{equation}
where $|B|$ denotes the number of non-empty masks in $B$.
Since a high SC score can still be attained when small objects are not segmented correctly, mSC is considered to be a more meaningful and robust metric across different datasets~\cite{engelcke2020genesis}.

Note that neither SC nor mSC are symmetric: Following \citet{engelcke2020genesis}, we consider $A$ to be the predicted segmentation masks and $B$ the ground-truth masks of the foreground objects.
As observed by \citet{engelcke2020genesis}, both SC and mSC penalize over-segmentation (segmenting one object into separate slots), unlike the ARI.
Both SC and mSC take values in $[0,1]$.

\subsection{Downstream property prediction}\label{app:evaluation_downstream}

Here we start by briefly summarizing the downstream property prediction task presented in the main text, and then provide additional details on the models and evaluation protocol.

\paragraph{Overview of the property prediction task.}
As outlined in \cref{sec:experimental_setup}, 
we evaluate scene representations by training downstream models to predict ground-truth object properties from the representations. 
Exploiting the fact that object slots share a common representational format, a single downstream model $f$ can be used to predict the properties of each object independently: for each slot representation $\zb_k$ we predict a vector of object properties $\ybpred_k = f(\zb_k)$. This vector represents predictions for \emph{all} properties of an object.
We then match each slot's prediction to a corresponding ground-truth object using \emph{mask matching} or \emph{loss matching} (see main text).
In non-slotted models such as the VAE baselines considered in this study, we do not have access to separate object representations $\{\zb_k\}_{k=1}^K$. Therefore, the downstream model $f$ in this case takes as input the overall distributed representation $\zb$, which is a flat vector, and outputs a prediction of \emph{all objects at once}: $\ybpred = f(\zb)$. This is then split into $K$ vectors, which are matched to ground-truth objects with either \emph{loss matching} or \emph{deterministic matching} (see main text).

\paragraph{Implementation details.}
% MODELS
We use 4 different downstream models: a linear model, and MLPs with up to 3 hidden layers of size 256 each.
Let $P$ be the size of the ground-truth property vector, which includes all numerical and categorical\footnote{Here we use the one-hot representation of categorical properties.} properties according to an order specified by the dataset.
We denote by $K$ be the number of slots and $d$ the dimensionality of a slot representation $\zb_k$ in object-centric models. Note that we must include in $\zb_k$ \emph{all} representations related to a slot, possibly including different latent variables that are explicitly responsible for modeling, e.g., the location, appearance, or presence of an object.
The downstream model $f$ has input size $d$ and output size $P$, and is applied in parallel (with shared weights) to all slots.
In non-slotted models, we always define the dimensionality of the distributed representation $\zb$ in terms of $K$ for fair comparison with slot-based models, hence we can write the latent dimensionality of such models as $d \cdot K$. In this case, the input and output sizes of the downstream model ($d$ and $P$, respectively) are multiplied by $K$, and we apply this model only once, to the entire scene representation.
The linear downstream model is implemented as a linear layer. MLP models (with at least one hidden layer) have hidden size 256 and LeakyReLU nonlinearities, as shown in \cref{tab:app_downstream_models}.

\begin{table}
\centering
\caption{Architecture of the downstream MLP models for property prediction. The third and fourth items are repeated 0 or more times, depending on the required number of hidden layers.}
\label{tab:app_downstream_models}
\phantom{repeated 0 or----}
\begin{tabular}{lccl}
\cmidrule[\heavyrulewidth]{1-3}
\textbf{Layer type} & \textbf{Input size} & \textbf{Output size} &                                                                    \\ 
\cmidrule{1-3}
Linear              & $d$ or $d \cdot K$  & 256                  &                                                                    \\
LeakyReLU($0.01$)   & 256                 & 256                  &                                                                    \\
Linear              & 256                 & 256                  & \rdelim\}{2}{30mm}[\ \ \parbox{24.5mm}{repeated 0 or\\more times}] \\
LeakyReLU($0.01$)   & 256                 & 256                  &                                                                    \\
Linear              & 256                 & $P$ or $P \cdot K$   &                                                                    \\ 
\cmidrule[\heavyrulewidth]{1-3}
\end{tabular}
\end{table}

\paragraph{Data splits.}
Let $\mathcal{D}_s$ be a source dataset and $\mathcal{D}_t$ a target dataset. When doing in-distribution evaluation, we train and test the downstream model without distribution shifts, so we simply have $\mathcal{D}_s = \mathcal{D}_t$.
Given a representation function $r$, and a matching strategy to match the slots with ground-truth objects, we consider:
\begin{itemize}
    \item a train split of $\num{10000}$ images from $\mathcal{D}_s$\ ,
    \item a validation split of $\num{1000}$ images from $\mathcal{D}_s$\ ,
    \item a test split of $\num{2000}$ images from $\mathcal{D}_t$\ .
\end{itemize}
The test split only contains images that were not used when training the upstream unsupervised models.

\paragraph{Training.}
We then train the downstream model to predict $\ybpred$ from $\zb=r(\xb)$ using the Adam optimizer with an initial learning rate of 1e-3 and a batch size of 64, for a maximum of $\num{6000}$ steps. The learning rate is halved every $\num{2000}$ steps. We perform early stopping as follows: We use the validation set to compute the (in-distribution) validation loss every $\num{250}$ training steps---if the loss does not decrease by more than $0.01$ for 3 evaluations (750 steps), training is interrupted.
In this stage, the representation for each image is fixed, i.e. the representation function $r$ is never updated.
The loss is computed independently for each object property, and is a sum of MSE and cross-entropy terms, depending on whether an object property is numerical or categorical.

\paragraph{Downstream training and evaluation under distribution shifts.}
As mentioned earlier, when doing in-distribution evaluation we simply have $\mathcal{D}_s = \mathcal{D}_t$. In the general case, we may for example train on the original Multi-dSprites dataset, and test on the Multi-dSprites variant that has an unseen shape or an occlusion. In the special case in which we allow retraining of the downstream model (see \cref{sec:results/hyp2,sec:results/hyp3}), we still have $\mathcal{D}_s = \mathcal{D}_t$, but they are both OOD with respect to the original ``clean'' dataset used for training the unsupervised models.

Under distribution shifts, the representations $r(\xb)$ might be inaccurate, which might bias our downstream results. Although there is no perfect solution to this issue, we attempt to reduce as much as possible the potential effect of distribution shifts on the training and evaluation of downstream models.
When distribution shifts affect global scene properties, there is no alternative but to train and evaluate the models as usual. 
When distribution shifts affect single objects, however, we can assume that the representations of the ID objects are not as severely affected by the shift, and only use these for training downstream models. 

Here we consider the case where the test dataset $\mathcal{D}_t$ has an object-level distribution shift, and the training dataset $\mathcal{D}_s$ is either the original ``clean'' dataset or the same as $\mathcal{D}_t$. 
At \textbf{train time}, we ignore OOD objects (if any) both when matching slots with objects and when training downstream property prediction models. Note that, when the training dataset $\mathcal{D}_s$ is the original ``clean'' dataset, the downstream models are always trained as usual because there are no OOD objects.
At \textbf{test time}, there are a few cases depending on the matching strategy:
\begin{itemize}
    \item When using mask matching, we consider \emph{all} objects for matching, and evaluate the downstream models on all objects. We then report test results on ID and OOD objects separately.

    \item When using loss matching, we cannot match all ground-truth objects, since the OOD objects might have OOD categorical properties (in our setup, the downstream models cannot predict classes that were not seen during training). Therefore, we resort to a two-step matching approach: we first match slots to all objects using the prediction loss computed only on the properties that are ID for \emph{all} objects. We then keep only the matches for OOD objects, and repeat the usual loss matching with the remaining slots and objects, using all properties. The OOD objects are thus matched in a relatively fair way, while the matching of the ID objects can be refined at a later step using all available properties.

    \item When using deterministic matching, we cannot exactly follow the two-step matching strategy presented above. Instead, we modify the lexicographic order to give a higher weight to OOD features of OOD objects, so the corresponding objects are pushed down in the order while maintaining the order given by more significant (according to the order) properties. Note that the downstream model in this case might be at a disadvantage if it is trained on a dataset with object-level distribution shifts: the model is now trained to predict only ID objects, so at test time there will be one more target object on average.
\end{itemize}

\subsection{Distribution shifts for OOD evaluation}\label{app:evaluation_shifts}

Here we present more in detail the distribution shifts we apply to images in order to test OOD generalization in different scenarios. Examples are shown in \cref{fig:transform/transforms_datasets}.

\begin{figure}
    \centering
    \includegraphics[width=0.82\linewidth]{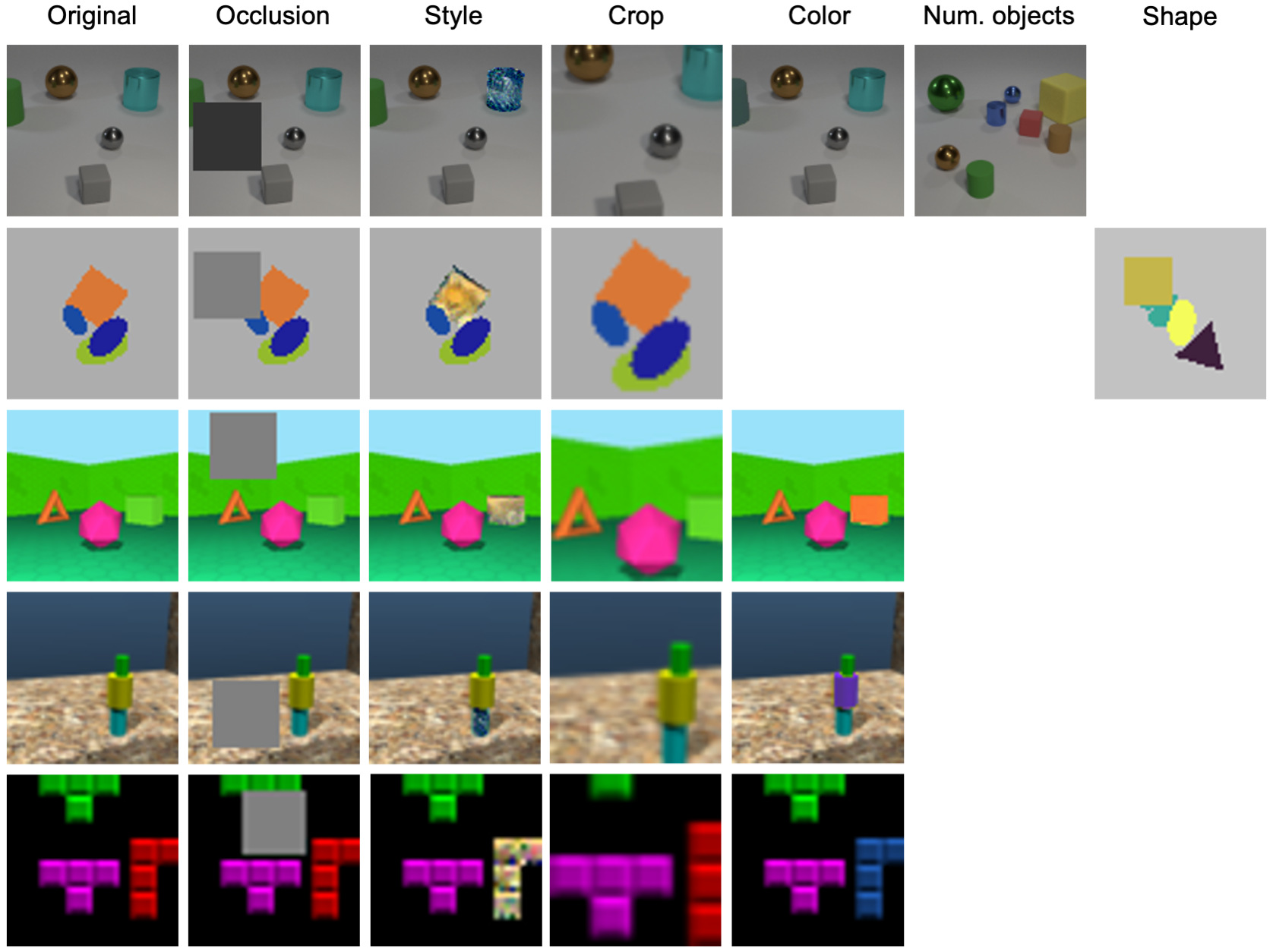}
    \caption{Distribution shifts applied to the different datasets to test generalization.}
    \label{fig:transform/transforms_datasets}
\end{figure}

\paragraph{Occlusion.} A gray square is placed on top of the scene. The position is determined by picking 5 locations uniformly at random (such that the entire square is in the image) and selecting the one that occludes less (in terms of total area) of the foreground objects. The size of the occlusion is $(\lfloor 0.4 \cdot H\rfloor, \lfloor0.4 \cdot W\rfloor)$ with $H$ and $W$ the height and width of the image, respectively. Occluded objects have their mask updated to reflect the occlusion. The occlusion is categorized as background (or first background object in case there are multiple background objects such as in Objects Room). The RGB color of the square is $[0.2, 0.2, 0.2]$ for CLEVR and $[0.5,0.5,0.5]$ for all other datasets.

\paragraph{Object color.} An object is selected uniformly at random and its color is changed by randomly adjusting its brightness, contrast, saturation, and hue, using torchvision's \code{ColorJitter} transform with arguments $[0.5, 0.5, 0.5, 0.5]$ for the four above-mentioned parameters. This transformation is not performed on Multi-dSprites, since the object colors in this dataset cover the entire RGB color space. The \texttt{color} and \texttt{material} properties (when relevant) are not used in downstream tasks.

\paragraph{Crop.} The image and mask are cropped at the center and resized to match their original size. The crop size is $(\lfloor\frac{2}{3}H\rfloor, \lfloor\frac{2}{3}W\rfloor)$ with $H$ and $W$ the original height and width of the image, respectively. When resizing, we use bilinear interpolation for the image and nearest neighbor for the mask.

\paragraph{Object style.} We implement style transfer based on \citet{gatys_neural_2016} and on the PyTorch tutorial by \citet{jacq_neural_nodate}. The first $100$k samples in all datasets are converted using as style image \emph{The Great Wave off Kanagawa} from Hokusai's series \emph{Thirty-six Views of Mount Fuji}. The style is applied only to one foreground object using the object masks. The \texttt{color} and \texttt{material} properties (when relevant) are not used in downstream tasks.

\paragraph{Object shape.} For the Multi-dSprites dataset, a triangle is placed on the scene with properties sampled according to the same distributions defined by the Multi-dSprites dataset. This is performed only on the images where at most 4 objects are present, to mimic changing the shape of an existing object. The depth of the triangle in the object stack is selected uniformly at random as an integer in $[1,5]$. All objects from the selected depth and upwards are moved up by one level to place the new shape underneath them. The objects masks are adjusted accordingly for both the added shape and the objects below it.

%%%%%%%%%%%%%%%%%%%%%%%%%%%%%%%%%%%%%%

\section{Additional results}
\label{app:results}

In this section, we report additional quantitative results and show qualitative performance
on all datasets for a selection of object-centric models and VAE baselines.

\subsection{Performance in the training distribution}

% Section 4.1

\cref{fig:indistrib/metrics/box_plots} shows the distributions of the reconstruction MSE and all the segmentation metrics, broken down by dataset and model.
The relationship between these metrics is also shown in scatter plots in \cref{fig:indistrib/metrics_metrics/aggregated_datasets/scatter}.
As discussed in \cref{sec:results/metrics}, we observe that segmentation covering metrics are correlated with the ARI only in some cases, and the models are ranked very differently depending on the chosen segmentation metric. In particular, we observe here that Slot Attention achieves a high ARI score and significantly lower (m)SC scores on CLEVR, Multi-dSprites, and Tetrominoes. This is because Slot Attention on these datasets tends to model the background across many slots (see \cref{app:results_qualitative}), which is penalized by the denominator of the IOU in the (m)SC scores (see \cref{eq:sc,eq:iou,eq:msc} in \cref{app:evaluation_metrics}). This behavior should not have a major effect on downstream performance, which is confirmed by the strong and consistent correlation between ARI and downstream performance (see also \cref{sec:results/hyp1} and \cref{fig:indistrib/downstream/corr_metrics_downstream_all}).

% Section 4.2

\cref{fig:indistrib/downstream/box_plots} shows an overview of downstream factor prediction performance on all labeled datasets (one per column), using as downstream predictors a linear model or an MLP with up to 3 hidden layers (one model per row). The MLP1 results are also shown in \cref{sec:results/hyp1} (\cref{fig:indistrib/downstream/barplots_loss_linear_main}). We report results separately for each object-centric model and for each ground-truth object property. The metrics used here are accuracy for categorical attributes and \rsq for numerical attributes.
We generally observe consistent trends across downstream models. 
In \cref{fig:indistrib/downstream/barplots_compare_downstream_models_loss}, we show the same results in a different way, to directly compare downstream models (here the median baselines for slot-based and distributed representations are shown as horizontal lines on top of the relevant bars). Using larger downstream models tends to slightly improve test performance but, interestingly, in many cases the effect is negligible. There are however a few cases in which using a larger model significantly boosts test performance in object property prediction. In some cases it seems sufficient to use a small MLP with one hidden layer instead of a linear model (e.g., color prediction in CLEVR with Slot Attention, shape prediction in CLEVR with MONet and Slot Attention, color prediction in Tetrominoes with MONet, GENESIS, and SPACE, or location prediction in Multi-dSprites with SPACE), while in other cases we get further gains by using even larger models (e.g., shape prediction in Multi-dSprites with SPACE, and shape prediction in Tetrominoes with all models except Slot Attention which already achieves a perfect score with a linear model).
Results for VAEs are generally less interpretable because the performance is often too close to the naive baseline. However, in some cases using deeper downstream models has clear benefits: e.g., shape prediction in Tetrominoes and color prediction in Shapestacks improve from baseline level when using a linear model to a relatively high accuracy when using one or two hidden layers. In other cases, a linear model already works relatively well even from distributed representations---although significantly worse than object-centric representations---and using deeper downstream models is not beneficial (e.g., color and size prediction in CLEVR). Finally, in many other cases, larger downstream models do not seem to be sufficient to improve performance from VAE representations, confirming that often the relevant information may not be easily accessible and suggesting that object-centric representations may be generally beneficial.

In \cref{fig:indistrib/downstream/corr_metrics_downstream_all} we show the Spearman rank correlations between evaluation metrics and downstream performance with all considered combinations of slot matching (loss- and mask-based) and downstream model (linear, MLP with 1, 2, or 3 hidden layers). The trends are broadly consistent in all combinations, except that correlations with ARI tend to be stronger (perhaps unsurprisingly) when using mask matching, and when using larger downstream models.

\begin{figure}[hb]
    \centering
    %    \vspace{15pt}
    \includegraphics[width=\textwidth]{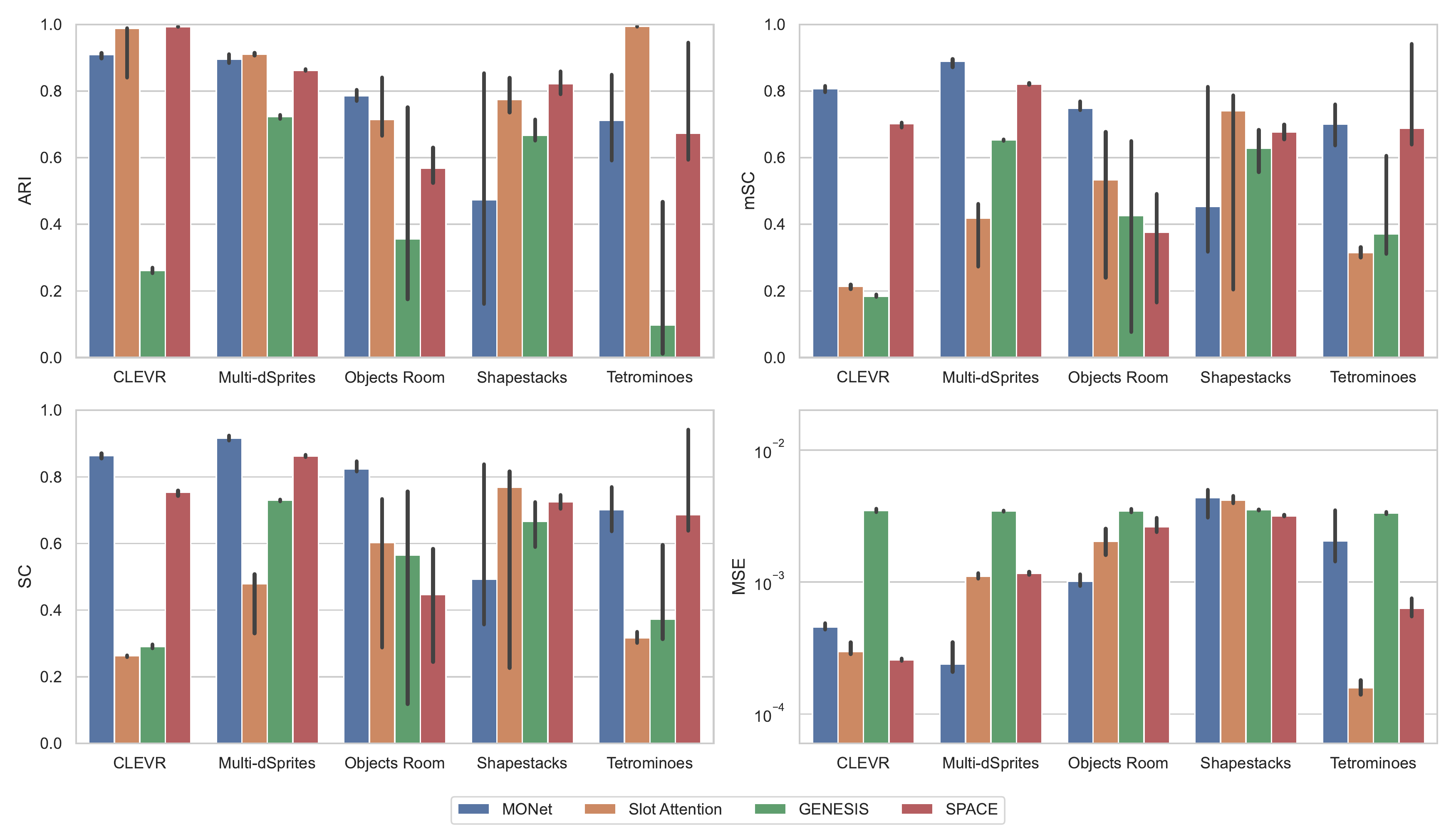}
    \caption{Overview of segmentation metrics (ARI \up, mSC \up, SC \up) and reconstruction MSE (\down) in distribution (test set of $\num{2000}$ images). The bars show medians and 95\% confidence intervals with 10 random seeds.}
    \label{fig:indistrib/metrics/box_plots}
\end{figure}

\begin{figure}
    \centering
    \includegraphics[width=\textwidth]{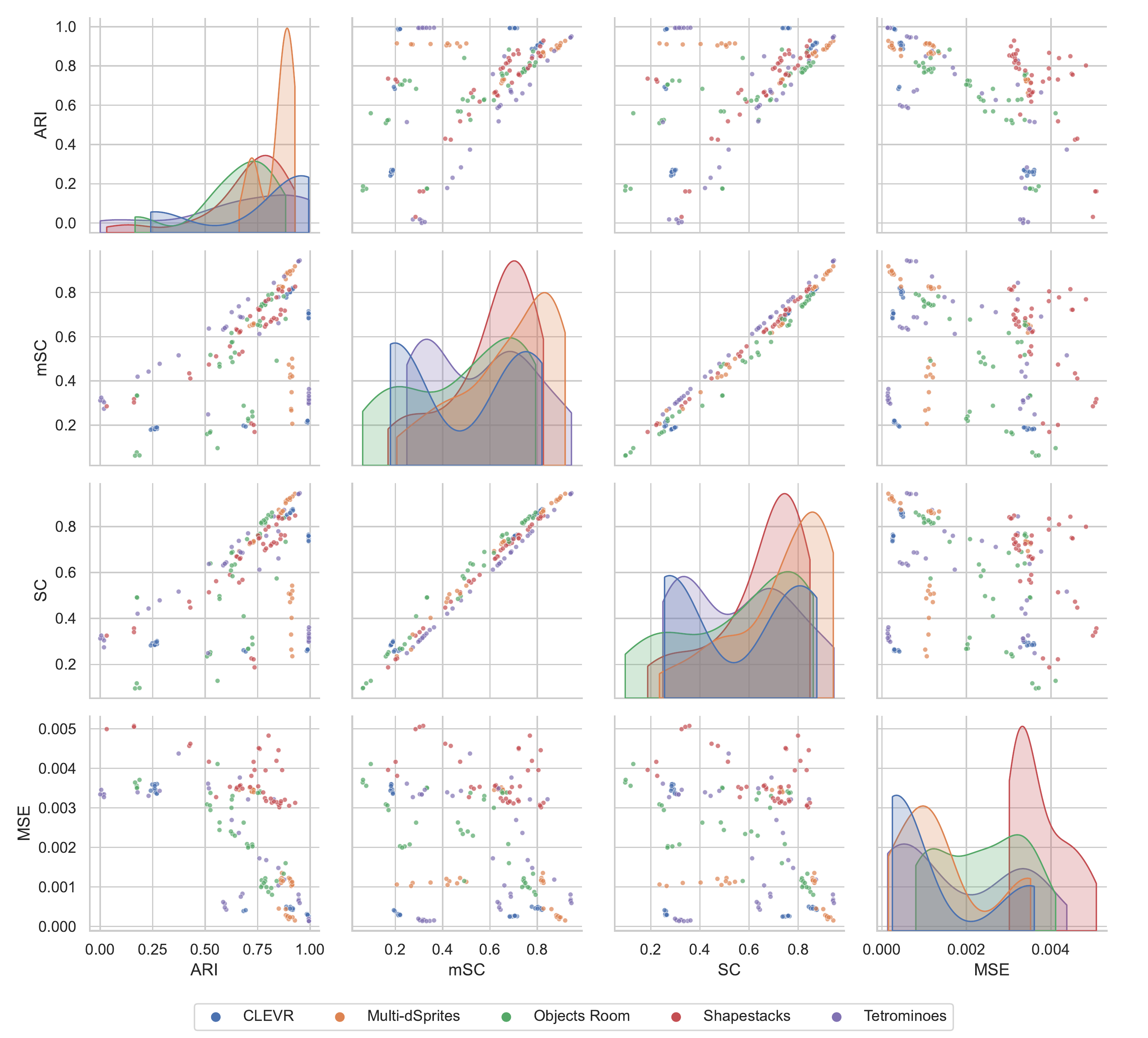}
    \caption{Scatter plots between metrics over all 200 object-centric models, color-coded by dataset. Diagonal plots: kernel density estimation (KDE) of the quantities on the x-axes.}
    \label{fig:indistrib/metrics_metrics/aggregated_datasets/scatter}
\end{figure}

\begin{figure}
    \centering
    \includegraphics[width=\textwidth]{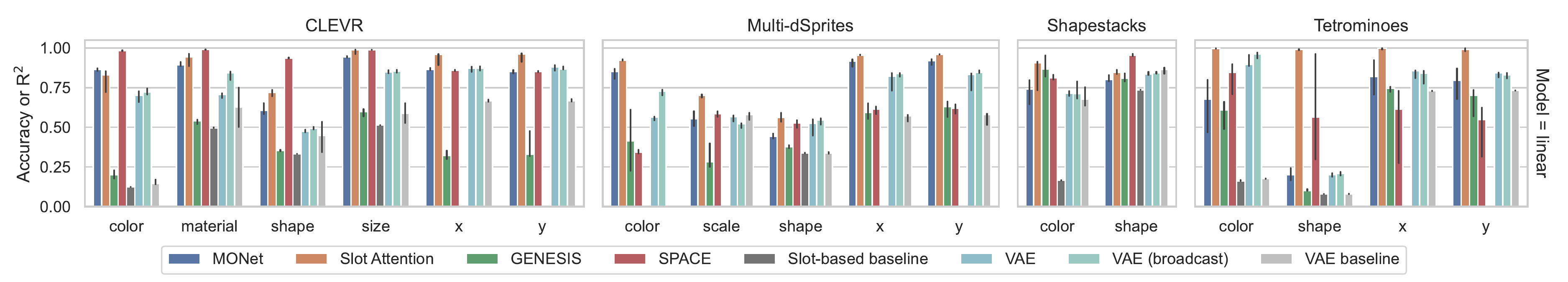}
    
    \vspace{10pt}
    \includegraphics[width=\textwidth]{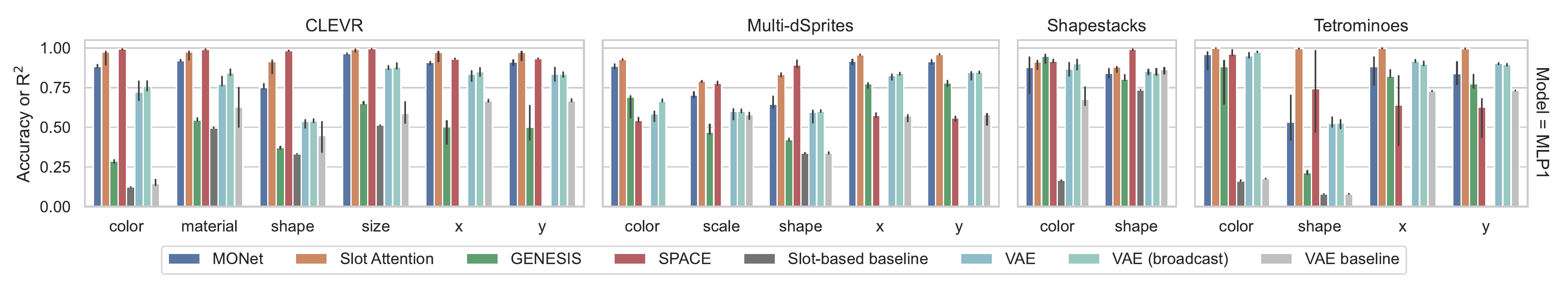}
    
    \vspace{10pt}
    \includegraphics[width=\textwidth]{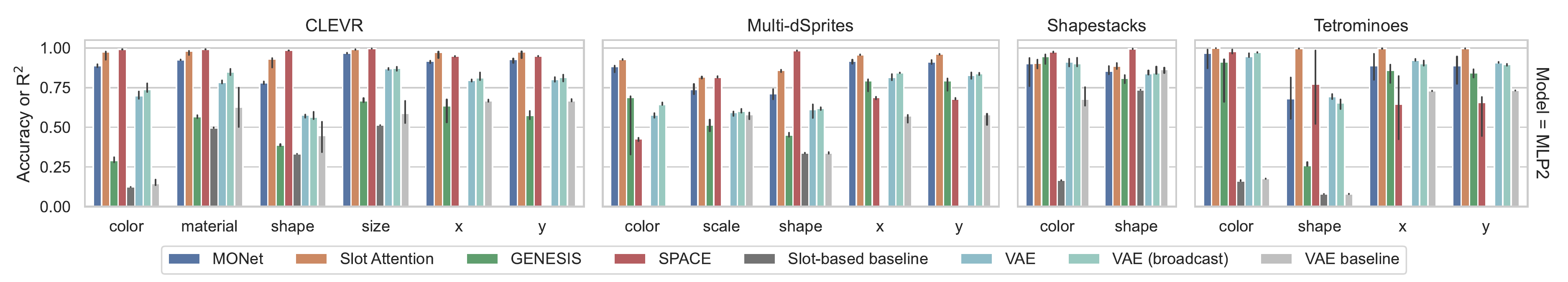}
    
    \vspace{10pt}
    \includegraphics[width=\textwidth]{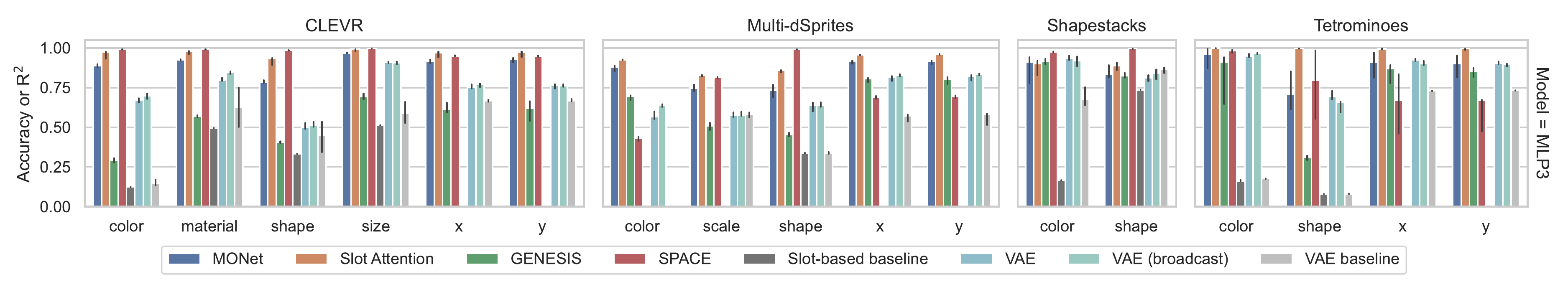}
    \caption[Overview of downstream performance in the training distribution (test set of $\num{2000}$ images) for object-centric models and VAEs, with respective baselines.]{Overview of downstream performance in the training distribution (test set of $\num{2000}$ images) for object-centric models and VAEs, with respective baselines. The metrics on the y-axes are accuracy (\up) for categorical properties and \rsq (\up) for numerical features. Each row show results for a different downstream prediction model. From top to bottom: linear, MLP with 1, 2, and 3 hidden layers (see annotation on the right). We use loss matching (see \cref{sec:experimental_setup}) for all models. The bars show medians and 95\% confidence intervals with 10 random seeds.}
    \label{fig:indistrib/downstream/box_plots}
\end{figure}

\begin{figure}
    \centering
    \includegraphics[height=2.82cm]{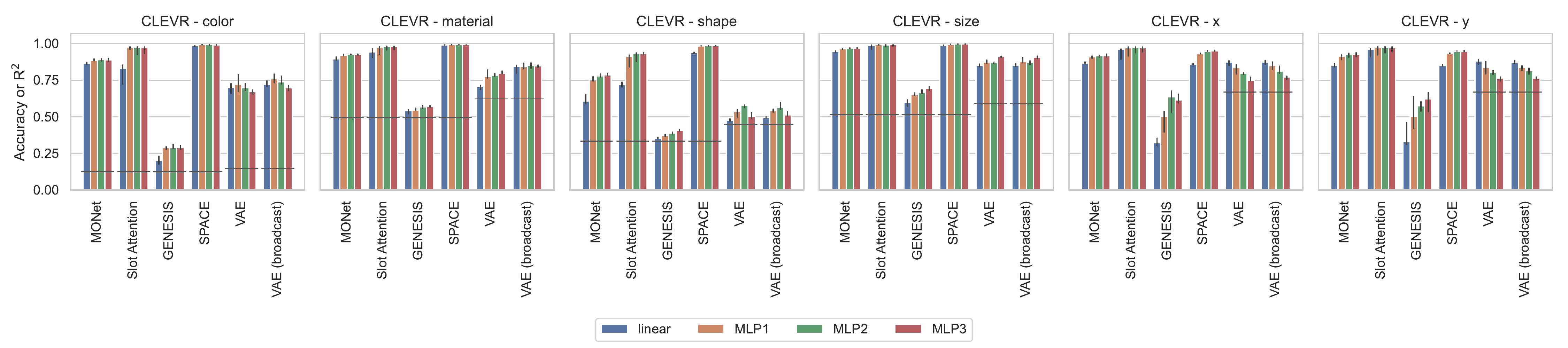}
    
    \vspace{20pt}
    \includegraphics[height=2.82cm]{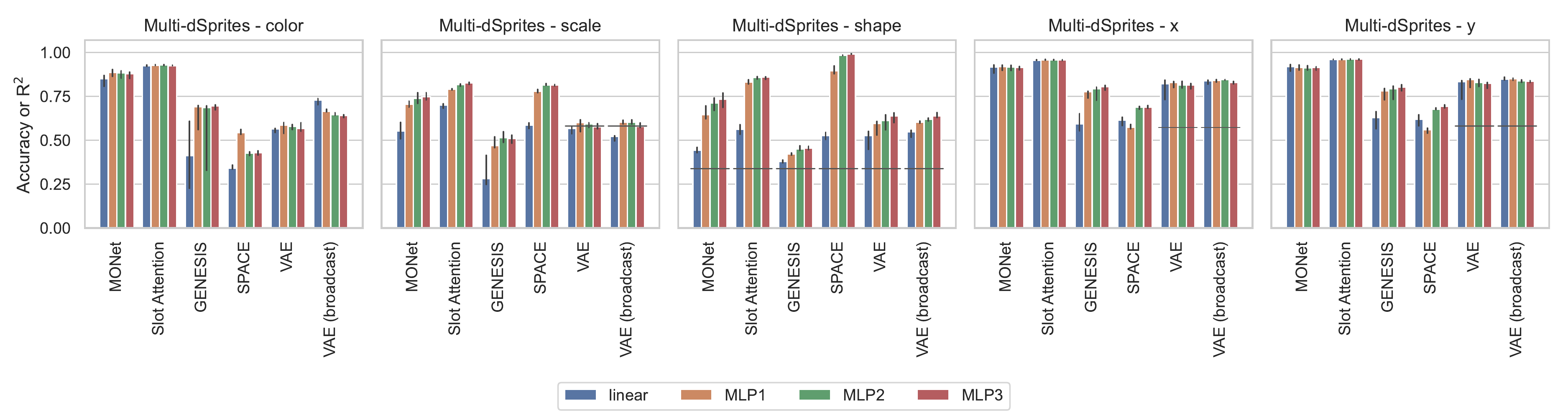}
    
    \vspace{20pt}
    \includegraphics[height=2.82cm]{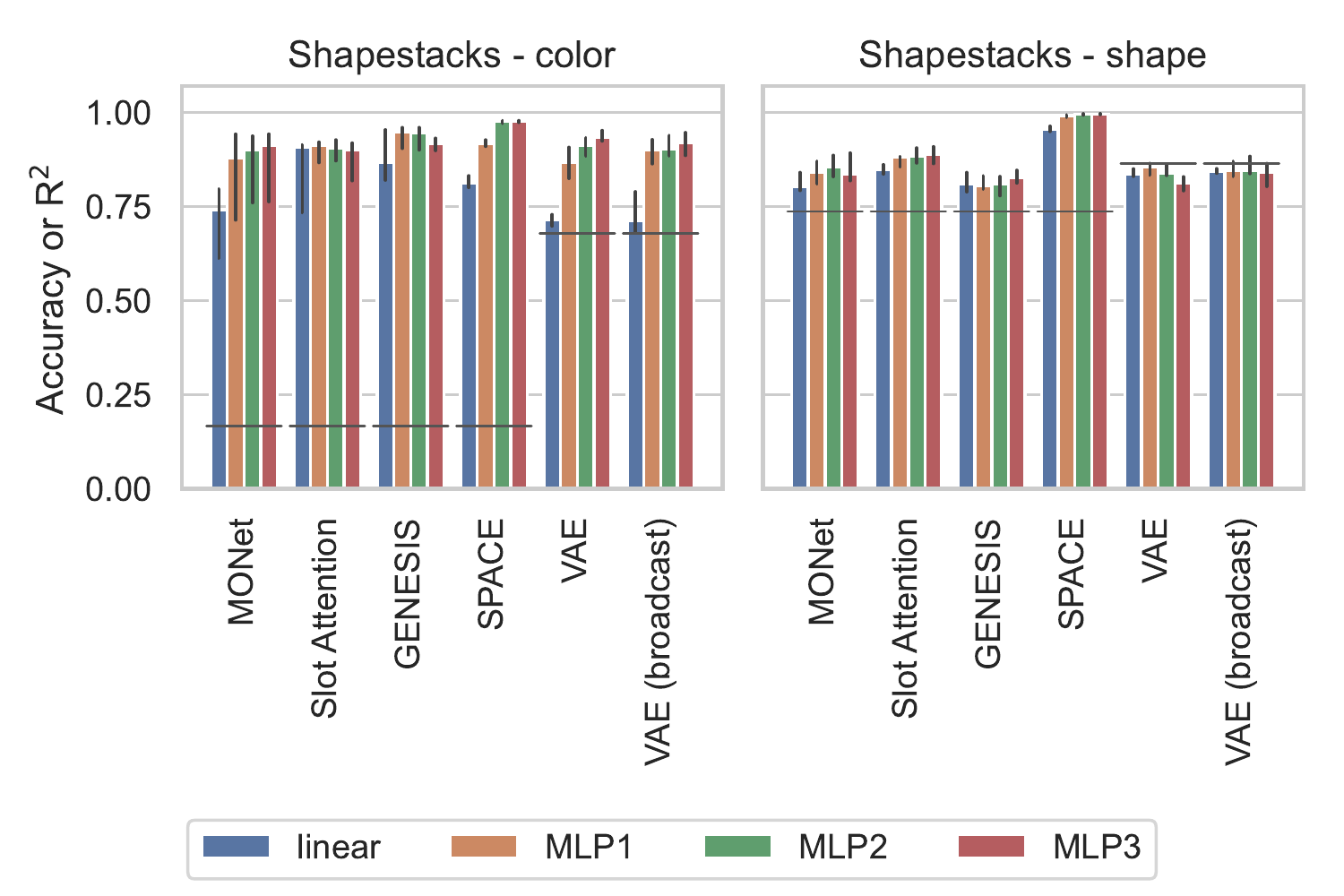}
    
    \vspace{20pt}
    \includegraphics[height=2.82cm]{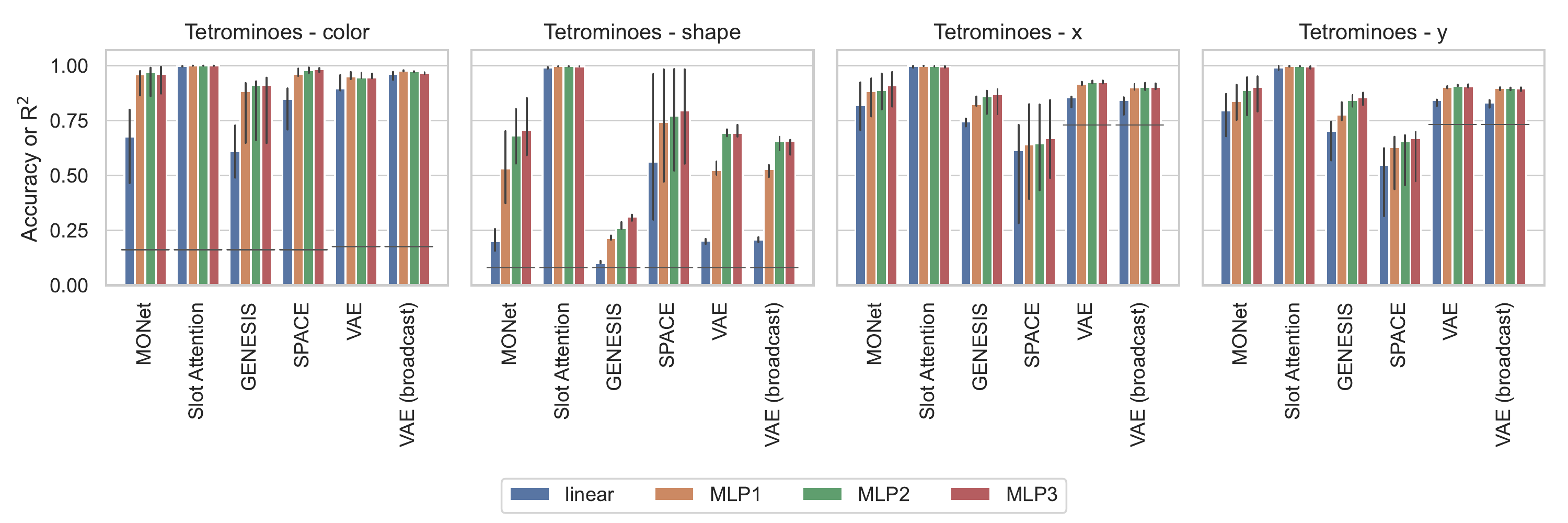}
    
    \caption[Comparing property prediction performance of different downstream models (linear, MLP with 1 to 3 hidden layers), using loss matching (see \cref{sec:experimental_setup}). Results on a test set of $\num{2000}$ images in the training distribution of the upstream unsupervised models.]{Comparing property prediction performance of different downstream models (linear, MLP with 1 to 3 hidden layers), using loss matching (see \cref{sec:experimental_setup}). Results on a test set of $\num{2000}$ images in the training distribution of the upstream unsupervised models. Each plot shows the test performance on one feature of a dataset. We show results for all object-centric models and VAEs, and indicate the baseline (see \cref{sec:experimental_setup}) with a horizontal line (not visible when the baseline is 0).
    The metrics on the y-axes are accuracy (\up) for categorical properties and \rsq (\up) for numerical features.
    The bars show medians and 95\% confidence intervals with 10 random seeds.}
    \label{fig:indistrib/downstream/barplots_compare_downstream_models_loss}
\end{figure}

\begin{figure}
    \centering
    \includegraphics[width=0.75\textwidth]{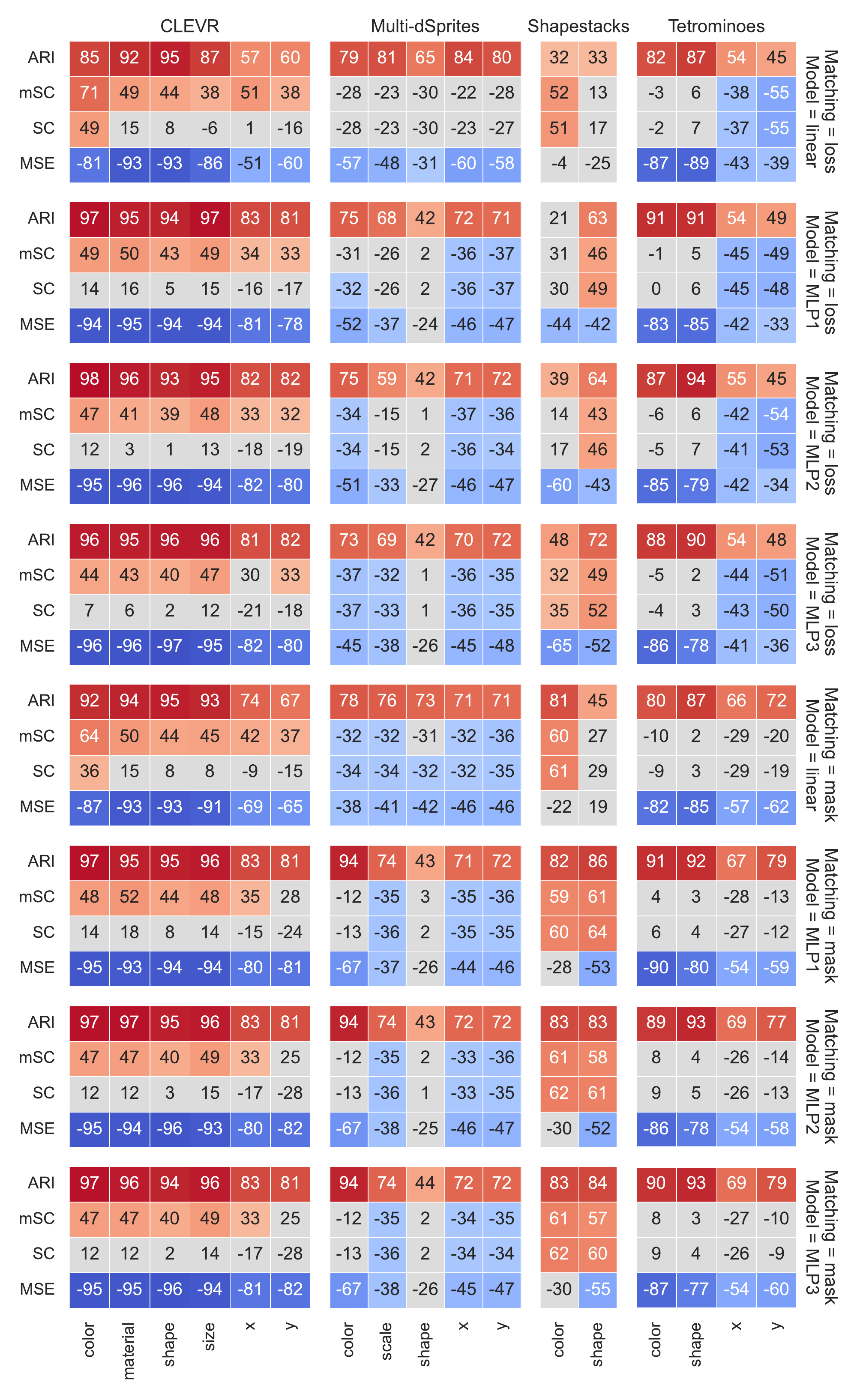}
    \caption{Spearman rank correlations between evaluation metrics and downstream performance with all considered combinations of slot matching (loss- and mask-based) and downstream model (linear, MLP with 1, 2, or 3 hidden layers). The correlations are color-coded only when p\textless 0.05.}
    \label{fig:indistrib/downstream/corr_metrics_downstream_all}
\end{figure}

\clearpage
\subsection{Performance under distribution shifts}

\subsubsection{Segmentation and reconstruction}

In \cref{fig:ood/metrics/generalization_compositional/box_plots}, we report the distributions of the reconstruction MSE and segmentation metrics in scenarios where one object is OOD. Results are split by dataset, model, and type of distribution shift.
As discussed in \cref{sec:results/hyp2}, the SC and mSC scores show a compatible but less pronounced trend, while the MSE more closely mirrors ARI. Notably, in many cases when we alter object style or color, the reconstruction MSE increases significantly while the ARI is only mildly affected. This suggests that the models are still capable of separating the objects but, unsurprisingly, they fail at reconstructing them properly as they have features that were never encountered during training.

\cref{fig:ood/metrics/generalization_global/box_plots} shows analogous results when the distribution shift affects global scene properties. Here we observe that segmentation performance is relatively robust to occlusion although the MSE increases significantly (as expected, the occlusion cannot be reconstructed properly). Segmentation metrics are also robust to increasing the number of objects in CLEVR---here the MSE also increases, but to a lesser extent, especially for SPACE.

\vspace{20pt}

\begin{figure}[hb]
    \centering
    \begin{minipage}{\linewidth}
        \includegraphics[width=\linewidth]{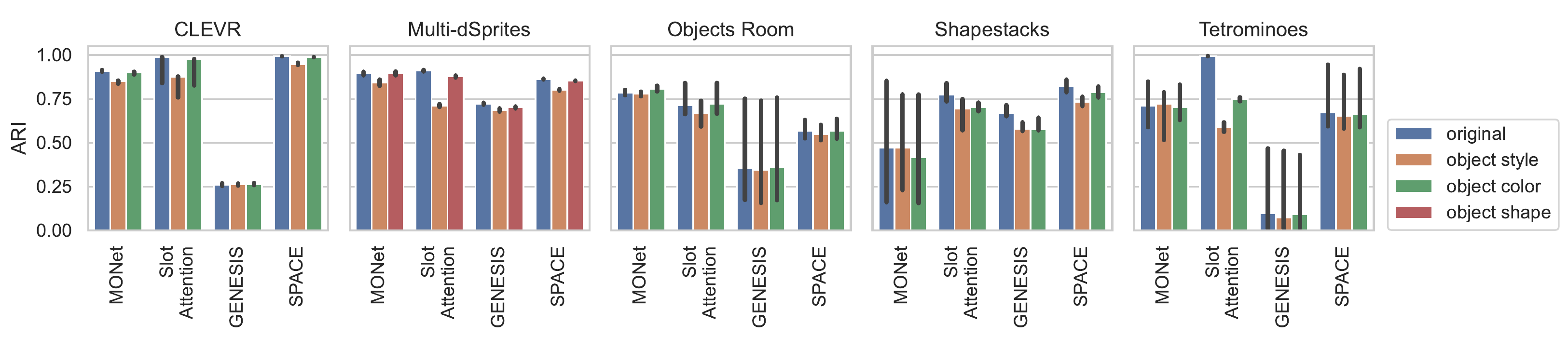}
        
        \vspace{10pt}
        \includegraphics[width=\linewidth]{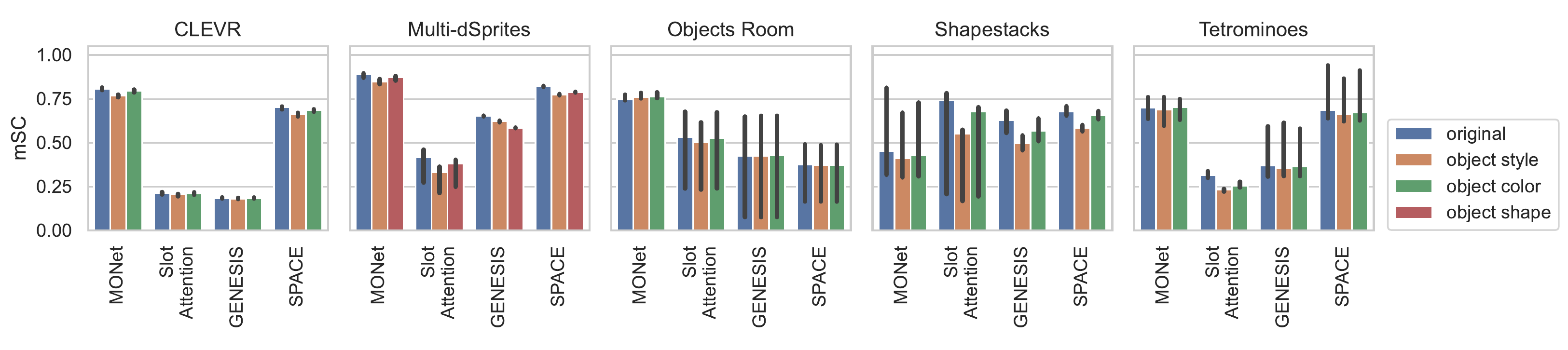}
        
        \vspace{10pt}
        \includegraphics[width=\linewidth]{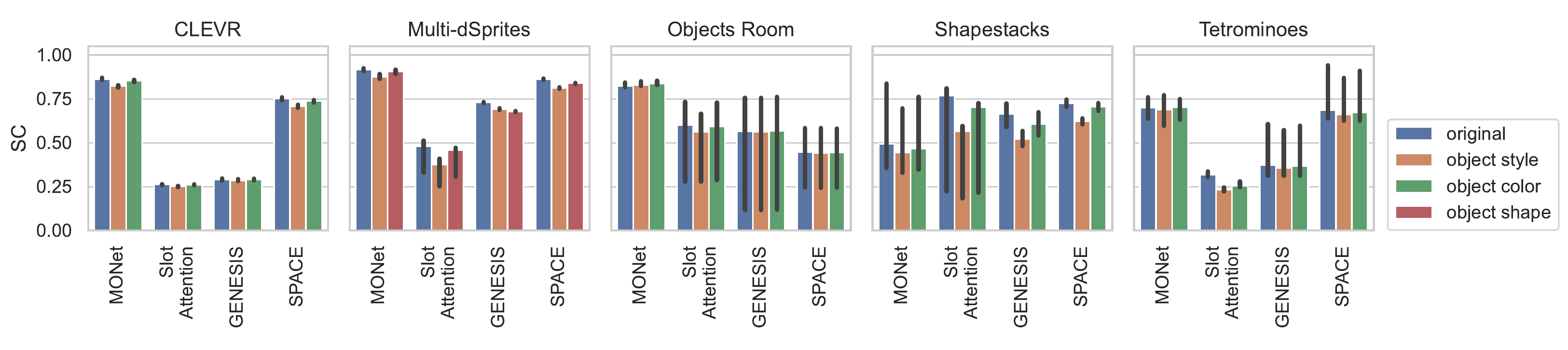}
        
        \vspace{10pt}
        \includegraphics[width=\linewidth]{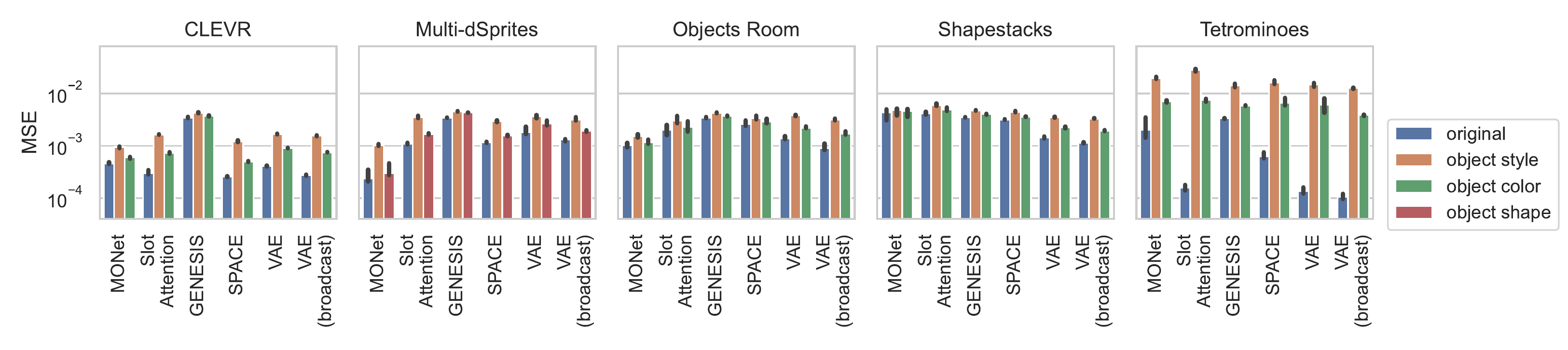}
    \end{minipage}
    \caption{Overview of segmentation metrics (ARI \up, mSC \up, SC \up) and reconstruction MSE (\down) on OOD dataset variants where \textbf{one object} undergoes distribution shift (test set of $\num{2000}$ images). The bars show medians and 95\% confidence intervals with 10 random seeds.}
    \label{fig:ood/metrics/generalization_compositional/box_plots}
\end{figure}

\begin{figure}
    \centering
    \begin{minipage}{\linewidth}
        \includegraphics[width=\linewidth]{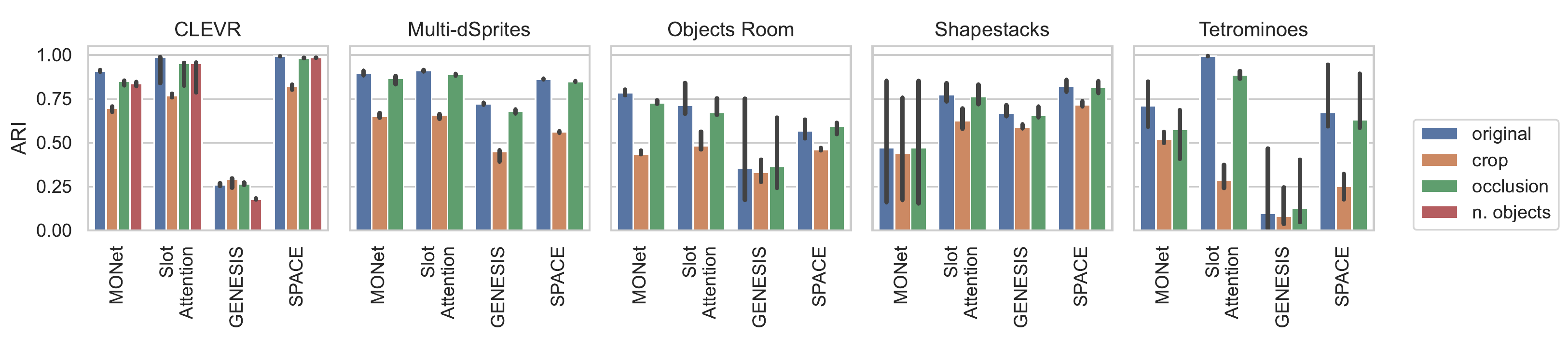}
        
        \vspace{10pt}
        \includegraphics[width=\linewidth]{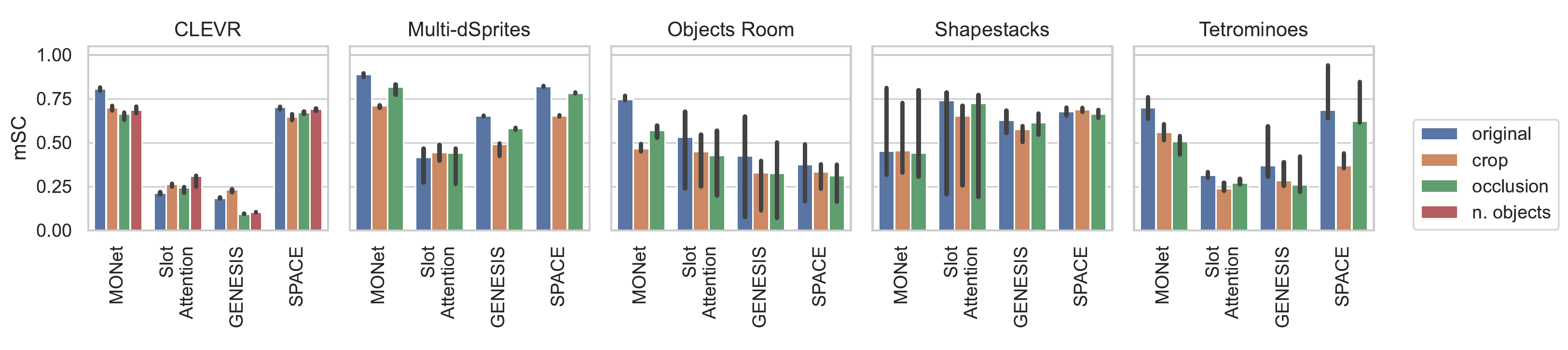}
        
        \vspace{10pt}
        \includegraphics[width=\linewidth]{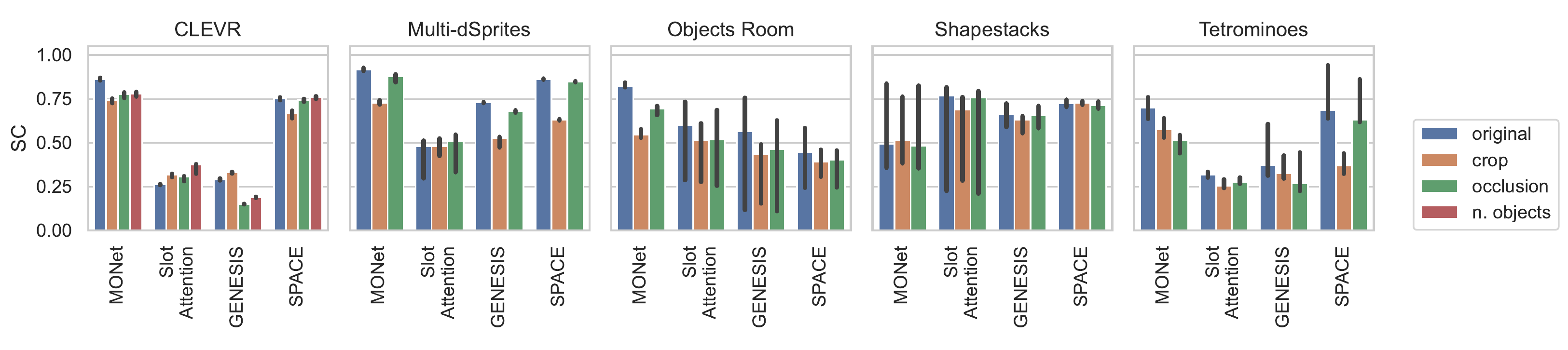}
        
        \vspace{10pt}
        \includegraphics[width=\linewidth]{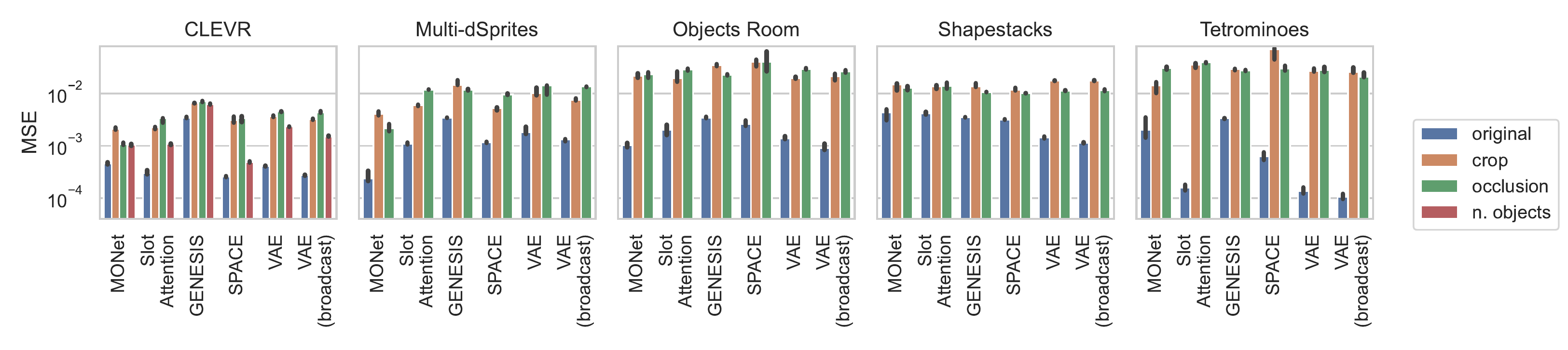}
    \end{minipage}
    \caption{Overview of segmentation metrics (ARI \up, mSC \up, SC \up) and reconstruction MSE (\down) on OOD dataset variants where \textbf{global properties} of the scene are altered (test set of $\num{2000}$ images). The bars show medians and 95\% confidence intervals with 10 random seeds.}
    \label{fig:ood/metrics/generalization_global/box_plots}
\end{figure}

\clearpage
\subsubsection{Downstream performance}

In \crefrange{fig:ood/downstream/slotted/match_loss/object/retrain_False}{fig:ood/downstream/vae/match_deterministic/global/retrain_True}, we show the relationship between ID and OOD downstream prediction performance for the same model, dataset, downstream predictor, and object property. Assume a pretrained unsupervised object discovery model is given, and a downstream model is trained from said model's representations to predict object properties. These plots answer the following question: given that the downstream model predicts a particular object property (e.g., size in CLEVR) with a certain accuracy (on average over all objects in all test images), how well is it going to predict the same property when the scene undergoes one of the possible distribution shifts considered in this study? And in case the distribution shift only affects one object, how well is it going to predict that property in the ID objects as opposed to the OOD objects?
These 16 figures show all combinations of the following 4 factors (hierarchically in this order): object-centric/distributed representations; loss/mask matching for object-centric representations or loss/deterministic for distributed representations; without/with retraining of the downstream model after the distribution shift has occurred; single-object/global distribution shifts.
In each figure, we show results for each of the 4 downstream models considered in this study (linear, and MLP with up to 3 hidden layers). For each of these, we show splits in terms of ID/OOD objects (when applicable), dataset, upstream model, type of distribution shift.

%%%%%%%%%%%%%%%%%%%%

\begin{figure}[hb]
    \centering
    \includegraphics[width=0.9\textwidth]{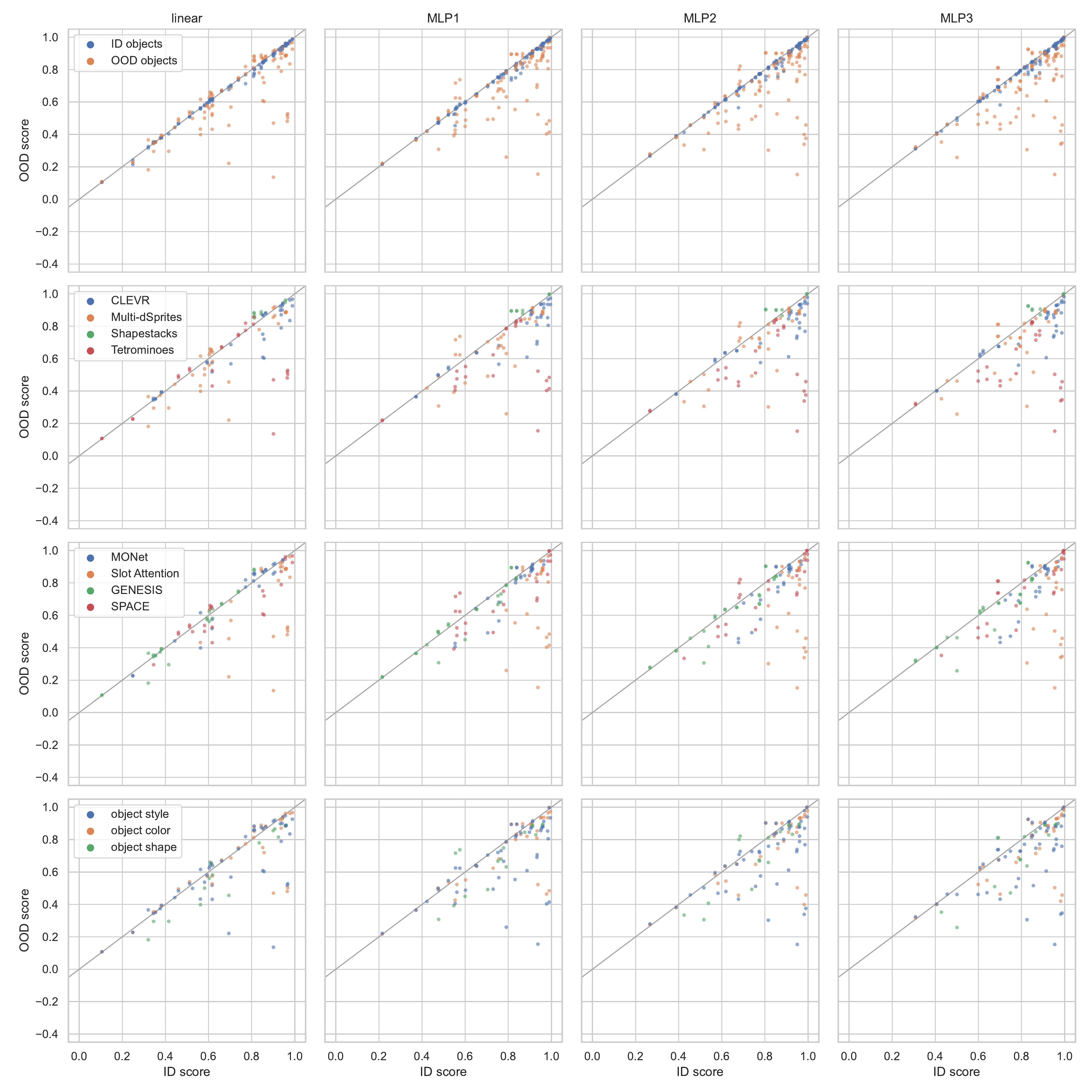}
    \caption[Generalization of \textbf{object-centric representations} in downstream prediction, using \textbf{loss matching} and \textbf{without retraining} the downstream model after the distribution shift. Here the distribution shift affects \textbf{one object}.]{Generalization of \textbf{object-centric representations} in downstream prediction, using \textbf{loss matching} and \textbf{without retraining} the downstream model after the distribution shift. Here the distribution shift affects \textbf{one object}.
    On the x-axis: prediction performance (accuracy or \rsq) for one object property on one dataset, averaged over all objects, on the original training set of the unsupervised object discovery model. On the y-axis: the same metric in OOD scenarios. Each data point corresponds to one representation model (e.g., MONet), one dataset, one object property, one type of distribution shift, and either ID or OOD objects. For each x (performance on one object feature in the training distribution, averaged over objects in a scene and over random seeds of the object-centric models) there are multiple y's, corresponding to different distribution shifts and to ID/OOD objects.
    In the top row, we separately report (color-coded) the performance over ID and OOD objects.
    In the following rows, we only show OOD objects and split according to dataset, model, or type of distribution shift.
    Each column shows analogous results for each of the 4 considered downstream models for property prediction (linear, and MLPs with up to 3 hidden layers).
    }
    \label{fig:ood/downstream/slotted/match_loss/object/retrain_False}
\end{figure}
\begin{figure}
    \centering
    \includegraphics[width=0.9\textwidth]{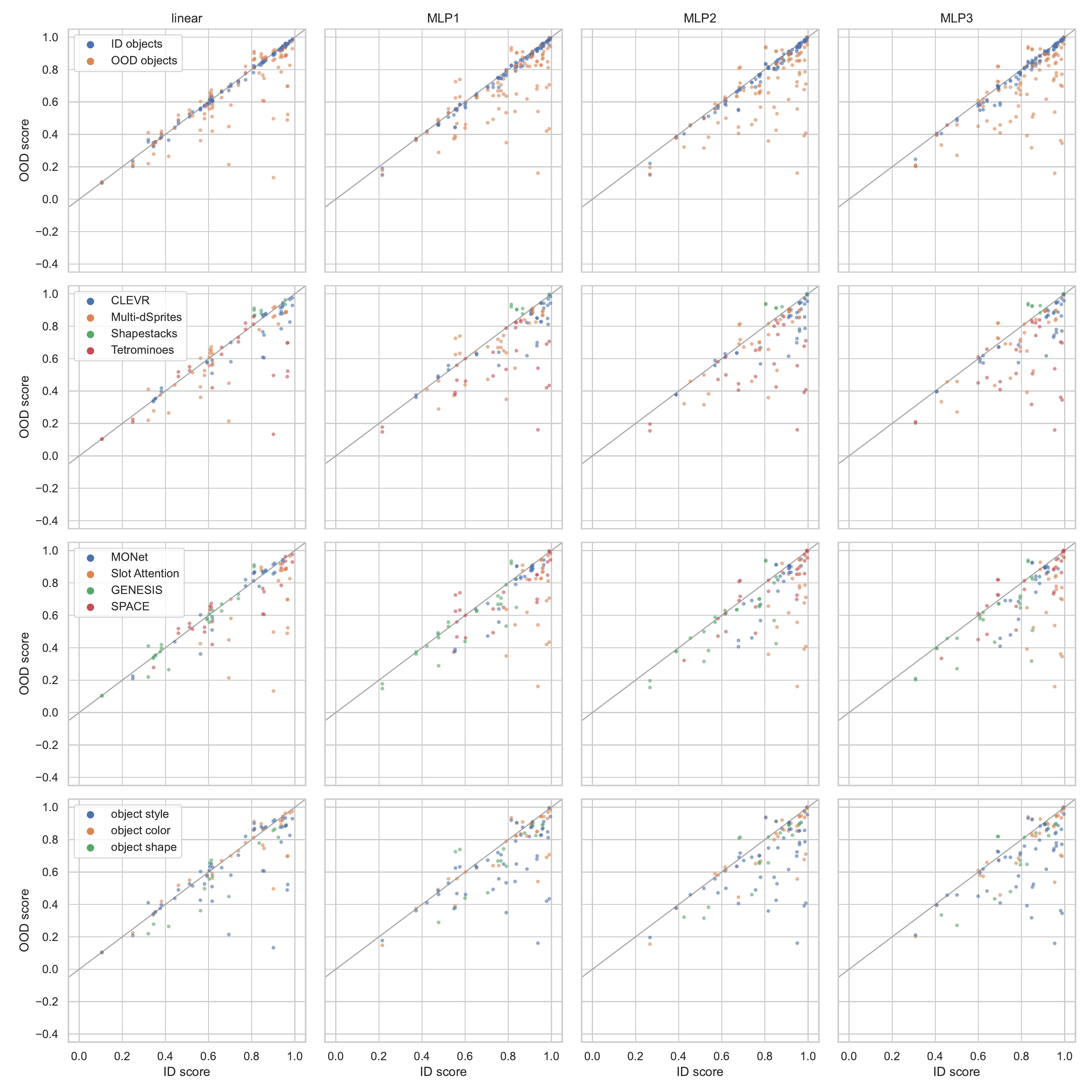}
    \caption[Generalization of \textbf{object-centric representations} in downstream prediction, using \textbf{loss matching} and \textbf{retraining} the downstream model after the distribution shift. Here the distribution shift affects \textbf{one object}.]{Generalization of \textbf{object-centric representations} in downstream prediction, using \textbf{loss matching} and \textbf{retraining} the downstream model after the distribution shift. Here the distribution shift affects \textbf{one object}.
    On the x-axis: prediction performance (accuracy or \rsq) for one object property on one dataset, averaged over all objects, on the original training set of the unsupervised object discovery model. On the y-axis: the same metric in OOD scenarios. Each data point corresponds to one representation model (e.g., MONet), one dataset, one object property, one type of distribution shift, and either ID or OOD objects. For each x (performance on one object feature in the training distribution, averaged over objects in a scene and over random seeds of the object-centric models) there are multiple y's, corresponding to different distribution shifts and to ID/OOD objects.
    In the top row, we separately report (color-coded) the performance over ID and OOD objects.
    In the following rows, we only show OOD objects and split according to dataset, model, or type of distribution shift.
    Each column shows analogous results for each of the 4 considered downstream models for property prediction (linear, and MLPs with up to 3 hidden layers).
    }
    \label{fig:ood/downstream/slotted/match_loss/object/retrain_True}
\end{figure}
\begin{figure}
    \centering
    \includegraphics[width=0.9\textwidth]{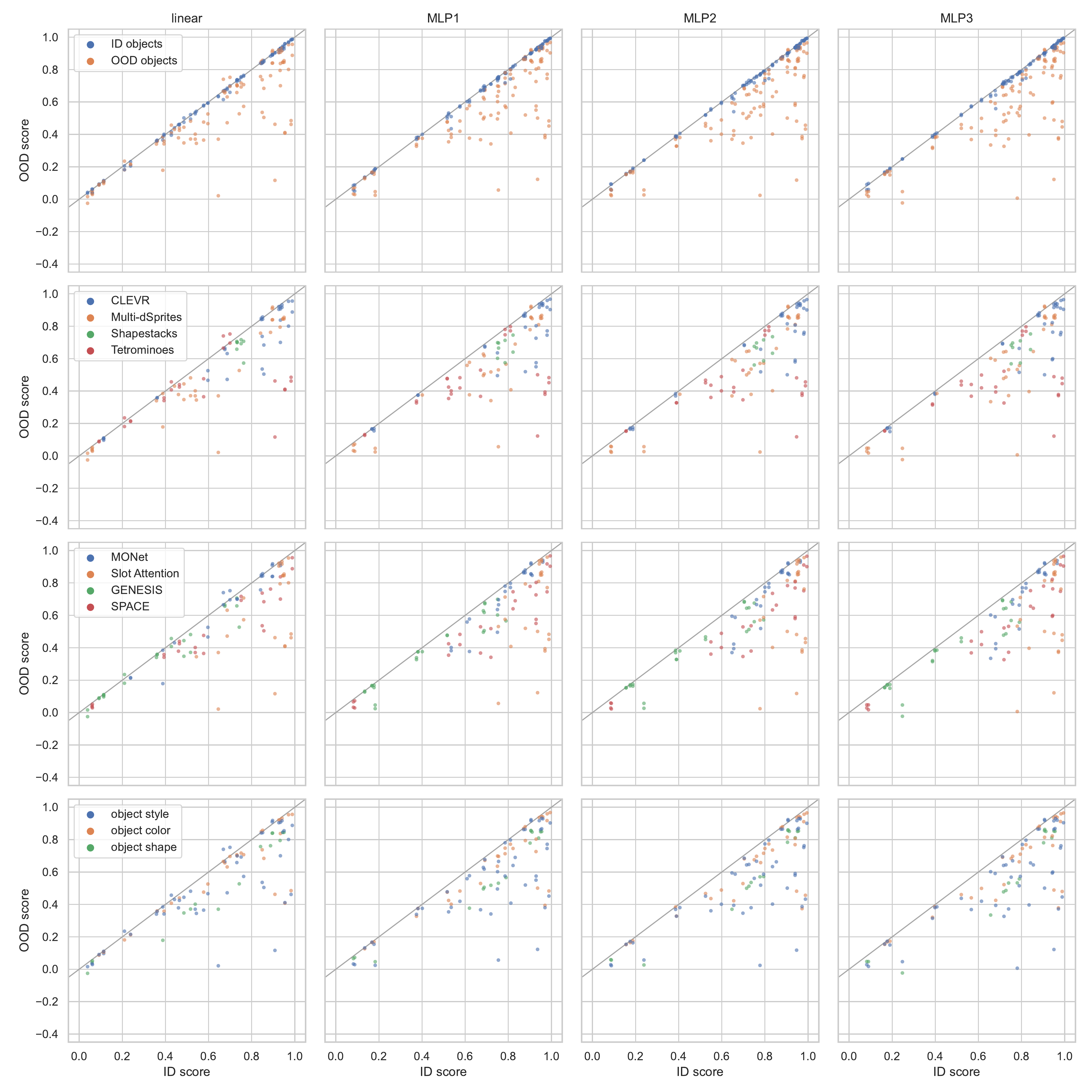}
    \caption[Generalization of \textbf{object-centric representations} in downstream prediction, using \textbf{mask matching} and \textbf{without retraining} the downstream model after the distribution shift. Here the distribution shift affects \textbf{one object}.]{Generalization of \textbf{object-centric representations} in downstream prediction, using \textbf{mask matching} and \textbf{without retraining} the downstream model after the distribution shift. Here the distribution shift affects \textbf{one object}.
    On the x-axis: prediction performance (accuracy or \rsq) for one object property on one dataset, averaged over all objects, on the original training set of the unsupervised object discovery model. On the y-axis: the same metric in OOD scenarios. Each data point corresponds to one representation model (e.g., MONet), one dataset, one object property, one type of distribution shift, and either ID or OOD objects. For each x (performance on one object feature in the training distribution, averaged over objects in a scene and over random seeds of the object-centric models) there are multiple y's, corresponding to different distribution shifts and to ID/OOD objects.
    In the top row, we separately report (color-coded) the performance over ID and OOD objects.
    In the following rows, we only show OOD objects and split according to dataset, model, or type of distribution shift.
    Each column shows analogous results for each of the 4 considered downstream models for property prediction (linear, and MLPs with up to 3 hidden layers).
    }
    \label{fig:ood/downstream/slotted/match_mask/object/retrain_False}
\end{figure}
\begin{figure}
    \centering
    \includegraphics[width=0.9\textwidth]{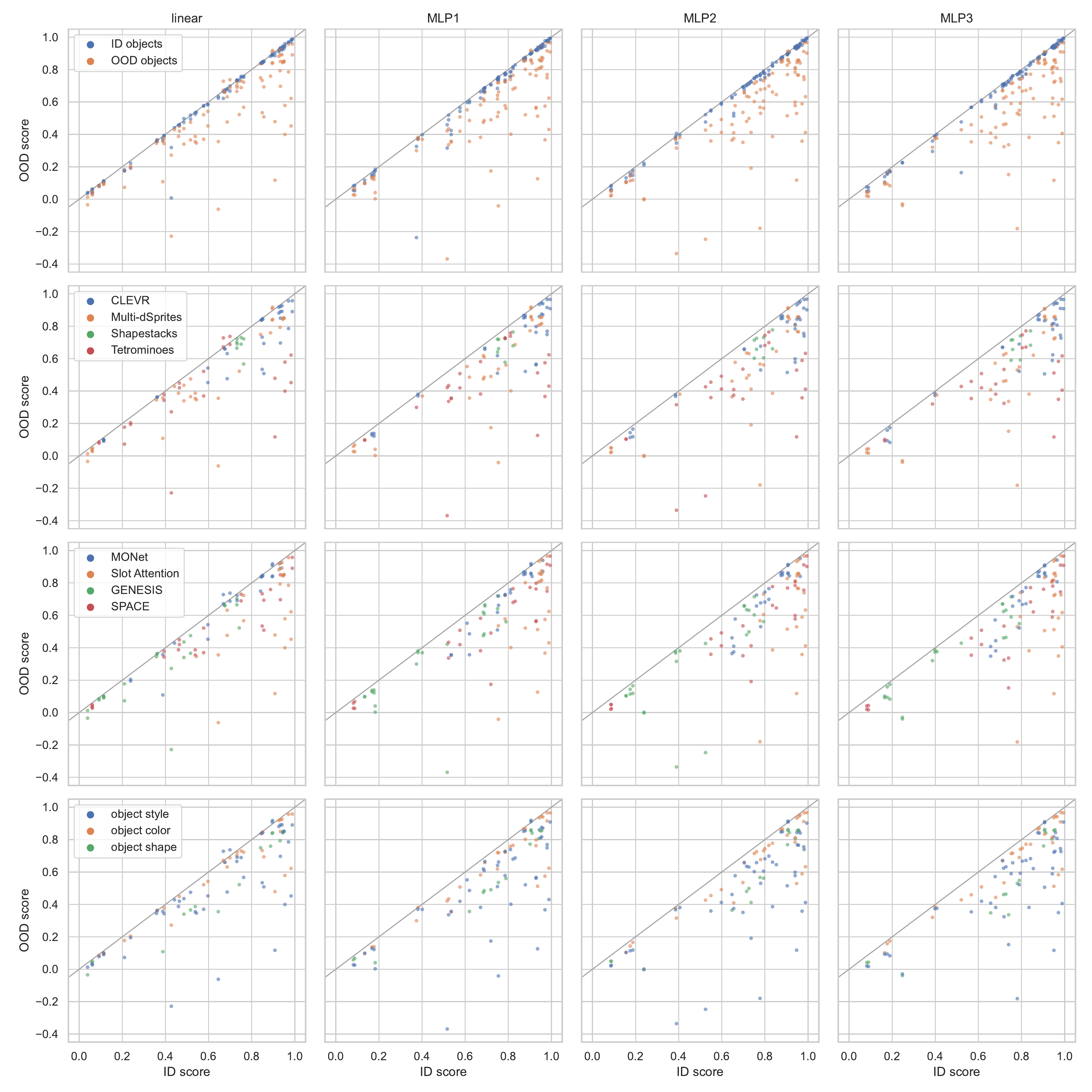}
    \caption[Generalization of \textbf{object-centric representations} in downstream prediction, using \textbf{mask matching} and \textbf{retraining} the downstream model after the distribution shift. Here the distribution shift affects \textbf{one object}.]{Generalization of \textbf{object-centric representations} in downstream prediction, using \textbf{mask matching} and \textbf{retraining} the downstream model after the distribution shift. Here the distribution shift affects \textbf{one object}.
    On the x-axis: prediction performance (accuracy or \rsq) for one object property on one dataset, averaged over all objects, on the original training set of the unsupervised object discovery model. On the y-axis: the same metric in OOD scenarios. Each data point corresponds to one representation model (e.g., MONet), one dataset, one object property, one type of distribution shift, and either ID or OOD objects. For each x (performance on one object feature in the training distribution, averaged over objects in a scene and over random seeds of the object-centric models) there are multiple y's, corresponding to different distribution shifts and to ID/OOD objects.
    In the top row, we separately report (color-coded) the performance over ID and OOD objects.
    In the following rows, we only show OOD objects and split according to dataset, model, or type of distribution shift.
    Each column shows analogous results for each of the 4 considered downstream models for property prediction (linear, and MLPs with up to 3 hidden layers).
    }
    \label{fig:ood/downstream/slotted/match_mask/object/retrain_True}
\end{figure}
\begin{figure}
    \centering
    \includegraphics[width=0.9\textwidth]{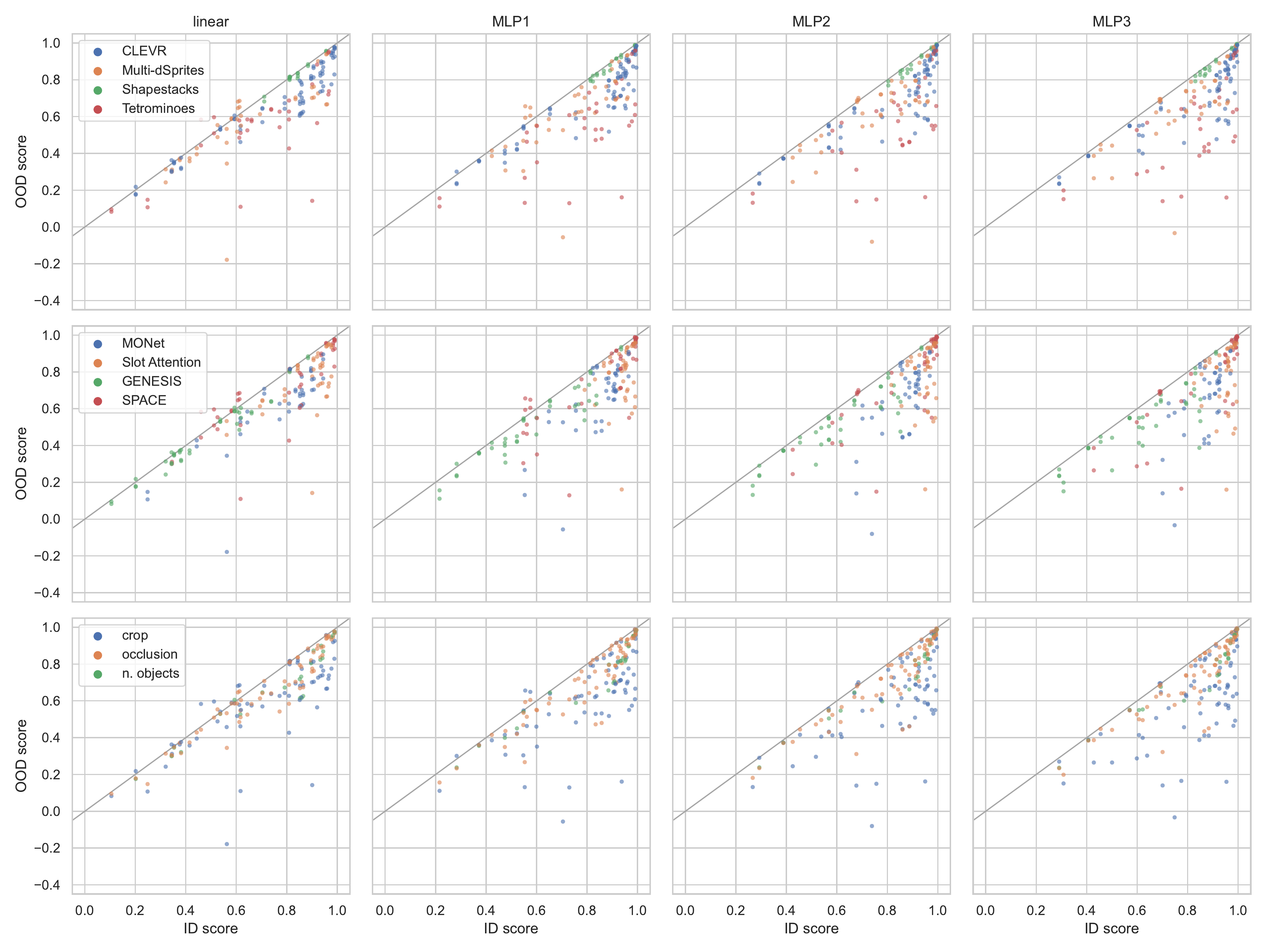}
    \caption[Generalization of \textbf{object-centric representations} in downstream prediction, using \textbf{loss matching} and \textbf{without retraining} the downstream model after the distribution shift. Here the distribution shift affects \textbf{global properties} of the scene.]{Generalization of \textbf{object-centric representations} in downstream prediction, using \textbf{loss matching} and \textbf{without retraining} the downstream model after the distribution shift. Here the distribution shift affects \textbf{global properties} of the scene.
    On the x-axis: prediction performance (accuracy or \rsq) for one object property on one dataset, averaged over all objects, on the original training set of the unsupervised object discovery model. On the y-axis: the same metric in OOD scenarios. Each data point corresponds to one representation model (e.g., MONet), one dataset, one object property, and one type of distribution shift. For each x (performance on one object feature in the training distribution, averaged over objects in a scene and over random seeds of the object-centric models) there are multiple y's, corresponding to different distribution shifts.
    In each row, we color-code the data according to dataset, model, or type of distribution shift.
    Each column shows analogous results for each of the 4 considered downstream models for property prediction (linear, and MLPs with up to 3 hidden layers).
    }
    \label{fig:ood/downstream/slotted/match_loss/global/retrain_False}
\end{figure}
\begin{figure}
    \centering
    \includegraphics[width=0.9\textwidth]{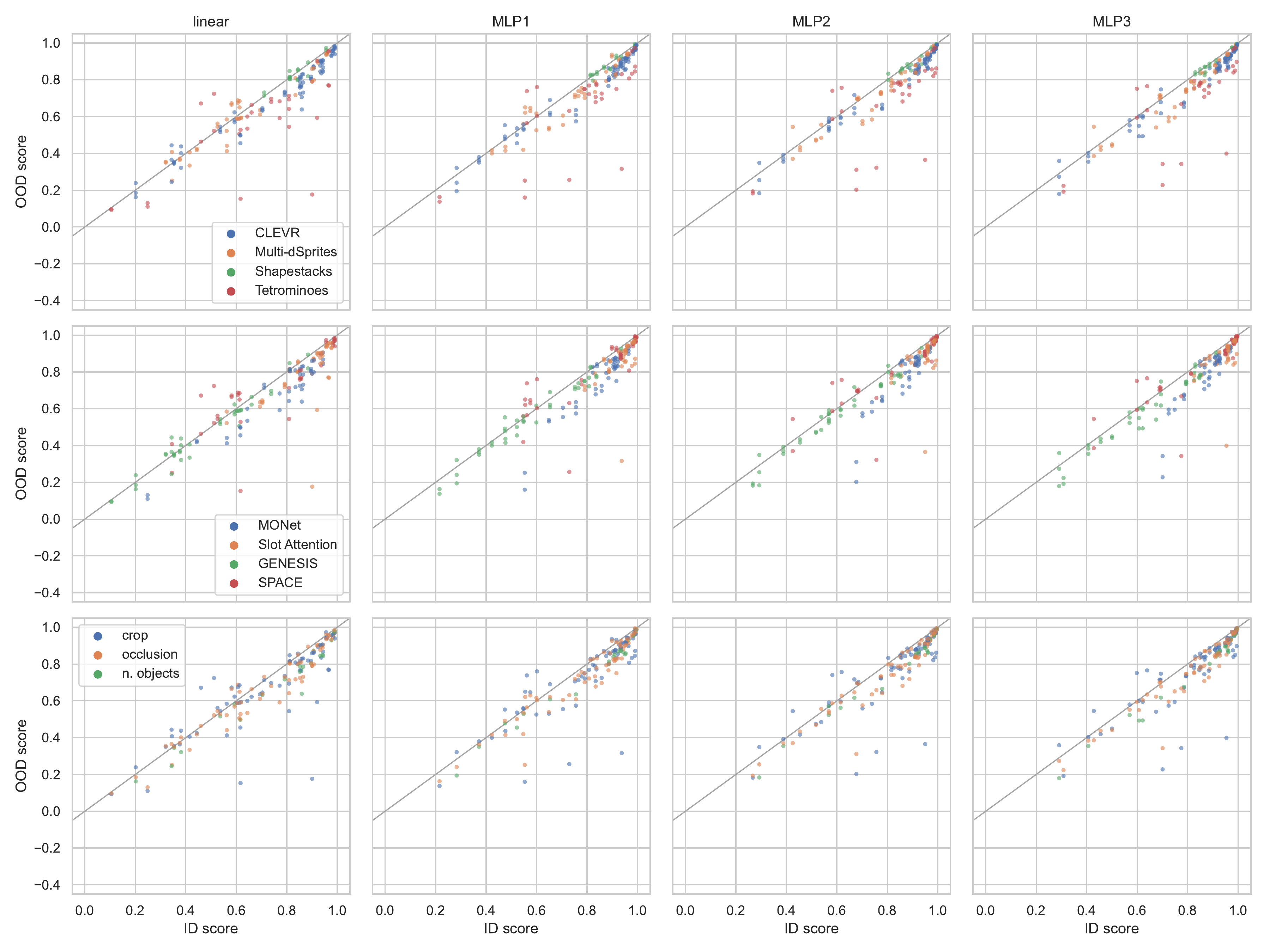}
    \caption[Generalization of \textbf{object-centric representations} in downstream prediction, using \textbf{loss matching} and \textbf{retraining} the downstream model after the distribution shift. Here the distribution shift affects \textbf{global properties} of the scene.]{Generalization of \textbf{object-centric representations} in downstream prediction, using \textbf{loss matching} and \textbf{retraining} the downstream model after the distribution shift. Here the distribution shift affects \textbf{global properties} of the scene.
        On the x-axis: prediction performance (accuracy or \rsq) for one object property on one dataset, averaged over all objects, on the original training set of the unsupervised object discovery model. On the y-axis: the same metric in OOD scenarios. Each data point corresponds to one representation model (e.g., MONet), one dataset, one object property, and one type of distribution shift. For each x (performance on one object feature in the training distribution, averaged over objects in a scene and over random seeds of the object-centric models) there are multiple y's, corresponding to different distribution shifts.
        In each row, we color-code the data according to dataset, model, or type of distribution shift.
        Each column shows analogous results for each of the 4 considered downstream models for property prediction (linear, and MLPs with up to 3 hidden layers).
    }
    \label{fig:ood/downstream/slotted/match_loss/global/retrain_True}
\end{figure}
\begin{figure}
    \centering
    \includegraphics[width=0.9\textwidth]{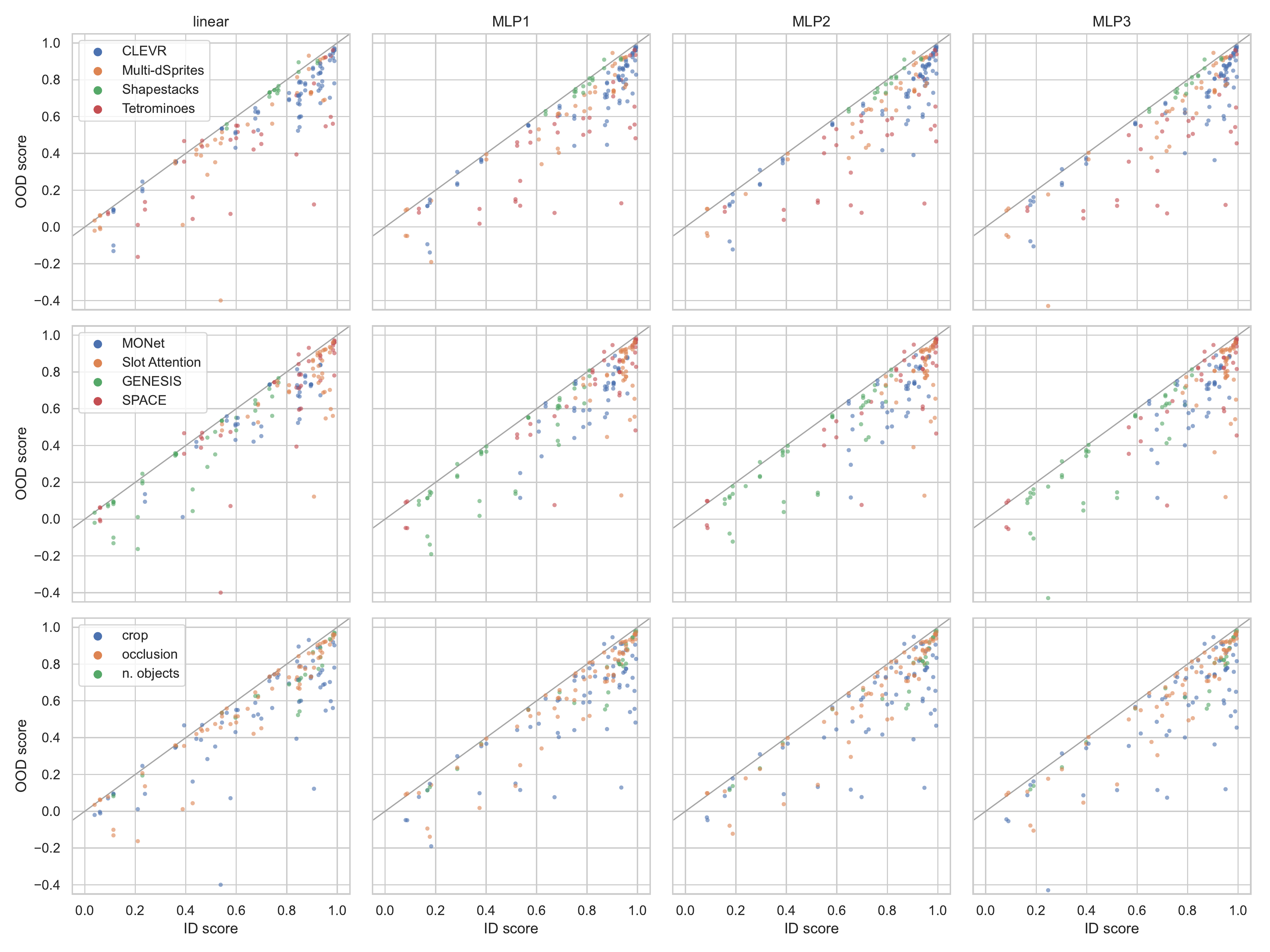}
    \caption[Generalization of \textbf{object-centric representations} in downstream prediction, using \textbf{mask matching} and \textbf{without retraining} the downstream model after the distribution shift. Here the distribution shift affects \textbf{global properties} of the scene.]{Generalization of \textbf{object-centric representations} in downstream prediction, using \textbf{mask matching} and \textbf{without retraining} the downstream model after the distribution shift. Here the distribution shift affects \textbf{global properties} of the scene.
        On the x-axis: prediction performance (accuracy or \rsq) for one object property on one dataset, averaged over all objects, on the original training set of the unsupervised object discovery model. On the y-axis: the same metric in OOD scenarios. Each data point corresponds to one representation model (e.g., MONet), one dataset, one object property, and one type of distribution shift. For each x (performance on one object feature in the training distribution, averaged over objects in a scene and over random seeds of the object-centric models) there are multiple y's, corresponding to different distribution shifts.
        In each row, we color-code the data according to dataset, model, or type of distribution shift.
        Each column shows analogous results for each of the 4 considered downstream models for property prediction (linear, and MLPs with up to 3 hidden layers).
    }
    \label{fig:ood/downstream/slotted/match_mask/global/retrain_False}
\end{figure}
\begin{figure}
    \centering
    \includegraphics[width=0.9\textwidth]{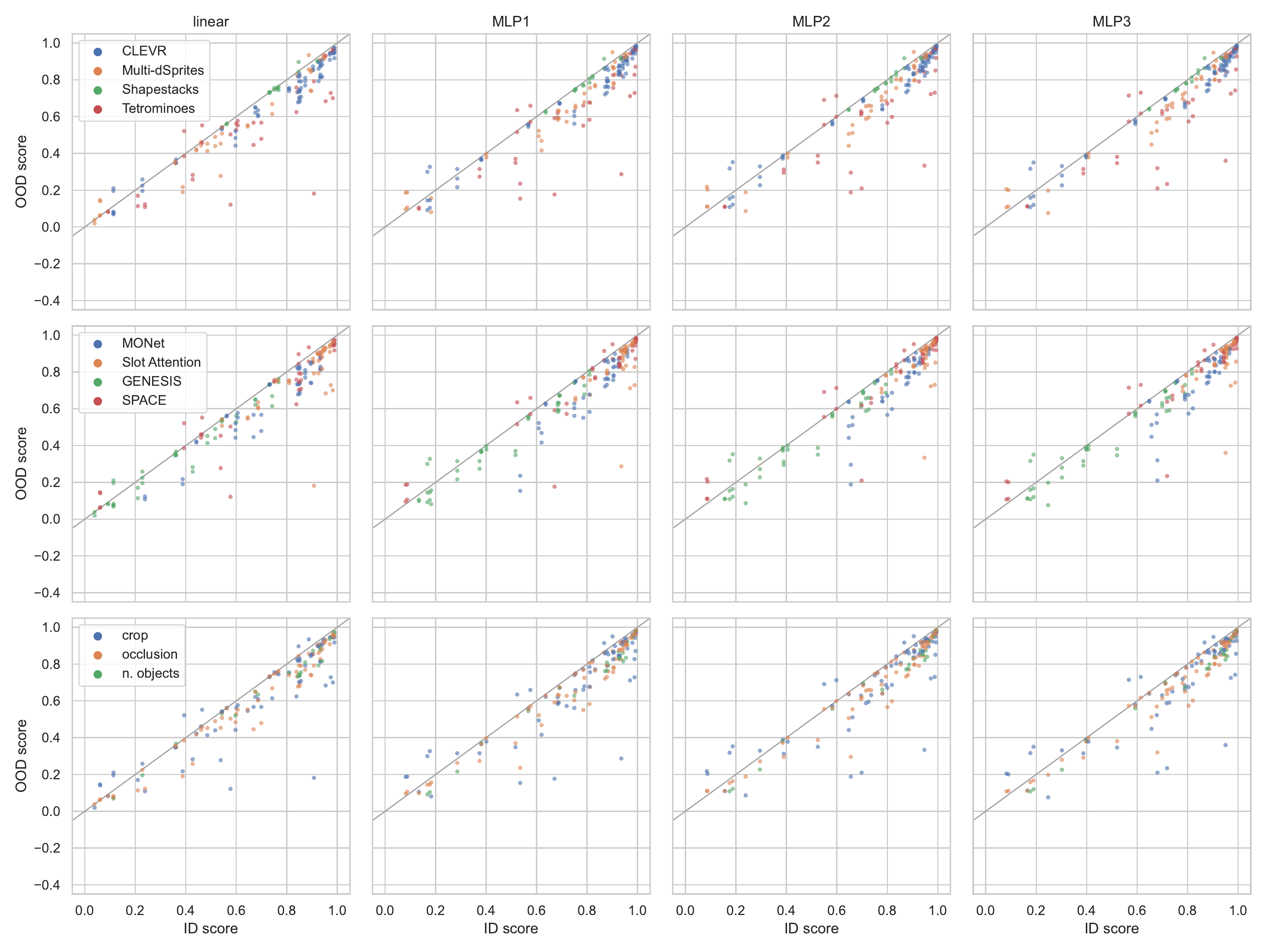}
    \caption[Generalization of \textbf{object-centric representations} in downstream prediction, using \textbf{mask matching} and \textbf{retraining} the downstream model after the distribution shift. Here the distribution shift affects \textbf{global properties} of the scene.]{Generalization of \textbf{object-centric representations} in downstream prediction, using \textbf{mask matching} and \textbf{retraining} the downstream model after the distribution shift. Here the distribution shift affects \textbf{global properties} of the scene.
        On the x-axis: prediction performance (accuracy or \rsq) for one object property on one dataset, averaged over all objects, on the original training set of the unsupervised object discovery model. On the y-axis: the same metric in OOD scenarios. Each data point corresponds to one representation model (e.g., MONet), one dataset, one object property, and one type of distribution shift. For each x (performance on one object feature in the training distribution, averaged over objects in a scene and over random seeds of the object-centric models) there are multiple y's, corresponding to different distribution shifts.
        In each row, we color-code the data according to dataset, model, or type of distribution shift.
        Each column shows analogous results for each of the 4 considered downstream models for property prediction (linear, and MLPs with up to 3 hidden layers).
    }
    \label{fig:ood/downstream/slotted/match_mask/global/retrain_True}
\end{figure}

\begin{figure}
    \centering
    \includegraphics[width=0.9\textwidth]{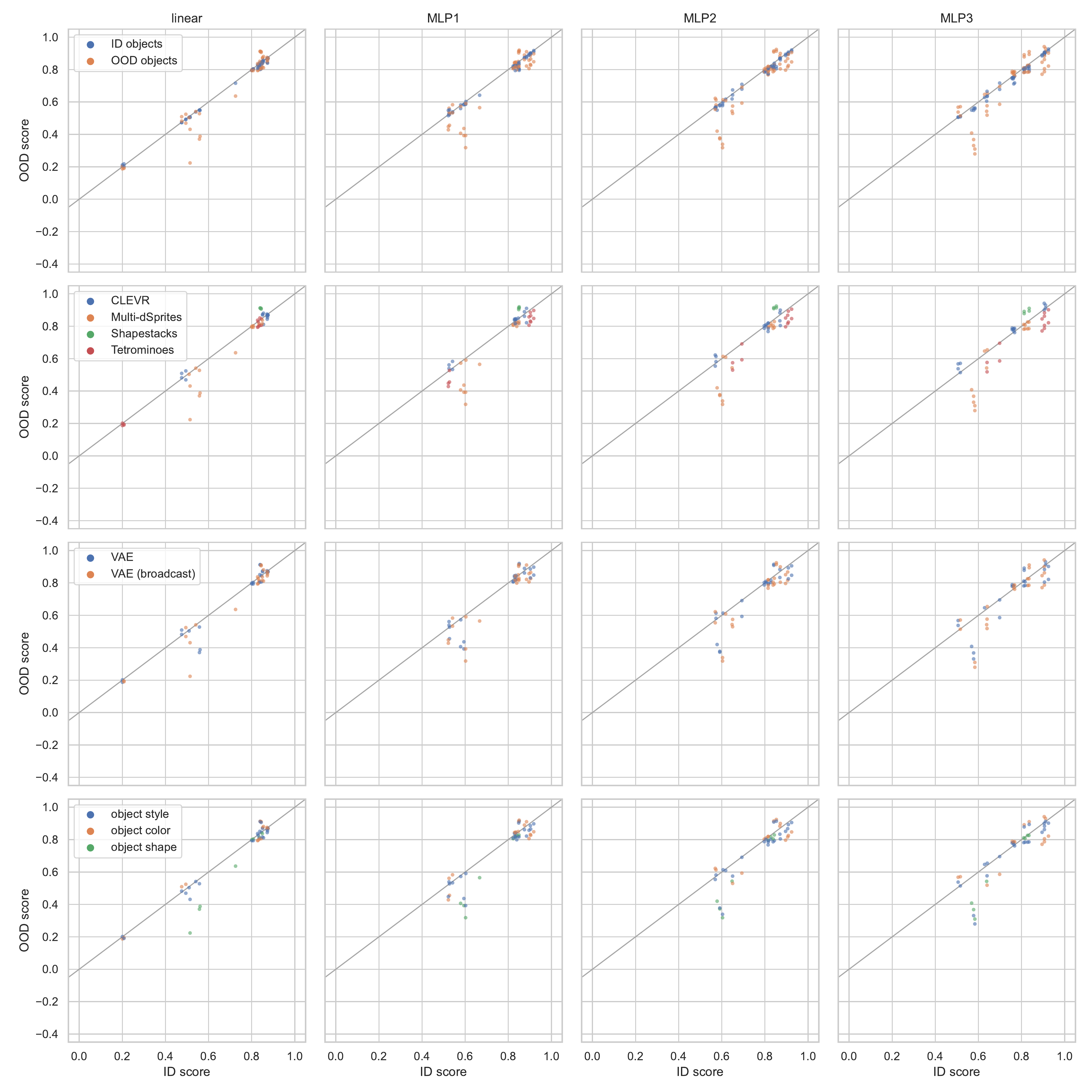}
    \caption[Generalization of \textbf{distributed representations} in downstream prediction, using \textbf{loss matching} and \textbf{without retraining} the downstream model after the distribution shift. Here the distribution shift affects \textbf{one object}.]{Generalization of \textbf{distributed representations} in downstream prediction, using \textbf{loss matching} and \textbf{without retraining} the downstream model after the distribution shift. Here the distribution shift affects \textbf{one object}.
    On the x-axis: prediction performance (accuracy or \rsq) for one object property on one dataset, averaged over all objects, on the original training set of the unsupervised object discovery model. On the y-axis: the same metric in OOD scenarios. Each data point corresponds to one representation model (e.g., MONet), one dataset, one object property, one type of distribution shift, and either ID or OOD objects. For each x (performance on one object feature in the training distribution, averaged over objects in a scene and over random seeds of the object-centric models) there are multiple y's, corresponding to different distribution shifts and to ID/OOD objects.
    In the top row, we separately report (color-coded) the performance over ID and OOD objects.
    In the following rows, we only show OOD objects and split according to dataset, model, or type of distribution shift.
    Each column shows analogous results for each of the 4 considered downstream models for property prediction (linear, and MLPs with up to 3 hidden layers).
    }
    \label{fig:ood/downstream/vae/match_loss/object/retrain_False}
\end{figure}
\begin{figure}
    \centering
    \includegraphics[width=0.9\textwidth]{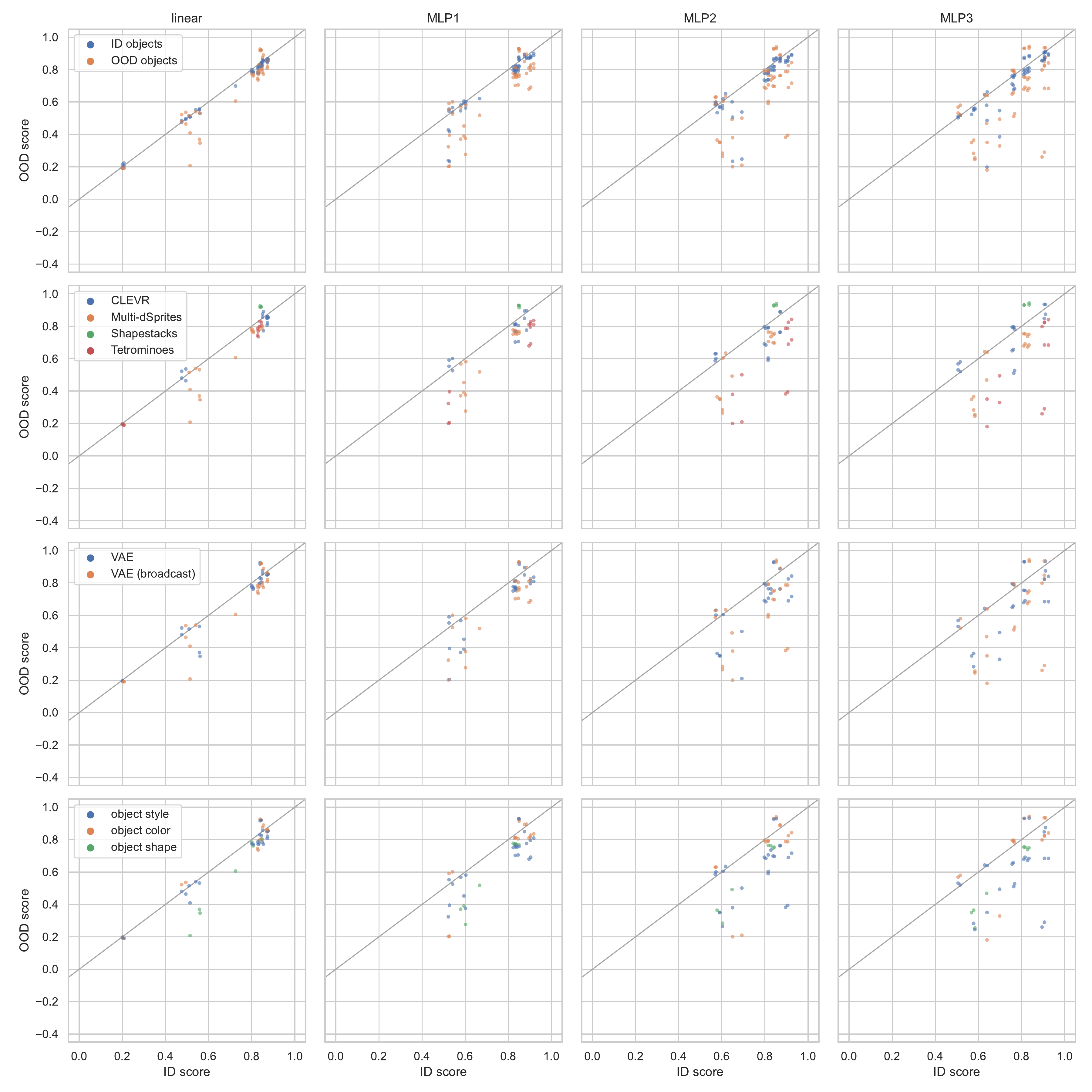}
    \caption[Generalization of \textbf{distributed representations} in downstream prediction, using \textbf{loss matching} and \textbf{retraining} the downstream model after the distribution shift. Here the distribution shift affects \textbf{one object}.]{Generalization of \textbf{distributed representations} in downstream prediction, using \textbf{loss matching} and \textbf{retraining} the downstream model after the distribution shift. Here the distribution shift affects \textbf{one object}.
    On the x-axis: prediction performance (accuracy or \rsq) for one object property on one dataset, averaged over all objects, on the original training set of the unsupervised object discovery model. On the y-axis: the same metric in OOD scenarios. Each data point corresponds to one representation model (e.g., MONet), one dataset, one object property, one type of distribution shift, and either ID or OOD objects. For each x (performance on one object feature in the training distribution, averaged over objects in a scene and over random seeds of the object-centric models) there are multiple y's, corresponding to different distribution shifts and to ID/OOD objects.
    In the top row, we separately report (color-coded) the performance over ID and OOD objects.
    In the following rows, we only show OOD objects and split according to dataset, model, or type of distribution shift.
    Each column shows analogous results for each of the 4 considered downstream models for property prediction (linear, and MLPs with up to 3 hidden layers).
    }
    \label{fig:ood/downstream/vae/match_loss/object/retrain_True}
\end{figure}
\begin{figure}
    \centering
    \includegraphics[width=0.9\textwidth]{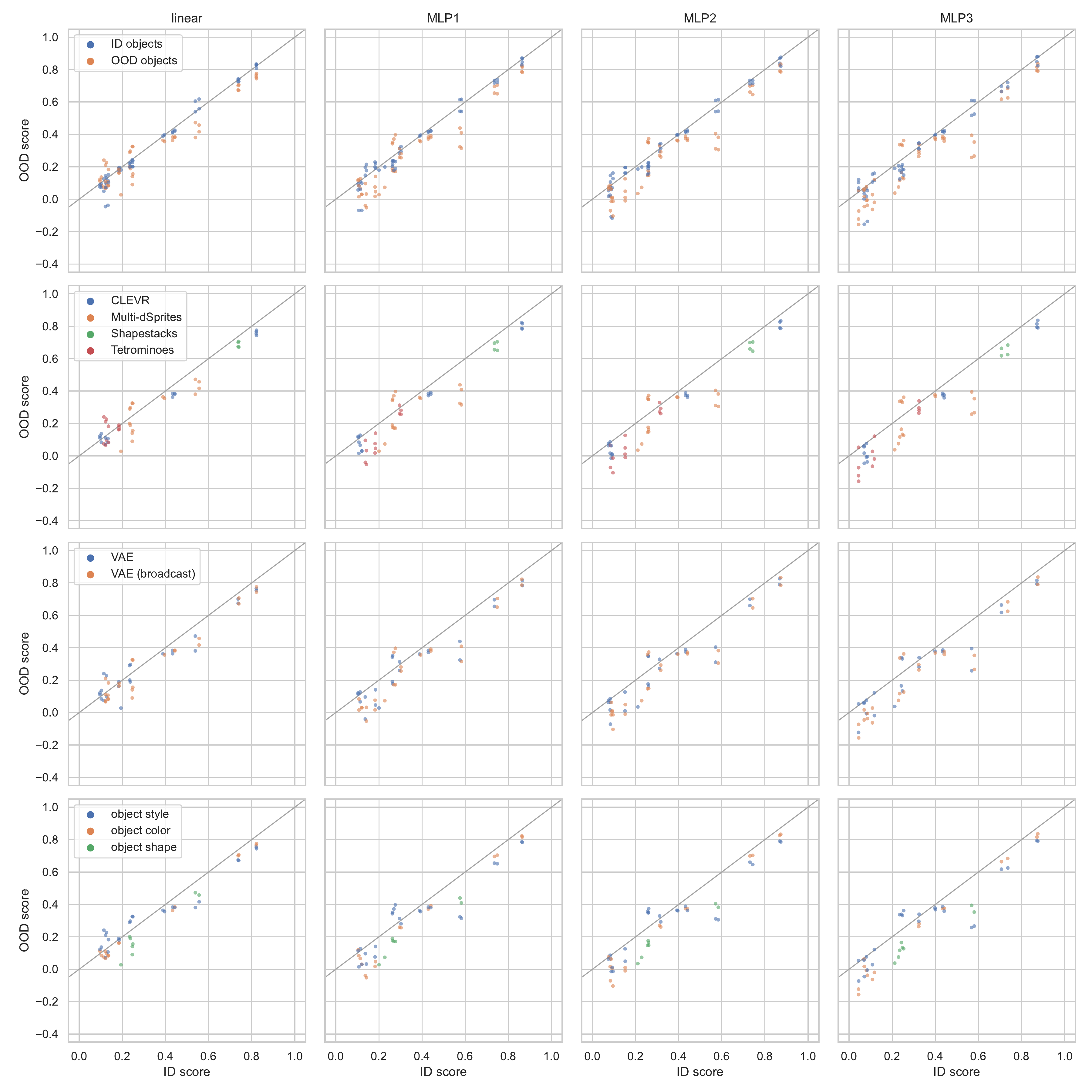}
    \caption[Generalization of \textbf{distributed representations} in downstream prediction, using \textbf{deterministic matching} and \textbf{without retraining} the downstream model after the distribution shift. Here the distribution shift affects \textbf{one object}.]{Generalization of \textbf{distributed representations} in downstream prediction, using \textbf{deterministic matching} and \textbf{without retraining} the downstream model after the distribution shift. Here the distribution shift affects \textbf{one object}.
    On the x-axis: prediction performance (accuracy or \rsq) for one object property on one dataset, averaged over all objects, on the original training set of the unsupervised object discovery model. On the y-axis: the same metric in OOD scenarios. Each data point corresponds to one representation model (e.g., MONet), one dataset, one object property, one type of distribution shift, and either ID or OOD objects. For each x (performance on one object feature in the training distribution, averaged over objects in a scene and over random seeds of the object-centric models) there are multiple y's, corresponding to different distribution shifts and to ID/OOD objects.
    In the top row, we separately report (color-coded) the performance over ID and OOD objects.
    In the following rows, we only show OOD objects and split according to dataset, model, or type of distribution shift.
    Each column shows analogous results for each of the 4 considered downstream models for property prediction (linear, and MLPs with up to 3 hidden layers).
    }
    \label{fig:ood/downstream/vae/match_deterministic/object/retrain_False}
\end{figure}
\begin{figure}
    \centering
    \includegraphics[width=0.9\textwidth]{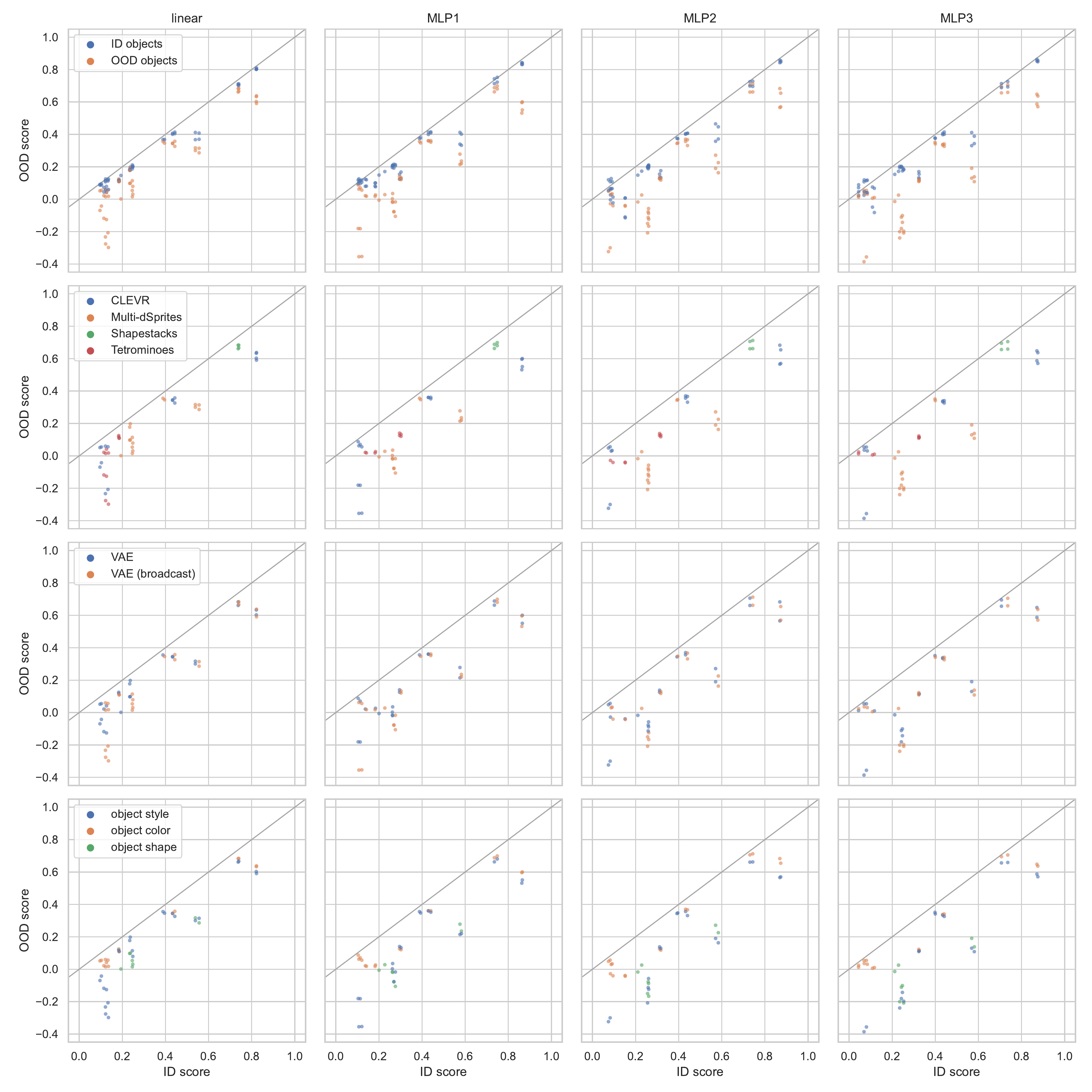}
    \caption[Generalization of \textbf{distributed representations} in downstream prediction, using \textbf{deterministic matching} and \textbf{retraining} the downstream model after the distribution shift. Here the distribution shift affects \textbf{one object}.]{Generalization of \textbf{distributed representations} in downstream prediction, using \textbf{deterministic matching} and \textbf{retraining} the downstream model after the distribution shift. Here the distribution shift affects \textbf{one object}.
    On the x-axis: prediction performance (accuracy or \rsq) for one object property on one dataset, averaged over all objects, on the original training set of the unsupervised object discovery model. On the y-axis: the same metric in OOD scenarios. Each data point corresponds to one representation model (e.g., MONet), one dataset, one object property, one type of distribution shift, and either ID or OOD objects. For each x (performance on one object feature in the training distribution, averaged over objects in a scene and over random seeds of the object-centric models) there are multiple y's, corresponding to different distribution shifts and to ID/OOD objects.
    In the top row, we separately report (color-coded) the performance over ID and OOD objects.
    In the following rows, we only show OOD objects and split according to dataset, model, or type of distribution shift.
    Each column shows analogous results for each of the 4 considered downstream models for property prediction (linear, and MLPs with up to 3 hidden layers).
    }
    \label{fig:ood/downstream/vae/match_deterministic/object/retrain_True}
\end{figure}
\begin{figure}
    \centering
    \includegraphics[width=0.9\textwidth]{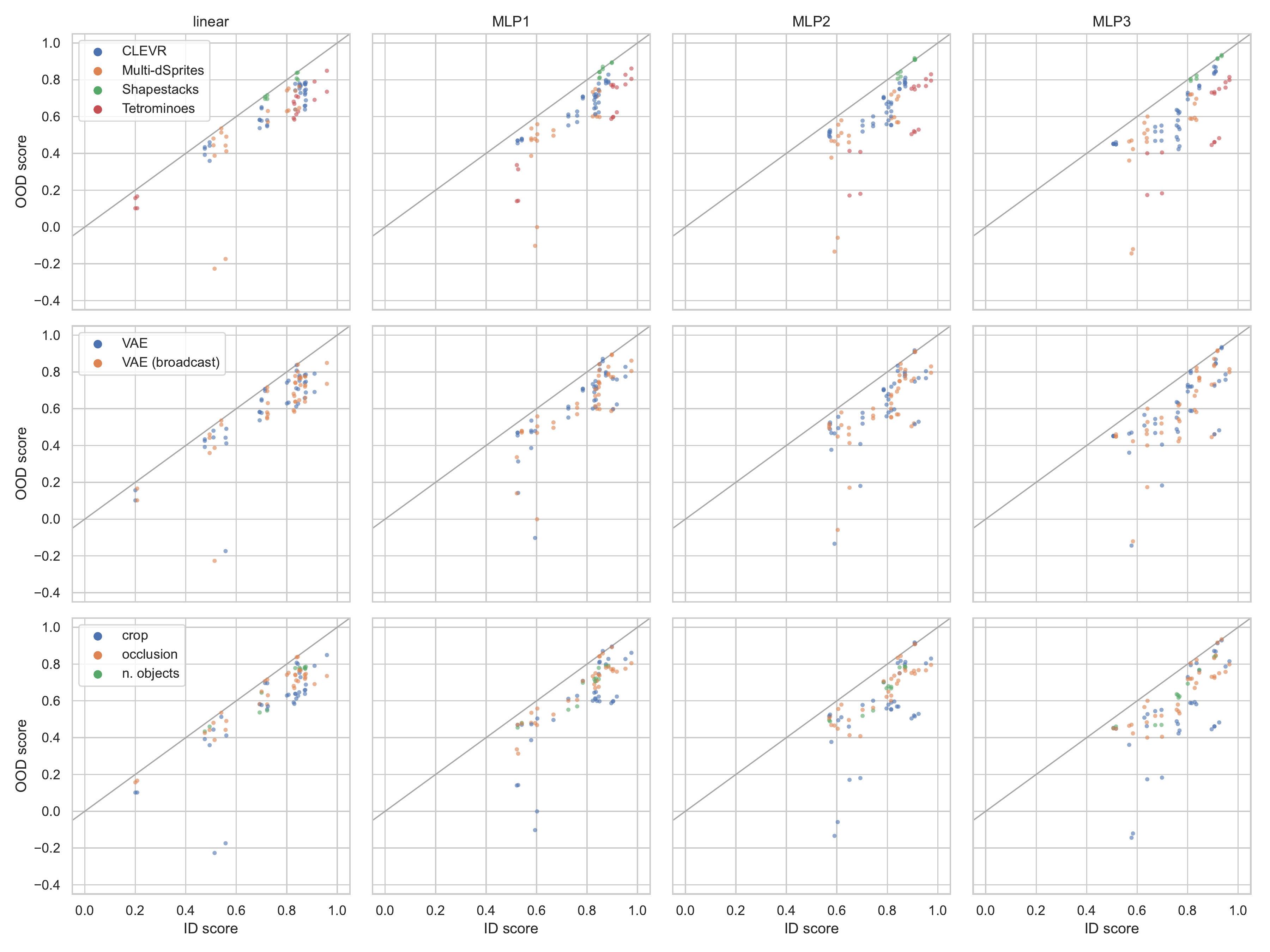}
    \caption[Generalization of \textbf{distributed representations} in downstream prediction, using \textbf{loss matching} and \textbf{without retraining} the downstream model after the distribution shift. Here the distribution shift affects \textbf{global properties} of the scene.]{Generalization of \textbf{distributed representations} in downstream prediction, using \textbf{loss matching} and \textbf{without retraining} the downstream model after the distribution shift. Here the distribution shift affects \textbf{global properties} of the scene.
        On the x-axis: prediction performance (accuracy or \rsq) for one object property on one dataset, averaged over all objects, on the original training set of the unsupervised object discovery model. On the y-axis: the same metric in OOD scenarios. Each data point corresponds to one representation model (e.g., MONet), one dataset, one object property, and one type of distribution shift. For each x (performance on one object feature in the training distribution, averaged over objects in a scene and over random seeds of the object-centric models) there are multiple y's, corresponding to different distribution shifts.
        In each row, we color-code the data according to dataset, model, or type of distribution shift.
        Each column shows analogous results for each of the 4 considered downstream models for property prediction (linear, and MLPs with up to 3 hidden layers).
    }
    \label{fig:ood/downstream/vae/match_loss/global/retrain_False/scatter}
\end{figure}
\begin{figure}
    \centering
    \includegraphics[width=0.9\textwidth]{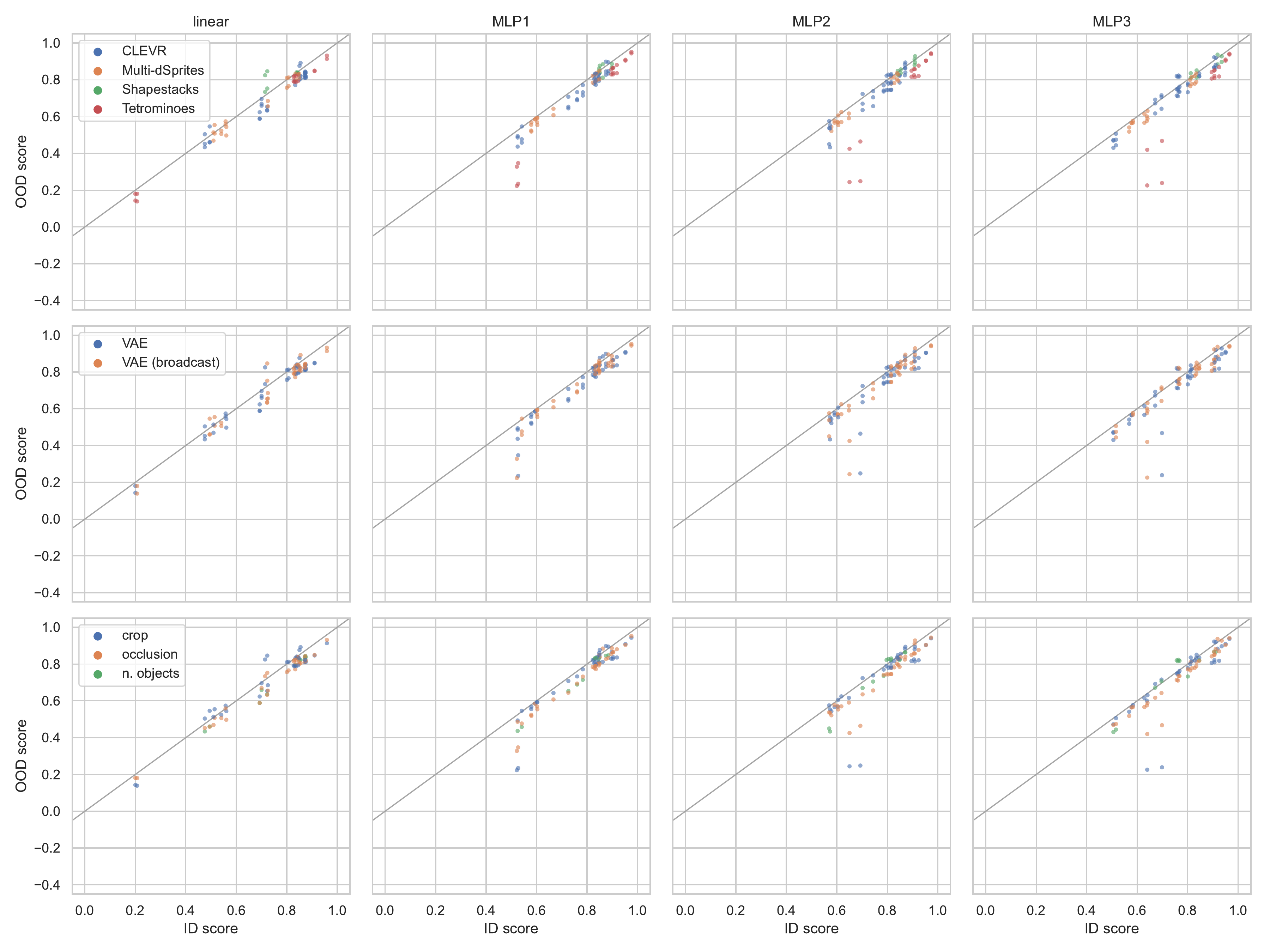}
    \caption[Generalization of \textbf{distributed representations} in downstream prediction, using \textbf{loss matching} and \textbf{retraining} the downstream model after the distribution shift. Here the distribution shift affects \textbf{global properties} of the scene.]{Generalization of \textbf{distributed representations} in downstream prediction, using \textbf{loss matching} and \textbf{retraining} the downstream model after the distribution shift. Here the distribution shift affects \textbf{global properties} of the scene.
        On the x-axis: prediction performance (accuracy or \rsq) for one object property on one dataset, averaged over all objects, on the original training set of the unsupervised object discovery model. On the y-axis: the same metric in OOD scenarios. Each data point corresponds to one representation model (e.g., MONet), one dataset, one object property, and one type of distribution shift. For each x (performance on one object feature in the training distribution, averaged over objects in a scene and over random seeds of the object-centric models) there are multiple y's, corresponding to different distribution shifts.
        In each row, we color-code the data according to dataset, model, or type of distribution shift.
        Each column shows analogous results for each of the 4 considered downstream models for property prediction (linear, and MLPs with up to 3 hidden layers).
    }
    \label{fig:ood/downstream/vae/match_loss/global/retrain_True}
\end{figure}
\begin{figure}
    \centering
    \includegraphics[width=0.9\textwidth]{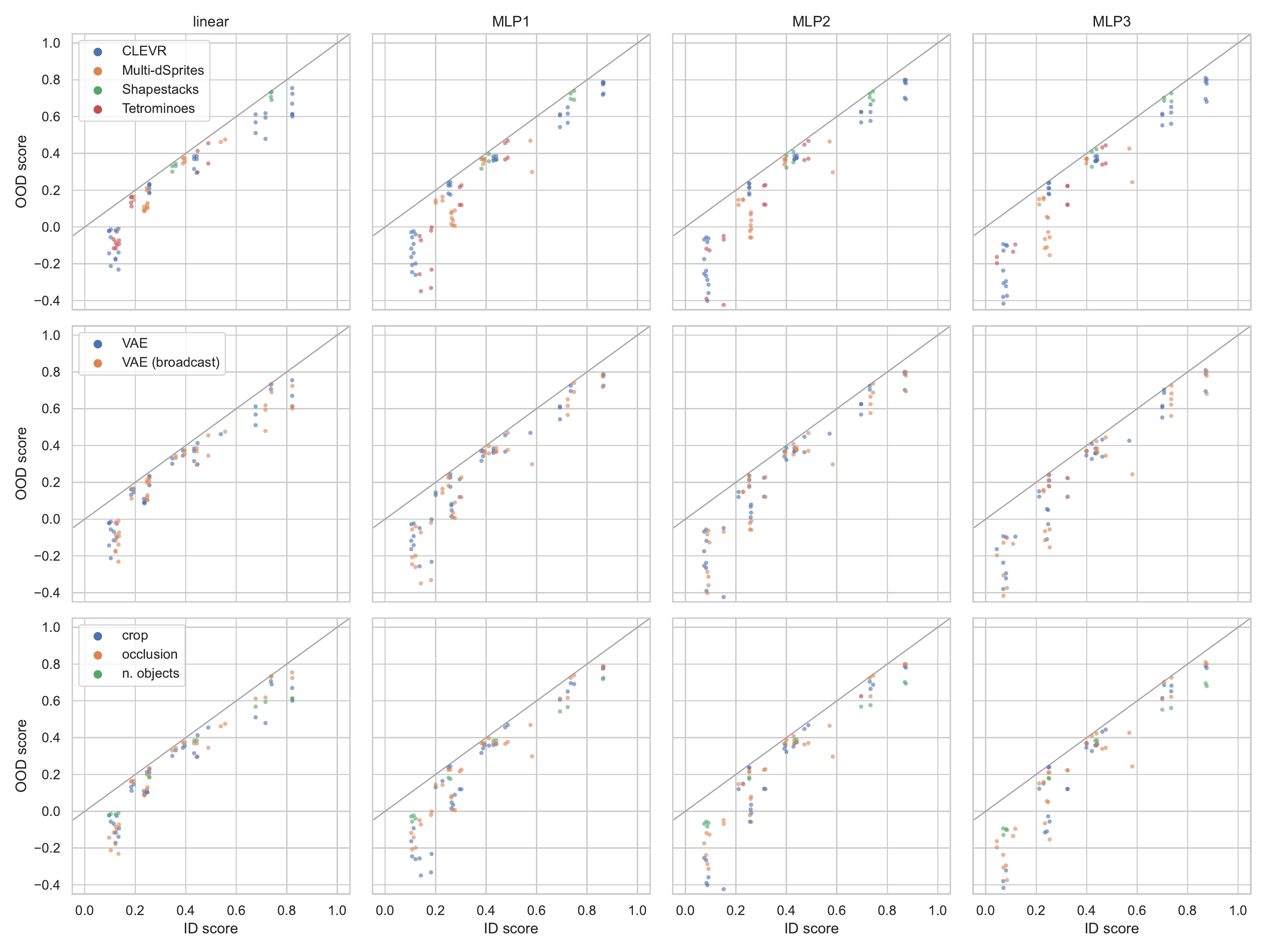}
    \caption[Generalization of \textbf{distributed representations} in downstream prediction, using \textbf{deterministic matching} and \textbf{without retraining} the downstream model after the distribution shift. Here the distribution shift affects \textbf{global properties} of the scene.]{Generalization of \textbf{distributed representations} in downstream prediction, using \textbf{deterministic matching} and \textbf{without retraining} the downstream model after the distribution shift. Here the distribution shift affects \textbf{global properties} of the scene.
        On the x-axis: prediction performance (accuracy or \rsq) for one object property on one dataset, averaged over all objects, on the original training set of the unsupervised object discovery model. On the y-axis: the same metric in OOD scenarios. Each data point corresponds to one representation model (e.g., MONet), one dataset, one object property, and one type of distribution shift. For each x (performance on one object feature in the training distribution, averaged over objects in a scene and over random seeds of the object-centric models) there are multiple y's, corresponding to different distribution shifts.
        In each row, we color-code the data according to dataset, model, or type of distribution shift.
        Each column shows analogous results for each of the 4 considered downstream models for property prediction (linear, and MLPs with up to 3 hidden layers).
    }
    \label{fig:ood/downstream/vae/match_deterministic/global/retrain_False}
\end{figure}
\begin{figure}
    \centering
    \includegraphics[width=0.9\textwidth]{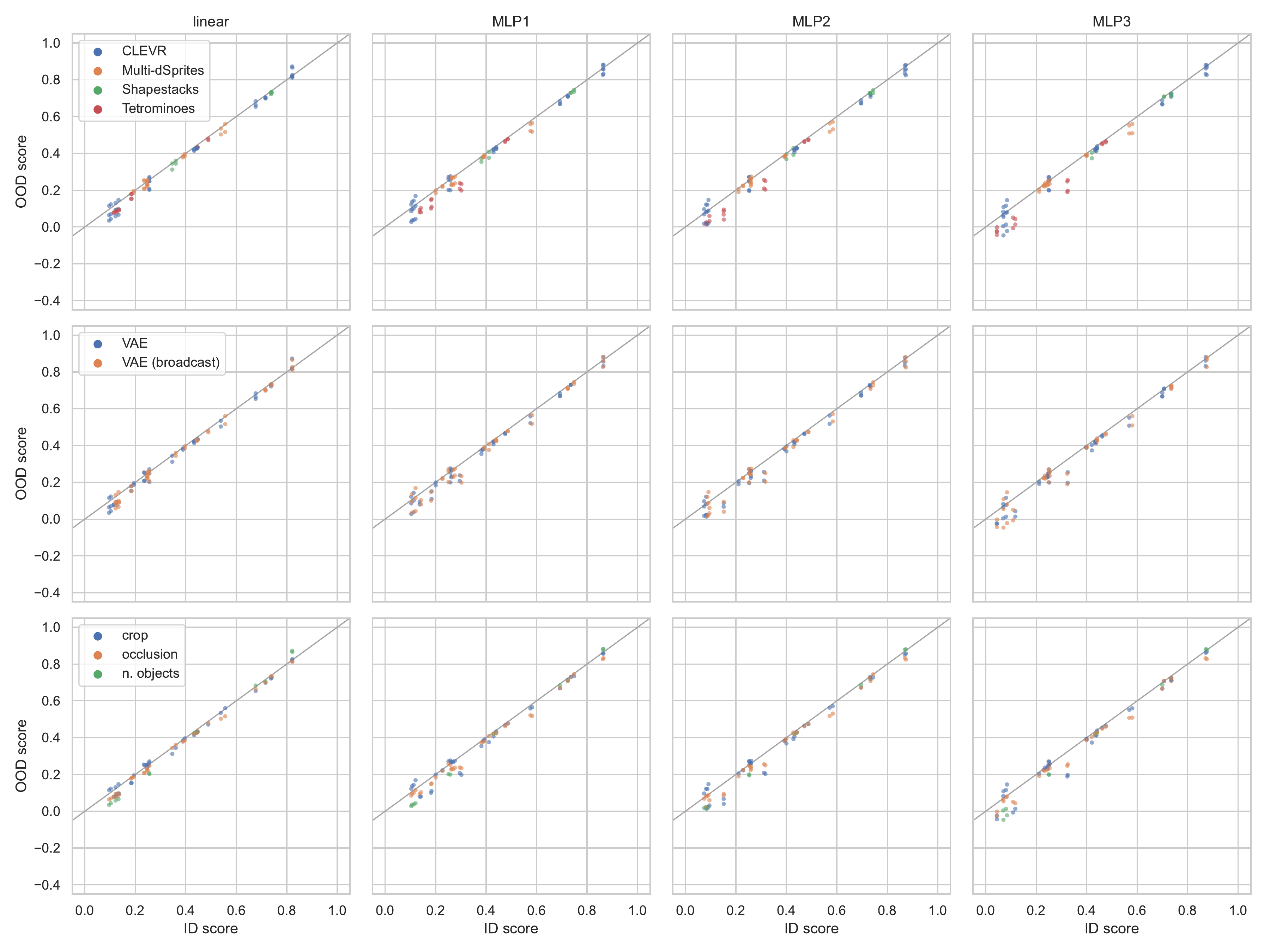}
    \caption[Generalization of \textbf{distributed representations} in downstream prediction, using \textbf{deterministic matching} and \textbf{retraining} the downstream model after the distribution shift. Here the distribution shift affects \textbf{global properties} of the scene.]{Generalization of \textbf{distributed representations} in downstream prediction, using \textbf{deterministic matching} and \textbf{retraining} the downstream model after the distribution shift. Here the distribution shift affects \textbf{global properties} of the scene.
        On the x-axis: prediction performance (accuracy or \rsq) for one object property on one dataset, averaged over all objects, on the original training set of the unsupervised object discovery model. On the y-axis: the same metric in OOD scenarios. Each data point corresponds to one representation model (e.g., MONet), one dataset, one object property, and one type of distribution shift. For each x (performance on one object feature in the training distribution, averaged over objects in a scene and over random seeds of the object-centric models) there are multiple y's, corresponding to different distribution shifts.
        In each row, we color-code the data according to dataset, model, or type of distribution shift.
        Each column shows analogous results for each of the 4 considered downstream models for property prediction (linear, and MLPs with up to 3 hidden layers).
    }
    \label{fig:ood/downstream/vae/match_deterministic/global/retrain_True}
\end{figure}

%%%%%%%%%%%%%%%%%%%%%%%%%%%%%%%%%%%%%%%%%%%%%%%%%%%%%%

\clearpage
\subsection{Qualitative results}\label{app:results_qualitative}

In \cref{fig:app_model_viz_clevr,fig:app_model_viz_multidsprites,fig:app_model_viz_tetrominoes,fig:app_model_viz_objectsroom,fig:app_model_viz_shapestacks} we show the reconstruction and segmentation performance of a selection of object-centric models on a random subset of held-out test images, for all 5 datasets. We select one object-centric model per type (MONet, Slot Attention, GENESIS, and SPACE) based on the ARI score on the validation set. The images we show were not used for model selection. For each model we show the following:
\begin{itemize}
    \item \emph{Input and reconstructed images}.
    \item \emph{Ground-truth and inferred segmentation maps}. Here we use a set of 8 colors and assign each object (or slot) to a color. If there are more than 8 slots, we loop over the 8 colors again (this does not happen here, except in SPACE, where it is not an issue in practice).
    Rather than taking hard masks, we treat the masks as ``soft'', such that a pixel's color is a weighted mean of the 8 colors according to the masks. This is evident in Slot Attention, which typically splits the background smoothly across slots (consistently with the qualitative results shown in \citet{locatello2020object}).
    For clarity, we match (with the Hungarian algorithm) the colors of the ground-truth and predicted masks using the cosine distance ($1$ minus the cosine similarity) between masks.
    \item \emph{Slot-wise reconstructions}. Each column corresponds to a slot in the object-centric representation of the model. Here we show the entire slot reconstruction with the inferred slot mask as alpha (transparency) channel. The overall reconstruction is the sum of these images.
    Since SPACE has in total up to 69 slots in our experiments ($K=5$ background slots, and a grid of foreground slots of size $G \times G$ with $G=8$), it is impractical to show all slots here. We choose instead to show the 10 most salient slots, selected according to the average mask value over the image. This number is sufficient as most slots are unused. When selecting slots this way, the selected slots are shown in their original order (in SPACE, the background slots are appended to the foreground slots).
\end{itemize}
For completeness, in \cref{fig:app_model_viz_vaes} we show inputs and reconstructions for one VAE baseline per type (convolutional and broadcast decoder), selected using the reconstruction MSE on the validation set.

Finally, \cref{fig:ood_visualizations_clevr,fig:ood_visualizations_multidsprites,fig:ood_visualizations_objects_room,fig:ood_visualizations_shapestacks,fig:ood_visualizations_tetrominoes} show input--reconstruction pairs for each dataset, model type, and distribution shift.
Note that the comparison is not necessarily fair, since object-centric models were chosen using the validation ARI on the training distribution, while VAEs were chosen in a similar way but using the MSE.
However, these qualitative results can still be highly informative. 
We report some examples:
\begin{itemize}
    \item Most object-centric models are relatively robust to shifts affecting a single object, as discussed in the main text based on quantitative results.
    
    \item On the other hand, they are often not robust to global shifts, especially when cropping and enlarging the scene.
    
    \item MONet achieves relatively good reconstructions even out of distribution, probably because images are segmented mostly based on color. This was suggested by \citet{papa2022inductive}, where the models are trained on objects with style transfer. However, we conjecture the behavior may be the same in our case, and that the argument should also apply to other distribution shifts, as seen by the relatively accurate reconstructions under both single-object and global distribution shifts. Note that, while reconstructions are potentially more accurate than for other models, this does not mean that MONet has segmented the object correctly.
    
    \item Although its ARI score does not decrease significantly, Slot Attention may not always handle more objects than in the training distribution, even when the number of slots in the model is increased. This is consistent with the results reported by \citet{locatello2020object}, and increasing the number of Slot Attention iterations at test time seems to be a promising approach \cite[Fig.~2]{locatello2020object}. 
    
    \item VAEs seem to be relatively good at generalizing to a greater number of objects in CLEVR. In particular, they reconstruct images with the correct number of objects, although a few details of the objects may not be inferred correctly (e.g. an object may be reconstructed with the wrong size, color, or shape). This is surprising, since VAEs do not have any inductive bias for this, and the fact that the encoder is OOD (i.e., the encoder input is OOD w.r.t. the distribution used to train the encoder itself) might lead us to expect poor generalization capabilities, as discussed by \citet{dittadi2021transfer} and \citet{trauble2022role} in the ``OOD2'' case.
    On the other hand, some object-centric models are remarkably robust to this shift (in particular SPACE, as confirmed by the ARI in \cref{fig:figures/ood/metrics/only_ARI_ood_global/barplots_ARI}).
    
\end{itemize}

\begin{figure}[hb]
    \centering
    \includegraphics[height=3.1cm]{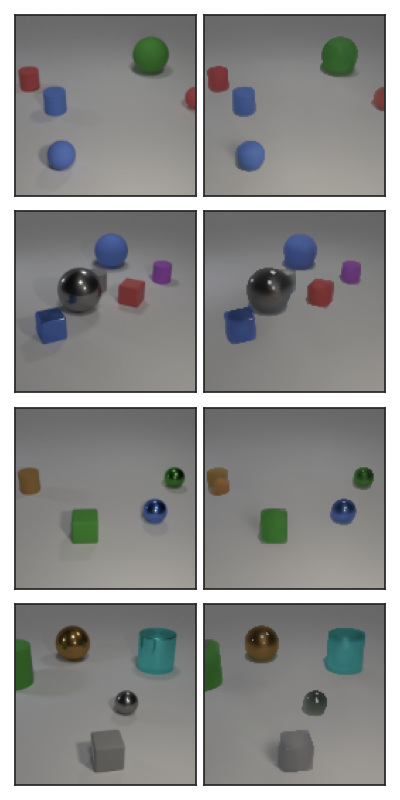}
    \hspace{0.2cm}
    \includegraphics[height=3.1cm]{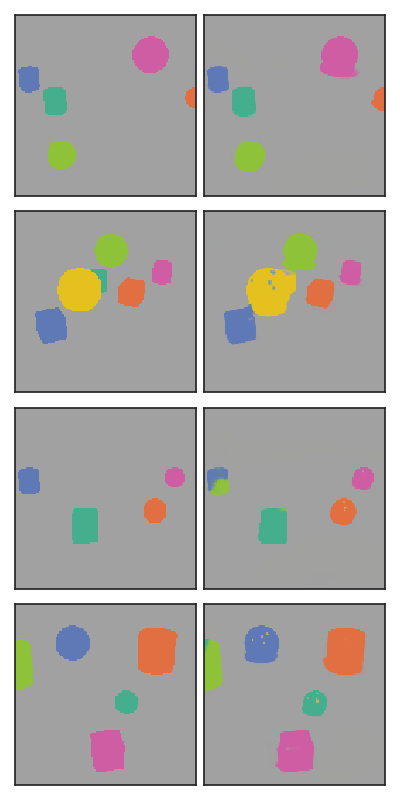}
    \hspace{0.2cm}
    \includegraphics[height=3.1cm]{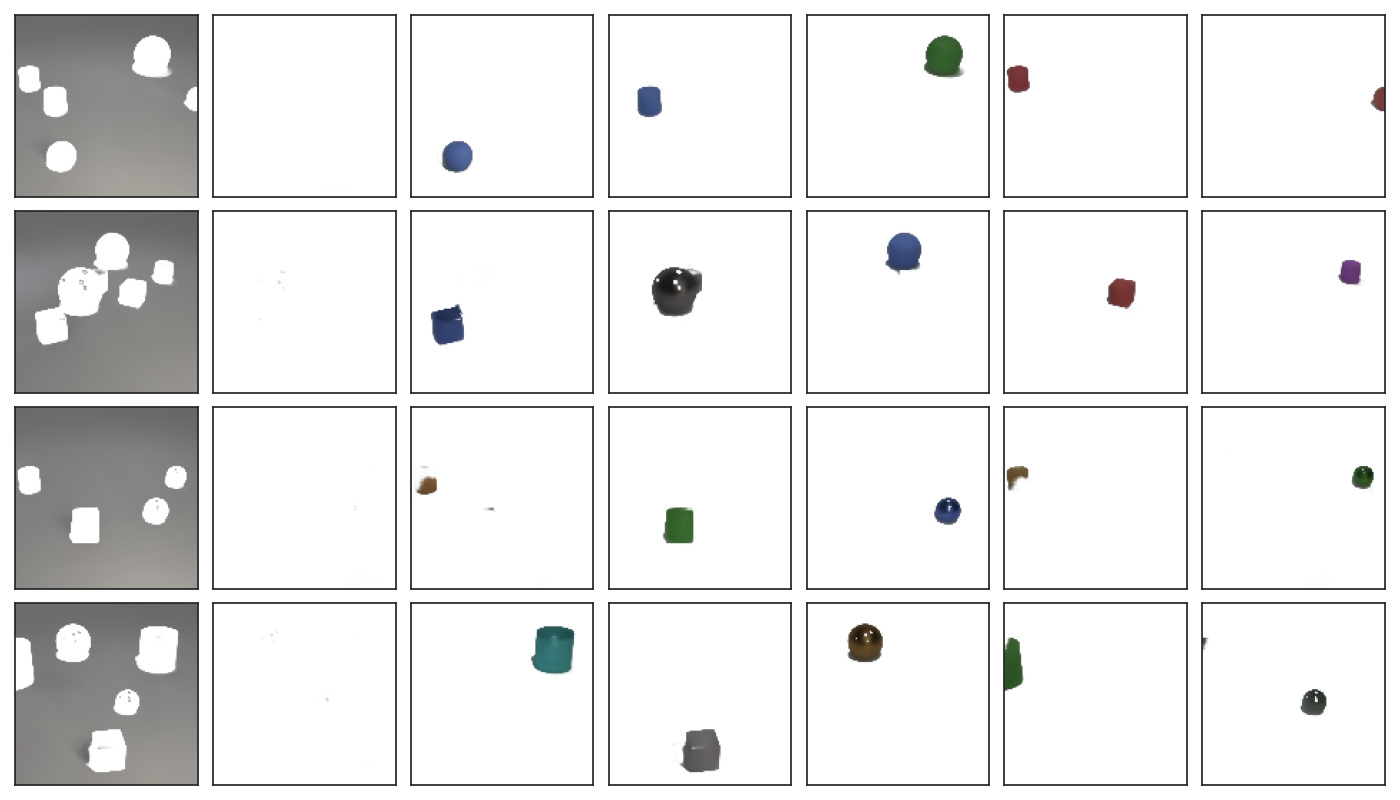}
    
    \vspace{0.45cm}
    
    \includegraphics[height=3.1cm]{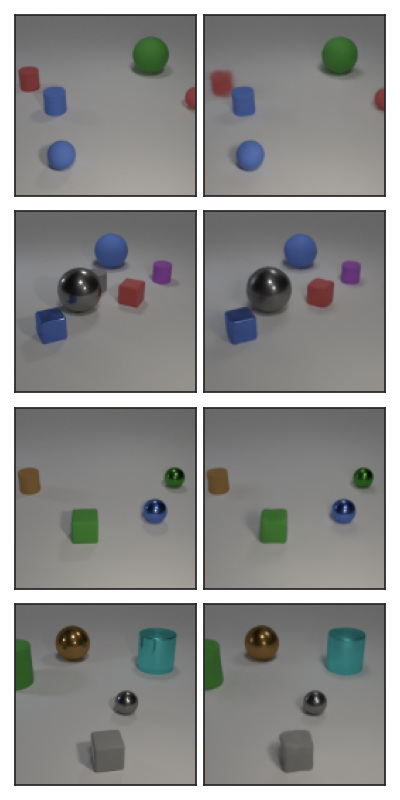}
    \hspace{0.2cm}
    \includegraphics[height=3.1cm]{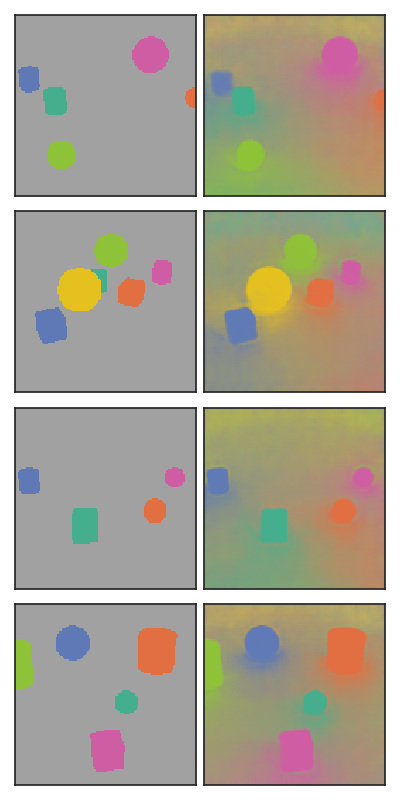}
    \hspace{0.2cm}
    \includegraphics[height=3.1cm]{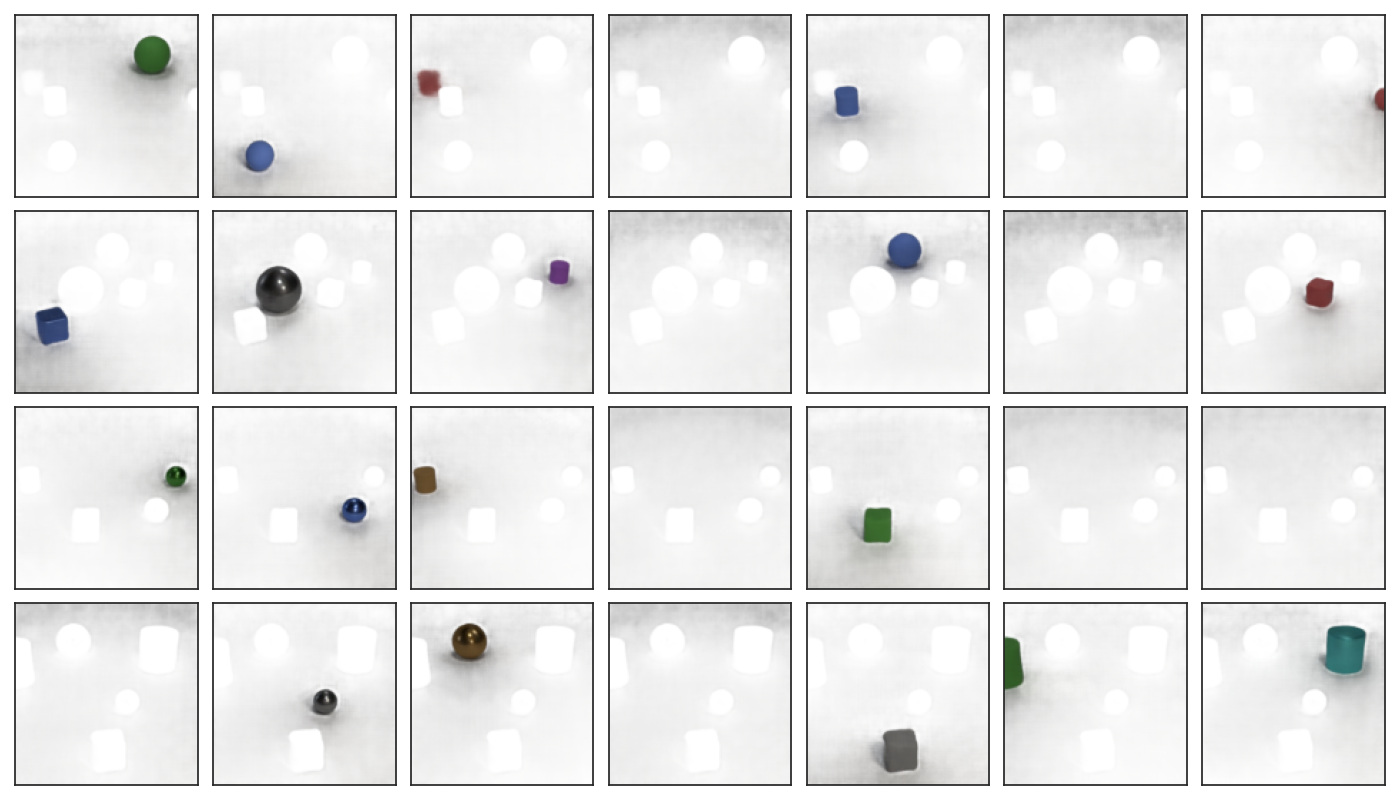}
    
    \vspace{0.45cm}
    
    \includegraphics[height=3.1cm]{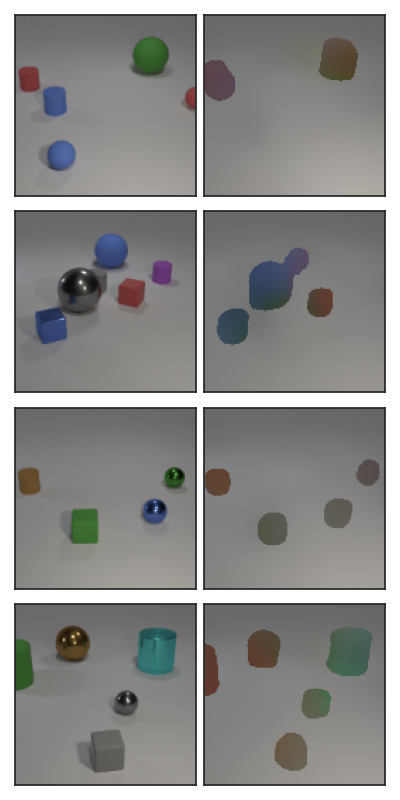}
    \hspace{0.2cm}
    \includegraphics[height=3.1cm]{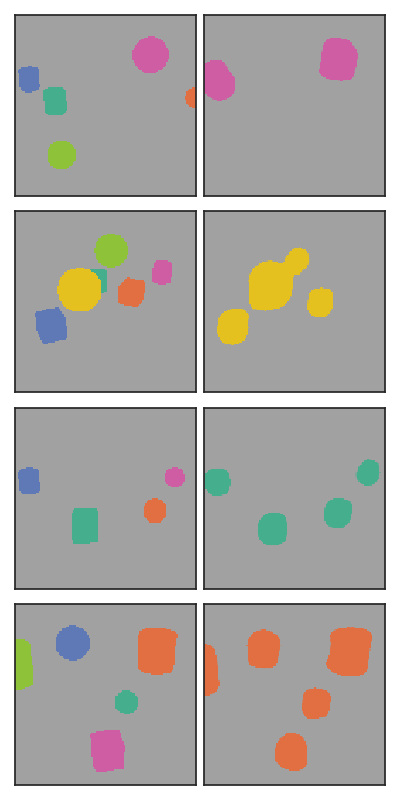}
    \hspace{0.2cm}
    \includegraphics[height=3.1cm]{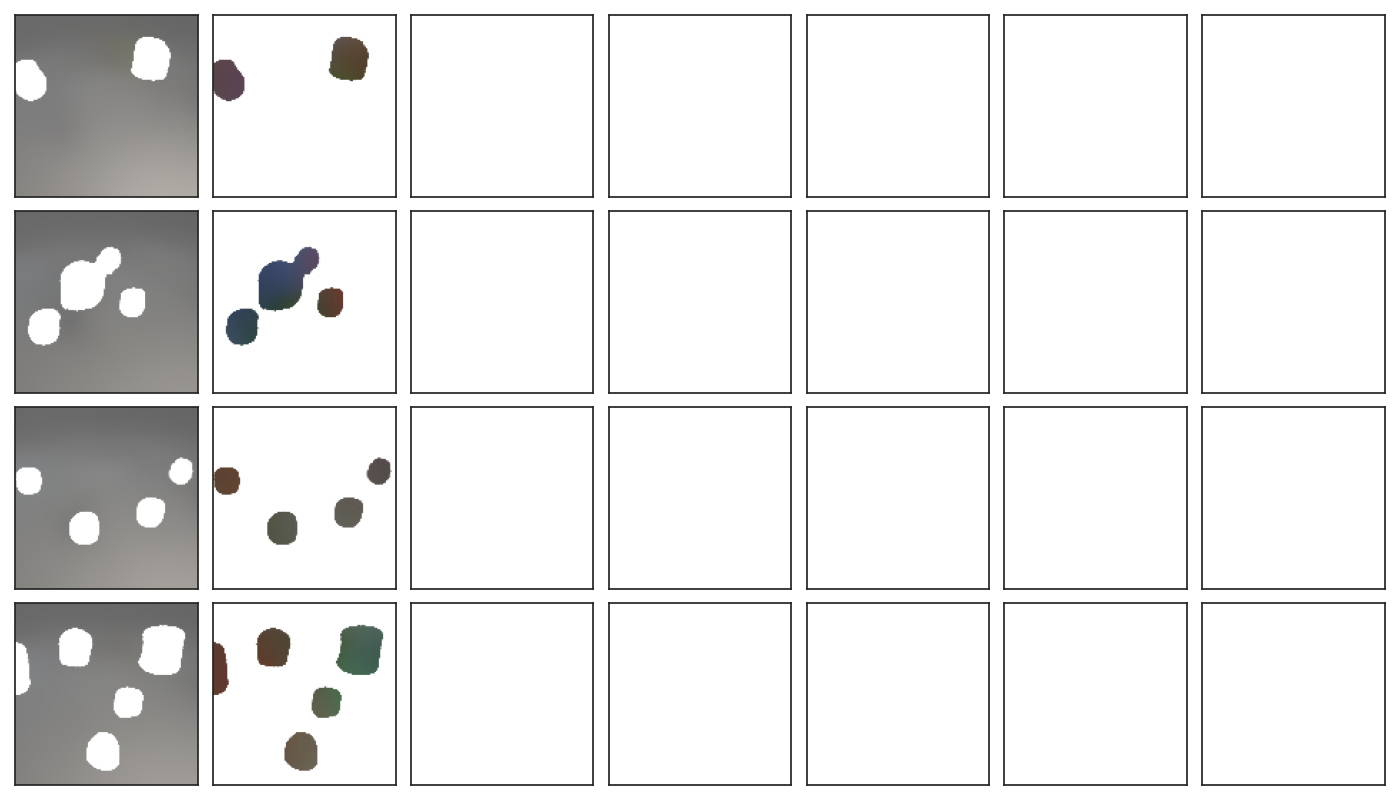}
    
    \vspace{0.45cm}
    
    \includegraphics[height=3.1cm]{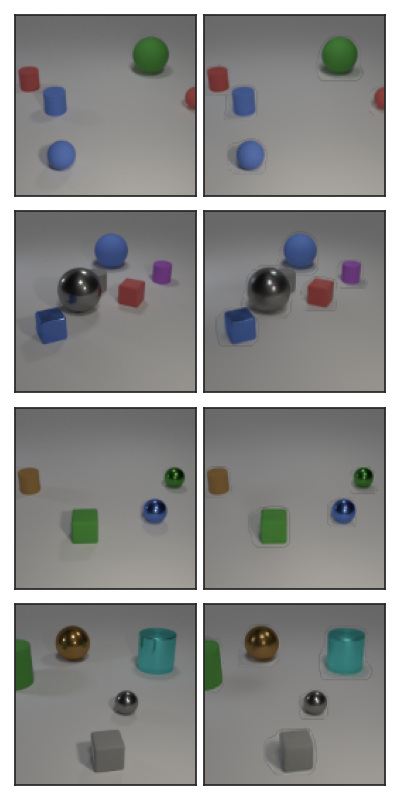}
    \hspace{0.2cm}
    \includegraphics[height=3.1cm]{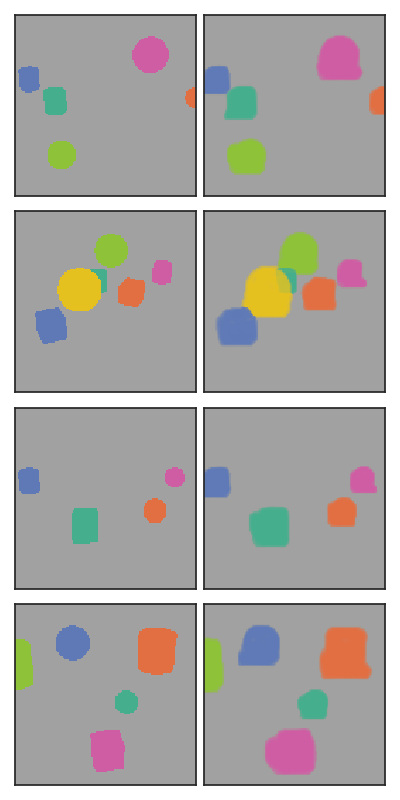}
    \hspace{0.2cm}
    \includegraphics[height=3.1cm]{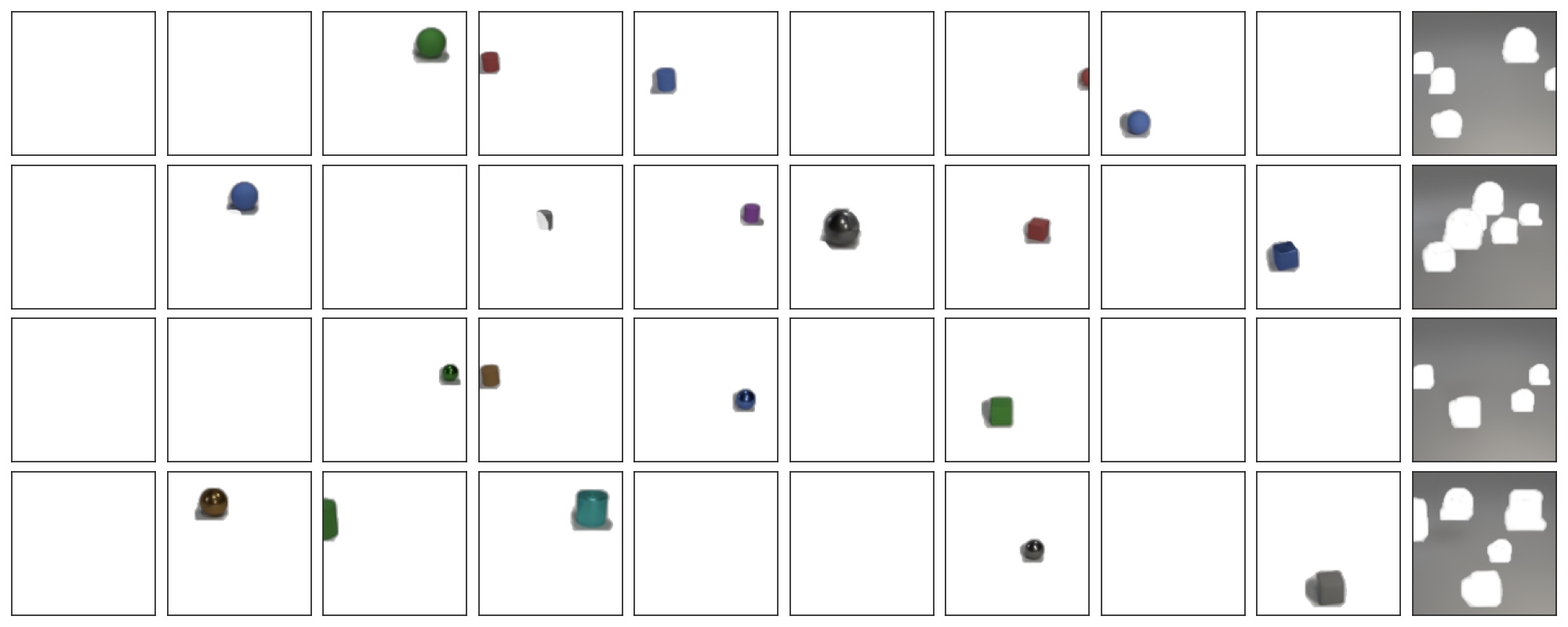}
    
    \vspace{0.45cm}
    \caption[\textbf{Reconstruction and segmentation} of 4 random images from the held-out test set of \textbf{CLEVR6}]{\textbf{Reconstruction and segmentation} of 4 random images from the held-out test set of \textbf{CLEVR6}. Top to bottom: MONet, Slot Attention, GENESIS, SPACE. Left to right: input, reconstruction, ground-truth masks, predicted (soft) masks, slot-wise reconstructions (masked with the predicted masks). As explained in the text, for SPACE we select the 10 most salient slots using the predicted masks. As explained in the text, for SPACE we select the 10 most salient slots using the predicted masks. For each model type, we visualize the specific model with the highest ARI score in the \textit{validation} set. The images shown here are from the \textit{test} set and were not used for model selection.}
    \label{fig:app_model_viz_clevr}
\end{figure}

\begin{figure}
    \centering
    \includegraphics[height=3.1cm]{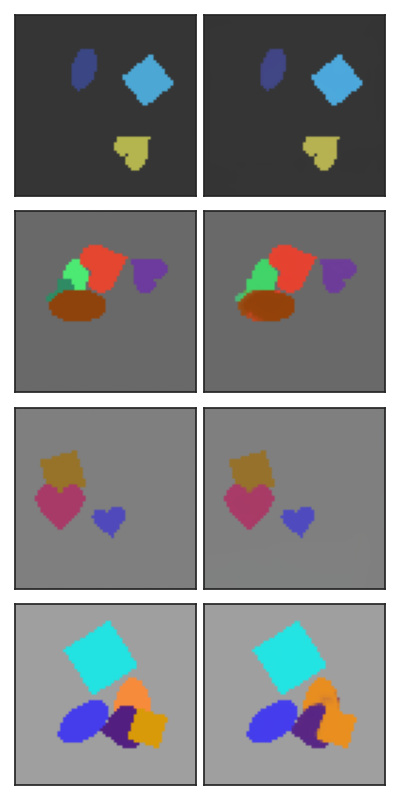}
    \hspace{0.2cm}
    \includegraphics[height=3.1cm]{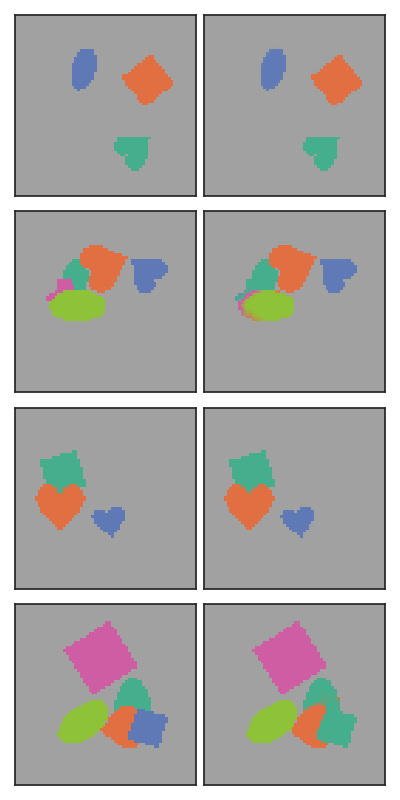}
    \hspace{0.2cm}
    \includegraphics[height=3.1cm]{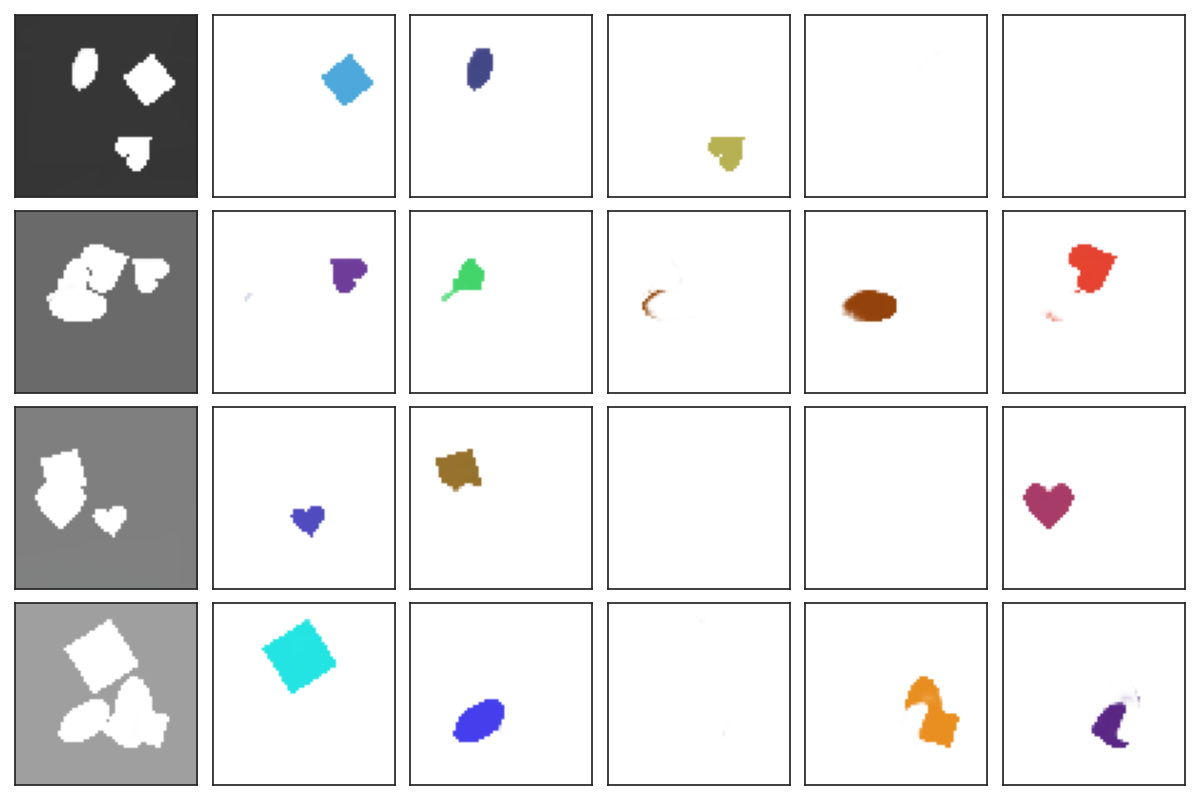}
    
    \vspace{0.45cm}
    
    \includegraphics[height=3.1cm]{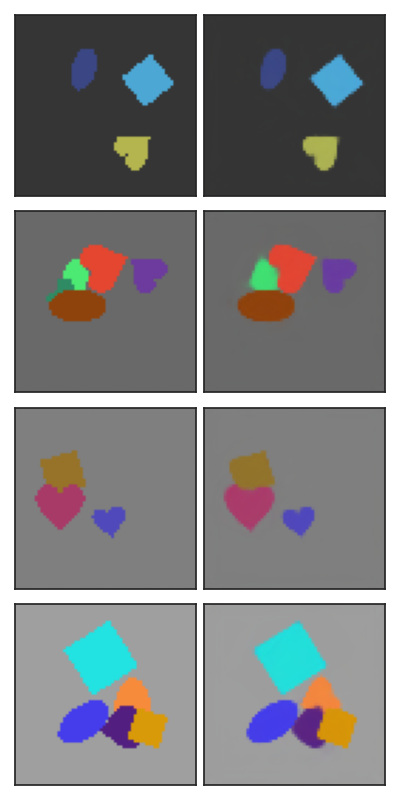}
    \hspace{0.2cm}
    \includegraphics[height=3.1cm]{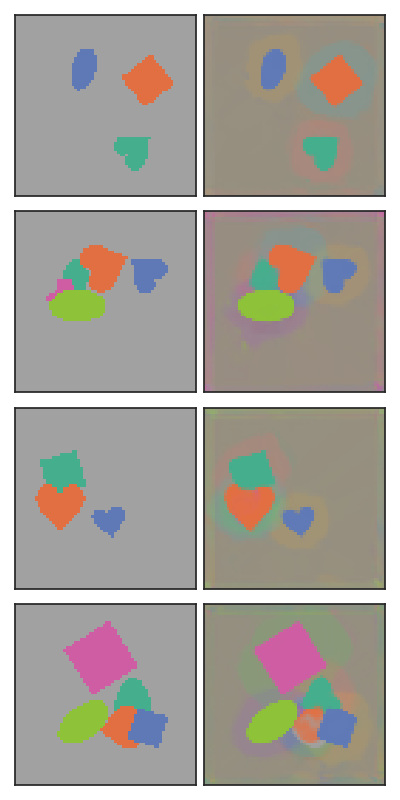}
    \hspace{0.2cm}
    \includegraphics[height=3.1cm]{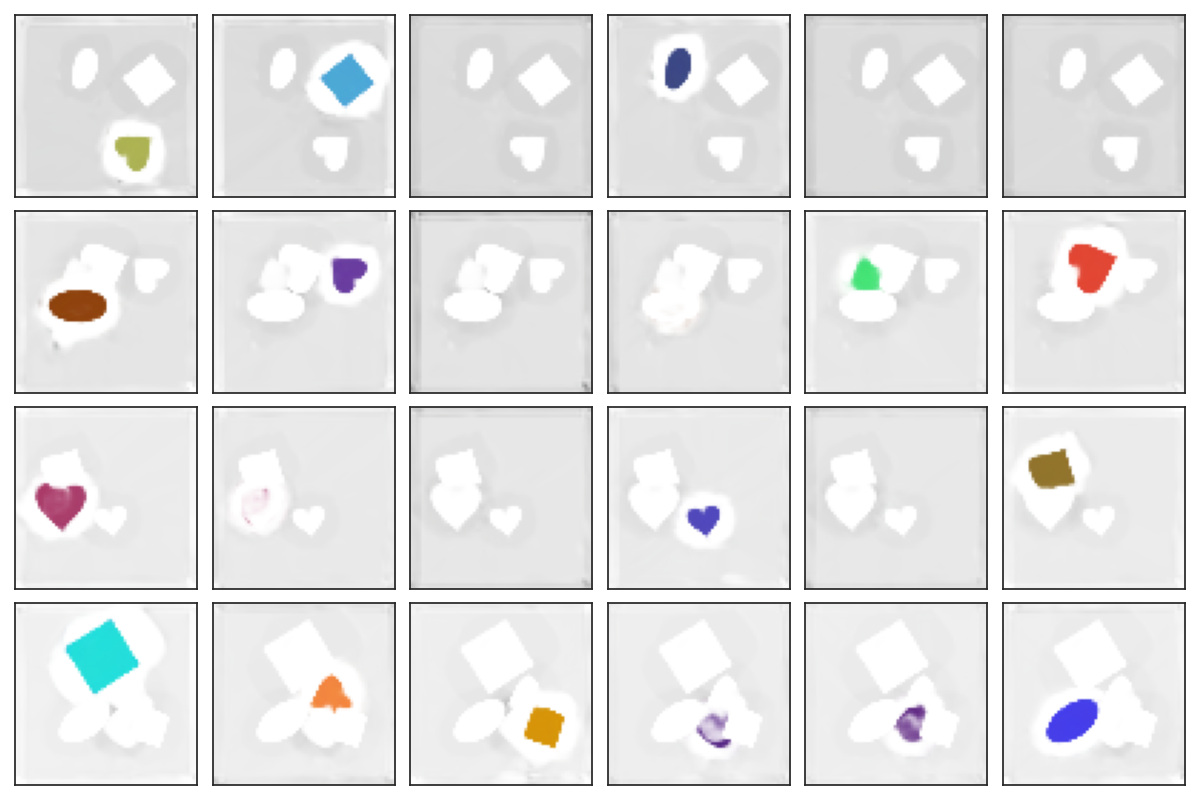}
    
    \vspace{0.45cm}
    
    \includegraphics[height=3.1cm]{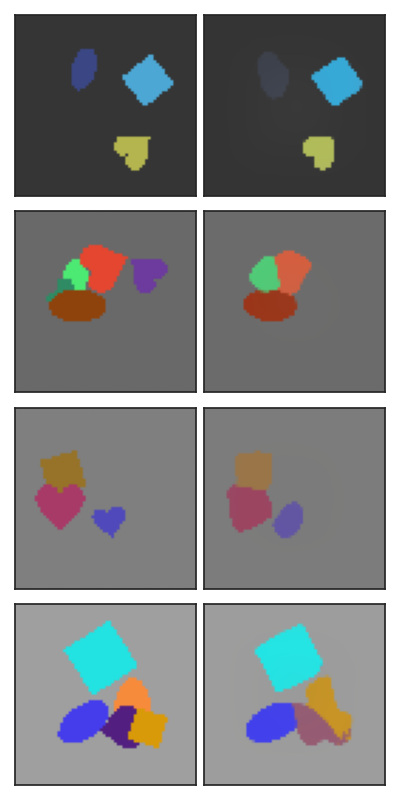}
    \hspace{0.2cm}
    \includegraphics[height=3.1cm]{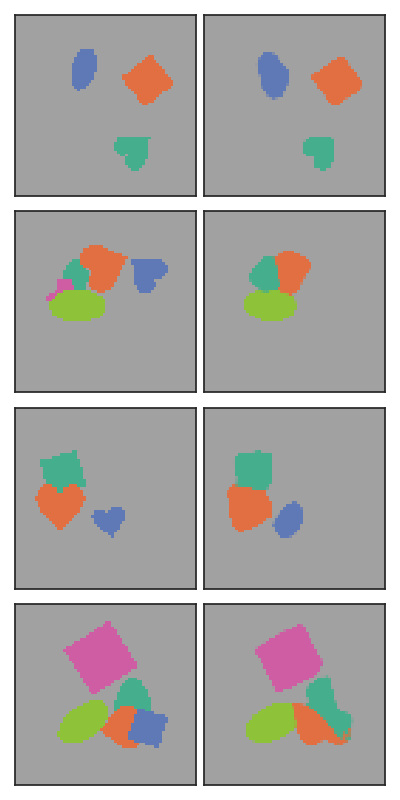}
    \hspace{0.2cm}
    \includegraphics[height=3.1cm]{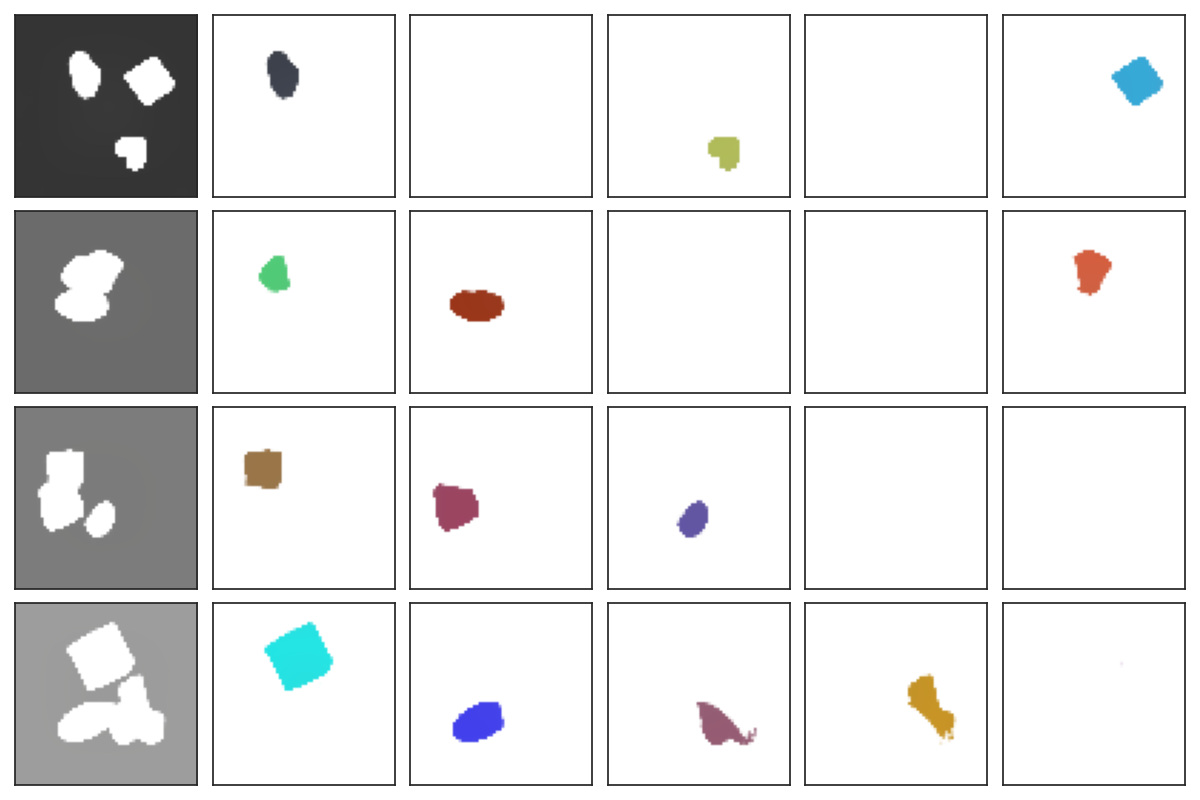}
    
    \vspace{0.45cm}
    
    \includegraphics[height=3.1cm]{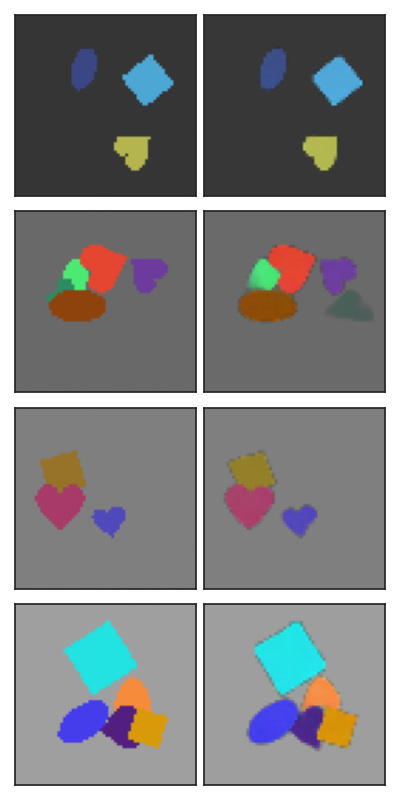}
    \hspace{0.2cm}
    \includegraphics[height=3.1cm]{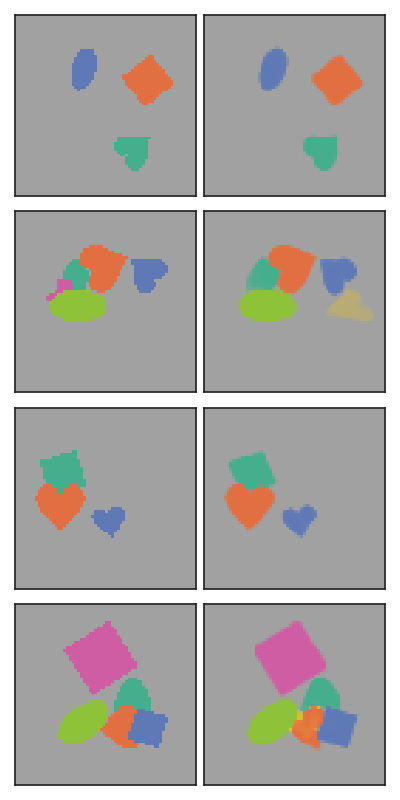}
    \hspace{0.2cm}
    \includegraphics[height=3.1cm]{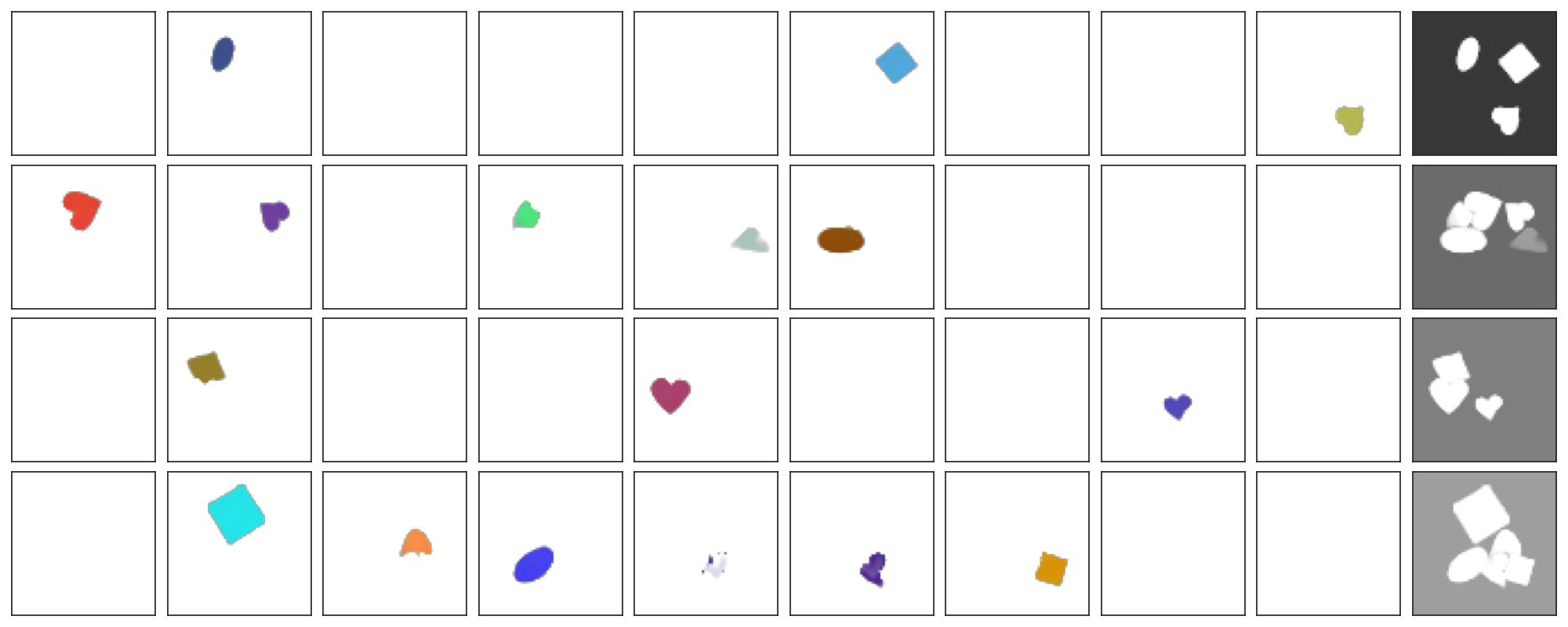}
    
    \vspace{0.45cm}
    \caption[\textbf{Reconstruction and segmentation} of 4 random images from the held-out test set of \textbf{Multi-dSprites}]{\textbf{Reconstruction and segmentation} of 4 random images from the held-out test set of \textbf{Multi-dSprites}. Top to bottom: MONet, Slot Attention, GENESIS, SPACE. Left to right: input, reconstruction, ground-truth masks, predicted (soft) masks, slot-wise reconstructions (masked with the predicted masks). As explained in the text, for SPACE we select the 10 most salient slots using the predicted masks. For each model type, we visualize the specific model with the highest ARI score in the \textit{validation} set. The images shown here are from the \textit{test} set and were not used for model selection.}
    \label{fig:app_model_viz_multidsprites}
\end{figure}
\begin{figure}
    \centering
    \includegraphics[height=3.1cm]{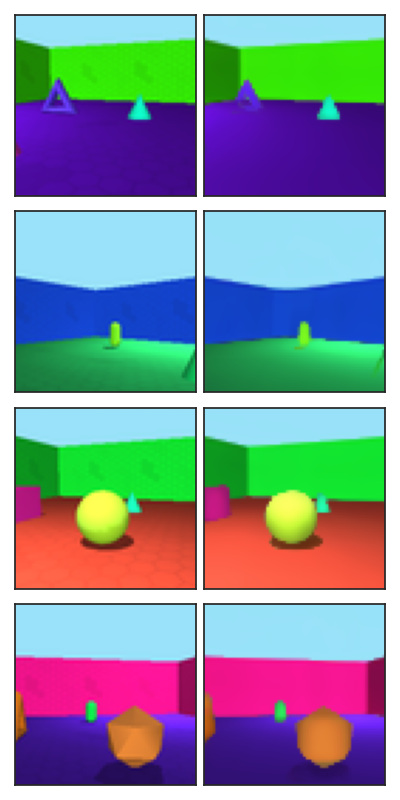}
    \hspace{0.2cm}
    \includegraphics[height=3.1cm]{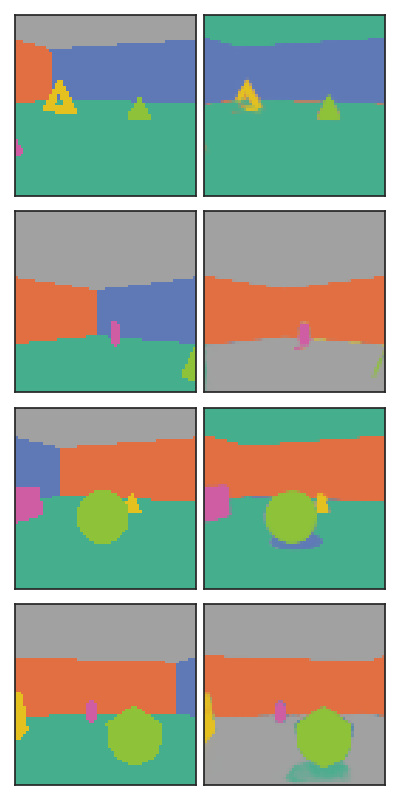}
    \hspace{0.2cm}
    \includegraphics[height=3.1cm]{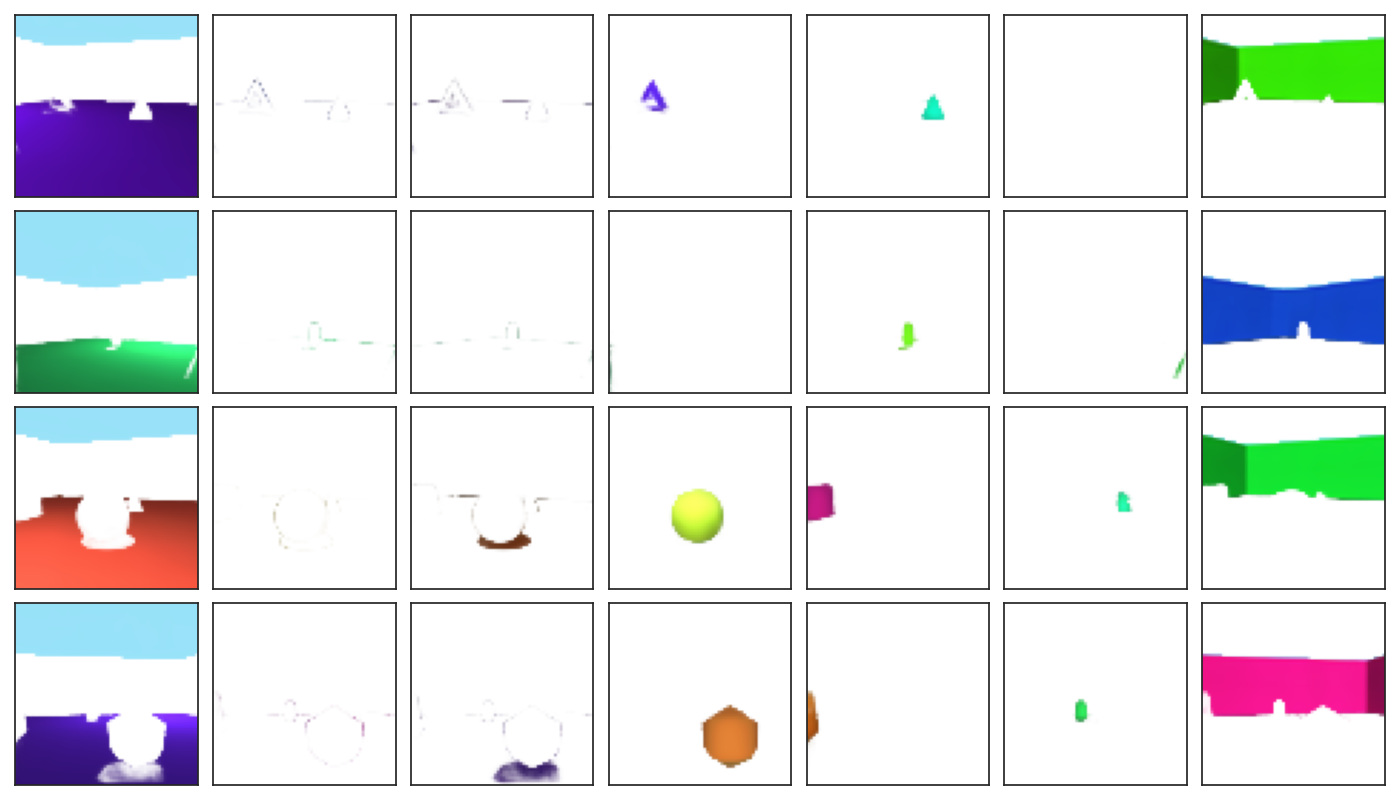}
    
    \vspace{0.45cm}
    
    \includegraphics[height=3.1cm]{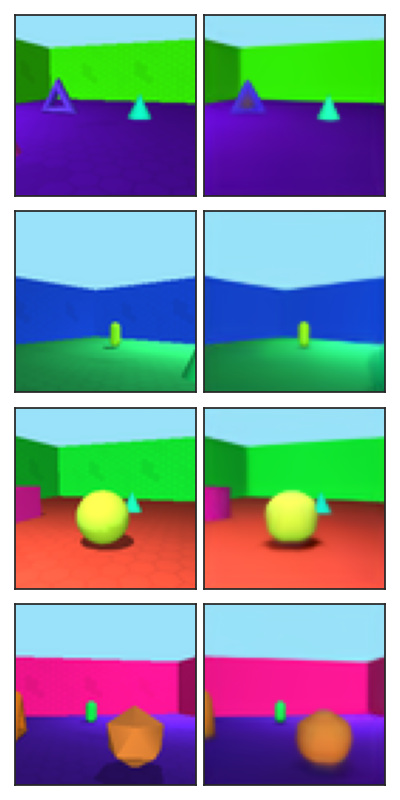}
    \hspace{0.2cm}
    \includegraphics[height=3.1cm]{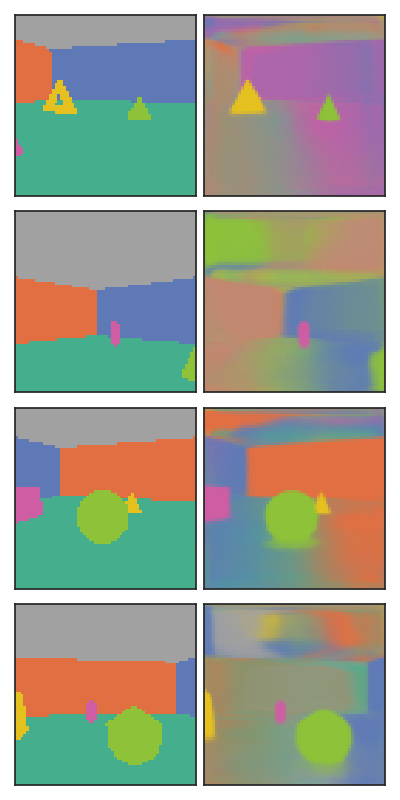}
    \hspace{0.2cm}
    \includegraphics[height=3.1cm]{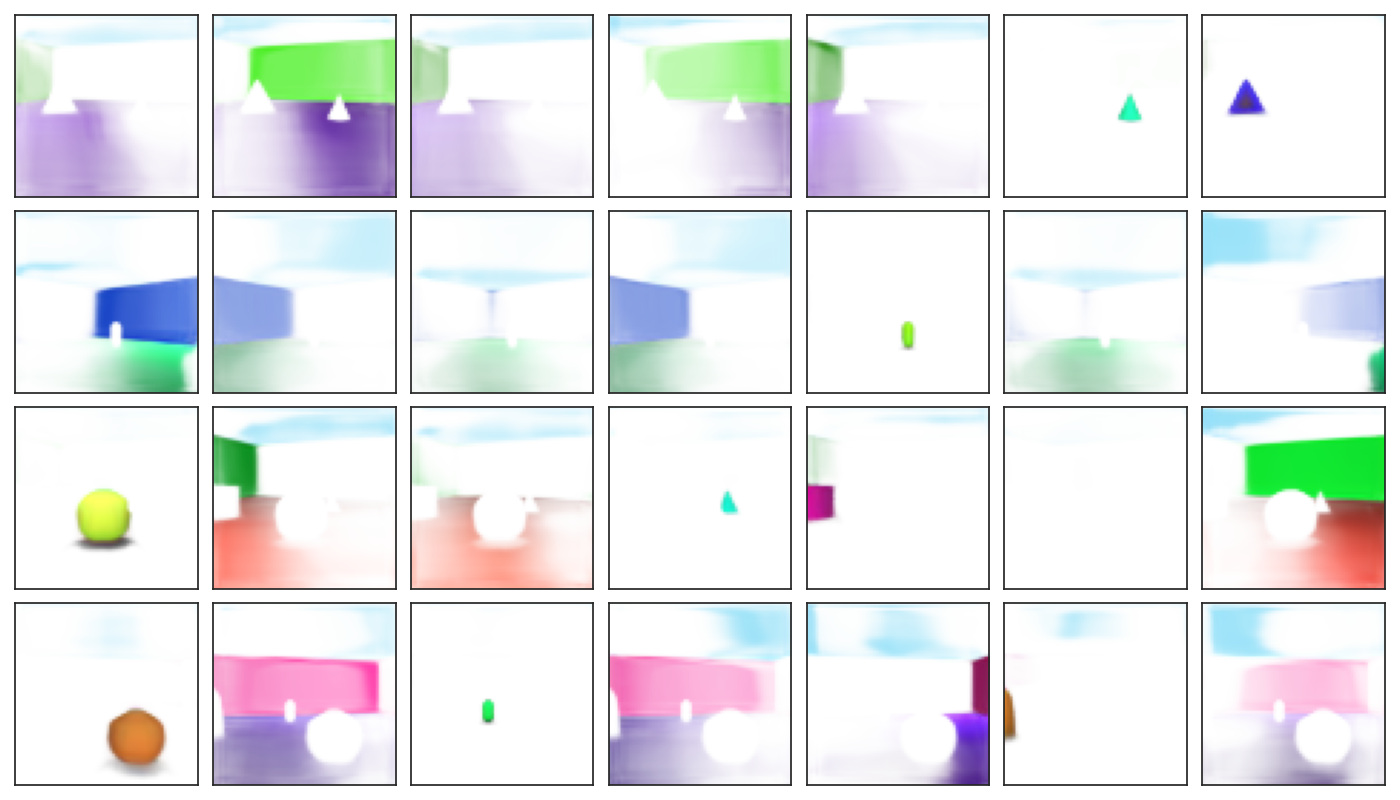}
    
    \vspace{0.45cm}
    
    \includegraphics[height=3.1cm]{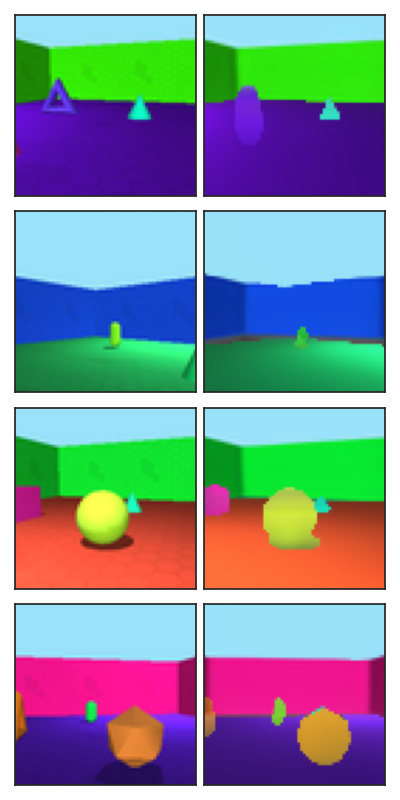}
    \hspace{0.2cm}
    \includegraphics[height=3.1cm]{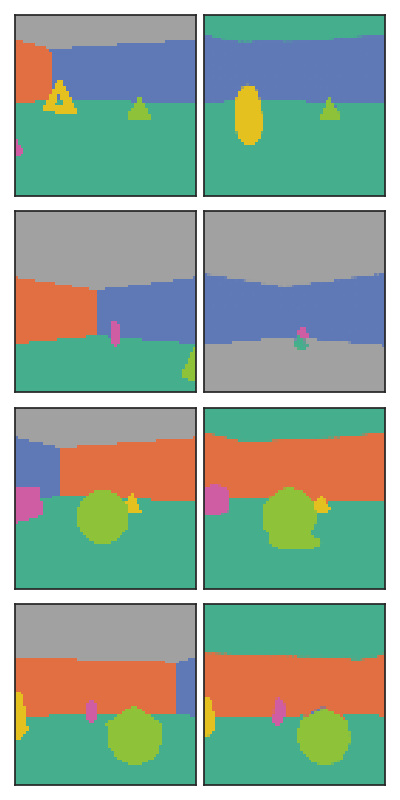}
    \hspace{0.2cm}
    \includegraphics[height=3.1cm]{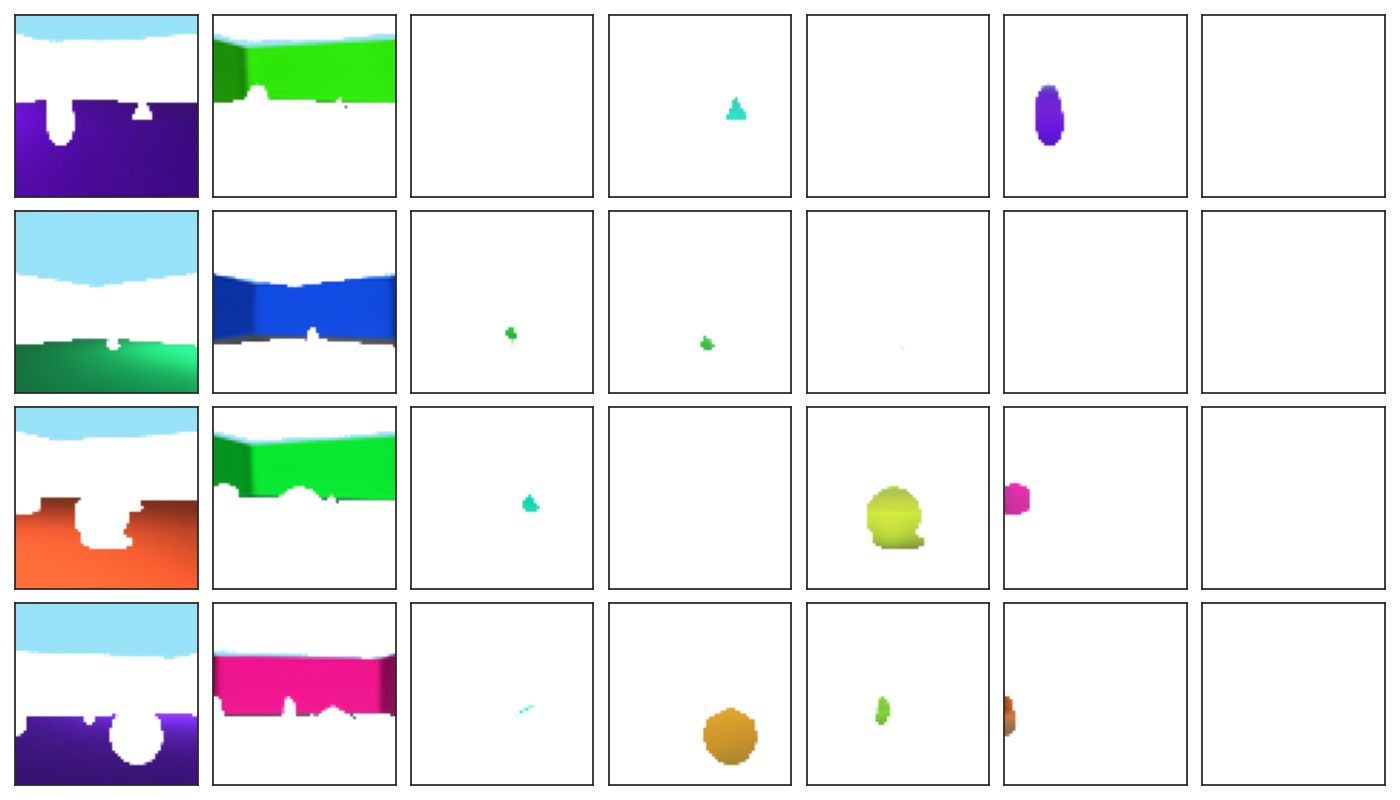}
    
    \vspace{0.45cm}
    
    \includegraphics[height=3.1cm]{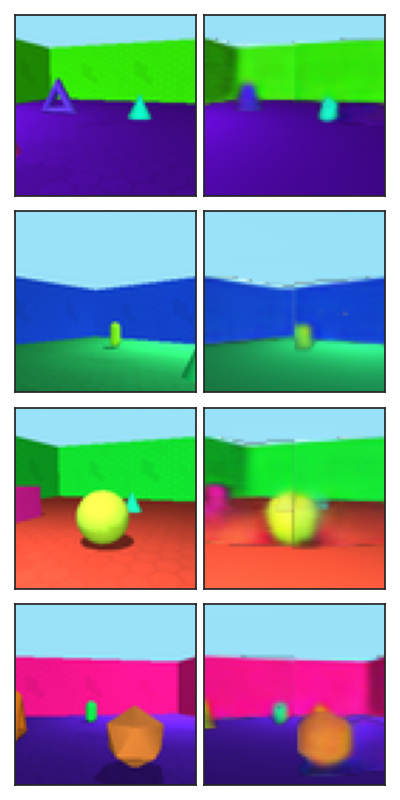}
    \hspace{0.2cm}
    \includegraphics[height=3.1cm]{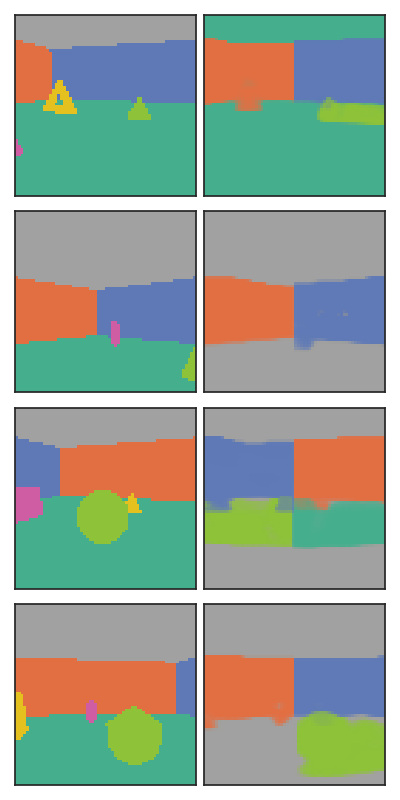}
    \hspace{0.2cm}
    \includegraphics[height=3.1cm]{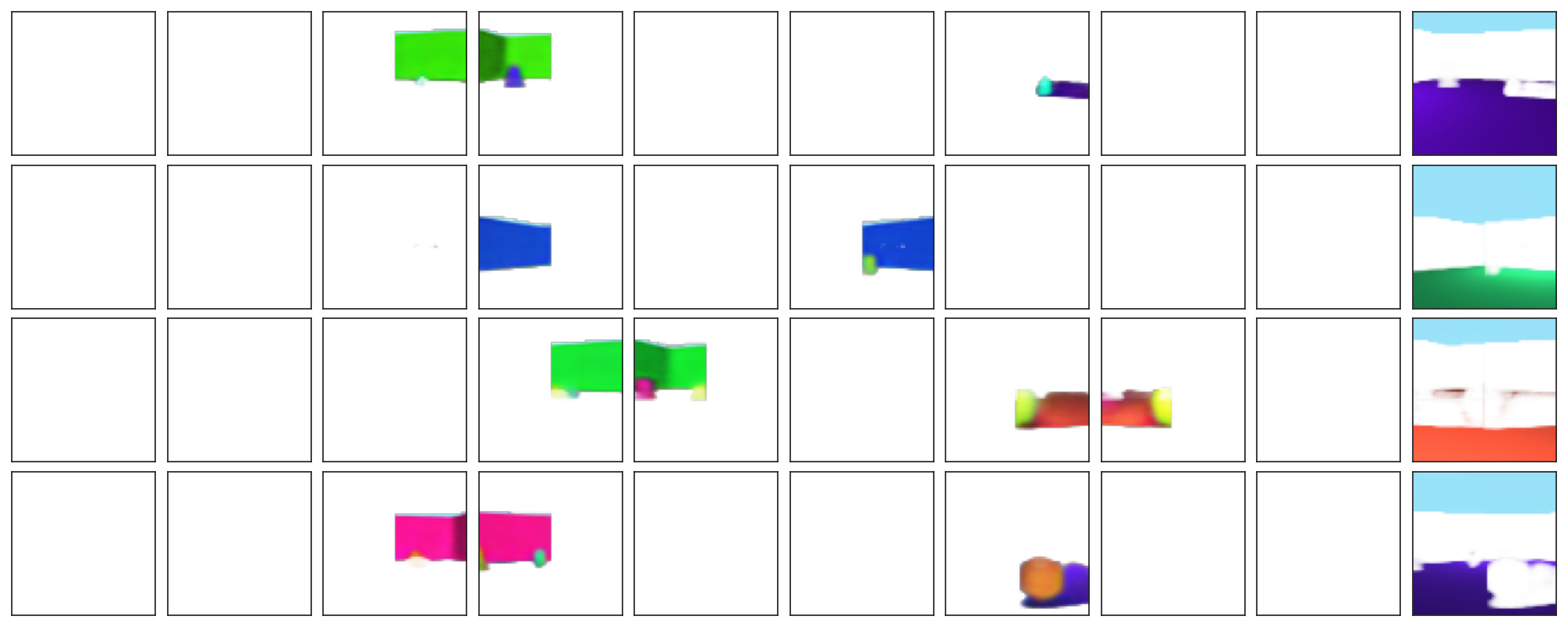}
    
    \vspace{0.45cm}
    \caption[\textbf{Reconstruction and segmentation} of 4 random images from the held-out test set of \textbf{Objects Room}]{\textbf{Reconstruction and segmentation} of 4 random images from the held-out test set of \textbf{Objects Room}. Top to bottom: MONet, Slot Attention, GENESIS, SPACE. Left to right: input, reconstruction, ground-truth masks, predicted (soft) masks, slot-wise reconstructions (masked with the predicted masks). As explained in the text, for SPACE we select the 10 most salient slots using the predicted masks. For each model type, we visualize the specific model with the highest ARI score in the \textit{validation} set. The images shown here are from the \textit{test} set and were not used for model selection.}
    \label{fig:app_model_viz_objectsroom}
\end{figure}
\begin{figure}
    \centering
    \includegraphics[height=3.1cm]{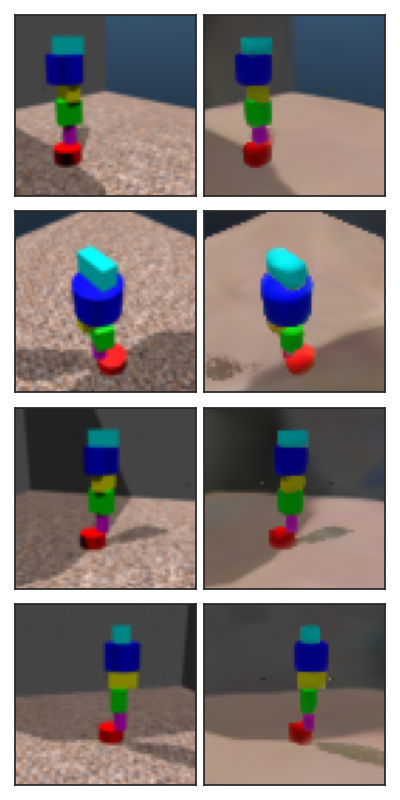}
    \hspace{0.2cm}
    \includegraphics[height=3.1cm]{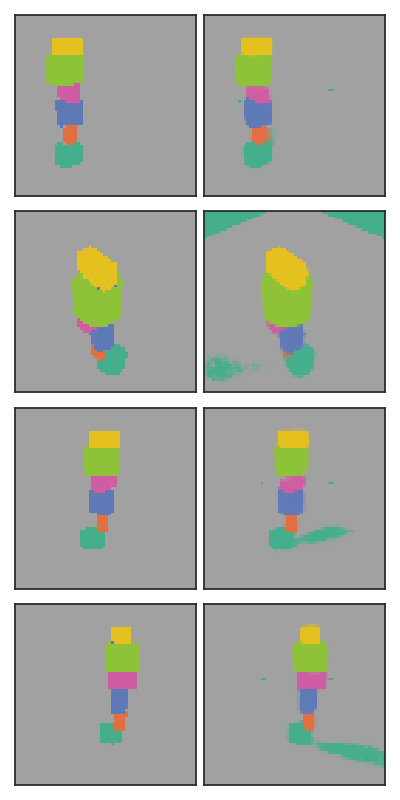}
    \hspace{0.2cm}
    \includegraphics[height=3.1cm]{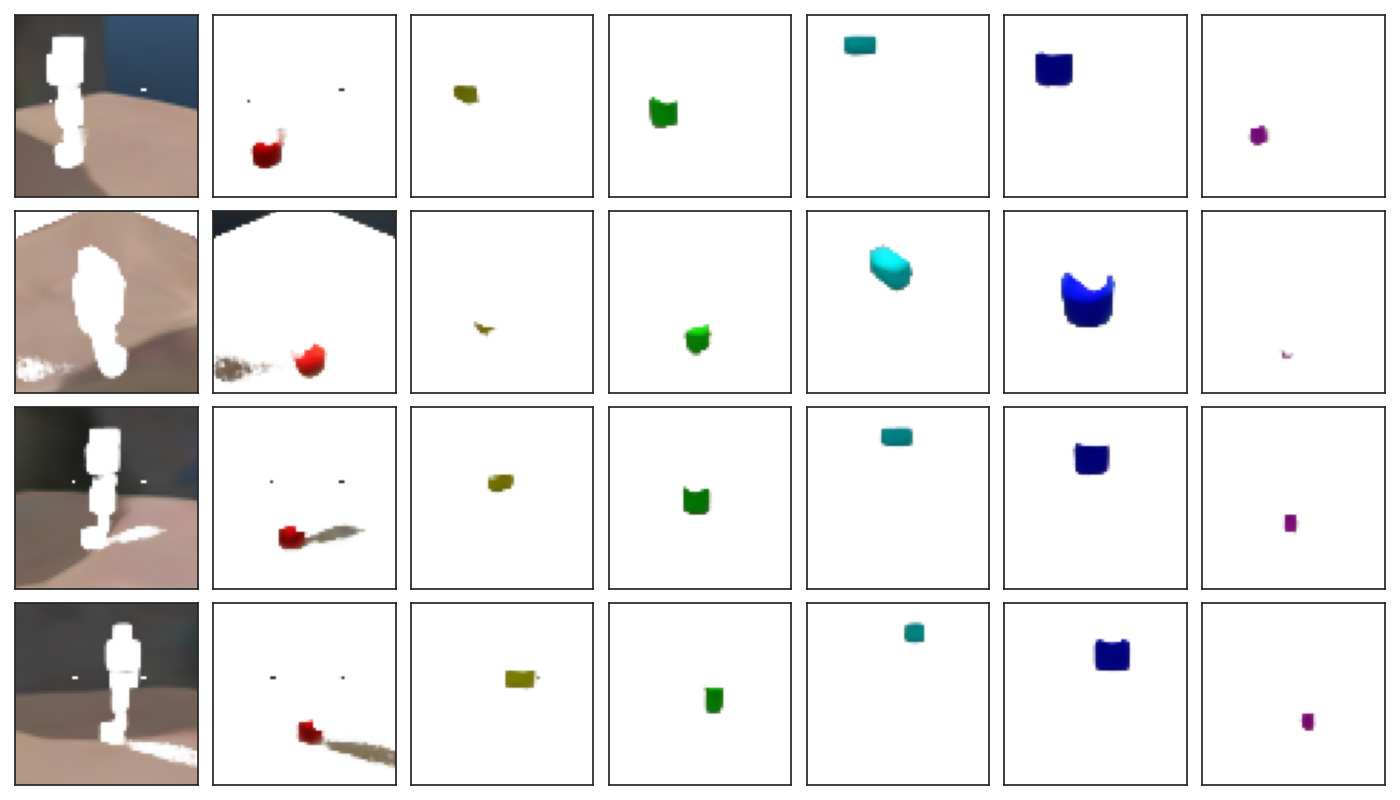}
    
    \vspace{0.45cm}
    
    \includegraphics[height=3.1cm]{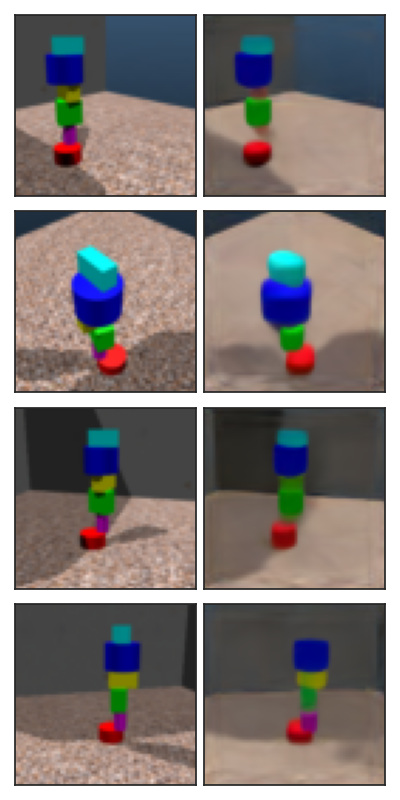}
    \hspace{0.2cm}
    \includegraphics[height=3.1cm]{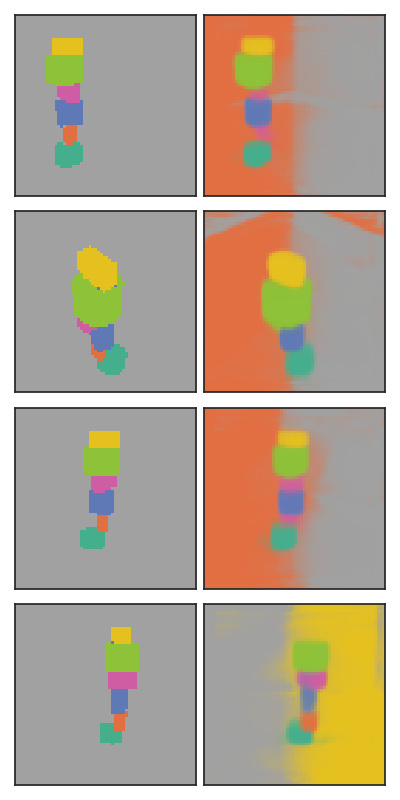}
    \hspace{0.2cm}
    \includegraphics[height=3.1cm]{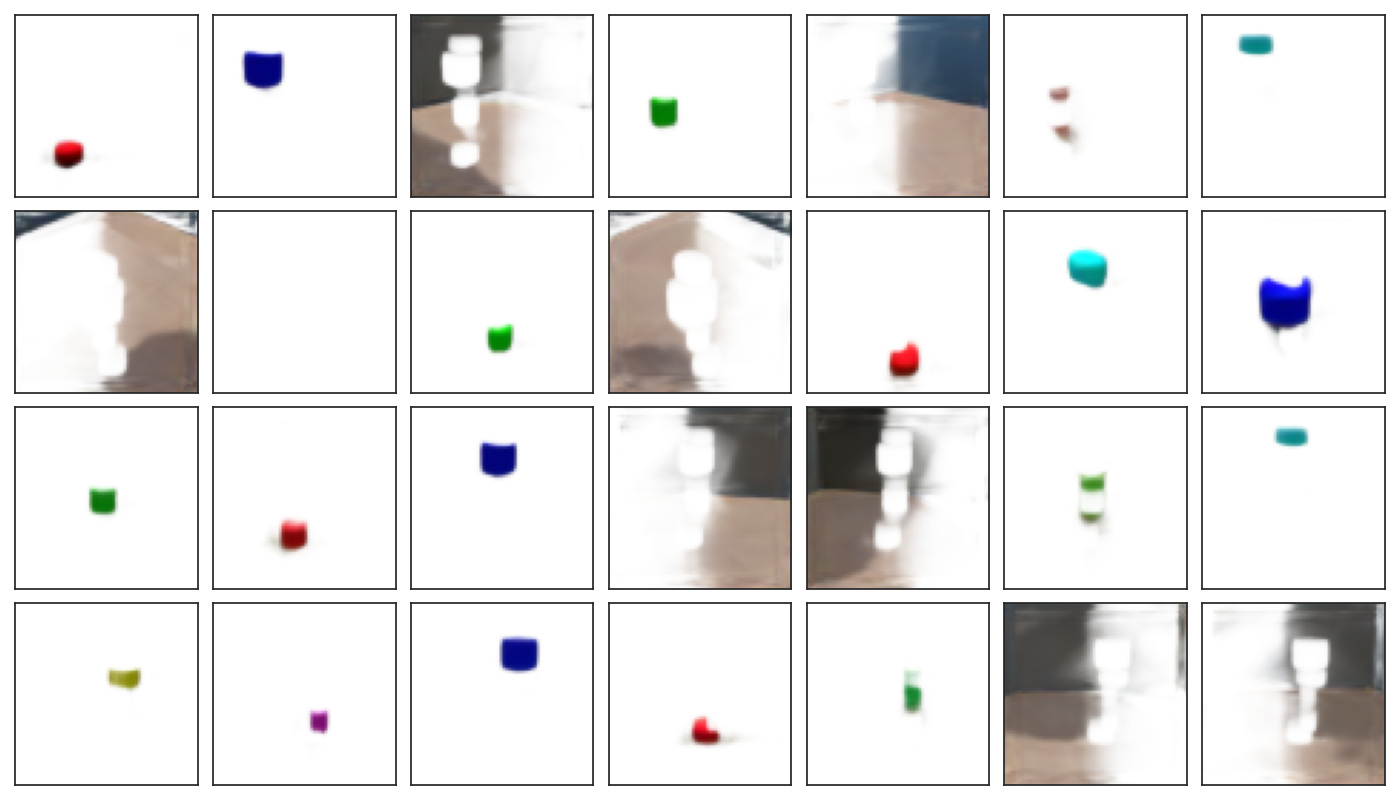}
    
    \vspace{0.45cm}
    
    \includegraphics[height=3.1cm]{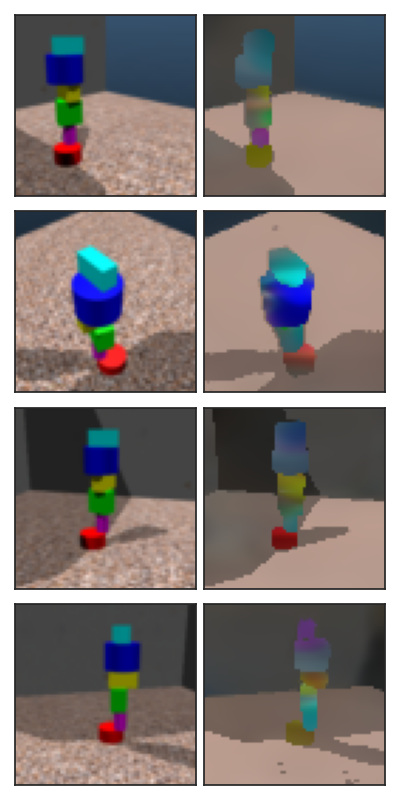}
    \hspace{0.2cm}
    \includegraphics[height=3.1cm]{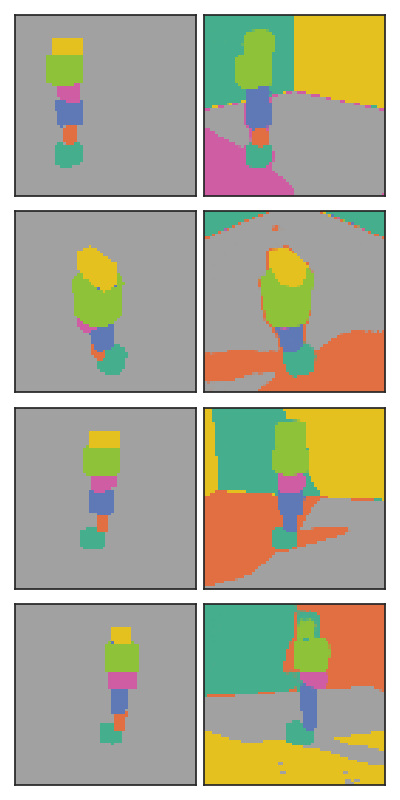}
    \hspace{0.2cm}
    \includegraphics[height=3.1cm]{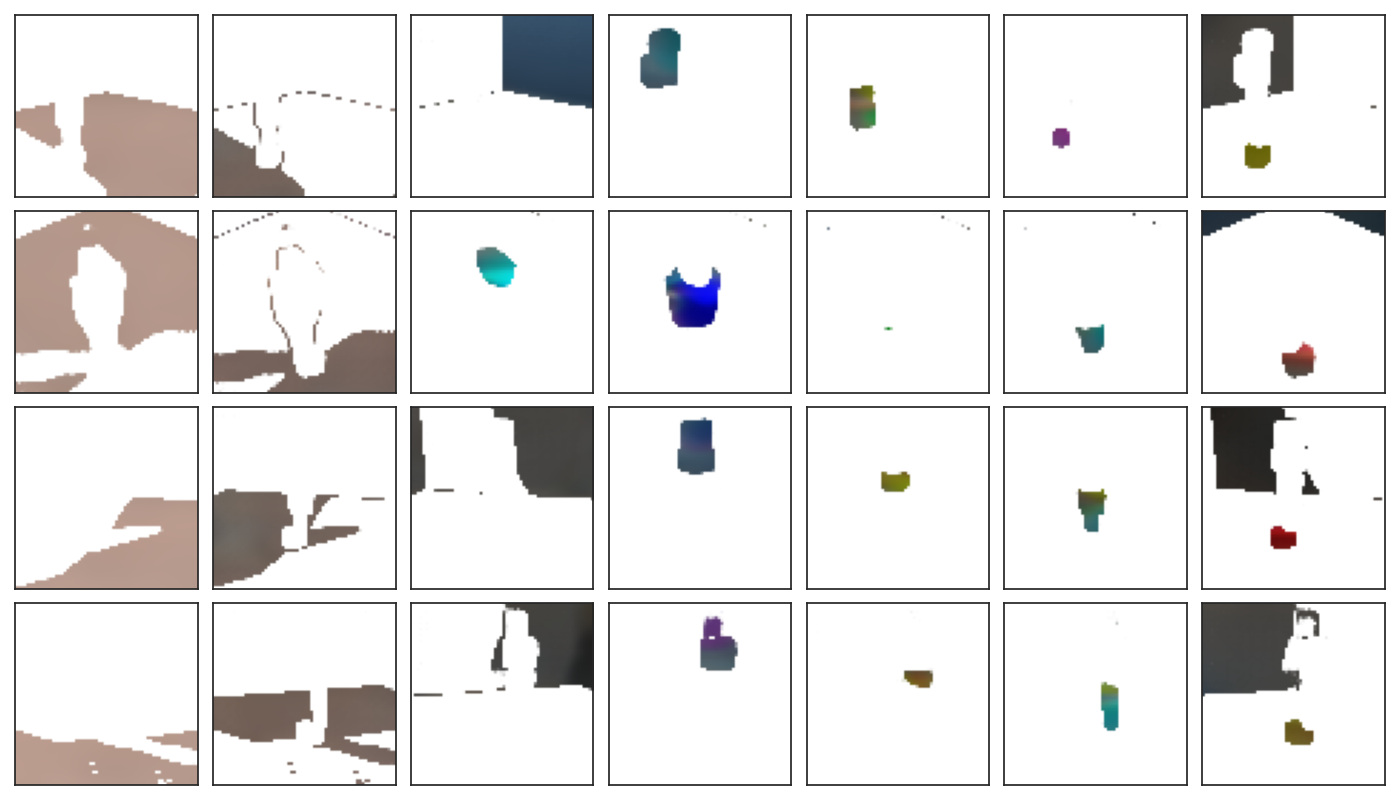}
    
    \vspace{0.45cm}
    
    \includegraphics[height=3.1cm]{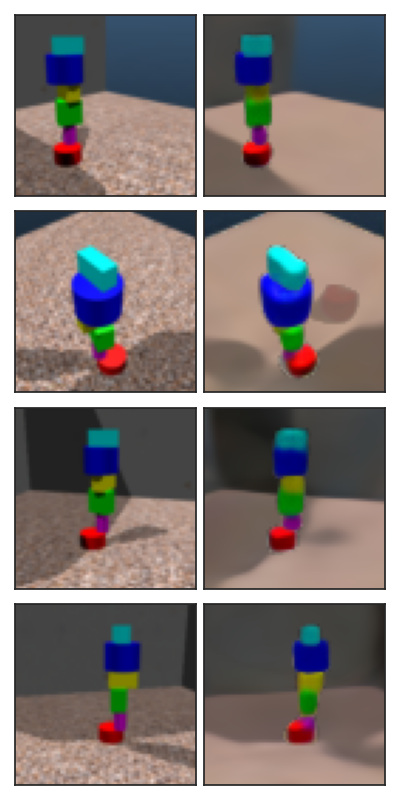}
    \hspace{0.2cm}
    \includegraphics[height=3.1cm]{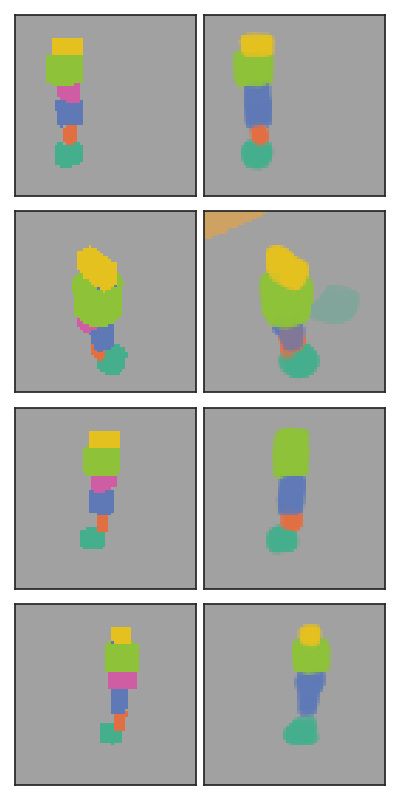}
    \hspace{0.2cm}
    \includegraphics[height=3.1cm]{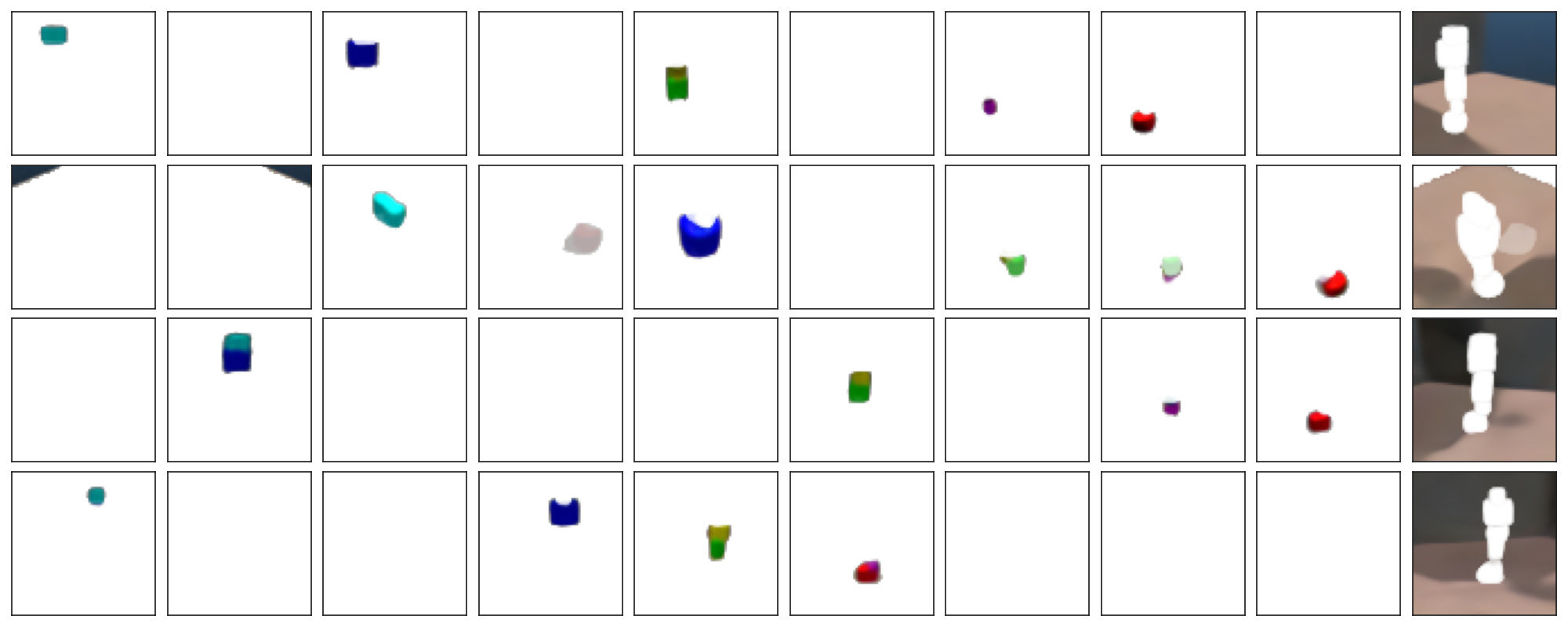}

    \vspace{0.45cm}
    \caption[\textbf{Reconstruction and segmentation} of 4 random images from the held-out test set of \textbf{Shapestacks}]{\textbf{Reconstruction and segmentation} of 4 random images from the held-out test set of \textbf{Shapestacks}. Top to bottom: MONet, Slot Attention, GENESIS, SPACE. Left to right: input, reconstruction, ground-truth masks, predicted (soft) masks, slot-wise reconstructions (masked with the predicted masks). As explained in the text, for SPACE we select the 10 most salient slots using the predicted masks. For each model type, we visualize the specific model with the highest ARI score in the \textit{validation} set. The images shown here are from the \textit{test} set and were not used for model selection.}
    \label{fig:app_model_viz_shapestacks}
\end{figure}
\begin{figure}
    \centering
    \includegraphics[height=3.1cm]{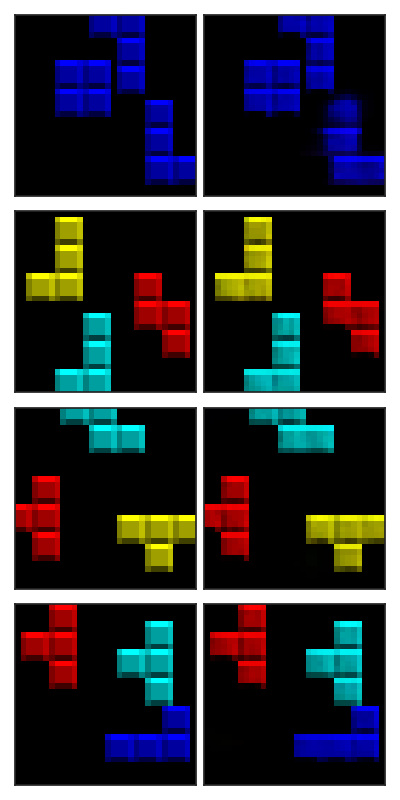}
    \hspace{0.2cm}
    \includegraphics[height=3.1cm]{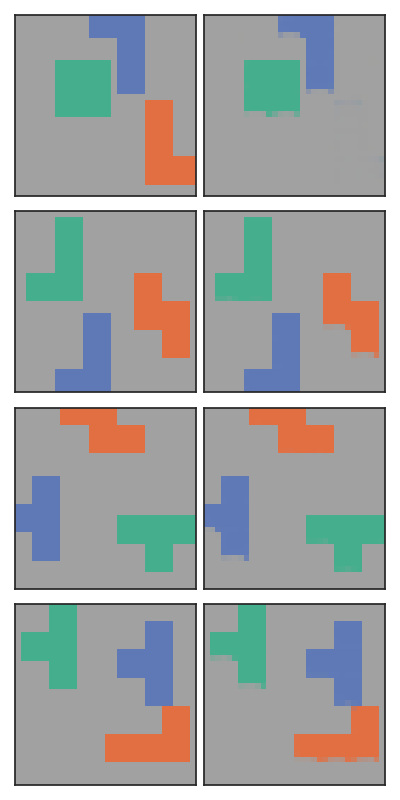}
    \hspace{0.2cm}
    \includegraphics[height=3.1cm]{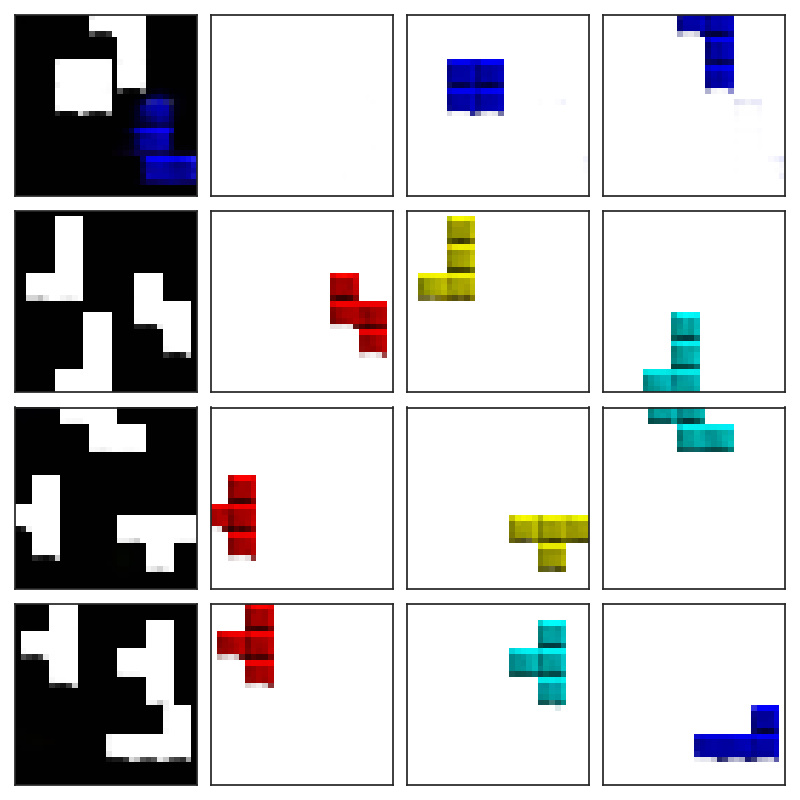}
    
    \vspace{0.45cm}
    
    \includegraphics[height=3.1cm]{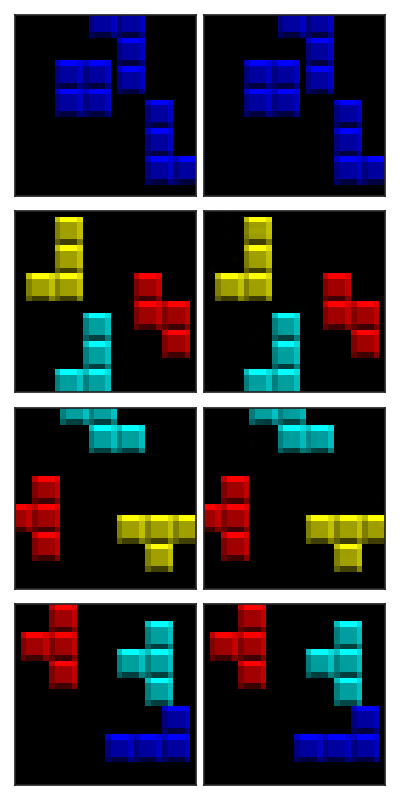}
    \hspace{0.2cm}
    \includegraphics[height=3.1cm]{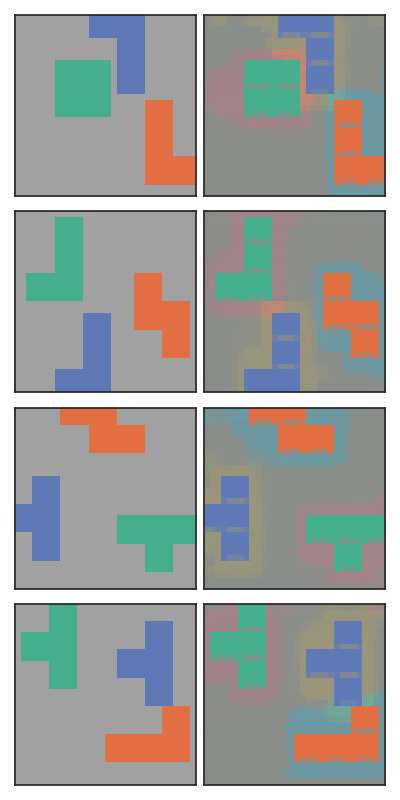}
    \hspace{0.2cm}
    \includegraphics[height=3.1cm]{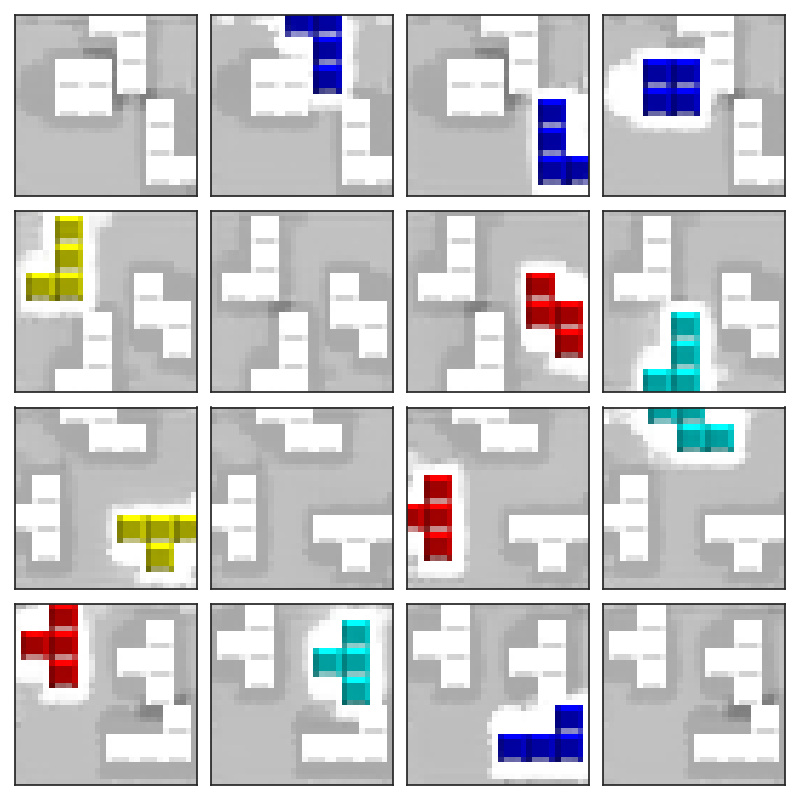}
    
    \vspace{0.45cm}
    
    \includegraphics[height=3.1cm]{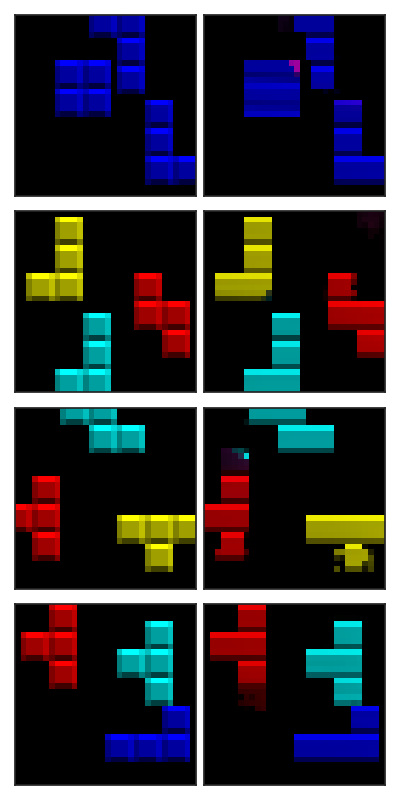}
    \hspace{0.2cm}
    \includegraphics[height=3.1cm]{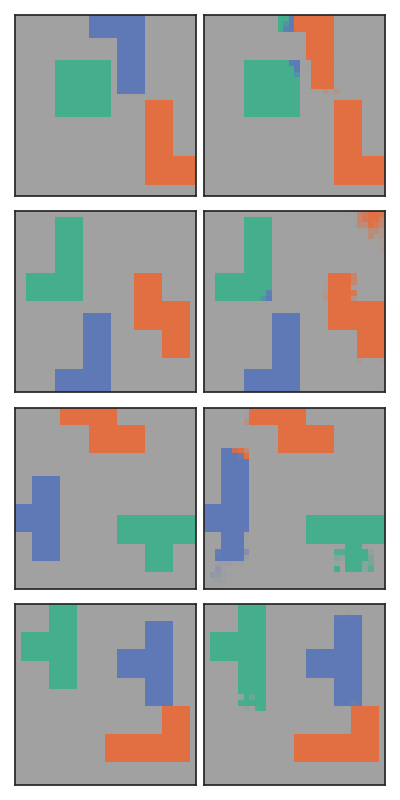}
    \hspace{0.2cm}
    \includegraphics[height=3.1cm]{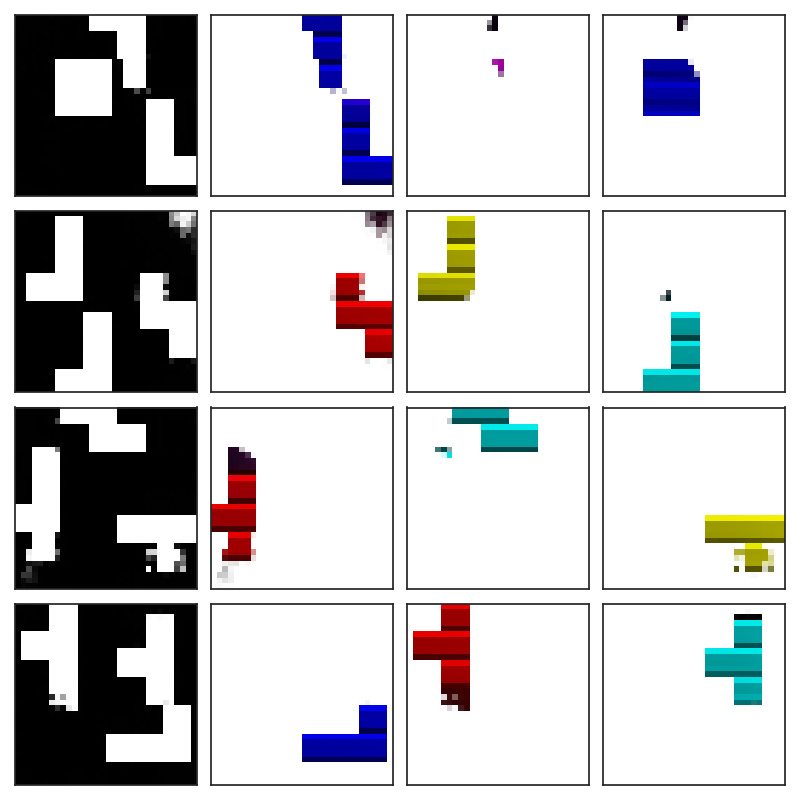}
    
    \vspace{0.45cm}

    \includegraphics[height=3.1cm]{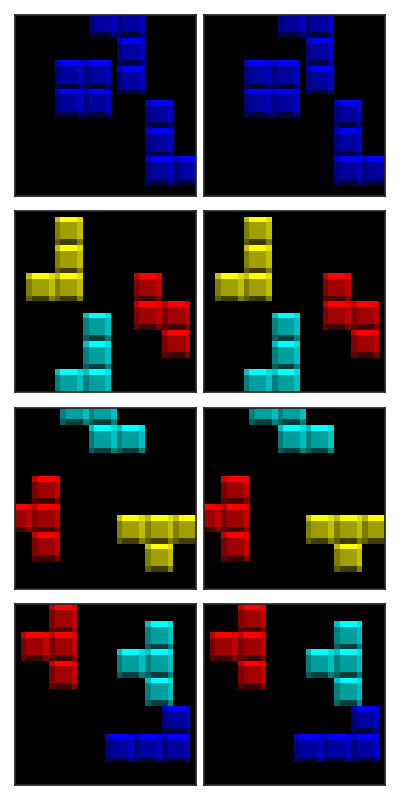}
    \hspace{0.2cm}
    \includegraphics[height=3.1cm]{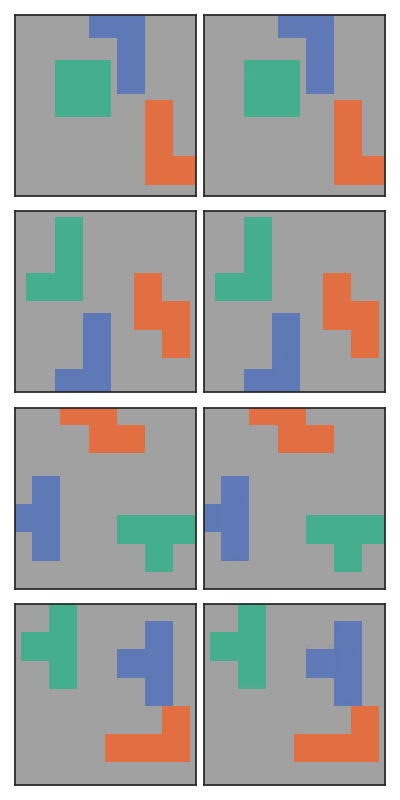}
    \hspace{0.2cm}
    \includegraphics[height=3.1cm]{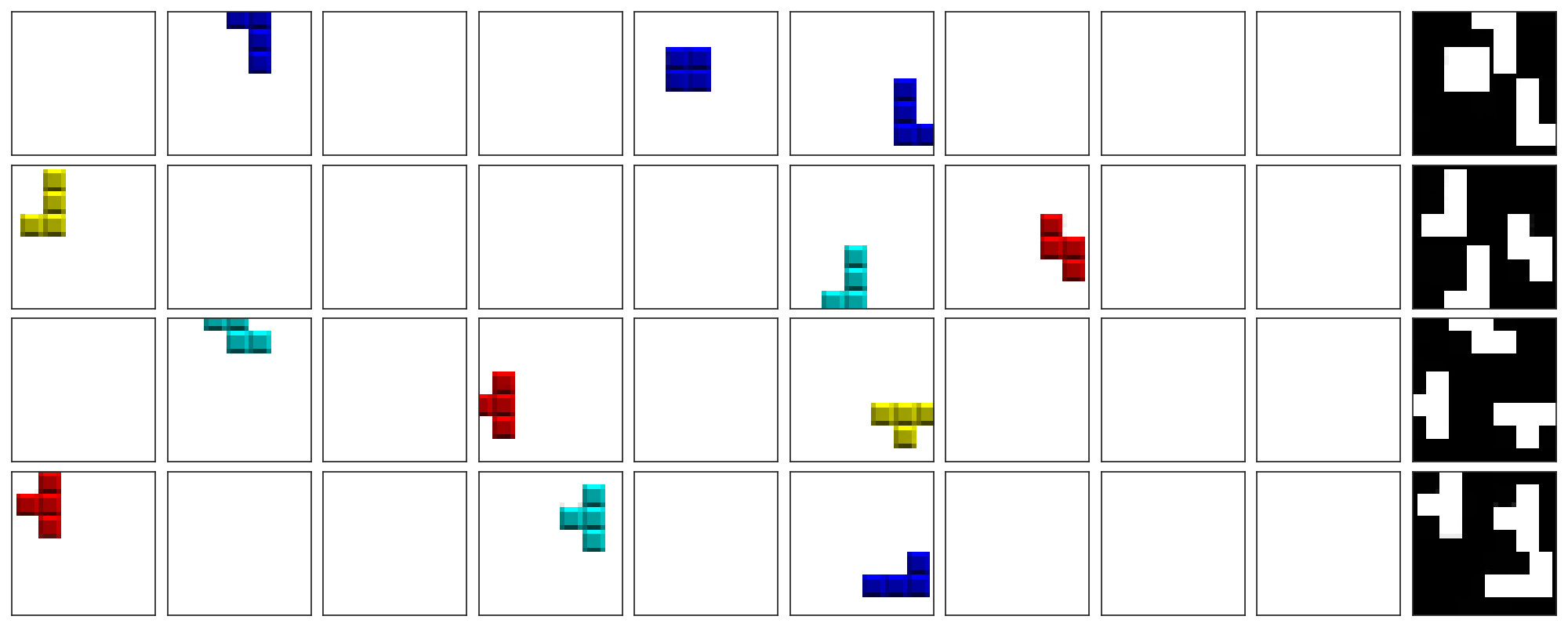}

    \vspace{0.45cm}
    \caption[\textbf{Reconstruction and segmentation} of 4 random images from the held-out test set of \textbf{Tetrominoes}]{\textbf{Reconstruction and segmentation} of 4 random images from the held-out test set of \textbf{Tetrominoes}. Top to bottom: MONet, Slot Attention, GENESIS, SPACE. Left to right: input, reconstruction, ground-truth masks, predicted (soft) masks, slot-wise reconstructions (masked with the predicted masks). As explained in the text, for SPACE we select the 10 most salient slots using the predicted masks. For each model type, we visualize the specific model with the highest ARI score in the \textit{validation} set. The images shown here are from the \textit{test} set and were not used for model selection.}
    \label{fig:app_model_viz_tetrominoes}
\end{figure}

%%% BASELINES

\begin{figure}
    \centering
    \includegraphics[height=4cm]{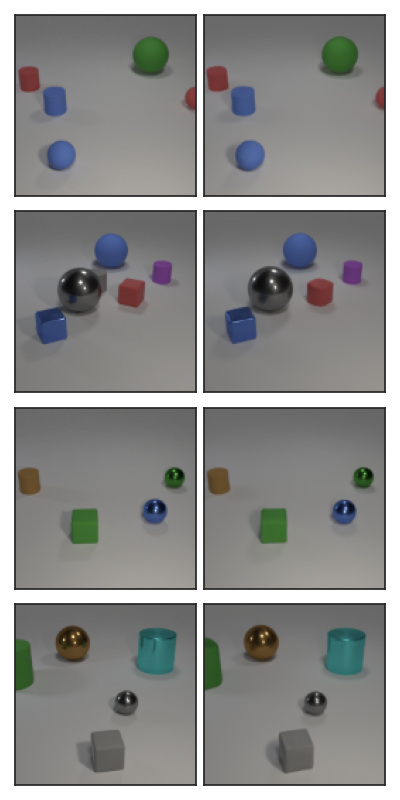}
    \hspace{0.2cm}
    \includegraphics[height=4cm]{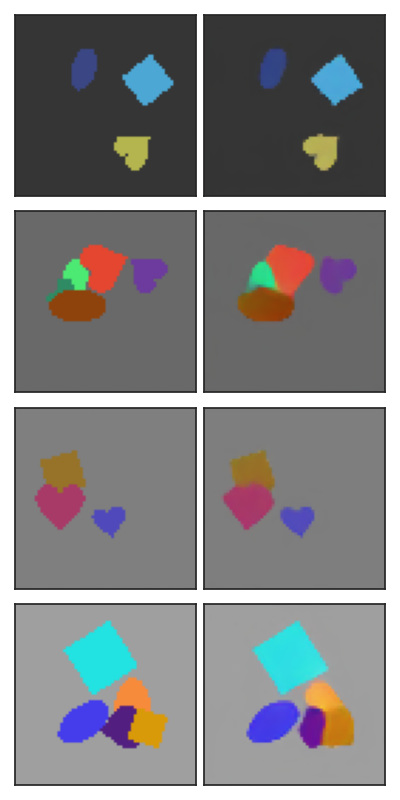}
    \hspace{0.2cm}
    \includegraphics[height=4cm]{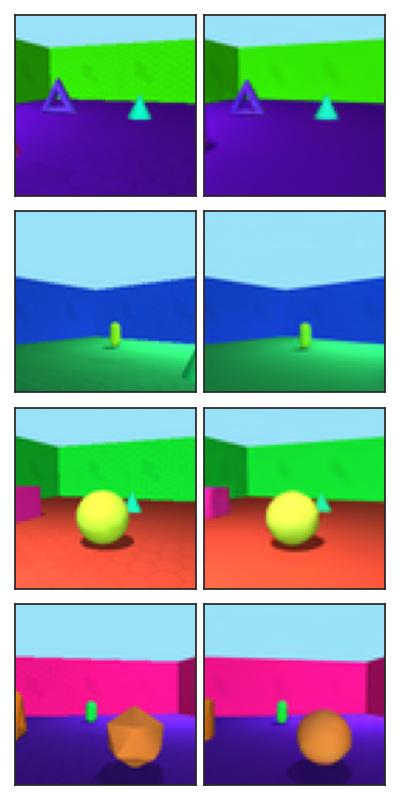}
    \hspace{0.2cm}
    \includegraphics[height=4cm]{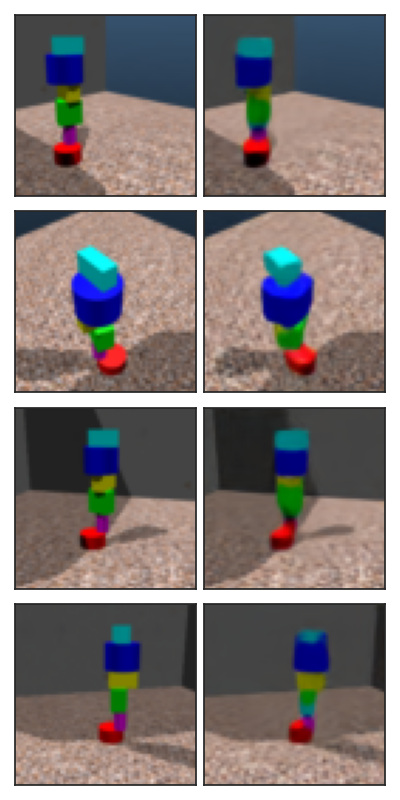}
    \hspace{0.2cm}
    \includegraphics[height=4cm]{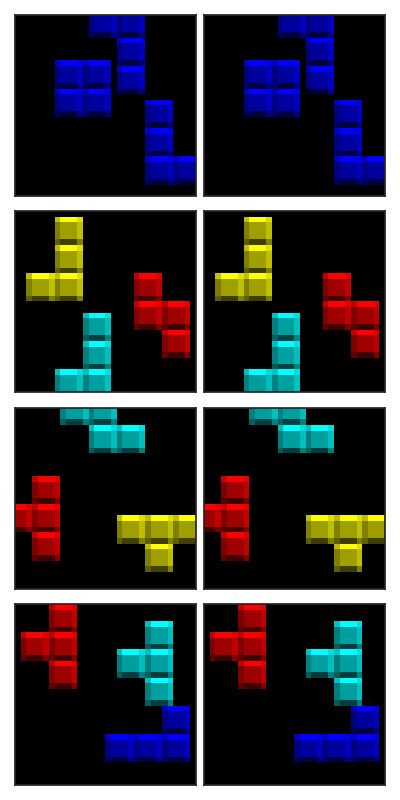}
    
    \vspace{0.45cm}
    
    \includegraphics[height=4cm]{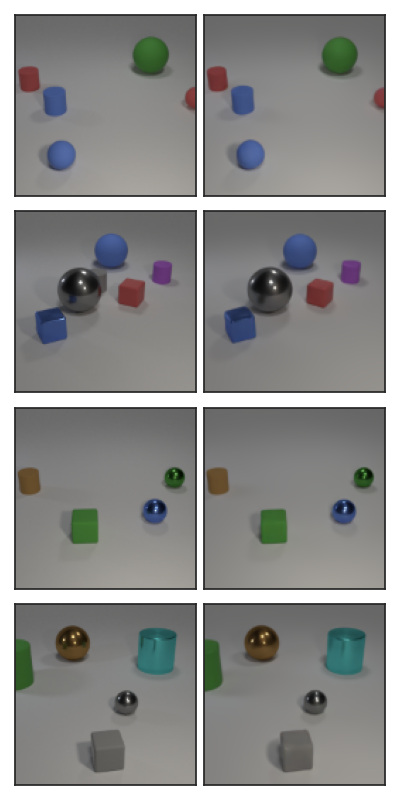}
    \hspace{0.2cm}
    \includegraphics[height=4cm]{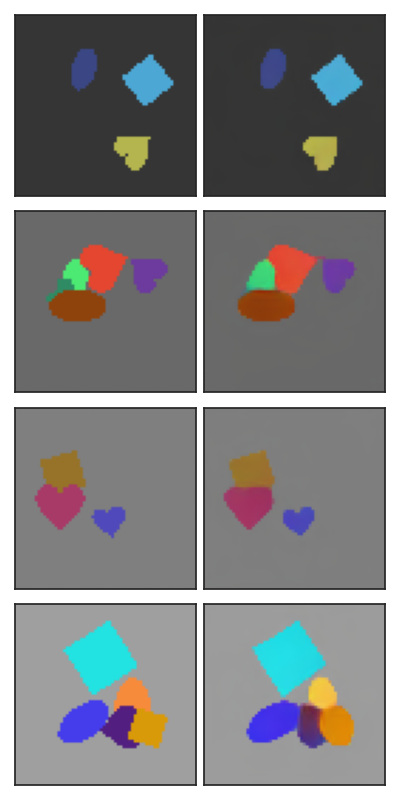}
    \hspace{0.2cm}
    \includegraphics[height=4cm]{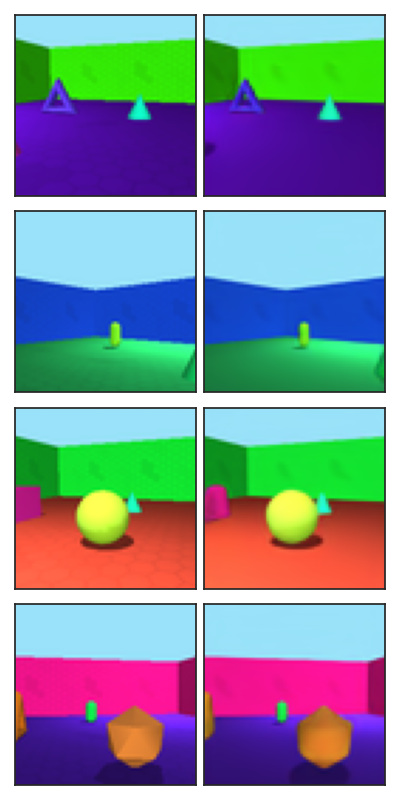}
    \hspace{0.2cm}
    \includegraphics[height=4cm]{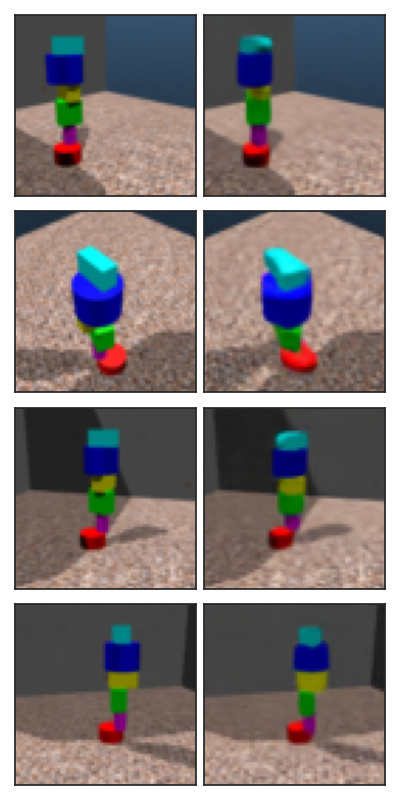}
    \hspace{0.2cm}
    \includegraphics[height=4cm]{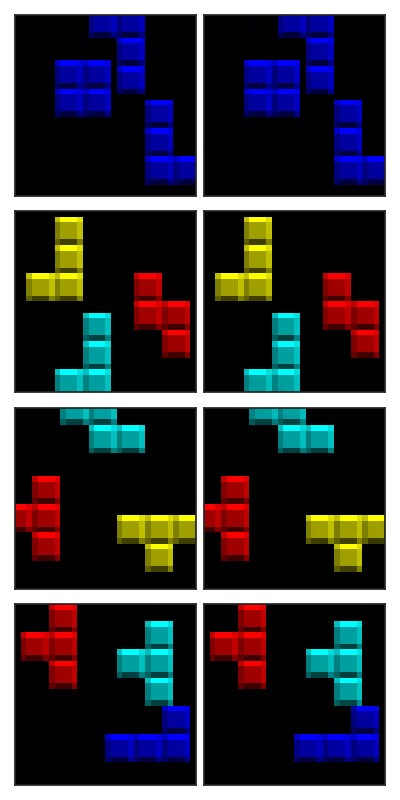}
    
    \vspace{0.45cm}
    \caption[\textbf{Input--reconstruction pairs} of 4 random images from the held-out \emph{test} set of all 5 datasets, for the VAE model with convolutional and broadcast decoder.]{\textbf{Input--reconstruction pairs} of 4 random images from the held-out \emph{test} set of all 5 datasets, for the VAE model with convolutional (top) and broadcast (bottom) decoder. Each VAE type was trained with 5 random seeds, and for each type we show here the model with the lowest MSE on the \textit{validation} set. The images shown here are from the \textit{test} set and were not used for model selection. For each image, we show the input on the left and the reconstruction on the right. As these are not slot-based models, segmentation masks and slot-wise reconstructions are not available.}
    \label{fig:app_model_viz_vaes}
\end{figure}

% OOD results

\def\OodVizVspace{\hspace{11pt}}
\def\OodVizHspace{\hspace{11pt}}
\newcommand{\OodViz}[3]{\includegraphics[height=0.17\textheight]{figures/objects/model_viz/#1/#2/#3/input_recon}}
\newcommand{\OodVizRow}[2]{%
    \OodViz{#1}{monet}{#2}
    \OodVizHspace
    \OodViz{#1}{slot-attention}{#2}
    \OodVizHspace
    \OodViz{#1}{genesis}{#2}
    \OodVizHspace
    \OodViz{#1}{space}{#2}
    \OodVizHspace
    \OodViz{#1}{baseline_vae_mlp}{#2}
    \OodVizHspace
    \OodViz{#1}{baseline_vae_broadcast}{#2}
}

\begin{figure}
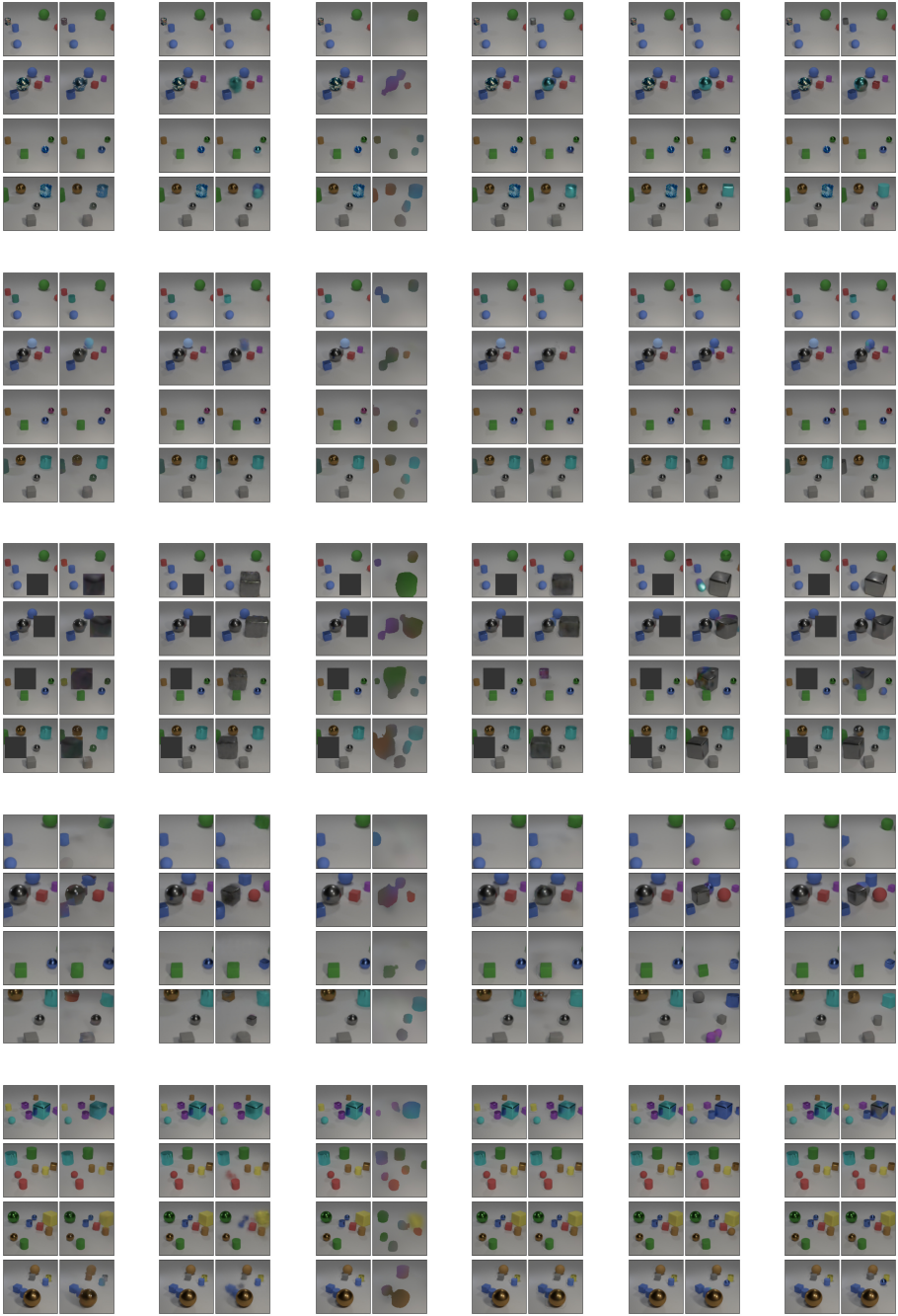

    \centering
    \OodVizRow{clevr}{object_style}
    
    \OodVizVspace
    
    \OodVizRow{clevr}{object_color}
    
    \OodVizVspace
    
    \OodVizRow{clevr}{occlusion}
    
    \OodVizVspace
    
    \OodVizRow{clevr}{crop}
    
    \OodVizVspace
    
    \OodVizRow{clevr}{num_objects}
    
    \caption[{Inputs and reconstructions for OOD images in CLEVR.}]{\textbf{Inputs and reconstructions for OOD images in CLEVR.} Columns from left to right: MONet, Slot Attention, GENESIS, SPACE, convolutional decoder VAE, broadcast decoder VAE. Rows from top to bottom: object style, object color, occlusion, crop, number of objects.}
    \label{fig:ood_visualizations_clevr}
\end{figure}

\begin{figure}
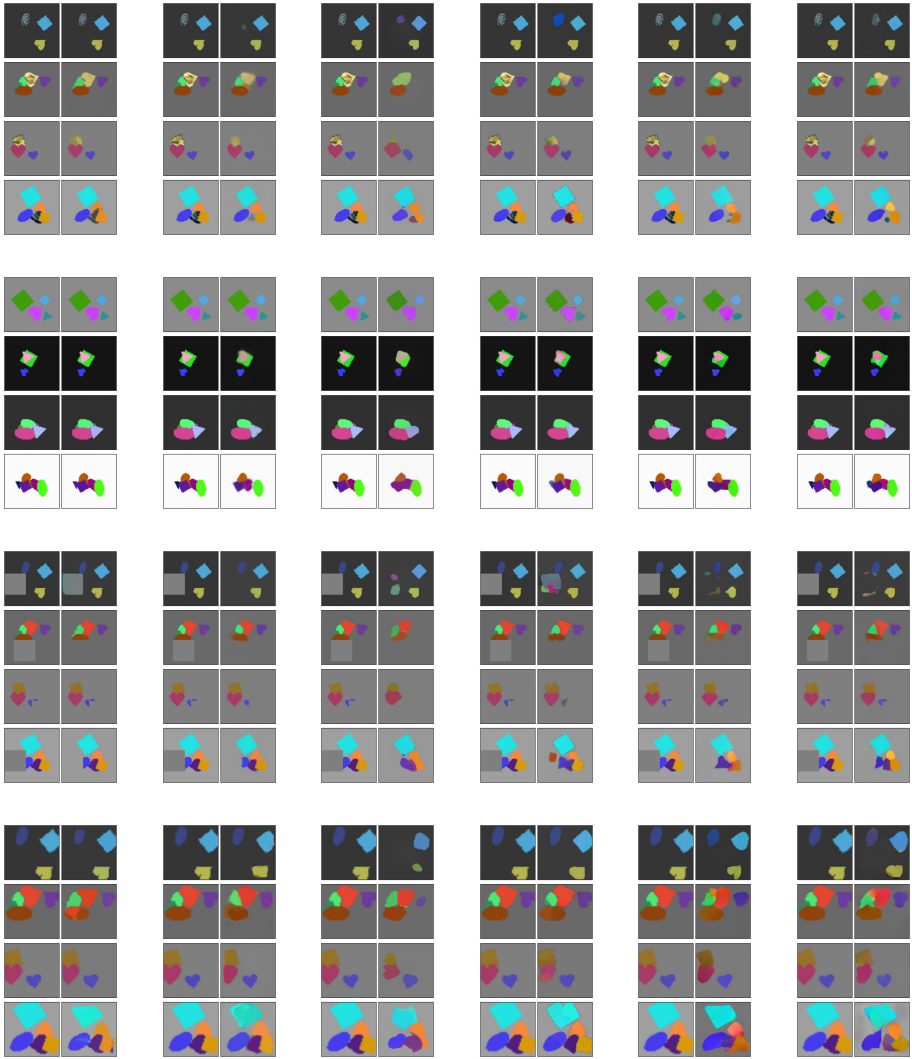

    \centering
    \OodVizRow{multidsprites}{object_style}
    
    \OodVizVspace
    
    \OodVizRow{multidsprites}{object_shape}
    
    \OodVizVspace
    
    \OodVizRow{multidsprites}{occlusion}
    
    \OodVizVspace
    
    \OodVizRow{multidsprites}{crop}
    
    \caption[{Inputs and reconstructions for OOD images in Multi-dSprites.}]{\textbf{Inputs and reconstructions for OOD images in Multi-dSprites.} Columns from left to right: MONet, Slot Attention, GENESIS, SPACE, convolutional decoder VAE, broadcast decoder VAE. Rows from top to bottom: object style, object shape, occlusion, crop.}
    \label{fig:ood_visualizations_multidsprites}
\end{figure}

\begin{figure}
    \centering
    \OodVizRow{objects_room}{object_style}
    
    \OodVizVspace
    
    \OodVizRow{objects_room}{object_color}
    
    \OodVizVspace
    
    \OodVizRow{objects_room}{occlusion}
    
    \OodVizVspace
    
    \OodVizRow{objects_room}{crop}
    
    \caption[{Inputs and reconstructions for OOD images in Objects Room.}]{\textbf{Inputs and reconstructions for OOD images in Objects Room.} Columns from left to right: MONet, Slot Attention, GENESIS, SPACE, convolutional decoder VAE, broadcast decoder VAE. Rows from top to bottom: object style, object color, occlusion, crop.}
    \label{fig:ood_visualizations_objects_room}
\end{figure}

\begin{figure}
    \centering
    \OodVizRow{shapestacks}{object_style}
    
    \OodVizVspace
    
    \OodVizRow{shapestacks}{object_color}
    
    \OodVizVspace
    
    \OodVizRow{shapestacks}{occlusion}
    
    \OodVizVspace
    
    \OodVizRow{shapestacks}{crop}
    
    \caption[{Inputs and reconstructions for OOD images in Shapestacks.}]{\textbf{Inputs and reconstructions for OOD images in Shapestacks.} Columns from left to right: MONet, Slot Attention, GENESIS, SPACE, convolutional decoder VAE, broadcast decoder VAE. Rows from top to bottom: object style, object color, occlusion, crop.}
    \label{fig:ood_visualizations_shapestacks}
\end{figure}

\begin{figure}
    \centering
    \OodVizRow{tetrominoes}{object_style}
    
    \OodVizVspace
    
    \OodVizRow{tetrominoes}{object_color}
    
    \OodVizVspace
    
    \OodVizRow{tetrominoes}{occlusion}
    
    \OodVizVspace
    
    \OodVizRow{tetrominoes}{crop}
    
    \caption[{Inputs and reconstructions for OOD images in Tetrominoes.}]{\textbf{Inputs and reconstructions for OOD images in Tetrominoes.} Columns from left to right: MONet, Slot Attention, GENESIS, SPACE, convolutional decoder VAE, broadcast decoder VAE. Rows from top to bottom: object style, object color, occlusion, crop.}
    \label{fig:ood_visualizations_tetrominoes}
\end{figure}

\fi

\backmatter
\pagestyle{myruled} %Restores the heading style as before "appruled". Remove if "appruled" is commented
\printbibliography
\end{document}